\documentclass[10pt, journal, compsoc]{IEEEtran}

\usepackage[algo2e, ruled, vlined]{algorithm2e}
\usepackage[utf8]{inputenc}
\usepackage{dblfloatfix}
\usepackage{fontawesome}
\usepackage[T1]{fontenc}
\usepackage{tcolorbox}
\usepackage{longtable}
\usepackage{multicol}
\usepackage{makecell}
\usepackage{scalerel}
\usepackage{amsfonts}
\usepackage{enumitem}
\usepackage{tabularx}
\usepackage{booktabs}
\usepackage{multirow}
\usepackage{flushend}
\usepackage{amsmath}
\usepackage{float}
\usepackage{tikz}
\usepackage{cite}
\usepackage{bm}

\definecolor{orcidlogocol}{HTML}{A6CE39}
\usetikzlibrary{svg.path}
\tikzset{
    orcidlogo/.pic={
        \fill[orcidlogocol] svg{M256,128c0,70.7-57.3,128-128,128C57.3,256,0,198.7,0,128C0,57.3,57.3,0,128,0C198.7,0,256,57.3,256,128z};
        \fill[white] svg{M86.3,186.2H70.9V79.1h15.4v48.4V186.2z}
                     svg{M108.9,79.1h41.6c39.6,0,57,28.3,57,53.6c0,27.5-21.5,53.6-56.8,53.6h-41.8V79.1z M124.3,172.4h24.5c34.9,0,42.9-26.5,42.9-39.7c0-21.5-13.7-39.7-43.7-39.7h-23.7V172.4z}
                     svg{M88.7,56.8c0,5.5-4.5,10.1-10.1,10.1c-5.6,0-10.1-4.6-10.1-10.1c0-5.6,4.5-10.1,10.1-10.1C84.2,46.7,88.7,51.3,88.7,56.8z};
    }
}

\newcommand\orcidicon[1]{\href{https://orcid.org/#1}{\mbox{\scalerel*{
    \begin{tikzpicture}[yscale=-1,transform shape]
        \pic{orcidlogo};
    \end{tikzpicture}
}{|}}}}

\usepackage{hyperref} 

\DeclareMathOperator*{\minimize}{\textbf{\textit{minimize~}}}
\DeclareMathAlphabet{\mathcal}{OMS}{cmsy}{m}{n}

\newcommand{\myset}[1]{\mathcal{\MakeUppercase #1}}
\newcommand{\mymtx}[1]{\mathbf{\MakeUppercase #1}}
\newcommand{\myvec}[1]{\mathbf{\MakeLowercase #1}}
\newcommand{\myshp}[1]{\mathbb{\MakeUppercase #1}}
\newcommand{\nomath}[1]{\mathrm{#1}}

\newcommand{\eg}{{\it e.g.}}
\newcommand{\ie}{{\it i.e.}}

\newcommand{\yhat}{\widehat{\mymtx{Y}}}
\newcommand{\y}{\mymtx{Y}}

\newcommand{\PyTorch}{{\sc PyTorch}}
\newcommand{\ReGENN}{{\sc ReGENN}}

\usepackage[
    justification=justified
]{caption}

\begin{document}

\title{Pay Attention to Evolution: Time Series Forecasting with Deep Graph-Evolution Learning}

\author{
    Gabriel~Spadon~\orcidicon{0000-0001-8437-4349},
    Shenda~Hong~\orcidicon{0000-0001-7521-5127},
    Bruno~Brandoli~\orcidicon{0000-0001-6167-8104},\\
    Stan~Matwin~\orcidicon{0000-0001-6629-8434},
    Jose F.~Rodrigues-Jr~\orcidicon{0000-0001-8318-1780},
    and~Jimeng~Sun~\orcidicon{0000-0003-1512-6426}
    \IEEEcompsocitemizethanks{%
        \IEEEcompsocthanksitem {\bf G. Spadon} and {\bf J. F. Rodrigues-Jr} are with the University of Sao Paulo, Institute of Mathematics and Computer Sciences, Sao Carlos -- SP, 13566-590, Brazil; {\bf G. Spadon} and {\bf J. Sun} are with the Georgia Institute of Technology, College of Computing, Atlanta -- GA, 30332-0365, USA; {\bf B. Brandoli} and {\bf S. Matwin} are with the Dalhousie University, Institute for Big Data Analytics, Halifax -- NS, B3H 1W5, Canada; {\bf S. Hong} is with the National Institute of Health Data Science and the Institute of Medical Technology, both at the Peking University, Beijing, 100191, China; {\bf S. Matwin} is with the Institute of Computer Science of the Polish Academy of Sciences, Warsaw, 525-000-94-01, Poland; {\bf J. F. Rodrigues-Jr} is with Université Grenoble Alpes, Faculty of Science, Saint-Martin-d'Hères, 38400, France; and, {\bf J. Sun} is with the University of Illinois Urbana-Champaign, Department of Computer Science, Champaign -- IL, 61801-2302, USA. \protect\\%
        \IEEEcompsocthanksitem Under a \href{https://creativecommons.org/licenses/by/4.0/}{Creative Commons License}.
        \IEEEcompsocthanksitem {\bf S. Hong} and {\bf B. Brandoli} contributed equally.
        \IEEEcompsocthanksitem Corresponding Author: {\bf J. Sun} (jimeng@illinois.edu).
        \IEEEcompsocthanksitem Digital Object Identifier no. \href{https://dx.doi.org/10.1109/TPAMI.2021.3076155}{10.1109/TPAMI.2021.3076155}.
    }%
}%

\markboth{IEEE Transactions on Pattern Analysis and Machine Intelligence}
{
    Spadon \MakeLowercase{\textit{et al.}}:
    Pay Attention to Evolution: Time Series Forecasting with Deep Graph-Evolution Learning
}

\IEEEtitleabstractindextext{%
    \begin{abstract}
        Time-series forecasting is one of the most active research topics in artificial intelligence.
        It has the power to bring light to problems in several areas of knowledge, such as epidemiological studies, healthcare inference, and climate change analysis.
        Applications in real-world time series should consider two factors for achieving reliable predictions: modeling dynamic dependencies among multiple variables and adjusting the model's intrinsic hyperparameters.
        An open gap in the literature is that statistical and ensemble learning approaches systematically present lower predictive performance than deep learning methods.
        The existing applications consistently disregard the data sequence aspect entangled with multivariate data represented in more than one time series.
        Conversely, this work presents a novel neural network architecture for time-series forecasting that combines the power of graph evolution with deep recurrent learning on distinct data distributions, named after Recurrent Graph Evolution Neural Network (\ReGENN).
        The idea is to infer multiple multivariate relationships between co-occurring time-series by assuming that the temporal data depends not only on inner variables and intra-temporal relationships (\ie, observations from itself) but also on outer variables and inter-temporal relationships (\ie, observations from other-selves).
        An extensive set of experiments was conducted comparing \ReGENN~with tens of ensemble methods and classical statistical ones.
        The results outperformed both statistical and ensemble-learning approaches, showing an improvement of 64.87\% over the competing algorithms on the SARS-CoV-2 dataset of the renowned John Hopkins University for 188 countries simultaneously.
        For further validation, we tested our architecture in two other public datasets of different domains, the PhysioNet Computing in Cardiology Challenge 2012 and Brazilian Weather datasets.
        We also analyzed the \emph{Evolution Weights} arising from the hidden layers of \ReGENN~to describe how the variables of the dataset interact with each other; and, as a result of looking at inter and intra-temporal relationships simultaneously, we concluded that time-series forecasting is majorly improved if paying attention to how multiple multivariate data synchronously evolve.
    \end{abstract}

    \begin{IEEEkeywords}
        Time Series, Graph Evolution, Representation Learning
    \end{IEEEkeywords}
}

\maketitle

\begin{figure*}[!h]
    \centering
    \includegraphics[width=\linewidth]{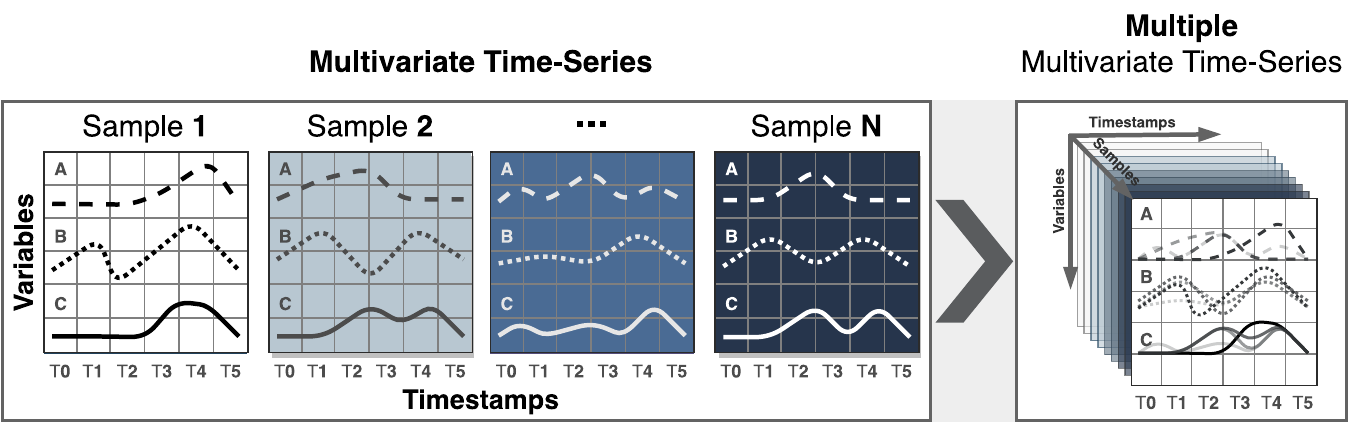}
    \caption{
    A multiple multivariate time-series forecasting problem, where each multivariate time-series (\ie, sample) shares the same domain, timestream, and variables.
    When stacking the time-series together, we assemble a tridimensional tensor with the axes describing samples, timestamps, and variables.
    The multiple samples have equal variables recorded during the same timestamps, meaning that samples are unique but all observed in the same way.
    By tackling the problem altogether, we leverage inner and outer variables besides intra- and inter-temporal relationships to improve forecasting.
    }
    \label{fig:problem-example}
\end{figure*}

\section{Introduction}

Time series refers to the persistent recording of a phenomenon along time, a continuous and intermittent unfolding of chronological events subdivided into past, present, and future.
In the last decades, time series analysis has been vital to predict dynamic phenomena on a wide range of applications, such as climate change~\cite{Lavender2018:Environment, Kao2020:Environment, Pryor2020:projectionsWindResources, Laurent2020:Environment}, financial market~\cite{Kelotra2020:Deep-ConvLSTM, Ananthi2020:CandlestickEegression, Jiao2021:ChangePoint}, land-use monitoring~\cite{Roy2020:Landsat, Zhu2020:Landsat, Yan2020:Landsat}, anomaly detection~\cite{Nikolay2015:Anomaly, Li2017:Markov, Hancock2020:FraudDetection}, energy consumption, and price forecasting~\cite{Deb2017:Environment, Orlov2020:renewableEnergyPrediction, Baratsas2021:energyPrediction}, apart from epidemiology and healthcare-related studies~\cite{Perros2019:TemporalPhenotyping, Rodrigues2019:PatientTrajectory, Rodrigues2020:LigDoctor, Afshar2020:TASTE, Yan2020:COVID19, Rodrigues2021:MinimalGRU}.
On such applications, an effective data-driven decision requires precise forecasting based on time series~\cite{Ienca2020:COVID19}.
A prime example is the SARS-CoV-2, COVID-19, or Coronavirus Pandemic~\cite{Velavan2020:Epidemic}, which is known to be highly contagious and cause increased pressure on healthcare systems worldwide~\cite{Omer2020:USPandemic}.
In this case, time-series analysis plays a vital role in planning a safe retake of fundamental activities by preventing economic systems' collapse.

Time series can be regarded as univariate or multivariate describing, respectively, single and multiple variables varying over time~\cite{Silvestrini2008:Aggregation}.
Recent techniques in time series have roots in the use of Artificial Neural Networks~\cite{Jain1996:ANN}, which contain a non-linear functioning that enables it to outperform classical algorithms~\cite{Zhang2001:ANN}.
Such techniques evolved into deep learning models for time-series forecasting, such as Haoyi {\it et al.}~\cite{informer2020} that used an informer component to enhance long-sequence time-series predictions but disregarded the inter-dependencies existing within different multivariate time-series, besides others from the spatiotemporal forecasting field.
For example, Seo {\it et al.}~\cite{seo2016} proposed the Graph Convolutional Recurrent Network (GCRN) from graph-structured and time-varying data by combining a Convolutional Neural Network (CNN)~\cite{Lecun1998:CNN} that identifies spatial structures and a Recurrent Neural Network (RNN)~\cite{Elman1990:RNN} that learns dynamic patterns; Li {\it et al.}~\cite{liYS018} introduced a spatiotemporal model for traffic forecasting with Diffusion over Convolutional Recurrent Neural Network (DCRNN); and, Zhang {\it et al.}~\cite{zhang18} presented a Gated Attention Network (GaAN) for forecasting traffic speed using a Graph Gated Recurrent Unit (GGRU) with a CNN controlling the attention head's importance.
Further contributions on the spatiotemporal field~\cite{Liang2018:GeoMAN, Bing2018:Traffic, Zhao2019:TGCN} employ traditional recurrent units, attention mechanisms, and even Graph Convolution Network (GCN)~\cite{Kipf2017:GCN}.
However, due to being designed to deal with spatial data, those models often fail to frame temporal dependencies as they aim to achieve state-of-the-art generalization across space and time at once.

Moreover, the LSTNet~\cite{Lai2018:LSTNet} encodes short-term data into low dimensional vectors by using a CNN for later decoding through an RNN; leverages from a recurrent-skip Gated Recurrent Unit for capturing long-term dependencies within the temporal data; and, incorporates an Autoregressive (AR) component in parallel to the non-linear neural network for preserving the scale of the output.
Similarly, the DSANet~\cite{Huang2019:DSANet} integrates an AR component with a dual self-attention network~\cite{Vaswani2017:Attention} with parallel convolutional components, a versatile idea for modeling global and local temporal patterns.
More recently, the MLCNN~\cite{Cheng2019:MLCNN} proposed the use of short and long-term prediction strategies for modeling temporal behavior through a multi-layer CNN together with Long Short-Term Memory (LSTM)~\cite{Hochreiter1997:LSTM} recurrent units.
However, although LSTNet, DSANet, and MLCNN are cutting-edge multivariate time-series forecasting algorithms, they do not explicitly address per-variable and inter-time-series dependencies, which weakens their forecasting ability in the face of higher-dimensional data.
Therefore, the state-of-the-art in time-series forecasting is bounded to a bidimensional space in which we understand the forecasting process by a non-linear function between time and variables.%

Differently, we hypothesize that {\it time-series are dependent on their inner variables, which are observations from themselves, and from outer variables provided by different time series that share the same timestream}.
For instance, the evolution of a biological species is not solely related to observations from itself but also from other species that share the same habitat, as they are all part of the same food chain.
The time series gains an increased dimensionality by considering the variables and the dependency aspect during the analysis.
Consequently, a previously considered bidimensional problem, in which a model's forecasting ability comes from observing relationships of variables over time, now becomes tridimensional, where forecasting means understanding the entanglement between variables of different time-series that co-occur in time.
Accordingly, time-series define an event that is not a consequence of a single chain of observations but a set of synchronous observations of many time-series.

\begin{table*}[!b]
\centering
\caption{Summary of context-specific notations.}
\begin{tabular}{r|l}                                                                                                                                           \toprule
\textbf{Notation}                                                         & \textbf{Definition}                                                             \\ \midrule
$\omega \in \mathbb{N}^{+}$                                               & Sliding window size                                                             \\
$w, z \in \mathbb{N}^{+}$                                                 & Number of training and testing (\ie, stride) timestamps                         \\
$s, t, v \in \mathbb{N}^{+}$                                              & Number of samples, timestamps, and variables                                    \\
$\mathbf{T} \in \mathbb{R}^{s \times t \times v}$                         & Tensor of multiple multivariate time-series                                     \\
$\mathbf{Y} \in \mathbb{R}^{s \times \omega \times v}$                    & Batched input of the first GSE and the Autoregression layers                    \\
$\mathbf{Y}_\alpha \in \mathbb{R}^{s \times \omega \times v}$             & Output of the first GSE and input of the encoder layers                         \\
$\mathbf{Y}_\varepsilon \in \mathbb{R}^{s \times \omega \times v}$        & Output of the encoder and input of the decoder layers                           \\
$\widetilde{\mathbf{Y}}_\varepsilon \in \mathbb{R}^{s \times z \times v}$ & Output from the first recurrent unit and input to the second one                \\
$\widetilde{\mathbf{Y}} \in \mathbb{R}^{s \times z \times v}$             & Output of the second recurrent unit and input of the second GSE layer           \\
$\mathbf{Y}_\psi \in \mathbb{R}^{s \times z \times v}$                    & Non-linear output yielded by the second GSE layer                               \\
$\mathbf{Y}_\lambda \in \mathbb{R}^{s \times z \times v}$                 & Linear output provided by the Autoregression layer                              \\
$\widehat{\mathbf{Y}} \in \mathbb{R}^{s \times z \times v}$               & The final result from the merging of the linear and non-linear outputs          \\
$\widehat{\mathbf{T}} \in \mathbb{R}^{s \times z \times v}$               & The ground truth expected from merging the linear and non-linear outputs        \\
$\myset{G}=\left<\myset{V}, \myset{E}\right>$                             & Graph in which $\myset{V}$ is the set of nodes and $\myset{E}$ the set of edges \\
$\mathbf{A} \in \mathbb{R}^{v \times v}$                                  & Adjacency matrix of co-occurring variables                                      \\
$\mathbf{A}_\mu \in \mathbb{R}^{v \times v}$                              & Neighbor-smoothed per-variable embedding shared between GSE layers              \\
$\mathbf{A}_\phi \in \mathbb{R}^{v \times v}$                             & Evolved adjacency matrix produced by the second GSE layer                       \\
$\mymtx{u} \circ \mymtx{v}$                                               & Batch-wise Hadamard product between matrices $\mymtx{u}$ and $\mymtx{v}$        \\
$\mymtx{u} \cdot \mymtx{v}$                                               & Batch-wise scalar product between matrices $\mymtx{u}$ and $\mymtx{v}$          \\
$\|\cdot\|_{\nomath{F}}$                                                  & The Frobenius norm of a given vector or matrix                                  \\
$\varphi(\cdot)$                                                          & Dropout regularization function                                                 \\
$\sigma_{g}(\cdot)$                                                       & Sigmoid activation function                                                     \\
$\sigma_{h}(\cdot)$                                                       & Hyperbolic tangent activation function                                          \\
$\cos_{\theta}(\cdot)$                                                    & Cosine matrix-similarity activation function                                    \\
\textsc{ReLU}$(\cdot)$                                                    & Rectified exponential linear unit activation function                           \\
\textsc{SOFTMAX}$(\cdot)$                                                 & Normalized exponential function activation function                             \\ \bottomrule
\end{tabular}
\label{tab:notations}
\end{table*}

For example, during the Coronavirus Pandemic, it is paramount to understand the disease's time-aware behavior in every country.
Despite progressing in different moments and locations, the pandemic's underlying mechanisms are supposed to follow similar (and probably interconnected) patterns.
Along these lines, looking individually at the development of the pandemic in each country, one can describe the problem in terms of multiple variables, like the number of confirmed cases, recovered people, and deaths.
However, when looking at all countries at once, the problem yields an additional data dimension, and each country becomes a multivariate sample of a broader problem, such as depicted in Fig.~\ref{fig:problem-example}.
In linguistic terms, we refer to such a problem as multiple multivariate time-series forecasting.

Along with these premises, in this study, we contribute with an unpreceded neural network that emerges from a graph-based time-aware auto-encoder with linear and non-linear components working in parallel to forecast multiple multivariate time-series simultaneously, named after Recurrent Graph Evolution Neural Network (\ReGENN).
{\it We refer to evolution as the natural progression of a process where the neural network iteratively optimizes a graph representing observations from the past until it reaches an evolved version of itself that generalizes on future data still to be observed}.
Accordingly, the underlying network structure of \ReGENN~is powered by two Graph Soft Evolution (GSE) layers, a further contribution of this study.
The GSE stands for a graph-based learning-representation layer that enhances the encoding and decoding processes by learning a shared graph across different time-series and timestamps.

The results we present are based on an extensive set of experiments, in which \ReGENN~surpassed a set of 49 competing algorithms from the fields of deep learning, machine learning, and time-series; among of which are single-target, multi-output, and multi-task regression algorithms in addition to univariate and multivariate time-series forecasting algorithms.
Aside from surpassing the state-of-the-art, \ReGENN~remained effective after three rounds of 30 ablation tests through distinct hyperparameters.
All experiments were carried out over the SARS-CoV-2, Brazilian Weather, and PhysioNet datasets.
In the task of epidemiology modeling on the SARS-CoV-2 dataset, we had improvements of at least 64.87\%.
We outperformed the task of climate forecasting on the Brazilian Weather dataset by at least 11.96\% and patient monitoring on intensive care units on the PhysioNet dataset by 7.33\%.
Furthermore, we analyzed the results using the \emph{Evolution Weights} from the GSE layers, which are the intermediate hidden adjacency matrices that arise from the graph-evolution process after going through the cosine similarity activation, showing that graphs shed new light on the understanding of non-linear black-box models.
Since multiple multivariate time-series is an ascending research topic, we understand \ReGENN~has implications in multiple domains, like economics, social sciences, and biology, in which different time-series share the same timestream and co-occur in time, mutually influencing one another.

In order to present our contributions, this paper is further divided into four sections.
We begin by proposing a layer and neural network architecture, besides detailing the methods used along with this study.
Subsequently, we display the experimental results compared to previous literature.
Next, we provide an overall discussion on our proposal and the achieved results.
Finally, we present the conclusions and final remarks.
Alongside, the Supplementary Material presents extended methods and additional results.

\section{Method}

\subsection{Preliminaries}
Hereinafter, we use bold uppercase letters to denote multidimensional matrices (\eg, $\mymtx{x}$), bold lowercase letters to vectors (\eg, $\myvec{x}$), and calligraphic letters to sets (\eg, $\myset{x}$).
Matrices, vectors, and sets can be used with subscripts.
For example, the element in the $i$-th row and $j$-th columns of a matrix is $\mymtx{x}_{ij}$, the $i$-th element of a vector is $\myvec{x}_i$, and the $j$-th element of a set is $\myset{x}_j$.
The transposed matrix of $\mymtx{x} \in \myshp{R}^{m \times n}$ is $\mymtx{x}^\nomath{T} \in \myshp{R}^{n \times m}$, and the transposed vector of $\myvec{x} \in \myshp{R}^{m \times 1}$ is $\myvec{x}^\nomath{T} \in \myshp{R}^{1 \times m}$, where $m$ and $n$ are arbitrary dimensions.
We display in Tab.~\ref{tab:notations} a summary of all context-specific notations.

\begin{figure*}[!h]
    \centering
    \includegraphics[width=\linewidth]{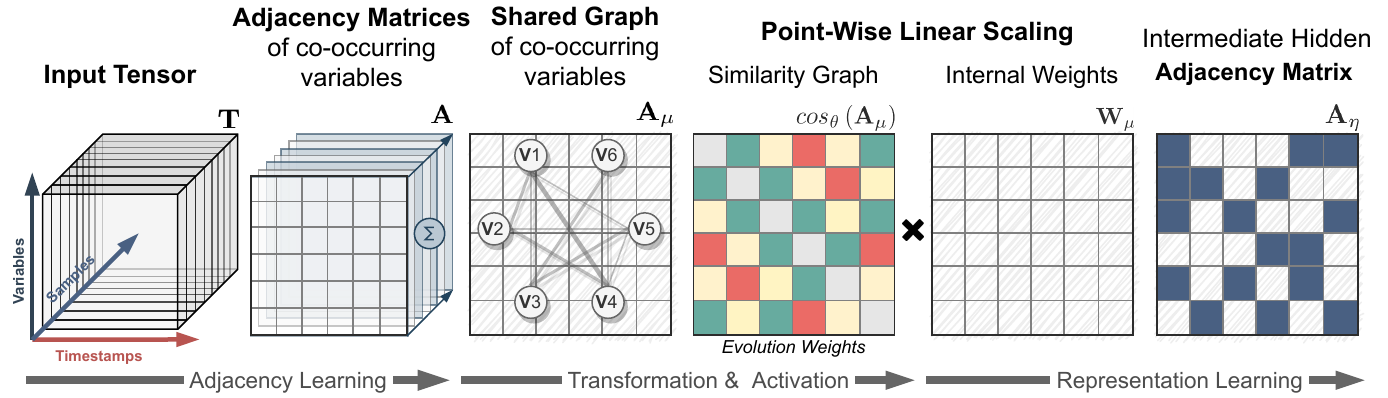}
    \caption{
    Graph Soft Evolution representation-learning, in which the set of multiple multivariate time-series is mapped into adjacency matrices of co-occurring variables.
    The matrices are element-wise summed to generate a shared graph among samples, which, after a linear transformation, goes through a similarity activation function and is scaled by an element-wise multiplication to produce an intermediate hidden adjacency-matrix with similarity properties inherent to the shared graph.
    }
    \label{fig:gse-workflow}
\end{figure*}

\begin{figure*}[!b]
    \centering
    \includegraphics[width=\linewidth]{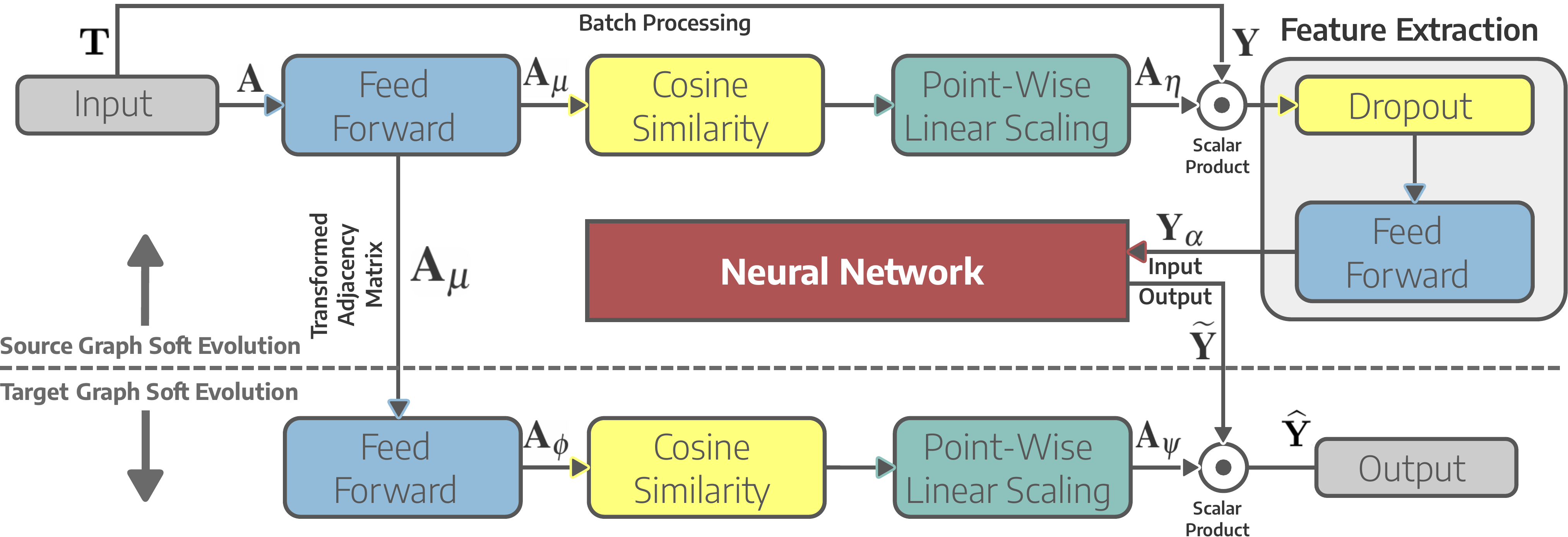}
    \caption{
    Graph Soft Evolution layers assembled for evolution-based learning.
    In such a case, the first GSE layer's output (\ie, source) will feed further layers of the neural network, whose result goes through the second GSE layer (\ie, target).
    The GSE, as the last layer, does not use regularizers nor linear transformations before the output.
    Contrarily, it outputs the result from the scalar product between the learned representation and the data propagated throughout the network.
    }
    \label{fig:gse-evolution}
\end{figure*}

\subsection{Graph Soft Evolution}
\label{sect:gse-formulation}
Graph Soft Evolution (GSE) stands for a representation-learning layer that, given a training dataset, builds a graph in the form of an adjacency matrix, as in Fig.~\ref{fig:gse-workflow}.
The GSE layer receives no graph as input but a set of multiple multivariate time-series.
The graph is built by tracking pairs of co-occurring variables, one sample at a time, and merging the results into a single co-occurrence graph shared among samples and timestamps.
We define co-occurring variables as two variables, from a multivariate time-series, with a non-zero value in the same timestamp -- in that case, we say one variable influences another and is influenced back.
The co-occurrence graph is the projection of a tridimensional tensor, $\mymtx{T}$, $\mymtx{T} \in \myshp{R}^{s \times w \times v}$, into a bidimensional one, $\mymtx{A}$, $\mymtx{A} \in \myshp{R}^{v \times v}$, describing variables' pair-wise time-invariant relationships.

\begin{figure*}[!b]
    \centering
    \includegraphics[width=\linewidth]{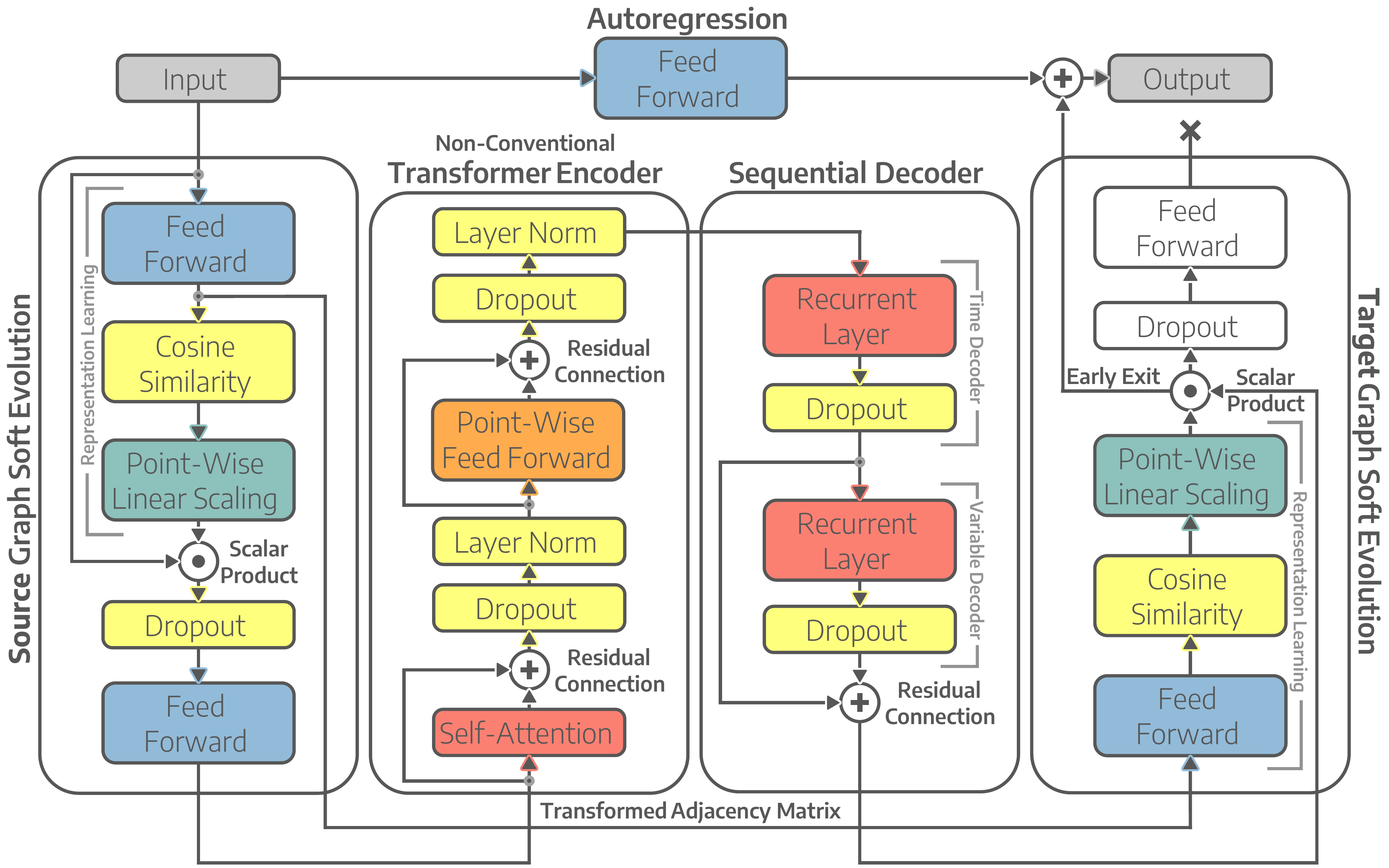}
    \caption{
    Data diagram of the Recurrent Graph Evolution Neural Network (\ReGENN), which has a linear component parallel to a non-linear one.
    The linear component has a feed-forward layer, and the non-linear one has an auto-encoder and two GSE layers.
    Although equal to the first, the last GSE layer yields an early output as it is not stacked with another layer.
    }
    \label{fig:regenn-architecture}
\end{figure*}

The co-occurrence graph $\myset{G}=\left<\myset{V}, \myset{E}\right>$ is symmetric and weighted.
It is composed of a set $\myset{V}$ of $|\myset{V}|$ nodes equal to the number of variables and another set $\myset{E}$ of $|\myset{E}|$ non-directed edges equal to the number of co-occurring variables.
A node $v \in \myset{V}$ corresponds to a variable from the time-series multivariate domain, and an edge $e \in \myset{e}$ is an ordered pair $\left<u,v\right> \equiv \left<v,u\right>$ of co-occurring variables $u,v \in \myset{V}$.
The edges' weight $f$ corresponds to the summation of the values of the variables $u,v \in \myset{V}$ whenever they co-occur in time, such that $f(u, v) = \sum_{i=0}^{s - 1} \sum_{j=0}^{w - 1} \mymtx{T}_{i,j,u} + \mymtx{T}_{i,j,v}$.
We use summation as the graph-merging operator when building $\mymtx{A}$ from $\myset{G}$ because, in the face of a zero-one input, a multiplicative operator would provide values close to zero, division would make those values overgrow toward infinity, subtraction would turn some of them into negative, while summation provides consistently positive values upper bounded to $2 \times (s \times w)$, helping to sustain the magnitude of variables' values.
This way, the whole graph is bounded to $w$, which is the number of timestamps existing in the training portion of the input tensor, and if a pair of variables never co-occur in the training data, no edge will be assigned to the graph, meaning that $\left<u, v\right> \not\in \myset{E}$, and $f(u, v) = 0$.

\noindent%
The GSE layer for an arbitrary graph is formulated as:
\begin{subequations}
    \renewcommand{\theequation}{\theparentequation.\arabic{equation}}
    \begin{align}
        \label{eq:S-GSE-1}
        \mymtx{A}_{\mu} & = \mymtx{W}_{\mu} \cdot \mymtx{A} + \textbf{b}_{\mu} \\
        \label{eq:S-GSE-2}
        \mymtx{A}_{\eta} & = \mymtx{W}_{\eta} \circ cos_{\theta}\left(\mymtx{A}_{\mu}\right) + \textbf{b}_{\eta} \\
        \label{eq:S-GSE-3}
        \y_{\alpha} & = \mymtx{W}_{\alpha} \cdot \varphi\left(\y \cdot \mymtx{A}_{\eta}\right) + \textbf{b}_{\alpha}
    \end{align}
\end{subequations}
where $\mymtx{W}_{\alpha},~\mymtx{W}_{\eta},~\mymtx{W}_{\mu} \in \myshp{R}^{v \times v}$ are symmetrically-unconstrained weights and $\myvec{b}_{\alpha},~\myvec{b}_{\eta},~\myvec{b}_{\mu} \in \myshp{R}^{v}$ the bias.
In Eq.~\ref{eq:S-GSE-1}, the layer starts by employing a linear transformation to the shared adjacency matrix $\mymtx{A}$, providing a per-variable embedding $\mymtx{A}_\mu$ after smoothing across neighbors.
Subsequently, in Eq.~\ref{eq:S-GSE-2}, it uses the cosine similarity (see the Supplementary Material) on the output of Eq.~\ref{eq:S-GSE-1}, \ie, $cos_\theta(\mymtx{A}_\mu)$, which is an intermediate activation function that provides the \emph{Evolution Weights} a symmetric per-variable similarity adjacency-matrix.
Under the same equation, the \emph{Evolution Weights} goes through a point-wise operator with a symmetrically-unconstrained weight matrix that enables the network to filter meaningful similarities from spurious similarities by learning that two variables influence one another on different scales and resulting in $\mymtx{A}_{\eta}$.
Next, in Eq.~\ref{eq:S-GSE-3}, it performs a batch-wise matrix-by-matrix multiplication between the adjacency matrix from Eq.~\ref{eq:S-GSE-2} and the batched input tensor $\y$ to combine the information from the graph, which generalizes samples and timestamps, with the time-series.
The result will be followed by a dropout regularizer~\cite{Srivastava2014:Dropout} and batch-wise matrix-by-matrix multiplication, where the final features from joining both tensors will be extracted and forwarded to the next layer of the network.%

The evolution concept comes from the cooperation between two GSE layers, one at the beginning (\ie, right after the input) and the other at the end (\ie, right before the output) of a neural network, such as in the example shown in Fig.~\ref{fig:gse-evolution}.
As evolution arises from sharing hidden weights between a pair of non-sequential layers, we named this process after \emph{Soft Evolution}.
Accordingly, the first layer (\ie, source) aims to learn the weights to scale the matrix and produce $\mymtx{A}_{\mu}$.
Such a result is the input of the second GSE layer (\ie, target), and it will be used for learning the evolved version of the adjacency matrix, referred to as $\mymtx{A}_{\phi}$ and produced as in Eq.~\ref{eq:S-GSE-1}.
Notice that in Fig.~\ref{fig:gse-evolution}, the source layer is different from the target one because we disregard the regularizer $\varphi$, trainable weights $\mymtx{W}_{\alpha}$, and bias $\myvec{b}_{\alpha}$ from Eq.~\ref{eq:S-GSE-3}.
They aim to enhance the feature-learning processes when multiple layers are stacked together and, as the last layer, GSE provides the output from already learned features through one last scalar product between the data propagated throughout the network, \ie, $\widetilde{\y}$, and the intermediate hidden adjacency-matrix, \ie, $\mymtx{A}_{\psi}$.

One can see that the source GSE layer has two constant inputs: the graph and input tensor.
The target GSE layer has two dynamic inputs, the shared graph from the source GSE layer and input propagated throughout the network.
In this work scope, we use an auto-encoder between GSE layers to learn data codings from the source layer's output, which will be decoded into a representation closest to the expected output and later re-scaled by the target layer.
In this sense, while the first layer learns a graph from the training data (\ie, past data) working as a pre-encoding feature-extraction layer, the second one re-learns (\ie, evolve) a graph at the end of the forecasting process based on future data, working as a post-decoding output-scaling layer.
When joining the GSE layers with the auto-encoder, we assemble the Recurrent Graph Evolution Neural Network (\ReGENN).

\subsection{Recurrent Graph Evolution Neural Network}

\ReGENN~is a graph-based time-aware auto-encoder with linear and non-linear components on parallel data-flows working together to provide future predictions based on past observations.
The linear component is the autoregression implemented as a feed-forward layer, and the non-linear component is made of an encoder and a decoder module powered by a pair of GSE layers.
Fig.~\ref{fig:regenn-architecture} shows how these components communicate from the input to the output, and, in the following, we detail their operation.

\subsubsection*{Autoregressive Component}
The non-periodical changes and constant progressions of the series across time usually decrease the performance of the network.
That is because the output scale loses significance compared to the input, which comes from the complexity and non-linear nature of neural networks in time-series forecasting tasks~\cite{Lai2018:LSTNet}.
Following a systematic strategy to deal with such a problem~\cite{Zilly2017:RecurrentHighway, Srivastava2015:Highway}, \ReGENN~leverages from an Autoregressive (AR) layer working as a linear feed-forward shortcut between the input and output, which for a tridimensional input, is algebraically defined as:
\begin{equation}
    \y_{\lambda} = \mymtx{W} \cdot \y + \myvec{b} \equiv
    \left(\sum_{i=1}^{\omega} \mymtx{W}_{i,z} \times \y_{s,i,v}\right) + \myvec{b}
    \label{eq:autoregression}
\end{equation}
where $\mymtx{W} \in \myshp{R}^{\omega \times z}$ are the weights and $\myvec{b} \in \myshp{R}^{z}$ the bias to be learned.
The output of the linear component, \ie, $\y_{\lambda} \in \myshp{R}^{s \times z \times v}$ as in Eq.~\ref{eq:autoregression}, is element-wise added to the non-linear component's output, \ie, $\y_{\psi} \in \myshp{R}^{s \times z \times v}$, so to produce the final predictions of the network $\yhat \in \myshp{R}^{s \times z \times v}$, formally given as $\yhat = \y_{\lambda} + \y_{\psi}$.
Subsequently, we describe the auto-encoder that produces the non-linear output of \ReGENN.

\subsubsection*{Encoder}
We use a non-conventional Transformer Encoder~\cite{Vaswani2017:Attention} that employs self-attention to learn an encoding from the features forwarded by the GSE layer.
It consists of multiple encoders joined through the scaled dot-product attention into a single set of encodings through multi-head attention.
The number of expected features by the Transformer Encoder must be a multiple of the number of heads in the multi-head attention.
Our encoder's non-conventionality comes from the fact that the first GSE layer's output goes through a single scaled dot-product attention on a single-head attention task.
That is because the number of features produced by the encoder is equal to the length of the sliding window, and through single-head attention, the window can assume any length.
The encoder module is defined as follows:
\begin{subequations}
    \begin{align}
        \label{eq:ENC-1}
        \mymtx{Y}_{\varepsilon} & = \text{\textsc{Self-Attention}}\left(\nomath{Q}: \mymtx{Y}_{\alpha}, \nomath{K}: \mymtx{Y}_{\alpha}, \nomath{V}: \mymtx{Y}_{\alpha}\right) \\
        \label{eq:ENC-2}
        \mymtx{Y}_{\varepsilon} & = \text{\textsc{Layer-Norm}}_{\gamma, \beta}\left(\mymtx{Y}_{\varepsilon} + \varphi\left(\mymtx{Y}_{\varepsilon}\right)\right) \\
        \label{eq:ENC-3}
        \mymtx{Y}_{\varepsilon} & = \mymtx{W}_{\varepsilon} \cdot \varphi\left(\text{\textsc{ReLU}}\left(\mymtx{W}_{\iota} \cdot \mymtx{Y}_{\varepsilon} + \textbf{b}_{\iota}\right)\right) + \textbf{b}_{\varepsilon} \\
        \label{eq:ENC-4}
        \mymtx{Y}_{\varepsilon} & = \text{\textsc{Layer-Norm}}_{\gamma, \beta}\left(\mymtx{Y}_{\varepsilon} + \varphi\left(\mymtx{Y}_{\varepsilon}\right)\right)
    \end{align}
\end{subequations}
where self-attention in Eq.~\ref{eq:ENC-1} is a particular case of the multi-head attention, in which the input \emph{query} Q, \emph{key} K, and \emph{value} V of the scaled dot-product attention, \ie, ${\text{\textsc{softmax}}}\left({\nomath{Q} \cdot \nomath{K}^{\mathrm{T}}} \div \sqrt {d_{k}}\right) \cdot \nomath{V}$, are equal; and $d_k$ is the dimension of the keys.
The attention results are followed by a dropout regularization~\cite{Srivastava2014:Dropout}, a residual connection~\cite{He2016:Residual}, and a layer normalization~\cite{Ba2016:LayerNormalization} as in Eq.~\ref{eq:ENC-2} to ensure generalization.
The first two layers work to avoid overfitting and gradient vanishing, while the last one normalizes the output such that the samples among the input have zero mean and unit variance $\gamma\left(\Delta\left(\mymtx{Y}_{\varepsilon} + \varphi\left(\mymtx{Y}_{\varepsilon}\right)\right)\right) + \beta$, where $\Delta$ is the normalization function, and $\gamma$ and $\beta$ are parameters to be learned.
After, in Eq.~\ref{eq:ENC-3}, the intermediate encoding goes through a double linear layer, a point-wise feed-forward layer, which, in this case, consists of two linear transformations in sequence with a ReLU activation in between, having the weights $\mymtx{W}_{\varepsilon}, \mymtx{W}_{\iota}$ and bias $\textbf{b}_{\varepsilon}, \textbf{b}_{\iota}$ as optimizable parameters.
Finally, the transformed encoding goes through one last set of generalizing operations, as shown in Eq.~\ref{eq:ENC-4}.
The resulting encoding $\mymtx{Y}_{\varepsilon} \in \myshp{R}^{s \times \omega \times v}$ is a tensor with the time-axis length matching the size of the sliding window $\omega$.

\subsubsection*{Decoder}
The previous encoding will be decoded by two sequence-to-sequence layers, which are Long Short Term Memory (LSTM)~\cite{Hochreiter1997:LSTM} units.
The decoder operates in two of the tridimensional axes of the encoding, the time-axis and variable-axis, once at a time.
Under the sample and time-axis, the time-axis decoder tracks temporal dependencies, looks for a set of weights that generalizes across variables, and translates the window-sized input into a stride-sized output:
\begin{equation}
    \hspace{-.2cm}
    \begin{aligned}
        \myvec{f}^{\left(t\right)}\left(v\right) & = \sigma_g\left(\mymtx{W}_\myvec{f}^{\left(t\right)}\cdot\left[\myvec{h}^{\left(t\right)}\left(v-1\right),\mymtx{Y}_{\varepsilon}^{\left(t\right)}\left(v\right)\right] + \myvec{b}_\myvec{f}^{\left(t\right)}\right) \\
        \myvec{i}^{\left(t\right)}\left(v\right) & = \sigma_g\left(\mymtx{W}_\myvec{i}^{\left(t\right)}\cdot\left[\myvec{h}^{\left(t\right)}\left(v-1\right),\mymtx{Y}_{\varepsilon}^{\left(t\right)}\left(v\right)\right] + \myvec{b}_\myvec{i}^{\left(t\right)}\right) \\
        \myvec{o}^{\left(t\right)}\left(v\right) & = \sigma_g\left(\mymtx{W}_\myvec{o}^{\left(t\right)}\cdot\left[\myvec{h}^{\left(t\right)}\left(v-1\right),\mymtx{Y}_{\varepsilon}^{\left(t\right)}\left(v\right)\right] + \myvec{b}_\myvec{o}^{\left(t\right)}\right) \\
        \widetilde{\mymtx{C}}^{\left(t\right)}\left(v\right) & = \sigma_h\left(\mymtx{W}_\mymtx{C}^{\left(t\right)}\cdot\left[\myvec{h}^{\left(t\right)}\left(v-1\right),\mymtx{Y}_{\varepsilon}^{\left(t\right)}\left(v\right)\right] + \myvec{b}_\mymtx{C}^{\left(t\right)}\right) \\
        \mymtx{C}^{\left(t\right)}\left(v\right) & = \myvec{f}^{\left(t\right)}\left(v\right) \circ \mymtx{C}^{\left(t\right)}\left(v-1\right) + \myvec{i}^{\left(t\right)}\left(v\right) \circ \widetilde{\mymtx{C}}^{\left(t\right)}\left(v-1\right) \\
        \myvec{h}^{\left(t\right)}\left(v\right) & = \varphi\left(\myvec{o}^{\left(t\right)}\left(v\right) \circ \sigma_h\left(\mymtx{C}^{\left(t\right)}\left(v\right)\right)\right)
    \end{aligned}
    \label{eq:LSTM-1}
\end{equation}
where $v$ is the $v$-th variable of the $t$-th time-series group, and the weights $\mymtx{W}_\myvec{f}^{(t)}, \mymtx{W}_\myvec{i}^{(t)}, \mymtx{W}_\mymtx{C}^{(t)}, \mymtx{W}_\myvec{o}^{(t)} \in \mathbb{R}^{\omega \times z}$ and bias $\myvec{b}_\myvec{f}^{(t)}, \myvec{b}_\myvec{i}^{(t)}, \myvec{b}_\mymtx{C}^{(t)}, \myvec{b}_\myvec{o}^{(t)} \in \mathbb{R}^{z}$ are parameters to be learned.
Along with Eq.~\ref{eq:LSTM-1}, we refer to $\myvec{f}$ as the forget gate's activation vector, $\myvec{i}$ as the input and update gate's activation vector, $\myvec{o}$ as the output gate's activation vector, $\widetilde{\mymtx{C}}$ as the cell input activation vector, $\mymtx{C}$ as the cell state vector, and $\myvec{h}$ as the hidden state vector.
The last hidden state vector goes through a dropout regularization $\varphi$ before the next LSTM in the sequence.

Under the time and variable-axis, the next recurrent unit decodes the variable-axis from the partially-decoded encoding, searches for a set of weights that generalizes across time, and translates patterns arising from different variables on the same timestamp.
The set of variables within the time-series does not necessarily imply a sequence, which does not interfere in the decoding process as long as the variables are always kept in the same order; a common approach used with boosting trees, such as the XGBoost~\cite{Chen2016:XGBoost} algorithm.
The second LSTM, in which $t$ is the $t$-th timestamp of the $v$-th variable group, is algebraically defined as follows:
\begin{equation}
    \hspace{-.2cm}
    \begin{aligned}
        \myvec{f}^{\left(v\right)}\left(t\right) & = \sigma_g\left(\mymtx{W}_\myvec{f}^{\left(v\right)}\cdot\left[\myvec{h}^{\left(v\right)}\left(t-1\right),\widetilde{\mymtx{Y}}_{\varepsilon}^{\left(v\right)}\left(t\right)\right] + \myvec{b}_\myvec{f}^{\left(v\right)}\right) \\
        \myvec{i}^{\left(v\right)}\left(t\right) & = \sigma_g\left(\mymtx{W}_\myvec{i}^{\left(v\right)}\cdot\left[\myvec{h}^{\left(v\right)}\left(t-1\right),\widetilde{\mymtx{Y}}_{\varepsilon}^{\left(v\right)}\left(t\right)\right] + \myvec{b}_\myvec{i}^{\left(v\right)}\right) \\
        \myvec{o}^{\left(v\right)}\left(t\right) & = \sigma_g\left(\mymtx{W}_\myvec{o}^{\left(v\right)}\cdot\left[\myvec{h}^{\left(v\right)}\left(t-1\right),\widetilde{\mymtx{Y}}_{\varepsilon}^{\left(v\right)}\left(t\right)\right] + \myvec{b}_\myvec{o}^{\left(v\right)}\right) \\
        \widetilde{\mymtx{C}}^{\left(v\right)}\left(t\right) & = \sigma_h\left(\mymtx{W}_\mymtx{C}^{\left(v\right)}\cdot\left[\myvec{h}^{\left(v\right)}\left(t-1\right),\widetilde{\mymtx{Y}}_{\varepsilon}^{\left(v\right)}\left(t\right)\right] + \myvec{b}_\mymtx{C}^{\left(v\right)}\right) \\
        \mymtx{C}^{\left(v\right)}\left(t\right) & = \myvec{f}^{\left(v\right)}\left(t\right) \circ \mymtx{C}^{\left(v\right)}\left(t-1\right) + \myvec{i}^{\left(v\right)}\left(t\right) \circ \widetilde{\mymtx{C}}^{\left(v\right)}\left(t-1\right) \\
        \myvec{h}^{\left(v\right)}\left(t\right) & = \widetilde{\mymtx{Y}}_{\varepsilon} + \varphi\left(\myvec{o}^{\left(v\right)}\left(t\right) \circ \sigma_h\left(\mymtx{C}^{\left(v\right)}\left(t\right)\right)\right)
    \end{aligned}
    \label{eq:LSTM-2}
\end{equation}
where $\widetilde{\mymtx{Y}}_{\varepsilon}$ is the partially-decoded encoding, and the weights $\mymtx{W}_\myvec{f}^{(v)}$, $\mymtx{W}_\myvec{i}^{(v)}$, $\mymtx{W}_\mymtx{C}^{(v)}$, $\mymtx{W}_\myvec{o}^{(v)} \in \mathbb{R}^{z \times z}$ and bias $\myvec{b}_\myvec{f}^{(v)}$, $\myvec{b}_\myvec{i}^{(v)}$, $\myvec{b}_\mymtx{C}^{(v)}$, $\myvec{b}_\myvec{o}^{(v)} \in \mathbb{R}^{z}$ are internal parameters.
The description of the notations within Eq.~\ref{eq:LSTM-1} holds for Eq.~\ref{eq:LSTM-2}.
Their difference is the residual connection with the partially-decoded encoding $\widetilde{\mymtx{Y}}_{\varepsilon}$ at the last hidden state vector after the dropout regularization $\varphi$.
After the decoding is complete, the last recurrent unit output $\widetilde{\mymtx{Y}}$ goes through the second GSE layer, producing the non-linear output $\mymtx{Y}_\psi$ of \ReGENN.

\subsection{Optimization Strategy}
\ReGENN~operates on a tridimensional space shared between samples, timestamps, and variables.
In such a space, it carries out a time-based optimization strategy.
The training process iterates over the time-axis of the dataset, showing to the network how the variables within a subset of time-series behave as time goes by and later repeating the process through subsets of different samples in distinct batches of preset window-size.
However, the shared graph $\mymtx{A}$ is built during the first epoch of the training phase.
Such a process happens in parallel to the optimization process, not impacting training stability nor convergence.
After that point, $\mymtx{A}$ will remain as it is during the training and testing.

The network's weights are shared among the entire dataset and optimized towards best generalization simultaneously across samples, timestamps, and variables.
We used Adam~\cite{Kingma2015:Adam}, a gradient descent-based algorithm, to optimize the model and, as the optimization criterion, the Mean Absolute Error (MAE), which is a generalization of the Support Vector Regression~\cite{Vapnik1997:SVR} with soft-margin criterion where $\Omega$ is the set of internal parameters of \ReGENN, $\widehat{\mathbf{Y}}$ is the network's output, and $\widehat{\mathbf{T}}$ the ground truth:
\begin{equation}
    \displaystyle{
        \minimize_{\Omega}
            \sum_{i=1}^{s} \sum_{j=1}^{w} \left|\widehat{\mathbf{Y}}_{i,j} - \widehat{\mathbf{T}}_{i,j}\right|
    }
\end{equation}

Due to the SARS-CoV-2 data behave as a streaming dataset, we adopted a transfer learning approach to train the network on that dataset.
Transfer learning shares knowledge across different domains by using pre-trained weights of another neural network.
The approach we adopted, although different, resembles Online Deep Learning~\cite{Sahoo2018:OnlineLearning}.
The main idea is to train the network on incremental slices of the time-axis, such that the pre-trained weights of a previous slice are used to initialize the weights of the network in the next slice.
This technique aims not only to achieve better forecasting performance but also to show that \ReGENN~is superior to other algorithms throughout the pandemic.

\section{Experiments}

\subsection{Experimental Setup}

\subsubsection*{Datasets}
The results are based on three datasets, all of which are multi-sample, multivariate, and vary over time.
The first dataset describes the Coronavirus Pandemic, referred to as SARS-CoV-2, made available by John Hopkins University~\cite{Dong2020:COVID19}.
It describes 3 variables through 120 days for 188 countries and varies from the first day of the pandemic to the day it completed four months of duration.
The second one is the Brazilian Weather dataset collected from 253 sensors during 1,095 days regarding 4 variables.
The third dataset is from the 2012 PhysioNet Computing in Cardiology Challenge~\cite{Silva2012:Physionet}, from which we are using 9 variables across 48 hours recorded from 11,988 ICU patients.

\noindent%
The variables within the datasets are:
\begin{itemize}
    \item{\bf SARS-CoV-2:} Number of Recovered, Number of Infected, and Number of Deaths;
    \item{\bf Brazilian Weather: } Minimum Temperature, Maximum Temperature, Solar Radiation, and Rainfall; and,
    \item{\bf PhysioNet: } Non-Invasive Diastolic Arterial Blood Pressure (mmHg), Non-Invasive Systolic Arterial Blood Pressure (mmHg), Invasive Diastolic Arterial Blood Pressure (mmHg), Non-Invasive Mean Arterial Blood Pressure (mmHg), Invasive Systolic Arterial Blood Pressure (mmHg), Invasive Mean Arterial Blood Pressure (mmHg), Urine Output (mL), Heart Rate (bpm), and Weight (Kg).
\end{itemize}

The number of datasets that can be used for multiple multivariate time-series forecasting is still limited.
Although the ones we experimented with have a small number of variables, there is no theoretical upper bound for the number of samples, timestamps, and variables \ReGENN~can handle.
However, as \ReGENN~deals with dense tridimensional tensors, increasing any of the axes $(samples \times time \times variables)$ in size will make the tensor grow at an almost-cubic rate, quickly overflowing the GPU Memory.
Accordingly, \ReGENN~ends up suffering from a trade-off between the dataset's size and the available hardware.%

\subsubsection*{Data Pre-processing}
The datasets were individually min-max normalized into a zero-one scale on the variable-axis to avoid gradient spikes and speed up training.
We used the training data to define the min-max parameters required for normalization.
Such parameters were applied for normalizing the test data as well.
As a consequence of normalizing the entire dataset, the network's output will follow a similar scale.
Therefore, the output was inversely transformed to what it was before the normalization for evaluating the results.

A simplistic yet effective approach to train time-series algorithms is through the Sliding Window technique~\cite{Keogh2004:SegmentingServey}, also referred to as Rolling Window.
The window size is well known to be a highly sensitive hyperparameter~\cite{Frank2000:WindowSize, Frank2001:SlidingWindow}.
Consequently, we followed a non-tunable approach, in which we set the window size before the experiments, just taking into consideration the context and domain of the datasets.
These values were used across all window-based experiments, including the baselines and ablation tests.
It is noteworthy that most of the machine-learning algorithms are not meant to handle time-variant data, such that no sliding window was used in those cases.
Conversely, we considered training timestamps as features and those reserved for testing as multi-task regression labels.

On the deep learning algorithms, we used a window size of 7 days for training and reserved 7 days for validation (between the training and test sets) to predict the last 14 days of the SARS-CoV-2 dataset.
The 7-7-14 split idea comes from the disease incubation period, which is of 14 days.
On the other hand, we used a window size of 84 days and reserved 28 days for validation to predict the last 56 days in the Brazilian Weather dataset.
The 84-28-56 split is based on the seasonality of the weather data, such that we will look to the previous 3 months (a weather-season window) to predict the last 2 months of the upcoming season.
Finally, we used a window size of 12 hours for training and 6 hours for validation to predict the last 6 hours of the PhysioNet dataset.
The 12-6-6 split comes from the fact that patients in ICUs are in a critical state, such that predictions within 24 hours are more useful than long-term predictions.

\subsubsection*{Algorithms \& Ensembles.}
Many existing algorithms are limited because they neither support multi-task nor multi-output regression, making these algorithms even more limited to tasks when data is tridimensional.
The most straightforward yet effective approach we followed to compare them to \ReGENN~was to create a chain of ensembles\footnote{See \textit{Regressor Chain} at \url{https://bit.ly/3hBfxTA}.}.
In such a case, each estimator makes predictions on order specified by the chain using all of the available features provided to the model plus the predictions of earlier models in the chain.
The number of estimators in each experiment varies according to the type of the ensemble and the type of the algorithms, and the final performance is the average of each estimator's performance.
For simplicity sake, we grouped the algorithms as follows:

\noindent%
\begin{itemize}[nolistsep, noitemsep]
    \item[\faCertificate] Corresponds to tridimensional compliant algorithms of single estimators;
    \item[\faCircle] Describes multivariate algorithms -- $s$ estimators, one estimator for each sample;
    \item[\faBullseye] Consists of multi-output and multi-task algorithms -- $v$ estimators, one estimator per variable;
    \item[\faDotCircleO] Indicates single-target algorithms -- $v \times z$ estimators, one estimator per variable and stride; and,
    \item[\faCircleO] Represents univariate algorithms -- $s \times v$ estimators, one estimator for each sample and variable.
\end{itemize}

\subsubsection*{Evaluation Metrics}
As time-series forecasting works as a time-aware regression problem, our goal remains in predicting values that resemble the ground truth the most.
As such, we used three evaluation metrics, the Mean Squared Logarithmic Error (MSLE), Mean Absolute Error (MAE), and Root Mean Squared Error (RMSE), that, despite different, have a complementary meaning.
The MSLE uses the logarithmic scale to assess how well the model can predict values close to zero.
Contrarily, the MAE uses the absolute operator to explain how the model fared among the median values.
In another perspective, the RMSE uses the square root to assess the model's ability to predict the larger values, which might be outliers.
The metrics' formal definition is as follows:

\begin{equation}
    MSLE = \frac{1}{n}~\sum^{n}_{i=1}log\left(\frac{\widehat{\mathbf{Y}}_i + 1}{\widehat{\mathbf{T}}_i + 1}\right)^2
\end{equation}

\begin{equation}
    MAE = \frac{1}{n}~{\sum^{n}_{i=1}\left|\widehat{\mathbf{Y}}_{i} - \widehat{\mathbf{T}}_{i}\right|}
\end{equation}

\begin{equation}
    RMSE = \sqrt{\frac{1}{n}~\sum^{n}_{i=1}\left(\widehat{\mathbf{Y}}_i - \widehat{\mathbf{T}}_i\right)^2}
\end{equation}

\subsubsection*{Hyperparameter Tuning}
\ReGENN~has two hyperparameters able to change the dimension of the weights' tensors, which are the window size (\ie, input size) and the stride size (\ie, output size).
As already discussed, both were set before the experiments, and none of them were tuned towards any performance improvement.
The trade-off of having fewer hyperparameters is to spend more energy on training the network towards a better performance.
We are focusing on the network optimizer, gradient clipping, learning rate scheduler, and dropout regularization when we refer to tunable hyperparameters.
Along these lines, we followed a controlled and limited exploratory approach similar to a random grid-search, starting with \PyTorch's defaults.
The tuning process was on the validation set, intentionally reserved for measuring the network improvement.
The tuning process follows by updating the hyperparameters whenever observing better results on the validation set, leading us to a set of optimized but no optimum hyperparameters.
We used the set of optimized hyperparameters to evaluate \ReGENN~on the test set and the default values for all the other algorithms~\cite{Pedregosa2011:Sklearn, Chen2016:XGBoost, Prokhorenkova2018:CatBoost} unless explicitly required for working with a particular dataset, as was the case of LSTNet~\cite{Lai2018:LSTNet}, DSANet~\cite{Huang2019:DSANet}, and MLCNN~\cite{Cheng2019:MLCNN}.
The complete list of hyperparameters is in the Supplementary Material.

\begin{figure*}[!h]
    \centering
    \includegraphics[width=\linewidth]{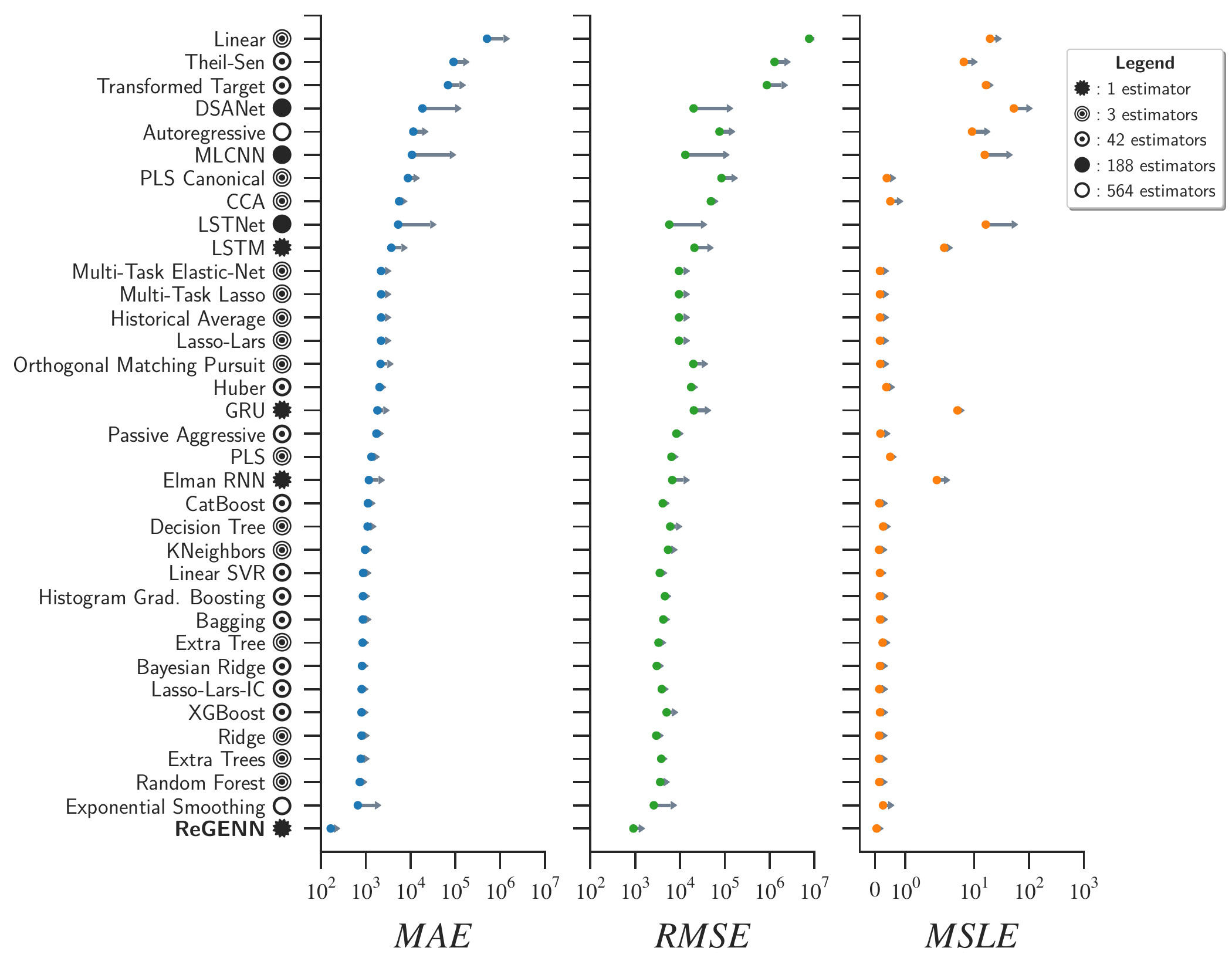}
    \vspace{-.72cm}\caption{
    Baseline results for the SARS-CoV-2 dataset over the Mean Absolute Error (MAE), Root Mean Square Error (RMSE), and Mean Squared Logarithmic Error (MSLE).
    The results are presented in descending order of MAE (\ie, worst performance at the top).
    The results confirmed \ReGENN's superior performance as it is the algorithm with the lowest error and standard deviation -- the improvement in the experiment was no lower than 64.87\%.
    In the image, the algorithms are symbol-encoded based on their type and number of estimators; we use gray arrows to report the standard deviation of the results.
    The negative deviation, which is equal to the positive one, was suppressed for improved readability.
    }
    \label{fig:covid-baselines}
\end{figure*}

\begin{table*}[!h]
\centering
\caption{
Ablation results for the SARS-CoV-2 dataset using ReGENN's data-flow but no GSE layer.
The experiment varies between Recurrent Unit (\textsc{RU}) and their directional flag, differing between Bidirectional (\textsc{B}) and Unidirectional (\textsc{U}).
We use \textsc{E} to designate the non-conventional Transformer Encoder and \textsc{AR} the Autoregression component along with the table.
}
\label{tab:covid-ablation}
\begin{tabular}{r|c|c|c|c|c|c|c|c|c}
\Xhline{3\arrayrulewidth}
\multicolumn{1}{r|}{\textsc{\textbf{Recurrent Unit (RU)}}} &
  \multicolumn{3}{c|}{\textbf{Elman RNN}} &
  \multicolumn{3}{c|}{\textbf{GRU}} &
  \multicolumn{3}{c}{\textbf{LSTM}} \\ \Xhline{3\arrayrulewidth}
\multicolumn{1}{r|}{\textsc{Network Architecture}} &
  \textsc{MAE} &
  \textsc{RMSE} &
  \textsc{MSLE} &
  \textsc{MAE} &
  \textsc{RMSE} &
  \textsc{MSLE} &
  \textsc{MAE} &
  \textsc{RMSE} &
  \textsc{MSLE} \\ \Xhline{3\arrayrulewidth}
\multicolumn{1}{r|}{${}_{\mathrm{B}}\text{\textsc{RU}}$} &
  424.32 &
  2819.11 &
  0.07 &
  3130.47 &
  28063.07 &
  7.78 &
  1800.23 &
  20264.27 &
  3.95 \\ \hline
\multicolumn{1}{r|}{$\mathrm{E} \rightarrow {}_{\mathrm{B}}\text{\textsc{RU}}$} &
  4161.35 &
  28755.33 &
  11.08 &
  596.77 &
  3446.67 &
  0.17 &
  704.80 &
  4088.05 &
  0.20 \\ \hline
\multicolumn{1}{r|}{$\left(\mathrm{E} \rightarrow {}_{\mathrm{B}}\text{\textsc{RU}} + {}_{\mathrm{B}}\text{\textsc{RU}}\right) + \mathrm{AR}$} &
  1624.44 &
  17245.24 &
  3.84 &
  388.93 &
  1888.70 &
  0.09 &
  1598.03 &
  19056.28 &
  3.89 \\ \hline
\multicolumn{1}{r|}{$\left(\mathrm{E} \rightarrow {}_{\mathrm{B}}\text{\textsc{RU}} + {}_{\mathrm{U}}\text{\textsc{RU}}\right) + \mathrm{AR}$} &
  \textbf{257.15} &
  \textbf{1438.72} &
  \textbf{0.09} &
  366.06 &
  2158.32 &
  0.10 &
  4064.52 &
  31196.10 &
  11.30 \\ \hline
\multicolumn{1}{r|}{$\left(\mathrm{E} \rightarrow {}_{\mathrm{B}}\text{\textsc{RU}}\right) + \mathrm{AR}$} &
  1471.49 &
  17050.08 &
  3.74 &
  1502.20 &
  16799.96 &
  3.67 &
  \textbf{211.93} &
  \textbf{1181.31} &
  \textbf{0.08} \\ \hline
\multicolumn{1}{r|}{$\mathrm{E} \rightarrow {}_{\mathrm{U}}\text{\textsc{RU}}$} &
  1949.65 &
  16925.69 &
  3.84 &
  968.83 &
  5982.35 &
  0.22 &
  2778.55 &
  12830.03 &
  0.21 \\ \hline
\multicolumn{1}{r|}{$\left(\mathrm{E} \rightarrow {}_{\mathrm{U}}\text{\textsc{RU}} + {}_{\mathrm{B}}\text{\textsc{RU}}\right) + \mathrm{AR}$} &
  372.80 &
  2155.36 &
  0.11 &
  \textbf{208.88} &
  \textbf{1141.56} &
  \textbf{0.08} &
  1380.99 &
  16370.41 &
  3.66 \\ \hline
\multicolumn{1}{r|}{$\left(\mathrm{E} \rightarrow {}_{\mathrm{U}}\text{\textsc{RU}} + {}_{\mathrm{U}}\text{\textsc{RU}}\right) + \mathrm{AR}$} &
  1472.05 &
  16491.81 &
  3.72 &
  4052.71 &
  31384.80 &
  11.30 &
  1713.72 &
  18531.68 &
  3.81 \\ \hline
\multicolumn{1}{r|}{$\left(\mathrm{E} \rightarrow {}_{\mathrm{U}}\text{\textsc{RU}}\right) + \mathrm{AR}$} &
  1350.35 &
  16673.11 &
  3.69 &
  2852.28 &
  25869.18 &
  7.60 &
  260.66 &
  1513.92 &
  0.10 \\ \hline
\multicolumn{1}{r|}{${}_{\mathrm{U}}\text{\textsc{RU}}$} &
  1175.99 &
  6716.42 &
  2.08 &
  1825.66 &
  20400.04 &
  4.95 &
  3723.88 &
  21022.60 &
  2.83  \\ \Xhline{2\arrayrulewidth}
\multicolumn{7}{r|}{\textsc{Recurrent Graph Evolution Neural Network (ReGENN)}} &
  \textbf{165.41} &
  \textbf{915.92} &
  \textbf{0.05} \\ \hline
\multicolumn{7}{r|}{\textsc{ReGENN without Variable Decoder}} &
   197.51 &
   1141.04 &
   0.15 \\ \hline
\multicolumn{7}{r|}{\textsc{ReGENN without Autoregression}} &
  8208.53 &
  50089.84 &
  19.19 \\ \hline
\multicolumn{7}{r|}{\textsc{Autoregression Only}} &
  11529.79 &
  76118.62 &
  9.15 \\ \Xhline{3\arrayrulewidth}
\multicolumn{7}{r|}{\textsc{Overall Improvement}} &
  20.81\% &
  19.77\% &
  35.72\% \\ \Xhline{3\arrayrulewidth}
\end{tabular}%
\end{table*}

\begin{figure*}[!h]
    \centering
    \includegraphics[width=\linewidth]{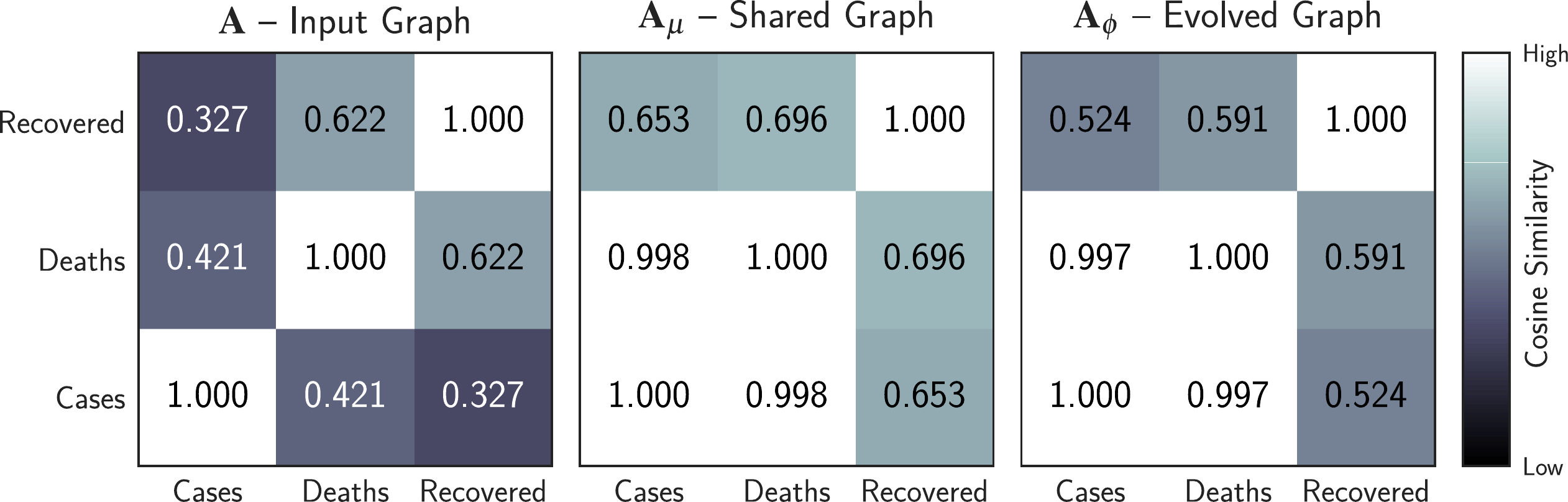}
    \caption{
    Set of \emph{Evolution Weights}, \ie, cosine-similarity activated hidden weights extracted from \ReGENN~at the end of the network training.
    The images compare the cosine similarity between the set of variables within the SARS-CoV-2 dataset.
    }
    \label{fig:covid-weights}
\end{figure*}

\subsubsection*{Computer Environment}
The experiments related to machine-learning and time-series algorithms were carried out on a Linux-based system with 64 CPUs and 750 GB of RAM.
The experiments related to deep-learning on the SARS-CoV-2 dataset were carried out on a Linux-based system with 56 CPUs, 256 GB of RAM, and 8 GPUs (Titan X -- Pascal).
The Brazilian Weather and PhysioNet datasets were tested on a different system with 32 CPUs, 512 GB of RAM, and 8 GPUs (Titan X -- Maxwell).
While CPU-based experiments are even across all CPU architectures, the same does not hold for GPUs, such that the GPU model and architecture must match to guarantee reproducibility.
Aiming at complete reproducibility, we disclose not only the source code of \ReGENN~on GitHub\footnote{Available at \url{https://bit.ly/2YDBrOo}.}, but also the scripts, pre-processed data, and snapshots of all trained networks on a public folder at Google Drive\footnote{Available at \url{https://bit.ly/30csLiJ}.}.

\subsection{Results}
Subsequently, we go through the experimental results in each one of the benchmark datasets.
Due to the fact that not all the algorithms perform evenly across them, we display the 34 most prominent ones out of the 49 tested algorithms; for the extended results, please refer to the Supplementary Material.
We also discuss the ablation experiments, which were carried out with \ReGENN's hyperparameters; in the Supplementary Material, we provide two other rounds of this same experiment using as hyperparameters \PyTorch's defaults\footnote{Available at \url{https://bit.ly/2QuWRsD}.} and other settings recurrently employed in the literature.
Additionally, we draw explanations about the \emph{Evolution Weights}, \ie, intermediate adjacency matrices from the GSE layers, by using the cosine similarity on the adjacency matrices of co-occurring variables (see Section~\ref{sect:gse-formulation}).%

\subsubsection*{Experiment on SARS-CoV-2 dataset}
The SARS-CoV-2 has being updated daily since the beginning of the Coronavirus Pandemic.
We used a self-to-self transfer-learning approach to train the network in slices of time due to such a dataset's streaming nature.
In short, the network was re-trained every 15 days with new incoming data, using as starting weights, the pre-trained weights from the network trained in the past 15 days so to evaluate \ReGENN's performance over the pandemic's progression.

In such a case, when the pandemic completed 45 days, the first time-slice in which \ReGENN~was trained, it outperformed the second-placed algorithm, the Orthogonal Matching Pursuit, by 27.27\% on the Mean Absolute Error (MAE), 16.50\% on the Root Mean Square Error (RMSE), and 38.87\% on the Mean Squared Logarithmic Error (MSLE).
For all subsequent time-slices (\ie, 60, 75, 90, 105, and 120 days), the second-placed algorithm was the Exponential Smoothing, which was outperformed with improvement no lower than 47.40\% on the MAE, 17.19\% on the RMSE, and 37.39\% on the MSLE.
We further detail the results on the time-slices mentioned above in the Supplementary Material.
However, in Fig.~\ref{fig:covid-baselines}, we detail the results on the complete dataset of 120 days, in which \ReGENN~surpassed the Exponential Smoothing, the second-best algorithm, by 75.21\% on the MAE, 64.87\% on the RMSE, and 79.61\% on the MSLE.

As a result of the analysis of the dataset in time-slices, we noticed that, as time goes by and more information is available on the SARS-CoV-2 dataset, the problem becomes more challenging to solve by looking individually at each country and more natural when looking at all of them together.
Although countries have their particularities, which make the disease spread in different ways, the main goal is to decrease the spreading, such that similarities between the historical data of different countries provide for finer predictions.
Furthermore, we observed that not all the estimators within an ensemble perform in the same way in the face of different countries.
Due to the \ReGENN~capability of observing inter-and intra-relationships between time-series, it performs better on highly uncertain cases like this one.

Subsequently, we present the ablation results, in which we utilized the same data-flow as \ReGENN~but no GSE layer while systematically changing the decoder architecture.
We provide results using different Recurrent Units (RU), including the Elman RNN~\cite{Elman1990:RNN}, LSTM~\cite{Hochreiter1997:LSTM}, and GRU~\cite{Chung2014:GRU}.
We also varied the recurrent unit's directional flag between Unidirectional (U) and Bidirectional (B).
That because a unidirectional recurrent unit tracks only forward dependencies while a bidirectional one tracks both forward and backward dependencies.
A summarized tag describes each test's network architecture; for example, $\left(\mathrm{E} \rightarrow {}_{\mathrm{U}}\text{\textsc{RU}} + {}_{\mathrm{B}}\text{\textsc{RU}}\right) + \mathrm{AR}$ means the model has a Transformer Encoder ($\mathrm{E}$) as the encoder, a Unidirectional Recurrent Unit as the time-axis decoder, and a Bidirectional Recurrent Unit as the variable-axis decoder.
Besides that, the output of the decoder is element-wise added to the Autoregression (AR) output.
The table shows results with and without the encoder and AR component, besides using a single recurrent unit only for time-axis decoding.

\begin{figure*}[!h]
    \centering
    \includegraphics[width=\linewidth]{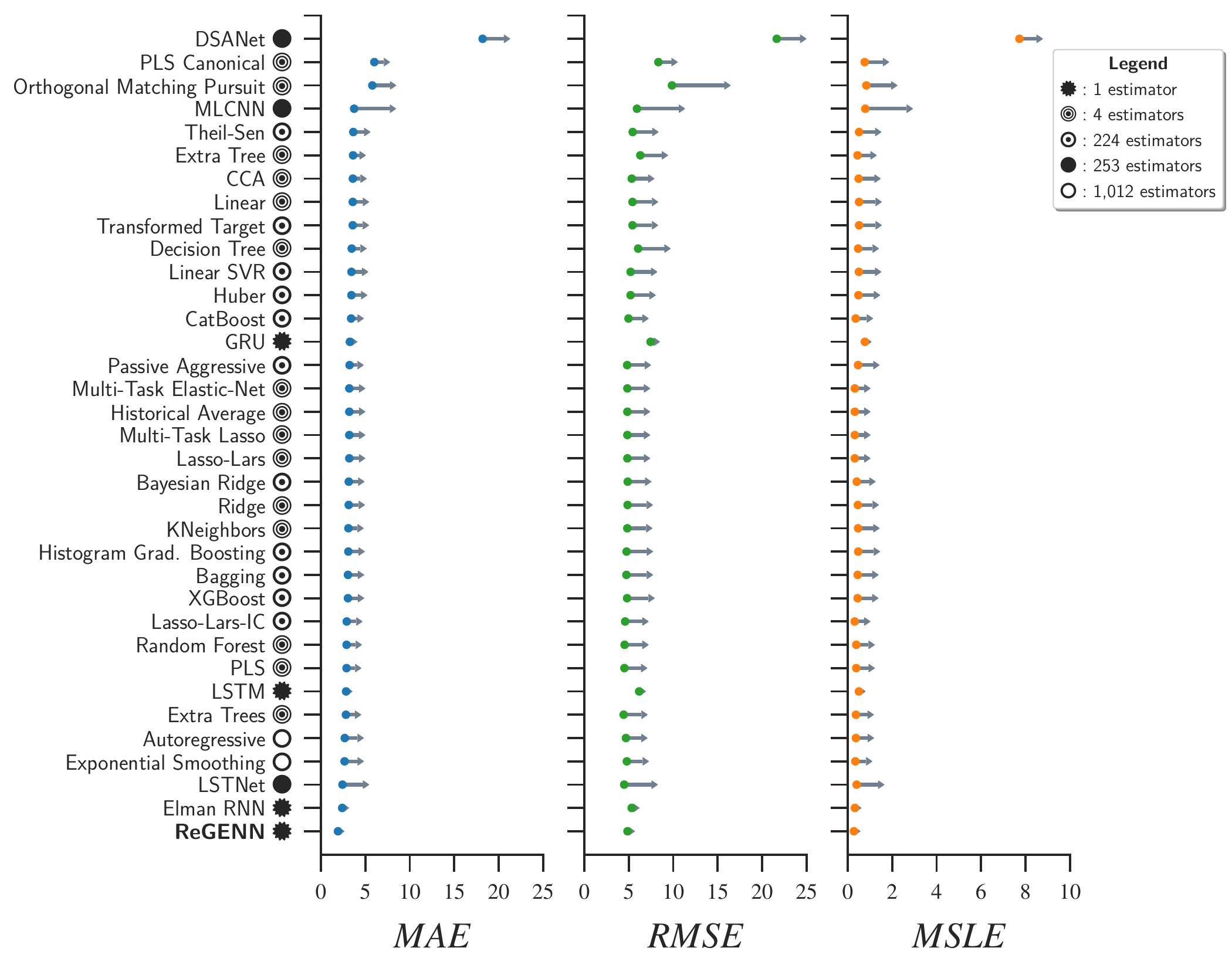}
    \vspace{-.72cm}\caption{
    Baseline results for the Brazilian Weather dataset presented in descending order of MAE.
    In this experiment, \ReGENN\ once more outperformed all the competing algorithms, demonstrating versatility by performing well even on a highly-seasonal dataset with improvement no lower than 11.95\%.
    In the face of seasonality, the Elman RNN surpassed the Exponential Smoothing, the previously second-best algorithm.
    In the image, the algorithms are symbol-encoded based on their type and number of estimators; we use gray arrows to report the results' standard deviation.
    The negative deviation, which is equal to the positive one, was suppressed for improved readability.
    }
    \label{fig:weather-baselines}
\end{figure*}

\begin{table*}[!h]
\centering
\caption{
Ablation results from experimenting over the Brazilian Weather dataset that was conducted by varying the Recurrent Unit (\textsc{RU}) and its directional flag between Bidirectional (\textsc{B}) and Unidirectional (\textsc{U}). We use \textsc{E} to designate the non-conventional Transformer Encoder and \textsc{AR} the feed-forward Autoregressive linear component.}

\label{tab:weather-ablation}
\begin{tabular}{r|c|c|c|c|c|c|c|c|c}
\Xhline{3\arrayrulewidth}
\multicolumn{1}{r|}{\textsc{\textbf{Recurrent Unit (RU)}}} &
  \multicolumn{3}{c|}{\textbf{Elman RNN}} &
  \multicolumn{3}{c|}{\textbf{GRU}} &
  \multicolumn{3}{c}{\textbf{LSTM}} \\ \Xhline{3\arrayrulewidth}
\multicolumn{1}{r|}{\textsc{Network Architecture}} &
  \textsc{MAE} &
  \textsc{RMSE} &
  \textsc{MSLE} &
  \textsc{MAE} &
  \textsc{RMSE} &
  \textsc{MSLE} &
  \textsc{MAE} &
  \textsc{RMSE} &
  \textsc{MSLE} \\ \Xhline{3\arrayrulewidth}
\multicolumn{1}{r|}{${}_{\mathrm{B}}\text{\textsc{RU}}$} &
  2.56 &
  5.90 &
  0.45 &
  2.19 &
  5.00 &
  0.28 &
  \textbf{2.20} &
  \textbf{5.04} &
  \textbf{0.28} \\ \hline
\multicolumn{1}{r|}{$\mathrm{E} \rightarrow {}_{\mathrm{B}}\text{\textsc{RU}}$} &
  2.70 &
  5.66 &
  0.37 &
  2.33 &
  5.19 &
  0.28 &
  2.57 &
  5.42 &
  0.29 \\ \hline
\multicolumn{1}{r|}{$\left(\mathrm{E} \rightarrow {}_{\mathrm{B}}\text{\textsc{RU}} + {}_{\mathrm{B}}\text{\textsc{RU}}\right) + \mathrm{AR}$} &
  \textbf{2.18} &
  \textbf{5.52} &
  \textbf{0.37} &
  2.38 &
  5.72 &
  0.41 &
  2.44 &
  5.75 &
  0.41 \\ \hline
\multicolumn{1}{r|}{$\left(\mathrm{E} \rightarrow {}_{\mathrm{B}}\text{\textsc{RU}} + {}_{\mathrm{U}}\text{\textsc{RU}}\right) + \mathrm{AR}$} &
  2.84 &
  6.47 &
  0.55 &
  2.24 &
  5.05 &
  0.28 &
  3.17 &
  7.42 &
  0.80 \\ \hline
\multicolumn{1}{r|}{$\left(\mathrm{E} \rightarrow {}_{\mathrm{B}}\text{\textsc{RU}}\right) + \mathrm{AR}$} &
  3.58 &
  8.13 &
  0.98 &
  3.31 &
  7.57 &
  0.82 &
  3.61 &
  8.19 &
  0.96 \\ \hline
\multicolumn{1}{r|}{$\mathrm{E} \rightarrow {}_{\mathrm{U}}\text{\textsc{RU}}$} &
  3.12 &
  5.86 &
  0.32 &
  2.81 &
  5.69 &
  0.34 &
  2.96 &
  5.54 &
  0.30 \\ \hline
\multicolumn{1}{r|}{$\left(\mathrm{E} \rightarrow {}_{\mathrm{U}}\text{\textsc{RU}} + {}_{\mathrm{B}}\text{\textsc{RU}}\right) + \mathrm{AR}$} &
  2.36 &
  5.69 &
  0.41 &
  2.23 &
  5.11 &
  0.28 &
  2.47 &
  5.77 &
  0.41 \\ \hline
\multicolumn{1}{r|}{$\left(\mathrm{E} \rightarrow {}_{\mathrm{U}}\text{\textsc{RU}} + {}_{\mathrm{U}}\text{\textsc{RU}}\right) + \mathrm{AR}$} &
  2.52 &
  5.83 &
  0.42 &
  \textbf{2.10} &
  \textbf{4.97} &
  \textbf{0.28} &
  4.88 &
  10.24 &
  1.61 \\ \hline
\multicolumn{1}{r|}{$\left(\mathrm{E} \rightarrow {}_{\mathrm{U}}\text{\textsc{RU}}\right) + \mathrm{AR}$} &
  4.30 &
  9.38 &
  1.34 &
  3.25 &
  7.55 &
  0.81 &
  5.25 &
  10.69 &
  1.75 \\ \hline
\multicolumn{1}{r|}{${}_{\mathrm{U}}\text{\textsc{RU}}$} &
  2.40 &
  5.32 &
  0.32 &
  3.26 &
  7.45 &
  0.77 &
  2.83 &
  6.17 &
  0.50  \\ \Xhline{2\arrayrulewidth}
\multicolumn{7}{r|}{\textsc{Recurrent Graph Evolution Neural Network (ReGENN)}} &
  \textbf{1.92} &
  \textbf{4.86} &
  \textbf{0.28} \\ \hline
\multicolumn{7}{r|}{\textsc{ReGENN without Variable Decoder}} &
   1.95 &
   4.85 &
   0.27 \\ \hline
\multicolumn{7}{r|}{\textsc{ReGENN without Autoregression}} &
  2.00 &
  4.95 &
  0.27 \\ \hline
\multicolumn{7}{r|}{\textsc{Autoregression Only}} &
  2.69 &
  4.68 &
  0.37 \\ \Xhline{3\arrayrulewidth}
\multicolumn{7}{r|}{\textsc{Overall Improvement}} &
  8.42\% &
  2.16\% &
  0.00\% \\ \Xhline{3\arrayrulewidth}
\end{tabular}%
\end{table*}

\begin{figure*}[!h]
    \centering
    \includegraphics[width=\linewidth]{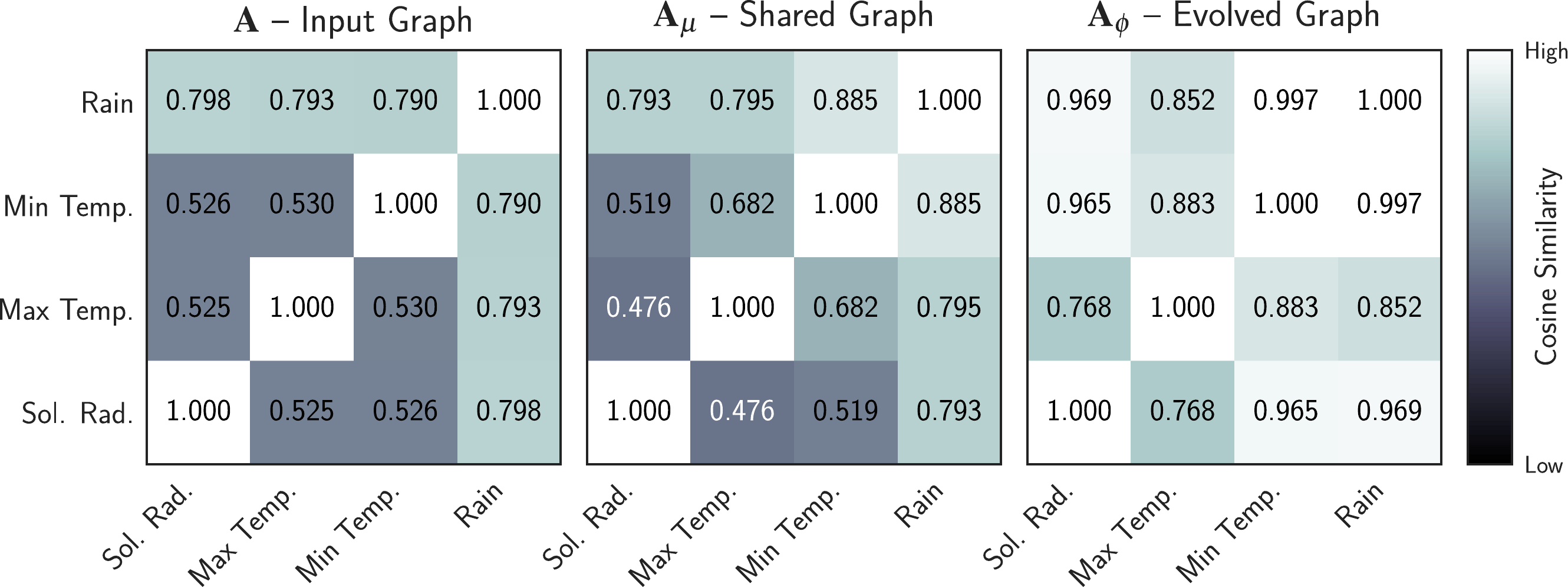}
    \caption{
    The \emph{Evolution Weights} from \ReGENN~for the Brazilian Weather dataset, in which we use the cosine similarity activation function on the network's hidden weights to compare the relationship between pairs of variables.
    The image uses "Sol. Rad." as a shortening for Solar Radiation, "Temp" for Temperature, "Max" for Maximum, and "Min" for Minimum.
    }
    \label{fig:weather-weights}
\end{figure*}

According to the ablation results in Tab.~\ref{tab:covid-ablation}, the improvement of \ReGENN~is slightly smaller than previously reported.
That is because its performance not only comes from the GSE layer but also from how the network handles the multiple multivariate time-series data.
Consequently, the ablation experiments reveal that some models without GSE layers are enough to surpass all the competing algorithms.
However, when using \ReGENN, we can improve them further and achieve 20.81\% of additional reduction on the MAE, 19.77\% on the RMSE, and 35.72\% on the MSLE.

Fig.~\ref{fig:covid-weights} shows the \emph{Evolution Weights} originated from applying the cosine similarity on the hidden adjacency matrices of \ReGENN.
When comparing the input and evolved graphs, the number of cases and deaths have a mild similarity.
That might come from the fact that diagnosing infected people was already a broad concern at the beginning of the pandemic.
The problem did not go away, but more infected people were discovered as more tests were made, and also because the disease spread worldwide.
A similar scenario can be drawn from the number of recovered people and the number of cases, as infected people with mild or no symptoms were unaware of being infected.
Contrarily, we can see that the similarity between recovered and deaths decreases over time, which comes from the fact that, as more tests are made, the mortality rate drops to a stable threshold due to the increased number of recovered people.

\subsubsection*{Experiment on Brazilian Weather dataset}
The Brazilian Weather dataset is a highly seasonal dataset with a higher number of samples, variables, and timestamps than the previous one.
For simplicity's sake, in this experiment, \ReGENN~was trained on the whole training set at once.
The results are in Fig.~\ref{fig:weather-baselines}, in which \ReGENN~was the first-placed algorithm, followed by the Elman RNN in second.
\ReGENN\ overcame the Elman RNN by 11.95\% on the MAE, 11.96\% on the RMSE, and 25.84\% on the MSLE.

We noticed that all the algorithms perform quite similarly for this dataset.
The major downside for most algorithms comes from predicting small values close to zero, as noted by the MSLE results.
In such a case, the ensembles showed a high variance when compared to \ReGENN.
We believe this is why the Elman RNN shows performance closer to \ReGENN~rather than to Exponential Smoothing, the third-placed algorithm, as \ReGENN~has a single estimator, while the Exponential Smoothing is an ensemble of estimators.
Another understanding of why some algorithms underperform on the MSLE is related to their difficulty to track temporal dependencies, such as weather seasonality.

The ablation results are in Tab.~\ref{tab:weather-ablation}, in which we observed again that the network without the GSE layers already surpasses the baselines.
When decommissioning the GSE layers of \ReGENN~and using GRU instead of LSTM on the decoder, we observed a 3.86\% improvement on the MAE, 10.02\% on the RMSE, and 25.34\% on the MSLE when compared to the Elman RNN results.
Using \ReGENN~instead, we achieve a further performance gain of 8.42\% on the MAE and 2.16\% on the RMSE over the ablation experiment.

\begin{figure*}[!h]
    \centering
    \includegraphics[width=\linewidth]{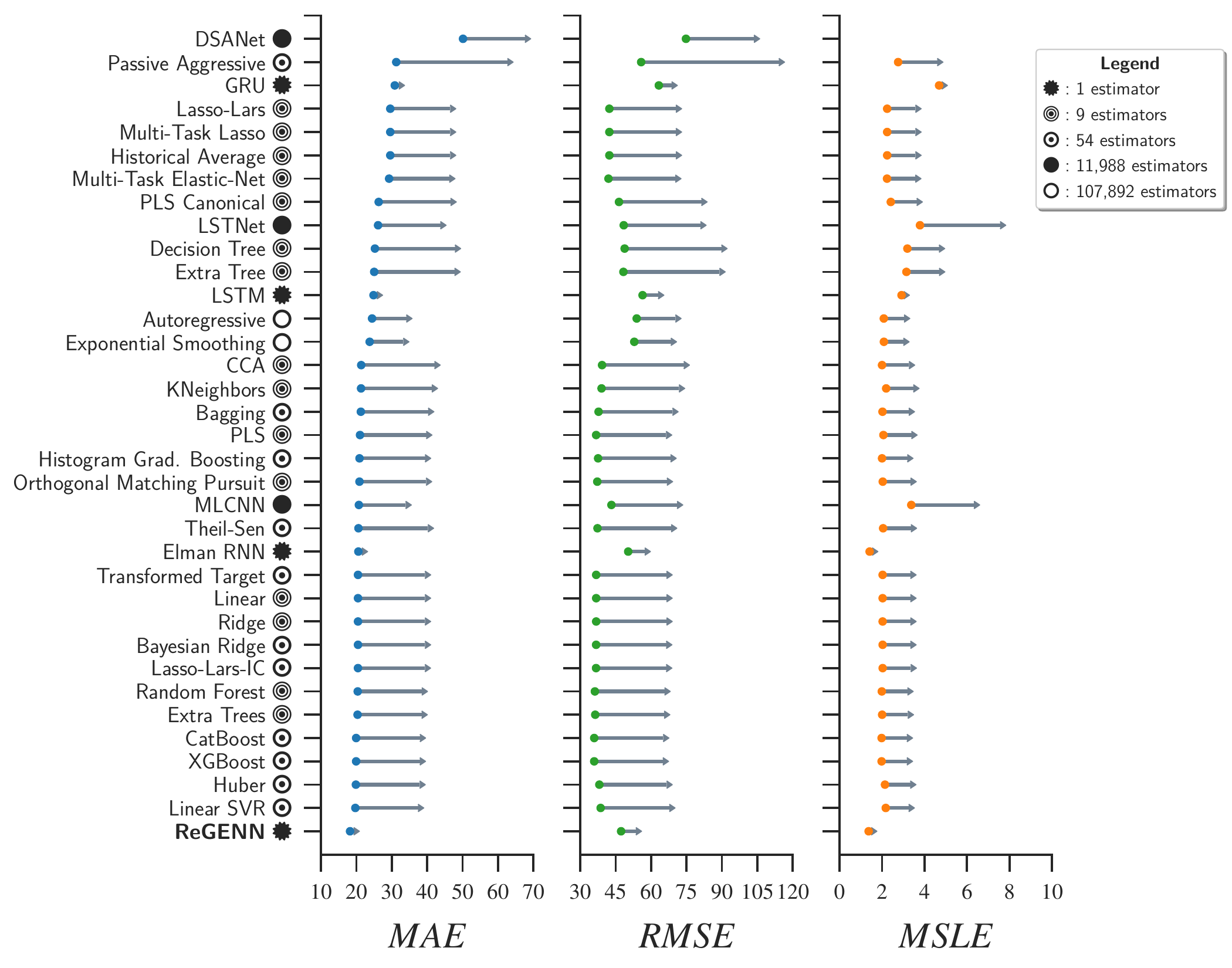}
    \vspace{-.72cm}\caption{
    Baseline results for the PhysioNet Computing in Cardiology Challenge 2012 dataset presented in descending order of MAE, in which \ReGENN~was the algorithm with the best performance followed by the Linear SVR.
    The comparative improvement was no lower than 7.33\%, but, in this case, \ReGENN~yielded an RMSE compatible with the Linear SVR.
    In the image, the algorithms are symbol-encoded based on their type and number of estimators; we use gray arrows to report standard deviation.
    The negative deviation, which is equal to the positive one, was suppressed for improved readability.
    }
    \label{fig:physionet-baselines}
\end{figure*}

\begin{table*}[!h]
\centering
\caption{
Ablation results over the PhysioNet Computing in Cardiology Challenge 2012 dataset, in which the Recurrent Unit (\textsc{RU}) is changed together with its directional flag, varying between Bidirectional (\textsc{B}) and Unidirectional (\textsc{U}). We use \textsc{E} as a shortening for the Transformer Encoder and \textsc{AR} for the Autoregressive linear component.
}
\label{tab:physionet-ablation}
\begin{tabular}{r|c|c|c|c|c|c|c|c|c}
\Xhline{3\arrayrulewidth}
\multicolumn{1}{r|}{\textsc{\textbf{Recurrent Unit (RU)}}} &
  \multicolumn{3}{c|}{\textbf{Elman RNN}} &
  \multicolumn{3}{c|}{\textbf{GRU}} &
  \multicolumn{3}{c}{\textbf{LSTM}} \\ \Xhline{3\arrayrulewidth}
\multicolumn{1}{r|}{\textsc{Network Architecture}} &
  \textsc{MAE} &
  \textsc{RMSE} &
  \textsc{MSLE} &
  \textsc{MAE} &
  \textsc{RMSE} &
  \textsc{MSLE} &
  \textsc{MAE} &
  \textsc{RMSE} &
  \textsc{MSLE} \\ \Xhline{3\arrayrulewidth}
\multicolumn{1}{r|}{${}_{\mathrm{B}}\text{\textsc{RU}}$} &
  19.95 &
  50.47 &
  1.38 &
  24.80 &
  56.42 &
  3.01 &
  19.49 &
  50.46 &
  1.40 \\ \hline
\multicolumn{1}{r|}{$\mathrm{E} \rightarrow {}_{\mathrm{B}}\text{\textsc{RU}}$} &
  19.11 &
  49.16 &
  1.41 &
  24.17 &
  54.81 &
  3.00 &
  18.88 &
  48.55 &
  1.40 \\ \hline
\multicolumn{1}{r|}{$\left(\mathrm{E} \rightarrow {}_{\mathrm{B}}\text{\textsc{RU}} + {}_{\mathrm{B}}\text{\textsc{RU}}\right) + \mathrm{AR}$} &
  \textbf{18.72} &
  \textbf{48.60} &
  \textbf{1.37} &
  18.55 &
  47.86 &
  1.39 &
  \textbf{18.42} &
  \textbf{47.78} &
  \textbf{1.37} \\ \hline
\multicolumn{1}{r|}{$\left(\mathrm{E} \rightarrow {}_{\mathrm{B}}\text{\textsc{RU}} + {}_{\mathrm{U}}\text{\textsc{RU}}\right) + \mathrm{AR}$} &
  18.97 &
  49.59 &
  1.38 &
  \textbf{18.54} &
  \textbf{48.90} &
  \textbf{1.37} &
  18.46 &
  48.33 &
  1.36 \\ \hline
\multicolumn{1}{r|}{$\left(\mathrm{E} \rightarrow {}_{\mathrm{B}}\text{\textsc{RU}}\right) + \mathrm{AR}$} &
  18.81 &
  50.20 &
  1.38 &
  18.67 &
  48.59 &
  1.39 &
  18.54 &
  47.35 &
  1.38 \\ \hline
\multicolumn{1}{r|}{$\mathrm{E} \rightarrow {}_{\mathrm{U}}\text{\textsc{RU}}$} &
  19.46 &
  50.24 &
  1.44 &
  19.42 &
  51.03 &
  1.44 &
  19.78 &
  51.29 &
  1.44 \\ \hline
\multicolumn{1}{r|}{$\left(\mathrm{E} \rightarrow {}_{\mathrm{U}}\text{\textsc{RU}} + {}_{\mathrm{B}}\text{\textsc{RU}}\right) + \mathrm{AR}$} &
  18.76 &
  48.81 &
  1.39 &
  18.65 &
  48.64 &
  1.38 &
  18.55 &
  48.28 &
  1.38 \\ \hline
\multicolumn{1}{r|}{$\left(\mathrm{E} \rightarrow {}_{\mathrm{U}}\text{\textsc{RU}} + {}_{\mathrm{U}}\text{\textsc{RU}}\right) + \mathrm{AR}$} &
  19.02 &
  49.75 &
  1.43 &
  18.64 &
  48.81 &
  1.38 &
  18.63 &
  48.93 &
  1.37 \\ \hline
\multicolumn{1}{r|}{$\left(\mathrm{E} \rightarrow {}_{\mathrm{U}}\text{\textsc{RU}}\right) + \mathrm{AR}$} &
  18.98 &
  48.67 &
  1.45 &
  18.82 &
  49.98 &
  1.41 &
  18.88 &
  50.31 &
  1.39 \\ \hline
\multicolumn{1}{r|}{${}_{\mathrm{U}}\text{\textsc{RU}}$} &
  20.58 &
  50.32 &
  1.42 &
  30.85 &
  63.30 &
  4.68 &
  24.85 &
  56.41 &
  2.92  \\ \Xhline{2\arrayrulewidth}
\multicolumn{7}{r|}{\textsc{Recurrent Graph Evolution Neural Network (ReGENN)}} &
  \textbf{18.22} &
  \textbf{47.31} &
  \textbf{1.37} \\ \hline
\multicolumn{7}{r|}{\textsc{ReGENN without Variable Decoder}} &
  18.23 &
  47.45 &
  1.37 \\ \hline
\multicolumn{7}{r|}{\textsc{ReGENN without Autoregression}} &
  18.48 &
  49.72 &
  1.37 \\ \hline
\multicolumn{7}{r|}{\textsc{Autoregression Only}} &
  24.44 &
  53.92 &
  2.08 \\ \Xhline{3\arrayrulewidth}
\multicolumn{7}{r|}{\textsc{Overall Improvement}} &
  1.05\% &
  0.98\% &
  0.0\% \\ \Xhline{3\arrayrulewidth}
\end{tabular}%
\end{table*}

\begin{figure*}[!h]
    \centering
    \includegraphics[width=\linewidth]{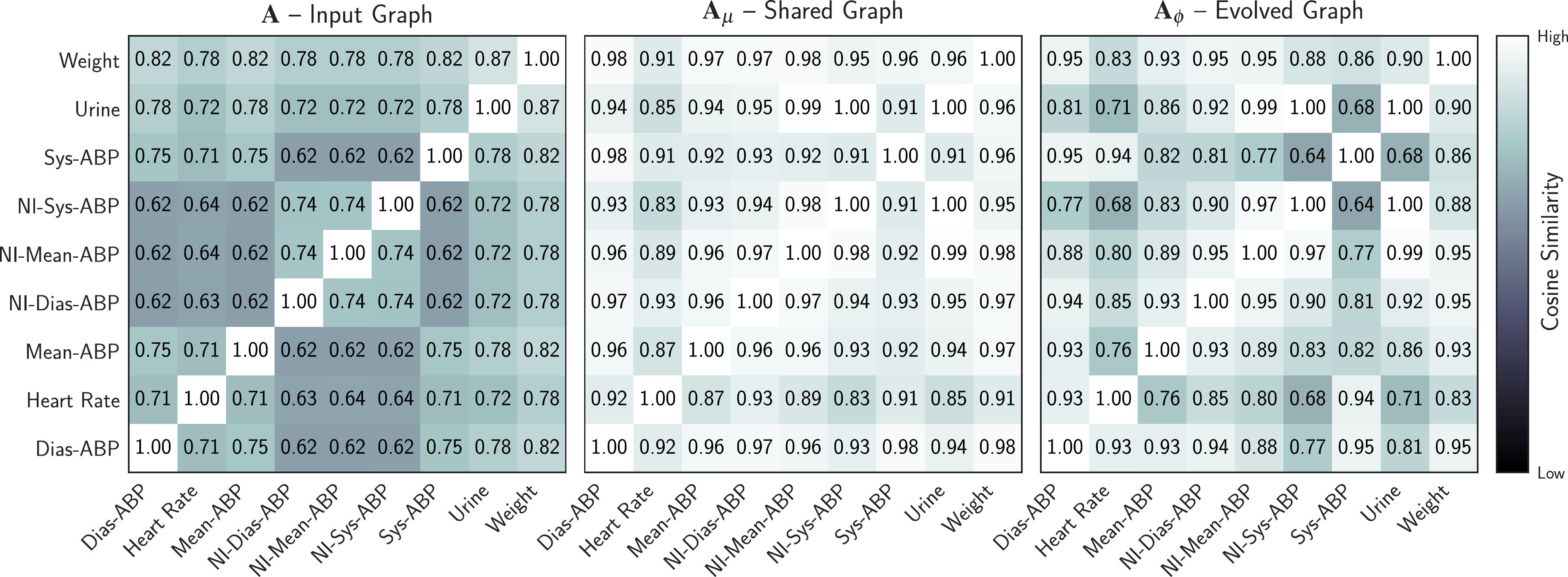}
    \caption{
    \emph{Evolution Weights} extracted from \ReGENN~after training on the PhysioNet Computing in Cardiology Challenge 2012 dataset, in which we use the cosine similarity to compare the relationship between pairs of variables.
    We use "ABP" as a shortening for \textit{Arterial Blood Pressure}, "NI" as \textit{Non-Invasive}, "Dias" as \textit{Diastolic}, and "Sys" as \textit{Systolic}.
    }
    \label{fig:physionet-weights}
\end{figure*}

Fig.~\ref{fig:weather-weights} depicts the \emph{Evolution Weights} for the current dataset, in which we can observe a consistent similarity between pairs of variables in the input graph, which does not repeat in the evolved graph, implying different relationships.
We observe that the similarity between all pairs of variables increased on the evolved graph.
The pairs \textit{Solar Radiation} and \textit{Rain}, \textit{Maximum Temperature} and \textit{Rain}, and \textit{Solar Radiation} and \textit{Minimum Temperature} stood out.
Those pairs are mutually related, which comes from solar radiation interfering in maximum and minimum temperature and in the precipitation factors, where the opposite relation holds.
What can be extracted from the \emph{Evolution Weights}, in this case, is the notion of importance between pairs of variables so that the pairs that stood out are more relevant and provide better information during the forecasting process.

\subsubsection*{Experiment on PhysioNet dataset}
The PhysioNet dataset presents a large number of samples and an increased number of variables, but little information on the time-axis, a setting in which ensembles still struggle to perform accurate predictions, as depicted in Fig.~\ref{fig:physionet-baselines}.
Once again, \ReGENN~keeps steady as the first-placed algorithm in performance, showing solid improvement over the Linear SVR, the second-placed algorithm.
The improvement was 7.33\% on the MAE and 35.13\% on the MSLE, while the RMSE achieved by \ReGENN~laid within the standard deviation of the Linear SVR, pointing out an equivalent performance between them.
The Linear SVR is an ensemble of multiple estimators, while \ReGENN~uses only one, making it better accurate for dealing with the current dataset.

As in Tab.~\ref{tab:physionet-ablation}, the ablation results reveal that a neural network architecture without the GSE layers can achieve a better performance than the baseline algorithms.
In this specific case, we see that by using a bidirectional LSTM instead of unidirectional on the decoder module of the neural network, we can achieve a performance almost as good as \ReGENN, but not enough to surpass it, as \ReGENN~still shows an improvement of 1.05\% on the MAE and 0.98\% on the RMSE over the experiment with bidirectional LSTM.

In this specific case, \ReGENN~learns by observing multiple ICU patients.
However, one cannot say that an ICU patient's state is somehow connected to another patient's state.
Contrarily, the idea holds as in the first experiment, where although the samples are different, they have the same domain, variables, and timestream, such that the information from one sample might help enhance future forecasting for another one.
That means \ReGENN~learns both from the past of the patient and from the past of other patients.
Nevertheless, we must be careful about drawing any understanding about these results, as the reason each patient is in the ICU is different, and while some explanations might be suited for a small set of patients, it tends not to generalize to a significant number of patients.
When analyzing the \emph{Evolution Weights} in Fig.~\ref{fig:physionet-weights} aided by a physician, we can say that there is a relationship between the amount of urine excreted by a patient and the arterial blood pressure, and also that there is a relation between the systolic and diastolic blood pressure.
However, even aided by the \emph{Evolution Weights}, we cannot further describe these relations once there are variables of the biological domain that are not being taken into consideration.

\section{Overall Discussions}
We refer to the \emph{Evolution Weights} as the intermediate weights of the representation-learning process optimized throughout the network's training.
Such weights are time-invariant and are a requirement for the feature-learning behind the GSE layer.
Although time does not flow through the adjacency matrix, the network is optimized as a whole, such that every operation influences the gradients of the backward propagation process.
That means the optimizer, influenced by the gradients of both time-variant and invariant data, will optimize the weights towards a better forecasting ability.
Such a process depends not only on the network architecture but also on the optimization process's reliability.

That increases uncertainty, which is the downside of \ReGENN, demanding more time to train the neural network and causing the improvement not to be strictly uprising.
Consequently, training might take long sessions, even with consistently reduced learning rates on plateaus or simulated annealing techniques; this is influenced by the fact that the second GSE layer has two dynamic inputs, which arise from the graph-evolution process.
However, we observed that throughout the epochs, the \emph{Evolution Weights} reach a stable point with no major updates.
As a result, the network demonstrates a remarkable improvement in its final iterations when the remaining weights intensely converge to a near-optimal configuration.

Even though \ReGENN~has a particular drawback, it demonstrates excellent versatility, which comes from its superior performance in the task of epidemiology modeling on the SARS-CoV-2 dataset, climate forecasting on the Brazilian Weather, and patient monitoring on intensive care units on the PhysioNet dataset.
Consequently, we see \ReGENN~as a tool to be used in data-driven decision-making tasks, helping prevent, for instance, natural disasters or preparing for an upcoming pandemic.
As a precursor in multiple multivariate time-series forecasting, there is still much to be improved.
For example, reducing the uncertainty that harms \ReGENN~without decreasing its performance should be the first step, followed by extending the proposal to handle problems in the spatiotemporal field of great interest to traffic forecasting and environmental monitoring.
Another possibility would be to remove the decoder's recurrent layers while tracking the temporal dependencies through multiple graphs, a new temporal-modeling perspective in which one could leverage from Graph Convolution Networks for extracting inter-variable relationships and using those as hidden-features during the time-series forecasting process, to which further hypothesis constraints may apply.%

Notwithstanding, in some cases, where extensive generalization is not required, the analysis of singular multivariate time-series may be preferred to multiple multivariate time-series.
That because, when focusing on a single series at a time, some but not all samples might yield a lower forecasting error, as the model will be driven to a single multivariate sample.
However, both approaches for tackling time-series forecasting can coexist in the state-of-the-art, and, as a consequence, the decision to work on a higher or lower dimensionality must relate to which problem is being solved and how much data is available to solve it.

\section{Conclusion}
This paper tackles multiple multivariate time-series forecasting tasks by proposing the Recurrent Graph Evolution Neural Network (\ReGENN), a graph-based time-aware auto-encoder powered by a pair of Graph Soft Evolution (GSE) layers, a further contribution of this study that stands for a graph-based learning-representation layer.

The literature handles multivariate time-series forecasting with outstanding performance, but up to this point, we lacked a technique with increased generalization over multiple multivariate time-series with sound performance.
Previous research might have avoided tackling such a problem as a neural network, to that matter, is challenging to train and usually yields poor results.
That because one aims to achieve good generalization on future observations for multivariate time-series that do not necessarily hold the same data distribution.

Because of that, \ReGENN~is a precursor in multiple multivariate time-series forecasting and, even though this is a challenging problem, \ReGENN~surpassed all the baselines and remained effective after three rounds of 30 ablation tests through distinct hyperparameters.
The experiments were carried out over the SARS-CoV-2, Brazilian Weather, and PhysioNet datasets with improvements, respectively, of at least 64.87\%, 11.96\%, and 7.33\%.
As a consequence of the results, \ReGENN~shows a new range of possibilities in time-series forecasting, starting by demonstrating that ensembles poorly perform if compared to a single model able to learn the entanglement between different variables by looking at how they interact as time goes by and how multiple multivariate time-series synchronously evolve.

\section*{Authors contributions statement}

GS conceived the experiments, GS prepared the data for experimentation, and GS conducted them.
SH and BB validated GS's source-code implementation.
SH and BB analyzed the preliminary results.
SM, JR, and JS validated the final results.
GS wrote the original draft.
SH, BB, SM, JR, and JS iteratively polished the main ideas and the paper.
All authors reviewed and approved the results and manuscript.

\section*{Acknowledgments}
This work was partially supported by the Coordena\c{c}\~ao de Aperfei\c{c}oamento de Pessoal de N\'ivel Superior -- Brazil (CAPES) -- Finance Code 001; Funda\c{c}\~ao de Amparo \`a Pesquisa do Estado de S\~ao Paulo (FAPESP), through grants 2014/25337-0, 2016/17078-0, 2017/08376-0, 2018/17620-5, 2019/04461-9, and 2020/07200-9; Conselho Nacional de Desenvolvimento Cient\'ifico e Tecnol\'ogico (CNPq) through grants 167967/2017-7, 305580/2017-5, and 406550/2018-2; National Science Foundation awards IIS-2014438, PPoSS-2028839, and IIS-1838042; and, the National Institute of Health awards NIH R01 1R01NS107291-01, and R56HL138415.
We thank Jeffrey Valdez for his aid with Sunlab's computer infrastructure, Lucas Scabora, for his careful review of the paper, and Gustavo Merchan, M.D., for his input on the \emph{Evolution Weights} of the PhysioNet dataset.

\bibliographystyle{IEEEtran}
\bibliography{3-references.bib}

\begin{thebibliography}{10}
\providecommand{\url}[1]{#1}
\csname url@samestyle\endcsname
\providecommand{\newblock}{\relax}
\providecommand{\bibinfo}[2]{#2}
\providecommand{\BIBentrySTDinterwordspacing}{\spaceskip=0pt\relax}
\providecommand{\BIBentryALTinterwordstretchfactor}{4}
\providecommand{\BIBentryALTinterwordspacing}{\spaceskip=\fontdimen2\font plus
\BIBentryALTinterwordstretchfactor\fontdimen3\font minus
  \fontdimen4\font\relax}
\providecommand{\BIBforeignlanguage}[2]{{%
\expandafter\ifx\csname l@#1\endcsname\relax
\typeout{** WARNING: IEEEtran.bst: No hyphenation pattern has been}%
\typeout{** loaded for the language `#1'. Using the pattern for}%
\typeout{** the default language instead.}%
\else
\language=\csname l@#1\endcsname
\fi
#2}}
\providecommand{\BIBdecl}{\relax}
\BIBdecl

\bibitem{Lavender2018:Environment}
S.~L. Lavender, K.~J.~E. Walsh, L.-P. Caron, M.~King, S.~Monkiewicz,
  M.~Guishard, Q.~Zhang, and B.~Hunt, ``{Estimation of the maximum annual
  number of North Atlantic tropical cyclones using climate models},''
  \emph{Science Advances}, vol.~4, no.~8, aug 2018.

\bibitem{Kao2020:Environment}
Y.-C. Kao, M.~W. Rogers, D.~B. Bunnell, I.~G. Cowx, S.~S. Qian, O.~Anneville,
  T.~D. Beard, A.~Brinker, J.~R. Britton, R.~Chura-Cruz, N.~J. Gownaris, J.~R.
  Jackson, K.~Kangur, J.~Kolding, A.~A. Lukin, A.~J. Lynch, N.~Mercado-Silva,
  R.~Moncayo-Estrada, F.~J. Njaya, I.~Ostrovsky, L.~G. Rudstam, A.~L.~E.
  Sandstr{\"{o}}m, Y.~Sato, H.~Siguayro-Mamani, A.~Thorpe, P.~A.~M. van
  Zwieten, P.~Volta, Y.~Wang, A.~Weiperth, O.~L.~F. Weyl, and J.~D. Young,
  ``{Effects of climate and land-use changes on fish catches across lakes at a
  global scale},'' \emph{Nature Communications}, vol.~11, no.~1, pp. 25--26,
  dec 2020.

\bibitem{Pryor2020:projectionsWindResources}
S.~C. Pryor, R.~J. Barthelmie, M.~S. Bukovsky, L.~R. Leung, and K.~Sakaguchi,
  ``{Climate change impacts on wind power generation},'' \emph{Nature Reviews
  Earth \& Environment}, vol.~1, no.~12, pp. 627--643, dec 2020.

\bibitem{Laurent2020:Environment}
L.~Laurent, J.-F. Buoncristiani, B.~Pohl, H.~Zekollari, D.~Farinotti, M.~Huss,
  J.-L. Mugnier, and J.~Pergaud, ``{The impact of climate change and glacier
  mass loss on the hydrology in the Mont-Blanc massif},'' \emph{Scientific
  Reports}, vol.~10, no.~1, dec 2020.

\bibitem{Kelotra2020:Deep-ConvLSTM}
A.~Kelotra and P.~Pandey, ``{Stock Market Prediction Using Optimized
  Deep-ConvLSTM Model},'' \emph{Big Data}, vol.~8, no.~1, pp. 5--24, feb 2020.

\bibitem{Ananthi2020:CandlestickEegression}
M.~Ananthi and K.~Vijayakumar, ``{Stock market analysis using candlestick
  regression and market trend prediction (CKRM)},'' \emph{Journal of Ambient
  Intelligence and Humanized Computing}, apr 2020.

\bibitem{Jiao2021:ChangePoint}
S.~Jiao, T.~Shen, Z.~Yu, and H.~Ombao, ``{Change-point detection using spectral
  PCA for multivariate time series},'' jan 2021.

\bibitem{Roy2020:Landsat}
D.~Roy and L.~Yan, ``{Robust Landsat-based crop time series modelling},''
  \emph{Remote Sensing of Environment}, vol. 238, mar 2020.

\bibitem{Zhu2020:Landsat}
Z.~Zhu, J.~Zhang, Z.~Yang, A.~H. Aljaddani, W.~B. Cohen, S.~Qiu, and C.~Zhou,
  ``{Continuous monitoring of land disturbance based on Landsat time series},''
  \emph{Remote Sensing of Environment}, vol. 238, pp. 111--116, mar 2020.

\bibitem{Yan2020:Landsat}
L.~Yan and D.~P. Roy, ``{Spatially and temporally complete Landsat reflectance
  time series modelling: The fill-and-fit approach},'' \emph{Remote Sensing of
  Environment}, vol. 241, may 2020.

\bibitem{Nikolay2015:Anomaly}
N.~Laptev, S.~Amizadeh, and I.~Flint, ``{Generic and Scalable Framework for
  Automated Time-series Anomaly Detection},'' in \emph{Proceedings of the 21th
  ACM SIGKDD International Conference on Knowledge Discovery and Data Mining},
  ser. KDD '15.\hskip 1em plus 0.5em minus 0.4em\relax New York, NY, USA: ACM,
  aug 2015, pp. 1939--1947.

\bibitem{Li2017:Markov}
J.~Li, W.~Pedrycz, and I.~Jamal, ``{Multivariate time series anomaly detection:
  A framework of Hidden Markov Models},'' \emph{Applied Soft Computing},
  vol.~60, pp. 229--240, nov 2017.

\bibitem{Hancock2020:FraudDetection}
J.~Hancock and T.~M. Khoshgoftaar, ``{Performance of CatBoost and XGBoost in
  Medicare Fraud Detection},'' in \emph{2020 19th IEEE International Conference
  on Machine Learning and Applications (ICMLA)}.\hskip 1em plus 0.5em minus
  0.4em\relax IEEE, dec 2020, pp. 572--579.

\bibitem{Deb2017:Environment}
C.~Deb, F.~Zhang, J.~Yang, S.~E. Lee, and K.~W. Shah, ``{A review on time
  series forecasting techniques for building energy consumption},''
  \emph{Renewable and Sustainable Energy Reviews}, vol.~74, pp. 902--924, jul
  2017.

\bibitem{Orlov2020:renewableEnergyPrediction}
A.~Orlov, J.~Sillmann, and I.~Vigo, ``{Better seasonal forecasts for the
  renewable energy industry},'' \emph{Nature Energy}, vol.~5, no.~2, pp.
  108--110, feb 2020.

\bibitem{Baratsas2021:energyPrediction}
S.~G. Baratsas, A.~M. Niziolek, O.~Onel, L.~R. Matthews, C.~A. Floudas, D.~R.
  Hallermann, S.~M. Sorescu, and E.~N. Pistikopoulos, ``{A framework to predict
  the price of energy for the end-users with applications to monetary and
  energy policies},'' \emph{Nature Communications}, vol.~12, no.~1, dec 2021.

\bibitem{Perros2019:TemporalPhenotyping}
I.~Perros, E.~E. Papalexakis, R.~Vuduc, E.~Searles, and J.~Sun, ``{Temporal
  phenotyping of medically complex children via PARAFAC2 tensor
  factorization},'' \emph{Journal of Biomedical Informatics}, vol.~93, no.
  September 2018, pp. 103--125, may 2019.

\bibitem{Rodrigues2019:PatientTrajectory}
J.~F. Rodrigues-Jr, G.~Spadon, B.~Brandoli, and S.~Amer-Yahia, ``{Patient
  trajectory prediction in the Mimic-III dataset, challenges and pitfalls},''
  \emph{arXiv}, sep 2019.

\bibitem{Rodrigues2020:LigDoctor}
J.~F. {Rodrigues}, G.~{Spadon}, B.~{Brandoli}, and S.~{Amer-Yahia},
  ``Lig-doctor: Real-world clinical prognosis using a bi-directional neural
  network,'' in \emph{2020 IEEE 33rd International Symposium on Computer-Based
  Medical Systems (CBMS)}, 2020, pp. 569--572.

\bibitem{Afshar2020:TASTE}
A.~Afshar, I.~Perros, H.~Park, C.~DeFilippi, X.~Yan, W.~Stewart, J.~Ho, and
  J.~Sun, ``{TASTE},'' in \emph{Proceedings of the ACM Conference on Health,
  Inference, and Learning}, ser. CHIL '20.\hskip 1em plus 0.5em minus
  0.4em\relax New York, NY, USA: ACM, apr 2020, pp. 193--203.

\bibitem{Yan2020:COVID19}
L.~Yan, H.-T. Zhang, J.~Goncalves, Y.~Xiao, M.~Wang, Y.~Guo, C.~Sun, X.~Tang,
  L.~Jing, M.~Zhang, X.~Huang, Y.~Xiao, H.~Cao, Y.~Chen, T.~Ren, F.~Wang,
  Y.~Xiao, S.~Huang, X.~Tan, N.~Huang, B.~Jiao, C.~Cheng, Y.~Zhang, A.~Luo,
  L.~Mombaerts, J.~Jin, Z.~Cao, S.~Li, H.~Xu, and Y.~Yuan, ``{An interpretable
  mortality prediction model for COVID-19 patients},'' \emph{Nature Machine
  Intelligence}, vol.~2, no.~5, pp. 283--288, may 2020.

\bibitem{Rodrigues2021:MinimalGRU}
J.~F. Rodrigues-Jr, M.~A. Gutierrez, G.~Spadon, B.~Brandoli, and S.~Amer-Yahia,
  ``{LIG-Doctor: Efficient patient trajectory prediction using bidirectional
  minimal gated-recurrent networks},'' \emph{Information Sciences}, vol. 545,
  pp. 813--827, feb 2021.

\bibitem{Ienca2020:COVID19}
M.~Ienca and E.~Vayena, ``{On the responsible use of digital data to tackle the
  COVID-19 pandemic},'' \emph{Nature Medicine}, vol.~26, no.~4, pp. 463--464,
  apr 2020.

\bibitem{Velavan2020:Epidemic}
T.~P. Velavan and C.~G. Meyer, ``{The COVID‐19 epidemic},'' \emph{Tropical
  Medicine \& International Health}, vol.~25, no.~3, pp. 278--280, mar 2020.

\bibitem{Omer2020:USPandemic}
S.~B. Omer, P.~Malani, and C.~del Rio, ``{The COVID-19 Pandemic in the US},''
  \emph{JAMA}, vol. 323, no.~18, pp. 1767--1768, apr 2020.

\bibitem{Silvestrini2008:Aggregation}
A.~Silvestrini and D.~Veredas, ``{Temporal aggregation of univariate and
  multivariate time series models: A survey},'' \emph{Journal of Economic
  Surveys}, vol.~22, no.~3, pp. 458--497, jul 2008.

\bibitem{Jain1996:ANN}
A.~Jain, {Jianchang Mao}, and K.~Mohiuddin, ``{Artificial neural networks: a
  tutorial},'' \emph{Computer}, vol.~29, no.~3, pp. 31--44, mar 1996.

\bibitem{Zhang2001:ANN}
G.~Zhang, B.~Patuwo, and M.~Y. Hu, ``{A simulation study of artificial neural
  networks for nonlinear time-series forecasting},'' \emph{Computers \&
  Operations Research}, vol.~28, no.~4, pp. 381--396, apr 2001.

\bibitem{informer2020}
H.~Zhou, S.~Zhang, J.~Peng, S.~Zhang, J.~Li, H.~Xiong, and W.~Zhang,
  ``{Informer: Beyond Efficient Transformer for Long Sequence Time-Series
  Forecasting},'' \emph{CoRR}, vol. abs/2012.0, dec 2020.

\bibitem{seo2016}
Y.~Seo, M.~Defferrard, P.~Vandergheynst, and X.~Bresson, ``{Structured sequence
  modeling with graph convolutional recurrent networks},'' in \emph{Lecture
  Notes in Computer Science (including subseries Lecture Notes in Artificial
  Intelligence and Lecture Notes in Bioinformatics)}, L.~Cheng, A.~C.~S. Leung,
  and S.~Ozawa, Eds., vol. 11301 LNCS.\hskip 1em plus 0.5em minus 0.4em\relax
  Springer International Publishing, 2018, pp. 362--373.

\bibitem{Lecun1998:CNN}
Y.~Lecun, L.~Bottou, Y.~Bengio, and P.~Haffner, ``{Gradient-based learning
  applied to document recognition},'' \emph{Proceedings of the IEEE}, vol.~86,
  no.~11, pp. 2278--2324, 1998.

\bibitem{Elman1990:RNN}
J.~Elman, ``{Finding structure in time},'' \emph{Cognitive Science}, vol.~14,
  no.~2, pp. 179--211, jun 1990.

\bibitem{liYS018}
Y.~Li, R.~Yu, C.~Shahabi, and Y.~Liu, ``{Diffusion Convolutional Recurrent
  Neural Network: Data-Driven Traffic Forecasting},'' in \emph{6th
  International Conference on Learning Representations, {ICLR} 2018, Vancouver,
  BC, Canada, April 30 - May 3, 2018, Conference Track Proceedings}, jul 2017.

\bibitem{zhang18}
J.~Zhang, X.~Shi, J.~Xie, H.~Ma, I.~King, and D.-Y. Yeung, ``{GaAN: Gated
  Attention Networks for Learning on Large and Spatiotemporal Graphs},'' in
  \emph{Proceedings of the Thirty-Fourth Conference on Uncertainty in
  Artificial Intelligence}, mar 2018, pp. 339--349.

\bibitem{Liang2018:GeoMAN}
Y.~Liang, S.~Ke, J.~Zhang, X.~Yi, and Y.~Zheng, ``{GeoMAN: Multi-level
  Attention Networks for Geo-sensory Time Series Prediction},'' in
  \emph{Proceedings of the Twenty-Seventh International Joint Conference on
  Artificial Intelligence}.\hskip 1em plus 0.5em minus 0.4em\relax California:
  International Joint Conferences on Artificial Intelligence Organization, jul
  2018, pp. 3428--3434.

\bibitem{Bing2018:Traffic}
B.~Yu, H.~Yin, and Z.~Zhu, ``{Spatio-Temporal Graph Convolutional Networks: A
  Deep Learning Framework for Traffic Forecasting},'' in \emph{Proceedings of
  the 27th International Joint Conference on Artificial Intelligence}, ser.
  IJCAI'18.\hskip 1em plus 0.5em minus 0.4em\relax AAAI Press, sep 2017, pp.
  3634--3640.

\bibitem{Zhao2019:TGCN}
L.~Zhao, Y.~Song, C.~Zhang, Y.~Liu, P.~Wang, T.~Lin, M.~Deng, and H.~Li,
  ``{T-GCN: A Temporal Graph Convolutional Network for Traffic Prediction},''
  \emph{IEEE Transactions on Intelligent Transportation Systems}, vol.~21,
  no.~9, pp. 3848--3858, sep 2020.

\bibitem{Kipf2017:GCN}
T.~N. Kipf and M.~Welling, ``{Semi-supervised classification with graph
  convolutional networks},'' in \emph{5th International Conference on Learning
  Representations, ICLR 2017 - Conference Track Proceedings}, 2017.

\bibitem{Lai2018:LSTNet}
G.~Lai, W.-C. Chang, Y.~Yang, and H.~Liu, ``{Modeling Long- and Short-Term
  Temporal Patterns with Deep Neural Networks},'' in \emph{The 41st
  International ACM SIGIR Conference on Research \& Development in Information
  Retrieval}.\hskip 1em plus 0.5em minus 0.4em\relax New York, NY, USA: ACM,
  jun 2018, pp. 95--104.

\bibitem{Huang2019:DSANet}
S.~Huang, X.~Wu, D.~Wang, and A.~Tang, ``{DSANet: Dual self-attention network
  for multivariate time series forecasting},'' in \emph{International
  Conference on Information and Knowledge Management, Proceedings}, ser.
  CIKM'19.\hskip 1em plus 0.5em minus 0.4em\relax New York, NY, USA: ACM, nov
  2019, pp. 2129--2132.

\bibitem{Vaswani2017:Attention}
A.~Vaswani, N.~Shazeer, N.~Parmar, J.~Uszkoreit, L.~Jones, A.~N. Gomez,
  {\L}.~Kaiser, and I.~Polosukhin, ``{Attention is all you need},''
  \emph{Advances in Neural Information Processing Systems}, vol. 2017-December,
  pp. 5999--6009, jun 2017.

\bibitem{Cheng2019:MLCNN}
J.~Cheng, K.~Huang, and Z.~Zheng, ``{Towards better forecasting by fusing near
  and distant future visions},'' \emph{arXiv}, 2019.

\bibitem{Hochreiter1997:LSTM}
S.~Hochreiter and J.~Schmidhuber, ``{Long Short-Term Memory},'' \emph{Neural
  Computation}, vol.~9, no.~8, pp. 1735--1780, nov 1997.

\bibitem{Srivastava2014:Dropout}
N.~Srivastava, G.~Hinton, A.~Krizhevsky, I.~Sutskever, and R.~Salakhutdinov,
  ``{Dropout: A Simple Way to Prevent Neural Networks from Overfitting},''
  \emph{Journal of Machine Learning Research}, vol.~15, no.~56, pp. 1929--1958,
  2014.

\bibitem{Zilly2017:RecurrentHighway}
J.~G. Zilly, R.~K. Srivastava, J.~Koutnik, and J.~Schmidhuber, ``{Recurrent
  highway networks},'' \emph{34th International Conference on Machine Learning,
  ICML 2017}, vol.~8, pp. 6346--6357, jul 2017.

\bibitem{Srivastava2015:Highway}
R.~K. Srivastava, K.~Greff, and J.~Schmidhuber, ``{Highway Networks},''
  \emph{arXiv}, 2015.

\bibitem{He2016:Residual}
K.~He, X.~Zhang, S.~Ren, and J.~Sun, ``{Deep Residual Learning for Image
  Recognition},'' in \emph{2016 IEEE Conference on Computer Vision and Pattern
  Recognition (CVPR)}.\hskip 1em plus 0.5em minus 0.4em\relax IEEE, jun 2016,
  pp. 770--778.

\bibitem{Ba2016:LayerNormalization}
J.~L. Ba, J.~R. Kiros, and G.~E. Hinton, ``{Layer Normalization},''
  \emph{arXiv}, vol. abs/1607.0, jul 2016.

\bibitem{Chen2016:XGBoost}
T.~Chen and C.~Guestrin, ``{XGBoost: A scalable tree boosting system},'' in
  \emph{Proceedings of the ACM SIGKDD International Conference on Knowledge
  Discovery and Data Mining}, vol. 13-17-August-2016, ACM.\hskip 1em plus 0.5em
  minus 0.4em\relax New York, NY, USA: ACM, aug 2016, pp. 785--794.

\bibitem{Kingma2015:Adam}
D.~P. Kingma and J.~L. Ba, ``{Adam: A method for stochastic optimization},'' in
  \emph{3rd International Conference on Learning Representations, ICLR 2015 -
  Conference Track Proceedings}, Y.~Bengio and Y.~LeCun, Eds., 2015.

\bibitem{Vapnik1997:SVR}
V.~Vapnik, S.~E. Golowich, and A.~Smola, ``{Support vector method for function
  approximation, regression estimation, and signal processing},'' in
  \emph{Advances in Neural Information Processing Systems}, M.~C. Mozer, M.~I.
  Jordan, and T.~Petsche, Eds.\hskip 1em plus 0.5em minus 0.4em\relax MIT
  Press, 1997, pp. 281--287.

\bibitem{Sahoo2018:OnlineLearning}
D.~Sahoo, Q.~Pham, J.~Lu, and S.~C.~H. Hoi, ``{Online Deep Learning: Learning
  Deep Neural Networks on the Fly},'' in \emph{Proceedings of the
  Twenty-Seventh International Joint Conference on Artificial
  Intelligence}.\hskip 1em plus 0.5em minus 0.4em\relax California:
  International Joint Conferences on Artificial Intelligence Organization, jul
  2018, pp. 2660--2666.

\bibitem{Dong2020:COVID19}
E.~Dong, H.~Du, and L.~Gardner, ``{An interactive web-based dashboard to track
  COVID-19 in real time},'' \emph{The Lancet Infectious Diseases}, vol.~20,
  no.~5, pp. 533--534, may 2020.

\bibitem{Silva2012:Physionet}
I.~Silva, G.~Moody, D.~J. Scott, L.~A. Celi, and R.~G. Mark, ``{Predicting
  in-hospital mortality of ICU patients: The PhysioNet/Computing in cardiology
  challenge 2012},'' in \emph{Computing in Cardiology}, vol.~39, 2012, pp.
  245--248.

\bibitem{Keogh2004:SegmentingServey}
E.~Keogh, S.~Chu, D.~Hart, and M.~Pazzani, ``{Segmenting Time Series: a Survey
  and Novel Approach},'' in \emph{Data Mining in Time Series Databases}, ser.
  Series in Machine Perception and Artificial Intelligence.\hskip 1em plus
  0.5em minus 0.4em\relax World Scientific, jun 2004, vol. Volume 57, pp.
  1--21.

\bibitem{Frank2000:WindowSize}
R.~Frank, N.~Davey, and S.~Hunt, ``{Input window size and neural network
  predictors},'' in \emph{Proceedings of the IEEE-INNS-ENNS International Joint
  Conference on Neural Networks. IJCNN 2000. Neural Computing: New Challenges
  and Perspectives for the New Millennium}, vol.~2.\hskip 1em plus 0.5em minus
  0.4em\relax IEEE, 2000, pp. 237--242.

\bibitem{Frank2001:SlidingWindow}
R.~J. Frank, N.~Davey, and S.~P. Hunt, ``{Time series prediction and neural
  networks},'' \emph{Journal of Intelligent and Robotic Systems: Theory and
  Applications}, vol.~31, no. 1-3, pp. 91--103, 2001.

\bibitem{Pedregosa2011:Sklearn}
P.~Fabian, {Michel Vincent}, G.~Olivier, B.~Mathieu, P.~Peter, W.~Ron,
  {Vanderplas Jake}, and D.~Cournapeau, ``{Scikit-learn: Machine Learning in
  Python},'' \emph{Journal of Machine Learning Research}, vol.~12, no.~85, pp.
  2825--2830, 2011.

\bibitem{Prokhorenkova2018:CatBoost}
A.~V. Dorogush, V.~Ershov, and A.~Gulin, ``{CatBoost: gradient boosting with
  categorical features support},'' in \emph{Advances in Neural Information
  Processing Systems 31}, S.~Bengio, H.~Wallach, H.~Larochelle, K.~Grauman,
  N.~Cesa-Bianchi, and R.~Garnett, Eds.\hskip 1em plus 0.5em minus 0.4em\relax
  Curran Associates, Inc., oct 2018, pp. 6638--6648.

\bibitem{Chung2014:GRU}
J.~Chung, C.~Gulcehre, K.~Cho, and Y.~Bengio, ``{Empirical Evaluation of Gated
  Recurrent Neural Networks on Sequence Modeling},'' \emph{arXiv}, dec 2014.

\end{thebibliography}


\begin{thebibliography}{10}
\providecommand{\url}[1]{#1}
\csname url@samestyle\endcsname
\providecommand{\newblock}{\relax}
\providecommand{\bibinfo}[2]{#2}
\providecommand{\BIBentrySTDinterwordspacing}{\spaceskip=0pt\relax}
\providecommand{\BIBentryALTinterwordstretchfactor}{4}
\providecommand{\BIBentryALTinterwordspacing}{\spaceskip=\fontdimen2\font plus
\BIBentryALTinterwordstretchfactor\fontdimen3\font minus
  \fontdimen4\font\relax}
\providecommand{\BIBforeignlanguage}[2]{{%
\expandafter\ifx\csname l@#1\endcsname\relax
\typeout{** WARNING: IEEEtran.bst: No hyphenation pattern has been}%
\typeout{** loaded for the language `#1'. Using the pattern for}%
\typeout{** the default language instead.}%
\else
\language=\csname l@#1\endcsname
\fi
#2}}
\providecommand{\BIBdecl}{\relax}
\BIBdecl

\bibitem{Tan2006:DataMining}
P.-N. Tan, M.~Steinbach, and V.~Kumar, ``{Data Mining Introduction},'' 2006.

\bibitem{Cheng2019:MLCNN}
J.~Cheng, K.~Huang, and Z.~Zheng, ``{Towards better forecasting by fusing near
  and distant future visions},'' \emph{arXiv}, 2019.

\bibitem{Lai2018:LSTNet}
G.~Lai, W.-C. Chang, Y.~Yang, and H.~Liu, ``{Modeling Long- and Short-Term
  Temporal Patterns with Deep Neural Networks},'' in \emph{The 41st
  International ACM SIGIR Conference on Research \& Development in Information
  Retrieval}.\hskip 1em plus 0.5em minus 0.4em\relax New York, NY, USA: ACM,
  jun 2018, pp. 95--104.

\bibitem{Huang2019:DSANet}
S.~Huang, X.~Wu, D.~Wang, and A.~Tang, ``{DSANet: Dual self-attention network
  for multivariate time series forecasting},'' in \emph{International
  Conference on Information and Knowledge Management, Proceedings}, ser.
  CIKM'19.\hskip 1em plus 0.5em minus 0.4em\relax New York, NY, USA: ACM, nov
  2019, pp. 2129--2132.

\bibitem{Keogh2004:SegmentingServey}
E.~Keogh, S.~Chu, D.~Hart, and M.~Pazzani, ``{Segmenting Time Series: a Survey
  and Novel Approach},'' in \emph{Data Mining in Time Series Databases}, ser.
  Series in Machine Perception and Artificial Intelligence.\hskip 1em plus
  0.5em minus 0.4em\relax World Scientific, jun 2004, vol. Volume 57, pp.
  1--21.

\bibitem{Egrioglu2020:FuzzyTimeSeries}
E.~Egrioglu, E.~Bas, U.~Yolcu, and M.~Y. Chen, ``{Picture fuzzy time series:
  Defining, modeling and creating a new forecasting method},''
  \emph{Engineering Applications of Artificial Intelligence}, vol.~88, feb
  2020.

\bibitem{Babii2020:NowCasting}
A.~Babii, R.~T. Ball, E.~Ghysels, and J.~Striaukas, ``{Machine learning panel
  data regressions with an application to nowcasting price earnings ratios},''
  \emph{arXiv}, 2020.

\bibitem{Frank2000:WindowSize}
R.~Frank, N.~Davey, and S.~Hunt, ``{Input window size and neural network
  predictors},'' in \emph{Proceedings of the IEEE-INNS-ENNS International Joint
  Conference on Neural Networks. IJCNN 2000. Neural Computing: New Challenges
  and Perspectives for the New Millennium}, vol.~2.\hskip 1em plus 0.5em minus
  0.4em\relax IEEE, 2000, pp. 237--242.

\bibitem{Frank2001:SlidingWindow}
R.~J. Frank, N.~Davey, and S.~P. Hunt, ``{Time series prediction and neural
  networks},'' \emph{Journal of Intelligent and Robotic Systems: Theory and
  Applications}, vol.~31, no. 1-3, pp. 91--103, 2001.

\bibitem{Kingma2015:Adam}
D.~P. Kingma and J.~L. Ba, ``{Adam: A method for stochastic optimization},'' in
  \emph{3rd International Conference on Learning Representations, ICLR 2015 -
  Conference Track Proceedings}, Y.~Bengio and Y.~LeCun, Eds., 2015.

\bibitem{Vapnik1997:SVR}
V.~Vapnik, S.~E. Golowich, and A.~Smola, ``{Support vector method for function
  approximation, regression estimation, and signal processing},'' in
  \emph{Advances in Neural Information Processing Systems}, M.~C. Mozer, M.~I.
  Jordan, and T.~Petsche, Eds.\hskip 1em plus 0.5em minus 0.4em\relax MIT
  Press, 1997, pp. 281--287.

\bibitem{Sahoo2018:OnlineLearning}
D.~Sahoo, Q.~Pham, J.~Lu, and S.~C.~H. Hoi, ``{Online Deep Learning: Learning
  Deep Neural Networks on the Fly},'' in \emph{Proceedings of the
  Twenty-Seventh International Joint Conference on Artificial
  Intelligence}.\hskip 1em plus 0.5em minus 0.4em\relax California:
  International Joint Conferences on Artificial Intelligence Organization, jul
  2018, pp. 2660--2666.

\end{thebibliography}

\begin{IEEEbiography}[{\includegraphics[width=1in, height=1.25in, clip, keepaspectratio]{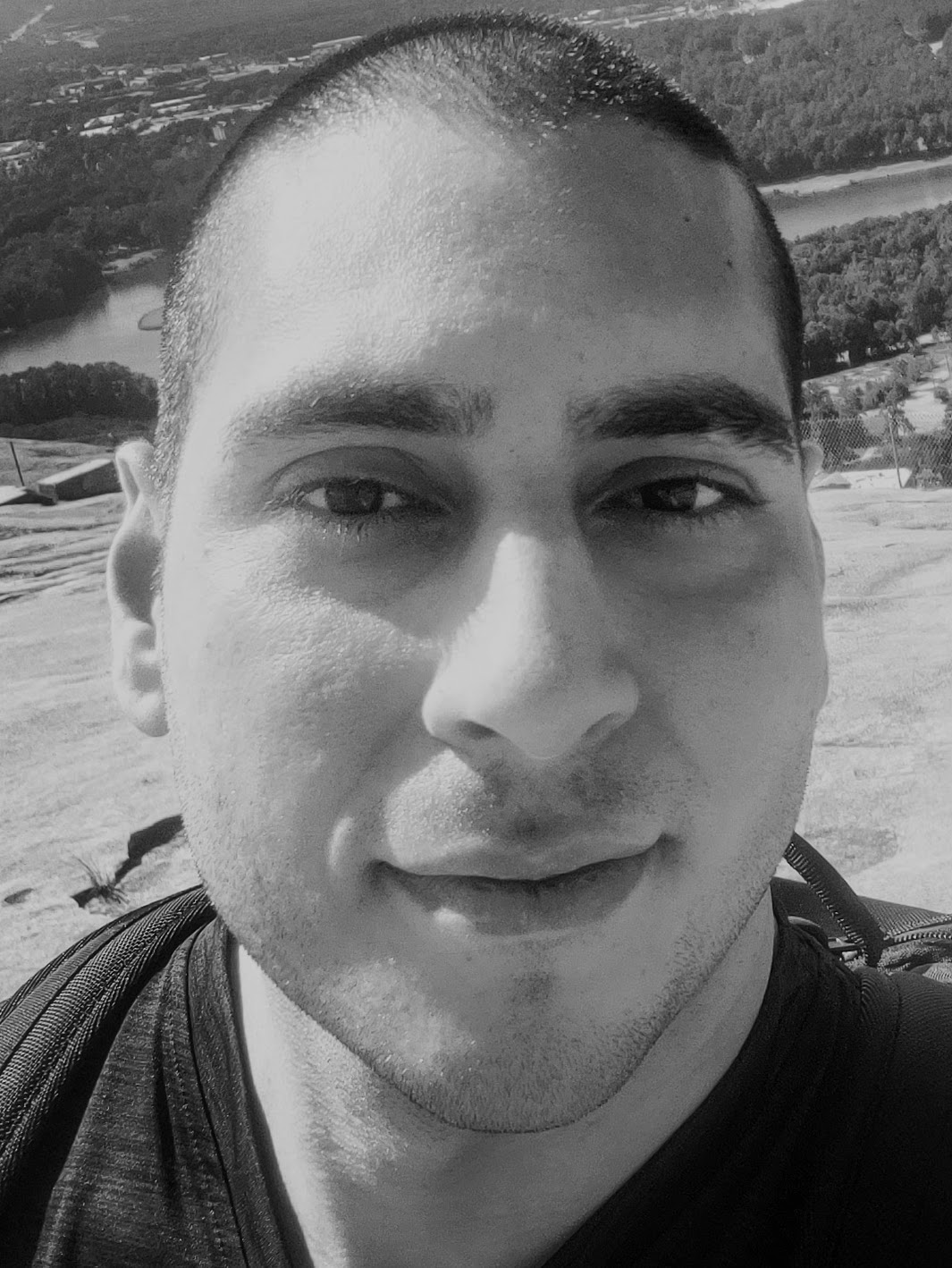}}]{Gabriel Spadon} is currently working toward the PhD degree in Computer Science at the University of Sao Paulo, Brazil, part of which was carried out with the Georgia Institute of Technology, USA. He has worked intensively on network science and artificial intelligence during the last years. He has authored or co-authored several research articles on knowledge discovery through complex networks and data mining. His current research interests include neural-inspired models, graph-based learning, and complex networks.
\end{IEEEbiography}
\vspace{-.3cm} 
\begin{IEEEbiography}[{\includegraphics[width=1in, height=1.25in, clip, keepaspectratio]{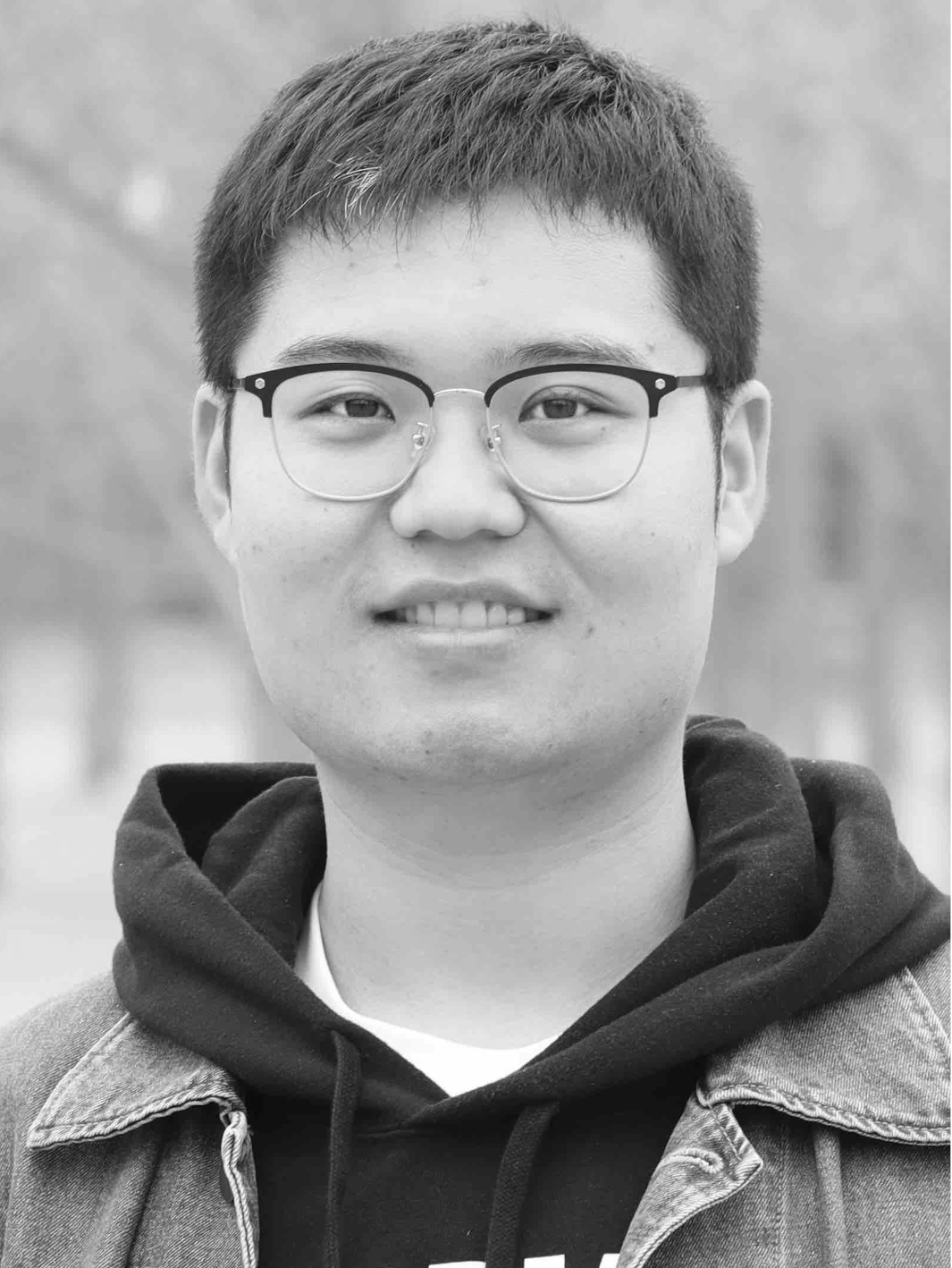}}]{Shenda Hong} received the PhD degree from the School of Electronics Engineering and Computer Sciences, Peking University, in 2019. He is currently a (Boya) postdoctoral fellow at the National Institute of Health Data Science, Peking University. From 2019 to 2020, he was a postdoctoral fellow with the College of Computing, Georgia Institute of Technology, and a visiting researcher with Harvard Medical School. His research interests include machine learning methods for temporal sequences and time series, especially deep-learning methods for electronic health records and biomedical signals, including ECG and EEG.
\end{IEEEbiography}
\vspace{-.5cm} 
\begin{IEEEbiography}[{\includegraphics[width=1in, height=1.25in, clip, keepaspectratio]{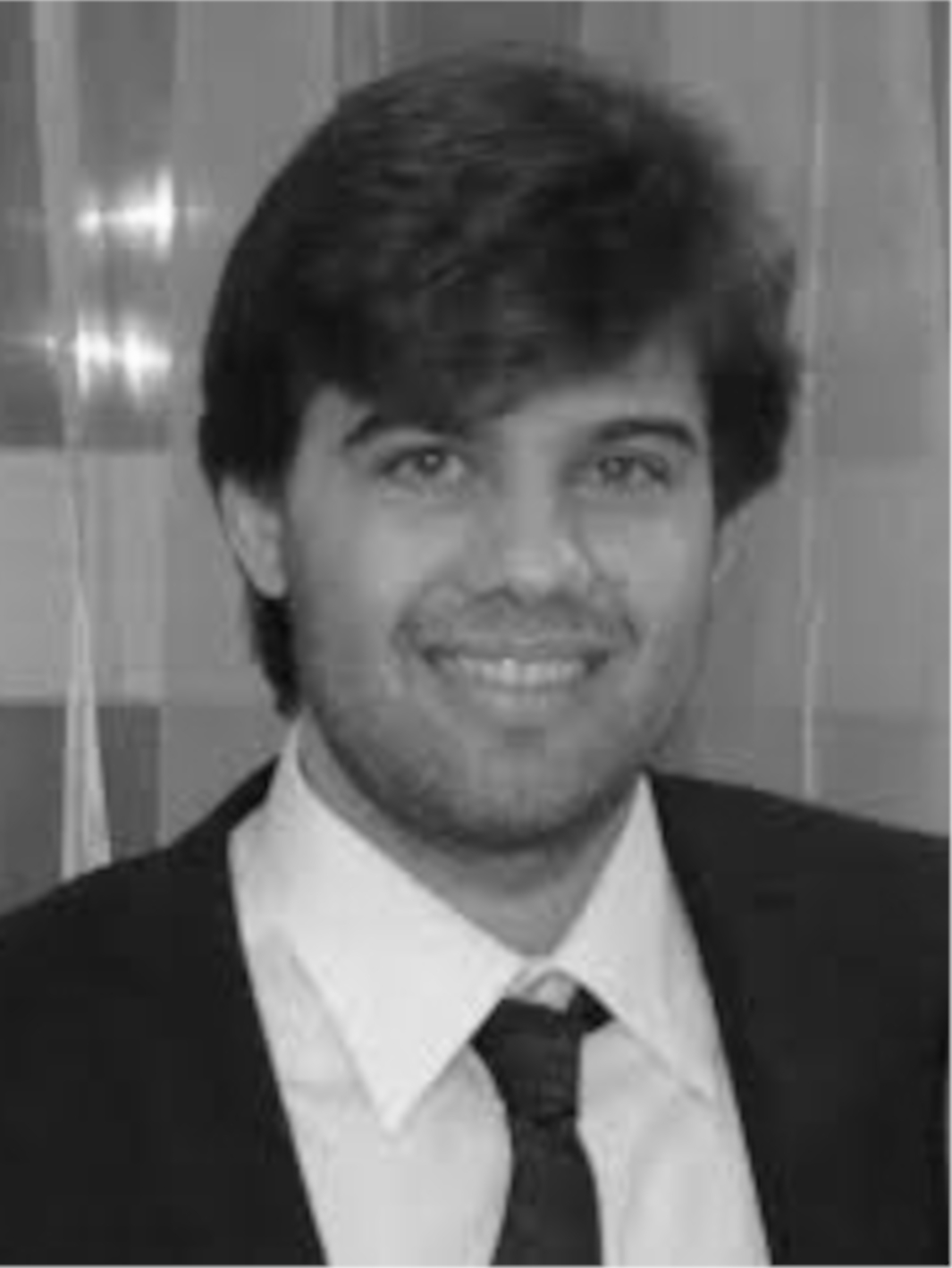}}]{Bruno Brandoli} received the PhD degree (with honors) in computer science from the University of Sao Paulo, Brazil, in 2016. He is currently a faculty at the Computer Science Department and a researcher with the Institute for Big Data Analytics, Dalhousie University, Canada. Over the past years, he has authored or coauthored AI-related articles in relevant journals and conferences. His main research interests include deep-learning-based models, computer vision, and data-driven learning. He is currently leading projects focused on the state-of-the-art AI models dedicated to marine biology.
\end{IEEEbiography}
\vfill 

\begin{IEEEbiography}[{\includegraphics[width=1in, height=1.25in, clip, keepaspectratio]{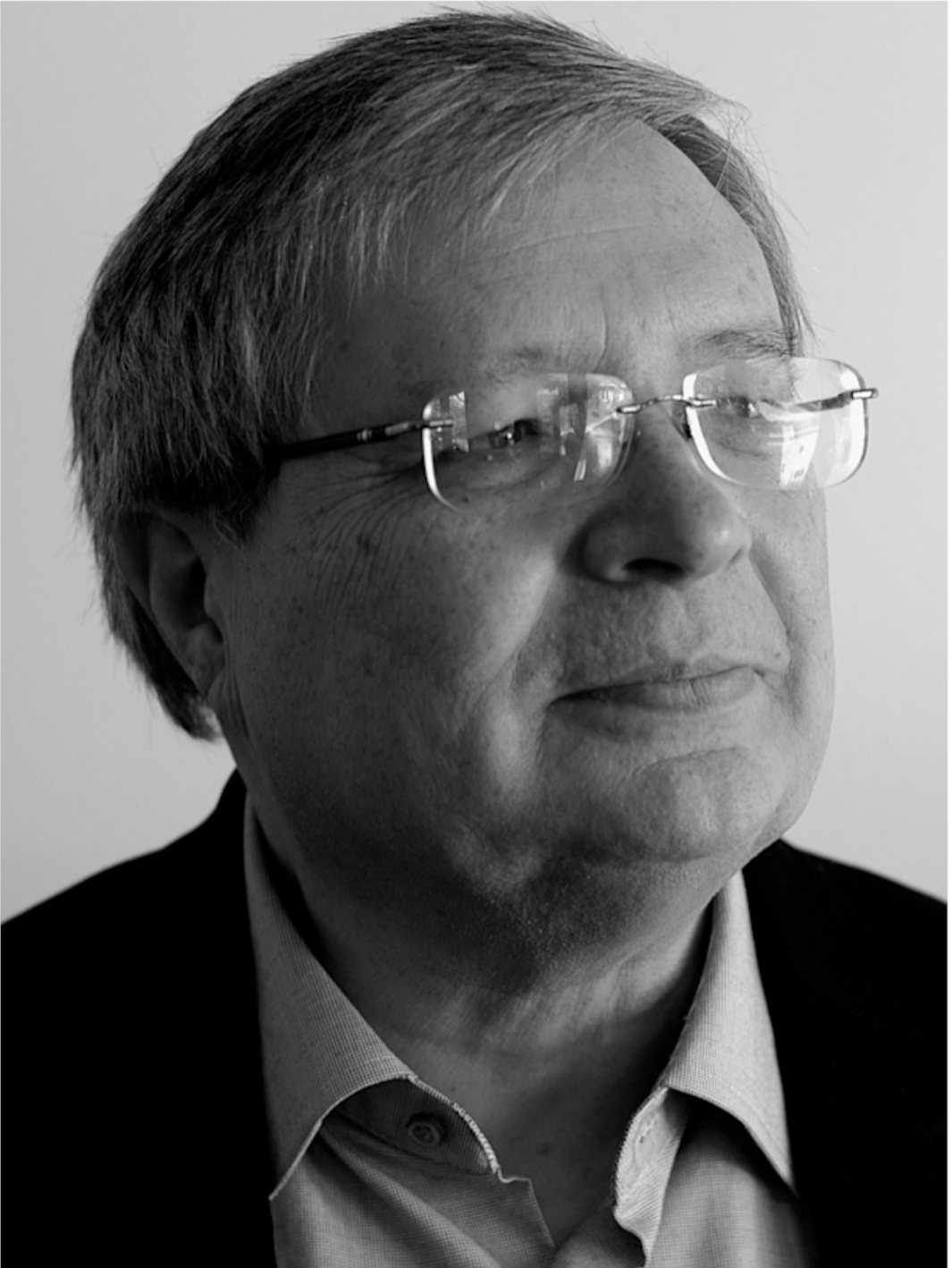}}]{Stan Matwin} is currently the director of the Institute for Big Data Analytics, Dalhousie University, Halifax, Nova Scotia, where he is a professor and Canada Research Chair (Tier 1) in Interpretability for Machine Learning. He is also a distinguished professor (Emeritus) at the University of Ottawa and a full professor with the Institute of Computer Science, Polish Academy of Sciences. His main research interests include big data, text mining, machine learning, and data privacy. He is a member of the Editorial Boards of \textit{IEEE Transactions on Knowledge and Data Engineering} and \textit{Journal of Intelligent Information Systems}. He was the recipient of the Lifetime Achievement Award of the Canadian AI Association (CAIAC).
\vspace{-1.35cm} 
\end{IEEEbiography}
\begin{IEEEbiography}[{\includegraphics[width=1in, height=1.25in, clip, keepaspectratio]{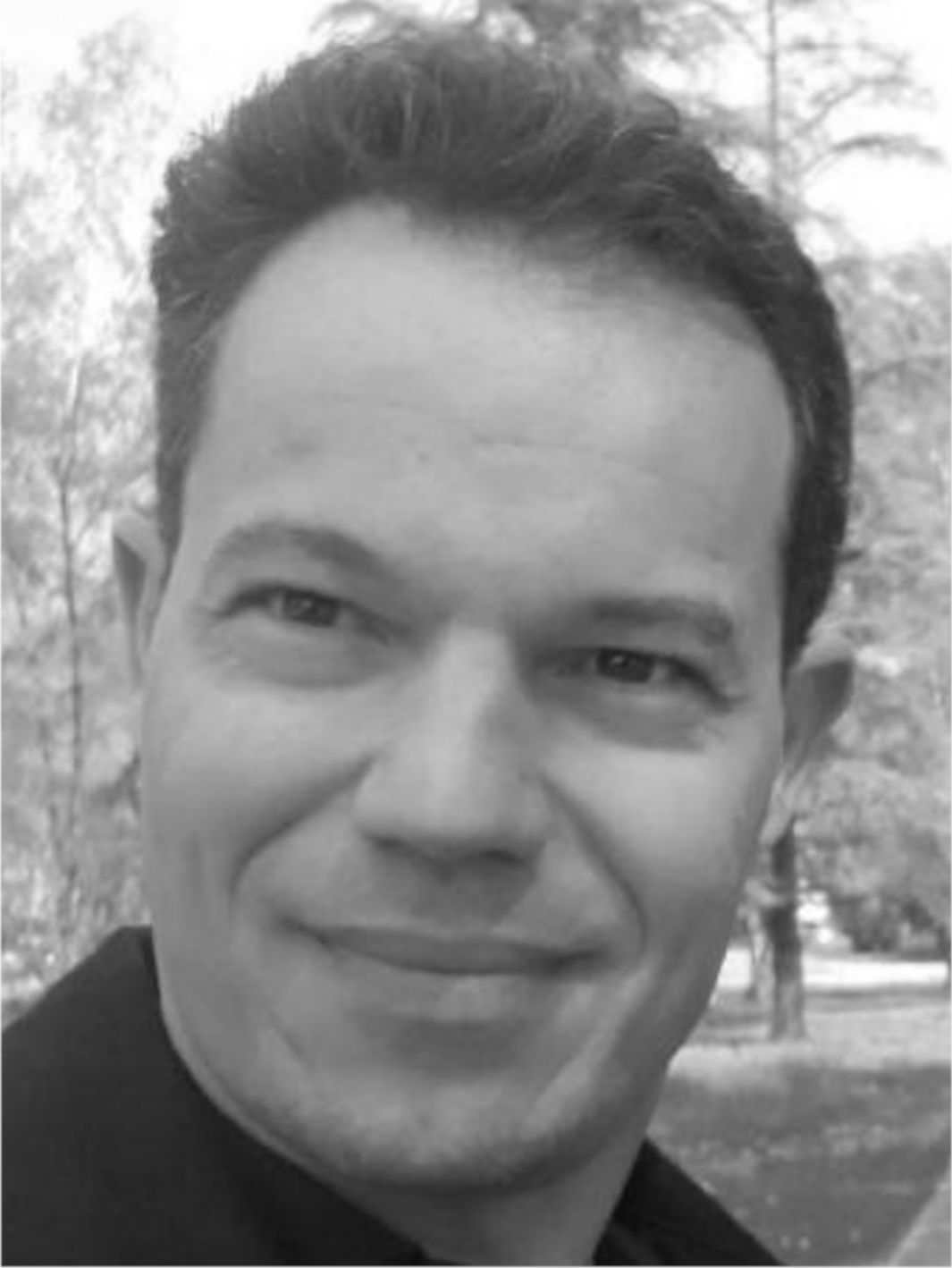}}]{Jose F. Rodrigues-Jr} received the PhD from the University of Sao Paulo, Brazil, part of which was carried out from Carnegie Mellon University, USA, in 2007. He is currently an associate professor at the University of Sao Paulo, Brazil. He is a regular reviewer and author in his research field, which includes data science, machine learning, content-based data retrieval, visualization, and the application of such techniques in the medic, agriculture, and e-learning domains, contributing to publications in major journals and conferences.
\vspace{.7cm} 
\end{IEEEbiography}
\begin{IEEEbiography}[{\includegraphics[width=1in, height=1.25in, clip, keepaspectratio]{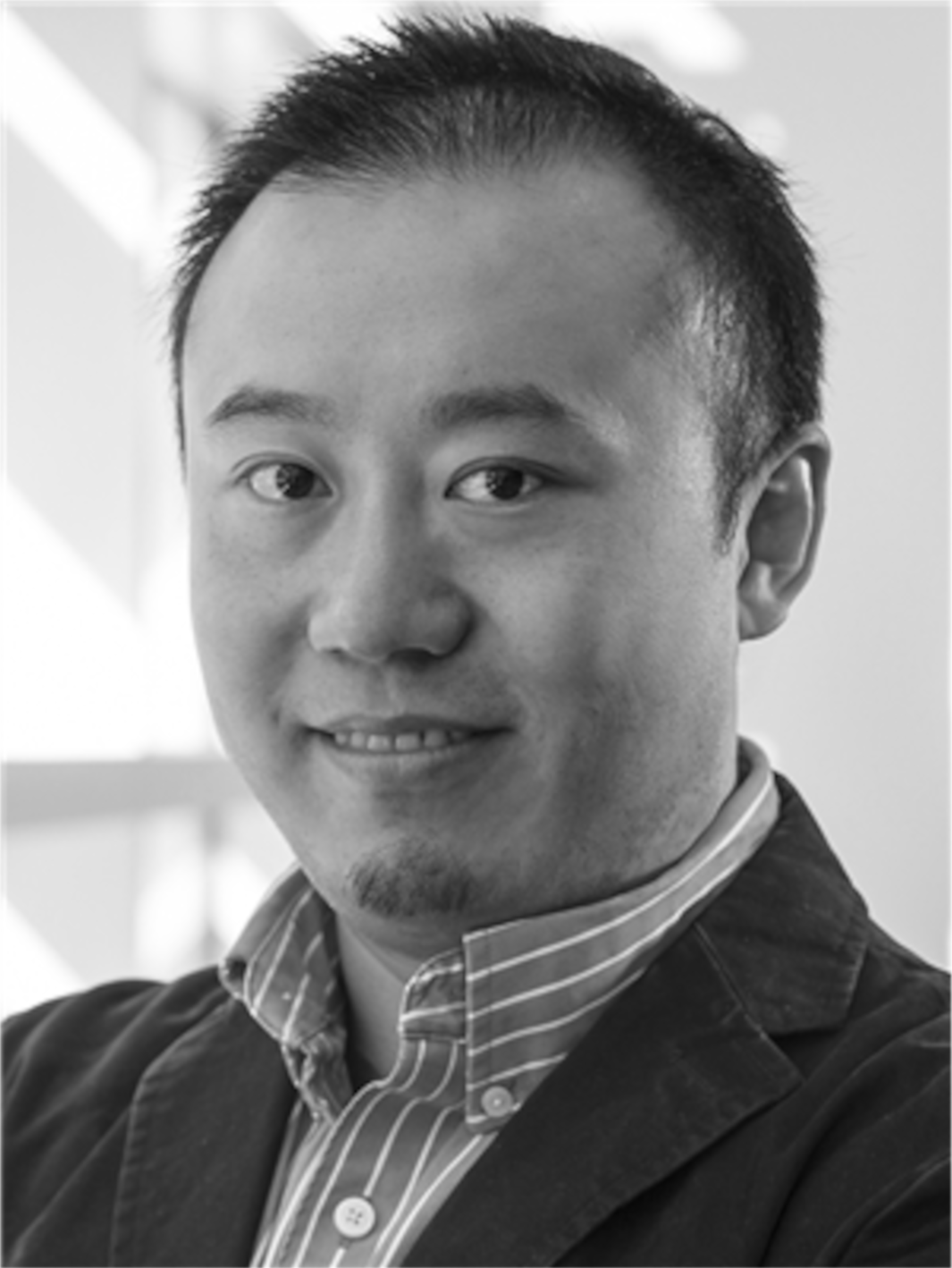}}]{Jimeng Sun} in currently a professor at the Computer Science Department, University of Illinois, Urbana-Champaing. Previously, he was with the College of Computing, Georgia Institute of Technology. His research interest includes Artificial Intelligence for healthcare, deep learning for drug discovery, clinical trial optimization, computational phenotyping, clinical predictive modeling, treatment recommendation, and health monitoring.
\vspace{.7cm} 
\end{IEEEbiography}
\vfill 

\end{document}


\lhead{\it Pay Attention to Evolution: Time Series forecasting with Deep Graph-Evolution Learning}
\rhead{}  

\title{
    {\bf Supplementary Material}\\
    Pay Attention to Evolution: Time Series\\
    Forecasting with Deep Graph-Evolution Learning
}
\author{
    Gabriel~Spadon~\orcidicon{0000-0001-8437-4349},
    Shenda~Hong~\orcidicon{0000-0001-7521-5127}, Bruno~Brandoli~\orcidicon{0000-0001-6167-8104},\\
    Stan~Matwin~\orcidicon{0000-0001-6629-8434},
    Jose F.~Rodrigues-Jr~\orcidicon{0000-0001-8318-1780},
    and~Jimeng~Sun~\orcidicon{0000-0003-1512-6426}
}

\date{} 
\maketitle 
\listoftables 

\paragraph{Notes.}
Table 1 list the acronym and full name of all algorithms we tested during the baselines' computation.
Tables 2 to 6 present detailed information from the experiments discussed along with the main manuscript.
The following tables regard the tests using Transfer Learning on the SARS-CoV-2 dataset, in which a new network was trained every 15 days starting on 45 days after the pandemic started and up to 120 days of its duration.

\section*{Extended Methods}

\paragraph{Cosine Similarity.}
The cosine similarity, which has been widely applied in learning approaches, accounts for the similarity between two non-zero vectors based on their orientation in an inner product space~\cite{Tan2006:DataMining}.
The underlying idea is that the similarity is a function of the cosine angle $\theta$ between vectors $\mathbf{u} = [u_1, u_2, \ldots, u_N] \in \mathbb{R}^{N \times 1}$ and $\mathbf{v} = [v_1, v_2, \ldots, v_N] \in \mathbb{R}^{N \times 1}$.
Hence, when $\theta=1$, the two vectors in the inner product space have the same orientation, when $\theta=0$, these vectors are oriented a $90^{\circ}$ relative to each other, and when $\theta=-1$, the vectors are diametrically opposed.
The cosine similarity between the vectors $\mathbf{u}$ and $\mathbf{v}$ is defined as:
\begin{equation}
    cos_\theta(\mathbf{u}, \mathbf{v}) = \frac{\mathbf{u} \cdot \mathbf{v}}{\|\mathbf{u}\| \circ \|\mathbf{v}\|}
\end{equation}
where $\mathbf{u} \cdot \mathbf{v} = \sum^{N}_{i=1} u_iv_i$ denotes the dot product between $\mathbf{u}$ and $\mathbf{v}$, and $\|\mathbf{u}\|$ represents the norm of the vector $\mathbf{u} = \sqrt{u\cdot u}$, while $u_i$ is the $i$-th variable of $u$.
In this work's scope, the cosine similarity is used to build similarity adjacency matrices, which measures per-nodes similarity in a variables' co-occurrence graph.
The similarity between two nodes in the graph describes how likely those two variables co-occur in time.
In this case, the similarity ends up acting as an intermediate activation function, enabling the graph evolution process by maintaining relationships' similarity between pairs of nodes.
Thus, the cosine-matrix similarity is defined as:
\begin{equation}
    cos_\theta(\mymtx{A}) = \frac{\mymtx{A} \cdot \mymtx{A}^\nomath{T}}{\|\mymtx{A}\| \circ \|\mymtx{A}\|^\nomath{T}}
\end{equation}
where $\mathbf{A} \cdot \mathbf{A}^\nomath{T}$ denotes the dot product between the adjacency matrix $\mathbf{A}$ and the transposed $\mathbf{A}^\nomath{T}$, while $\|\mathbf{A}\|$ represents the norm of that same matrix with respect to any of its ranks. The resulting matrix of using the cosine-similarity activation is symmetric and referred to along with the main manuscript as \emph{Evolution Weights}.

\paragraph{Horizon Forecasting.}
It stands for an approach used for making non-continuous predictions by accounting for a future gap in the data.
It is useful in a range of applications by considering, for instance, that recent data is not available or too costly to be collected.
Thereby, it is possible to optimize a model that disregards the near future and focuses on the far-away future.
However, such an approach abdicates from additional information that could be learned from continuous timestamp predictions~\cite{Cheng2019:MLCNN}.
By not considering the near past as a variable that influences the near future, we might result in a non-stochastic view of time, meaning that the algorithm focuses on long-term dependencies rather than both long-and short-term dependencies.
Along these lines, both the LSTNet~\cite{Lai2018:LSTNet} and DSANet~\cite{Huang2019:DSANet} comply with horizon forecasting, and to make our results comparable, we set the horizon to one on both of them.
Thus, we started assessing the test results right after the algorithms' last validation step because as closer to the horizon, the more accurate the horizon-based models should be.

\begin{figure}[b!]
    \centering
    \includegraphics[width=.6\linewidth]{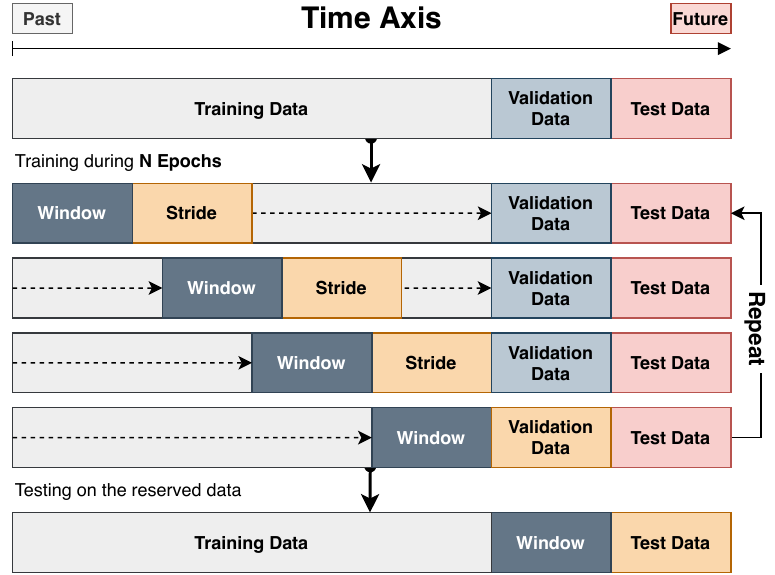}
    \caption{Sliding Window technique for training neural networks.}
    \label{fig:sliding-window}
\end{figure}

\paragraph{Time-Series Segmentation.}
A simplistic yet effective approach to train time-series algorithms is through the Sliding Window technique~\cite{Keogh2004:SegmentingServey}, which is also referred to as Rolling Window (see Fig.~\ref{fig:sliding-window}).
Such a technique fixes a window size, which slides over the time axis, predicting a predefined number of future steps, referred to as stride.
Some time-series studies have been using a variant technique known as Expanding Sliding Window~\cite{Egrioglu2020:FuzzyTimeSeries, Babii2020:NowCasting}.
This variant starts with a prefixed window size, which grows as it slides, showing more information to the algorithm as time goes by.
\ReGENN~holds for the traditional technique as it is bounded to the tensor weights dimension.
Those dimensions are of a preset size and cannot be effortlessly changed during training, as it comes with increased uncertainty by continuously changing the number of internal parameters, such that a conventional neural network optimizer cannot handle it properly.
Nevertheless, the window size of the Sliding Window is well known to be a highly sensitive hyperparameter~\cite{Frank2000:WindowSize, Frank2001:SlidingWindow}; to avoid an increased number of hyperparameter, we followed a non-tunable approach, in which we set the window size before the experiments taking into consideration the context of the datasets; such values were even across all window-based trials, including the baselines and ablation.

\paragraph{Optimization Strategy.}
\ReGENN~operates on a three-dimensional space shared between samples, time, and variables.
In such a space, it carries out a time-based optimization strategy.
The training process iterates over the time-axis of the dataset, showing to the network how the variables within a subset of time-series behave as time goes by, and later repeating the process through subsets of different samples.
The network's weights are shared among the entire dataset and optimized towards best generalization simultaneously across samples, time, and variables.
The dataset $\mathbf{T} \in \mathbb{R}^{s \times t \times v}$ is sliced into training $\widetilde{\mathbf{T}} \in \mathbb{R}^{s \times w \times v}$ and testing $\widehat{\mathbf{T}} \in \mathbb{R}^{s \times z \times v}$ as follows:

\vspace{.5cm}\noindent%
\begin{minipage}{0.4\textwidth}
    \begin{equation*}
    \widetilde{\mathbf{T}} = 
        \begin{bmatrix}
        \mathbf{T}_{1,1,v} & \mathbf{T}_{1,2,v} & \mathbf{T}_{1,2,v} & \dots  & \mathbf{T}_{1,w,v}\\
        \mathbf{T}_{2,1,v} & \mathbf{T}_{2,2,v} & \mathbf{T}_{2,2,v} & \dots  & \mathbf{T}_{2,w,v}\\
        \mathbf{T}_{3,1,v} & \mathbf{T}_{3,2,v} & \mathbf{T}_{3,2,v} & \dots  & \mathbf{T}_{3,w,v}\\
        \vdots     & \vdots     & \vdots     & \ddots & \vdots    \\
        \mathbf{T}_{b,1,v} & \mathbf{T}_{b,2,v} & \mathbf{T}_{b,2,v} & \dots  & \mathbf{T}_{b,w,v}\\
        \end{bmatrix}
    \end{equation*}
\end{minipage}
\hfill%
\begin{minipage}{0.5\textwidth}
    \begin{equation*}
    \widehat{\mathbf{T}} = 
        \begin{bmatrix}
        \mathbf{T}_{1,1+z,v} & \mathbf{T}_{1,2+z,v} & \mathbf{T}_{1,2+z,v} & \dots  & \mathbf{T}_{1,w+z,v}\\
        \mathbf{T}_{2,1+z,v} & \mathbf{T}_{2,2+z,v} & \mathbf{T}_{2,2+z,v} & \dots  & \mathbf{T}_{2,w+z,v}\\
        \mathbf{T}_{3,1+z,v} & \mathbf{T}_{3,2+z,v} & \mathbf{T}_{3,2+z,v} & \dots  & \mathbf{T}_{3,w+z,v}\\
        \vdots       & \vdots       & \vdots       & \ddots & \vdots      \\
        \mathbf{T}_{b,1+z,v} & \mathbf{T}_{b,2+z,v} & \mathbf{T}_{b,2+z,v} & \dots  & \mathbf{T}_{b,w+z,v}\\
        \end{bmatrix}
    \end{equation*}
\end{minipage}
\vspace{.5cm}%

Once the data is sliced, we follow by using a gradient descent-based algorithm to optimize the model.
 In this work's scope, we used Adam~\cite{Kingma2015:Adam} as the optimizer, as it is the most common optimizer among time-series forecasting problems.
As the optimization criterion, we used the Mean Absolute Error (MAE), which is a generalization of the Support Vector Regression~\cite{Vapnik1997:SVR} with soft-margin criterion formalized as it follows:
\begin{equation*}
    \displaystyle{
        \minimize_{\myvec{w}} 
        \left(\frac{1}{2}\left\|{\myvec{w}}\right\|^{2}_{\nomath{F}} + \myset{C}\right) \times \sum_{i=1}^{n} (\xi_{i} + \xi^*_{i})
        \hspace{1cm}%
        \begin{aligned}
            \subject
            & y_i-(\myvec{w} \cdot \myvec{x}_i)-b \leq \rho + \xi_{i},\\
            & (\myvec{w} \cdot \myvec{x}_i)+b-y_i \leq \rho + \xi^*_{i},\\
            & \xi_{i},\xi^*_{i} \geq 0.
        \end{aligned}
    }%
\end{equation*}
\noindent%
where $\myvec{w}$ is the set of optimizable parameters, $\|\cdot\|_{\nomath{F}}$ is the Frobenius norm, and both $\myset{C}$ and $\rho$ are hyperparameters.
The idea, then, is to find $\myvec{w}$ that better fit $y_i, \myvec{x}_i \forall i \in [1,n]$ so that all values are in $[\rho + \xi_{i}, \rho + \xi^*_{i}]$, where $\xi_{i}$ and $\xi^*_{i}$ are the two farther opposite points in the dataset.
A similar formulation on the Linear SVR implementation for horizon forecasting was presented by {\it Lai et al.}~\cite{Lai2018:LSTNet}.
Due to the higher-dimensionality among the multiple multivariate time-series used in this study, in which we consider the time to be continuous, the problem becomes:
\begin{equation*}
    \displaystyle{
        \minimize_{\Omega} 
        \left(\frac{1}{2}\left\|{\Omega}\right\|^{2}_{\nomath{F}} + \myset{C}\right) \times  \sum_{i=1}^{s} \sum_{j=1}^{w} \xi_{i}
        \hspace{1cm}%
        \begin{aligned}
            \subject
            &\left|\widehat{\mathbf{Y}}_{i,j} - \widehat{\mathbf{T}}_{i,j}\right| \leq \rho + \xi_{i,j},~\xi_{i,j} \geq 0.
        \end{aligned}
    }%
\end{equation*}
\noindent%
where $\Omega$ is the set of internal parameters of \ReGENN, $\widehat{\mathbf{Y}}$ is the network's output and $\widehat{\mathbf{T}}$ the ground truth.
When disregarding $\myset{C}$ and setting $\rho$ as zero, we can reduce the problem to the MAE loss formulation:
\begin{equation*}
    \displaystyle{%
        \minimize_{\Omega}%
            \sum_{i=1}^{s} \sum_{j=1}^{w} \left|\widehat{\mathbf{Y}}_{i,j} - \widehat{\mathbf{T}}_{i,j}\right|
    }%
\end{equation*}

Square-and logarithm-based criteria can also be used with \ReGENN.
We avoid doing so, as this is a decision that should be made based on each dataset.
Contrarily, we follow the SVR path towards the evaluation of absolute values, which is less sensitive to outliers and enables \ReGENN~to be applied to a range of applications.

\begin{figure}[hbt!]
    \centering
    \includegraphics[width=.7\linewidth]{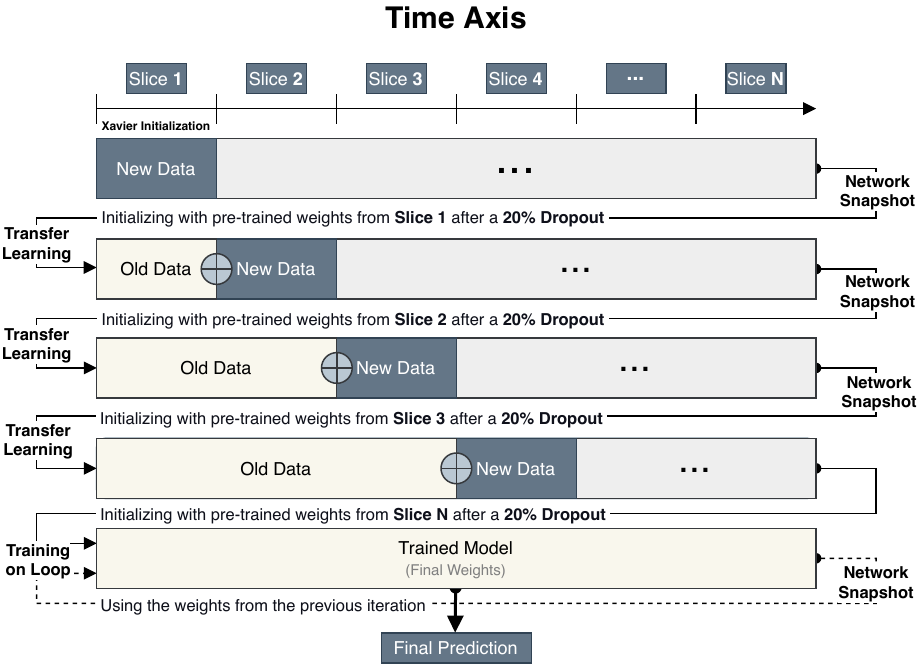}
    \caption{Transfer Learning used for streaming time-series.}
    \label{fig:transfer-learning}
\end{figure}

\paragraph{Transfer-Learning Approach.}
We adopted a Transfer Learning approach to train the network on the SARS-CoV-2 dataset that, although different, resembles Online Deep Learning~\cite{Sahoo2018:OnlineLearning}.
The idea is to train the network on incremental slices of the time-axis, such that the pre-trained weights of a previous slice are used to initialize the weights of the network in the next slice (see Fig.~\ref{fig:transfer-learning}).
This technique aims not only to achieve better performance towards the network but also to show that \ReGENN~is useful throughout the pandemic.

Hyperparameters adjustment is usually required when transferring the weights from one network to another, mainly the learning rate; for the list of hyperparameters we used, see Tab.~\ref{tab:hyperparameters-2}.
Besides, we deliberately applied a 20\% Dropout on all tensor weights outside the network architecture and before starting the training.
The aim behind that decision was to insert randomness in the pipeline and avoid local optima.
It worth mentioning that we did not observe any decrease in performance, but the optimizer's convergence was slower in some cases.

\paragraph{Baselines Algorithms.}
Open-source Python libraries provided the time series and machine learning algorithms used along with the experiments.
Time series algorithms came from the statsmodels\footnote{Available at \url{https://www.statsmodels.org/stable/index.html}.}, while the machine learning ones majorly from the Scikit-Learn\footnote{Available at \url{https://scikit-learn.org/stable/}.}.
Further algorithms, such as XGBoost\footnote{Available at \url{https://xgboost.readthedocs.io/en/latest/}.}, LGBM\footnote{Available at \url{https://lightgbm.readthedocs.io/en/latest/}.}, and CatBoost\footnote{Available at \url{https://catboost.ai/}.}, have a proprietary open-source implementation, which was preferred over the others.
We used the default hyperparameters over all the experiments, performing no fine-tuning.
However, because all the datasets we tested are strictly positive, we forced all the negative output to become zero, such as made by a ReLU activation function.

A list with names and algorithms tested along with the experiments is provided in Tab~\ref{tab:algorithms}, which contains more algorithms than we reported in the main paper.
That because we are listing all algorithms, even those removed from the pipeline due to being incapable of working with the input data and yielding exceptions.

\begin{longtable}[c]{|c|l|l|}
\caption
[List of tested algorithms]%
{List of algorithms tested during the baselines' computation. The \textbf{Acronym} column presents each algorithm's short name, and the \textbf{Name} column shows the full name of all algorithms (when applicable).}
\label{tab:algorithms}\\
\hline
\textbf{\#} & \textbf{Acronym}            & \textbf{Name}                                           \\ \hline
\endfirsthead
\endhead
\textbf{0}  & AdaBoost                    & Adaptive Boosting                                       \\ \hline
\textbf{1}  & ARD                         & Automatic Relevance Determination                       \\ \hline
\textbf{2}  & ARIMA                       & Autoregressive Integrated Moving Average                \\ \hline
\textbf{3}  & ARMA                        & Autoregressive Moving Average                           \\ \hline
\textbf{4}  & Autoregressive              & ---                                                     \\ \hline
\textbf{5}  & Bagging                     & ---                                                     \\ \hline
\textbf{6}  & BayesianRidge               & ---                                                     \\ \hline
\textbf{7}  & CatBoost                    & ---                                                     \\ \hline
\textbf{8}  & CCA                         & Canonical Correlation Analysis                          \\ \hline
\textbf{9}  & Decision Tree               & ---                                                     \\ \hline
\textbf{10} & DSANet                      & Dual Self-Attention Network                             \\ \hline
\textbf{11} & Elman RNN                   & Elman's Recurrent Neural Network                        \\ \hline
\textbf{12} & Exponential Smoothing       & Single Exponential Smoothing                            \\ \hline
\textbf{13} & Extra Tree                  & Extremely Randomized Tree                               \\ \hline
\textbf{14} & Extra Trees                 & Extremely Randomized Trees                              \\ \hline
\textbf{15} & Gaussian Process            & ---                                                     \\ \hline
\textbf{16} & Gradient Boosting           & ---                                                     \\ \hline
\textbf{17} & GRU                         & Gated Recurrent Unit                                    \\ \hline
\textbf{18} & Histogram Grad. Boosting    & Histogram Gradient Boosting                             \\ \hline
\textbf{19} & Historical Average          & Dummy Regressor                                         \\ \hline
\textbf{20} & Huber                       & ---                                                     \\ \hline
\textbf{21} & Isotonic                    & ---                                                     \\ \hline
\textbf{22} & Kernel Ridge                & ---                                                     \\ \hline
\textbf{23} & KNeighbors                  & k-Nearest Neighbors                                     \\ \hline
\textbf{24} & Lars                        & Least Angle Regression                                  \\ \hline
\textbf{25} & Lasso-Lars                  & Least Absolute Shrinkage and Selection Operator w/ Lars \\ \hline
\textbf{26} & Lasso-Lars-IC               & Lasso-Lars w/ Information Criterion                     \\ \hline
\textbf{27} & LGBM                        & Light Gradient Boosting Machine                         \\ \hline
\textbf{28} & Linear Regression           & ---                                                     \\ \hline
\textbf{29} & Linear SVR                  & Linear Support Vector Regression                        \\ \hline
\textbf{30} & LSTM                        & Long Short Term Memory                                  \\ \hline
\textbf{31} & LSTNet                      & Long-and Short Term time-series Network                 \\ \hline
\textbf{32} & MLCNN                       & Multi-Level Construal Neural Network                    \\ \hline
\textbf{33} & Moving Average              & ---                                                     \\ \hline
\textbf{34} & Multi-Task Elastic-Net      & ---                                                     \\ \hline
\textbf{35} & Multi-Task Lasso            & ---                                                     \\ \hline
\textbf{36} & NuSVR                       & Nu-Support Vector Regression                            \\ \hline
\textbf{37} & Orthogonal Matching Pursuit & ---                                                     \\ \hline
\textbf{38} & Passive Aggressive          & ---                                                     \\ \hline
\textbf{39} & PLS Canonical               & Partial Least Squares Canonical                         \\ \hline
\textbf{40} & PLS                         & Partial Least Squares                                   \\ \hline
\textbf{41} & Radius Neighbors            & ---                                                     \\ \hline
\textbf{42} & Random Forest               & ---                                                     \\ \hline
\textbf{43} & RANSAC                      & Random Sample Consensus Regressor                       \\ \hline
\textbf{44} & ReGENN                      & Recurrent Graph Evolution Neural Network                \\ \hline
\textbf{45} & Ridge                       & ---                                                     \\ \hline
\textbf{46} & SARIMA                      & Seasonal Autoregressive Integrated Moving Average       \\ \hline
\textbf{47} & SGD                         & Stochastic Gradient Descent                             \\ \hline
\textbf{48} & SVR                         & Support Vector Regression                               \\ \hline
\textbf{49} & Theil-Sen                   & ---                                                     \\ \hline
\textbf{50} & Transformed Target          & ---                                                     \\ \hline
\textbf{51} & Vector Autoregression       & ---                                                     \\ \hline
\textbf{52} & XGBoost                     & Extreme Gradient Boosting                               \\ \hline
\end{longtable}

\bibliographystyle{IEEEtran}
\bibliography{3-references.bib}

\clearpage\newpage%
\begin{titlepage}
    \vspace*{\fill}
        \begin{center}
          \Huge{\bf Hyperparameters\\\rule[5px]{2cm}{.1cm}~for the~\rule[5px]{2cm}{.1cm}\\Main Experiments}
        \end{center}
    \vspace*{\fill}
\end{titlepage}

\savegeometry{geom}
\newgeometry{margin=45pt}
\thispagestyle{empty}\setcounter{page}{9}\clearpage
\begin{longtable}[c]{rccc}
\caption[Hyperparameters for the main experiments.]%
{List of hyperparameters used during the main experiments.}
\label{tab:hyperparameters-1}\\
\cline{2-4}
\multicolumn{1}{l|}{} &
  \multicolumn{3}{c|}{\multirow{2}{*}{\textbf{---------------| ReGENN |---------------}}} \\
\multicolumn{1}{r|}{\textbf{}} &
  \multicolumn{3}{c|}{} \\ \cline{2-4} 
\endfirsthead
\multicolumn{4}{c}%
{{\bfseries Table \thetable\ continued from previous page}} \\
\endhead
\multicolumn{1}{l|}{} &
  \multicolumn{1}{c|}{\textbf{SARS-CoV-2}} &
  \multicolumn{1}{c|}{\textbf{Brazilian Weather}} &
  \multicolumn{1}{c|}{\textbf{PhysioNet}} \\ \cline{2-4} 
\multicolumn{1}{r|}{\textbf{autoregression}} &
  \multicolumn{1}{c|}{True} &
  \multicolumn{1}{c|}{True} &
  \multicolumn{1}{c|}{True} \\ \cline{2-4} 
\multicolumn{1}{r|}{\textbf{batch-size}} &
  \multicolumn{1}{c|}{32} &
  \multicolumn{1}{c|}{64} &
  \multicolumn{1}{c|}{256} \\ \cline{2-4} 
\multicolumn{1}{r|}{\textbf{bias}} &
  \multicolumn{1}{c|}{True} &
  \multicolumn{1}{c|}{True} &
  \multicolumn{1}{c|}{True} \\ \cline{2-4} 
\multicolumn{1}{r|}{\textbf{bidirectional-gate}} &
  \multicolumn{1}{c|}{False} &
  \multicolumn{1}{c|}{False} &
  \multicolumn{1}{c|}{False} \\ \cline{2-4} 
\multicolumn{1}{r|}{\textbf{bidirectional-sequencer}} &
  \multicolumn{1}{c|}{False} &
  \multicolumn{1}{c|}{False} &
  \multicolumn{1}{c|}{False} \\ \cline{2-4} 
\multicolumn{1}{r|}{\textbf{clip-norm}} &
  \multicolumn{1}{c|}{85.0} &
  \multicolumn{1}{c|}{10.0} &
  \multicolumn{1}{c|}{25.0} \\ \cline{2-4} 
\multicolumn{1}{r|}{\textbf{criterion}} &
  \multicolumn{1}{c|}{MAE} &
  \multicolumn{1}{c|}{MAE} &
  \multicolumn{1}{c|}{MAE} \\ \cline{2-4} 
\multicolumn{1}{r|}{\textbf{dropout}} &
  \multicolumn{1}{c|}{0.0} &
  \multicolumn{1}{c|}{0.0} &
  \multicolumn{1}{c|}{0.0} \\ \cline{2-4} 
\multicolumn{1}{r|}{\textbf{early-stop}} &
  \multicolumn{1}{c|}{250} &
  \multicolumn{1}{c|}{100} &
  \multicolumn{1}{c|}{10} \\ \cline{2-4} 
\multicolumn{1}{r|}{\textbf{epochs}} &
  \multicolumn{1}{c|}{2500} &
  \multicolumn{1}{c|}{1000} &
  \multicolumn{1}{c|}{100} \\ \cline{2-4} 
\multicolumn{1}{r|}{\textbf{evolution-function}} &
  \multicolumn{1}{c|}{Identity} &
  \multicolumn{1}{c|}{Identity} &
  \multicolumn{1}{c|}{Identity} \\ \cline{2-4} 
\multicolumn{1}{r|}{\textbf{gate}} &
  \multicolumn{1}{c|}{LSTM} &
  \multicolumn{1}{c|}{LSTM} &
  \multicolumn{1}{c|}{LSTM} \\ \cline{2-4} 
\multicolumn{1}{r|}{\textbf{iterator}} &
  \multicolumn{1}{c|}{Time} &
  \multicolumn{1}{c|}{Time} &
  \multicolumn{1}{c|}{Time} \\ \cline{2-4} 
\multicolumn{1}{r|}{\textbf{learning-rate}} &
  \multicolumn{1}{c|}{0.001} &
  \multicolumn{1}{c|}{0.0001} &
  \multicolumn{1}{c|}{0.001} \\ \cline{2-4} 
\multicolumn{1}{r|}{\textbf{load-weights}} &
  \multicolumn{1}{c|}{True} &
  \multicolumn{1}{c|}{False} &
  \multicolumn{1}{c|}{False} \\ \cline{2-4} 
\multicolumn{1}{r|}{\textbf{no-encoder}} &
  \multicolumn{1}{c|}{False} &
  \multicolumn{1}{c|}{False} &
  \multicolumn{1}{c|}{False} \\ \cline{2-4} 
\multicolumn{1}{r|}{\textbf{no-sequencer}} &
  \multicolumn{1}{c|}{False} &
  \multicolumn{1}{c|}{False} &
  \multicolumn{1}{c|}{False} \\ \cline{2-4} 
\multicolumn{1}{r|}{\textbf{normalization-axis}} &
  \multicolumn{1}{c|}{2} &
  \multicolumn{1}{c|}{2} &
  \multicolumn{1}{c|}{2} \\ \cline{2-4} 
\multicolumn{1}{r|}{\textbf{normalization-type}} &
  \multicolumn{1}{c|}{Maximum} &
  \multicolumn{1}{c|}{Maximum} &
  \multicolumn{1}{c|}{Maximum} \\ \cline{2-4} 
\multicolumn{1}{r|}{\textbf{optimizer}} &
  \multicolumn{1}{c|}{Adam} &
  \multicolumn{1}{c|}{Adam} &
  \multicolumn{1}{c|}{Adam} \\ \cline{2-4} 
\multicolumn{1}{r|}{\textbf{output-function}} &
  \multicolumn{1}{c|}{ReLU} &
  \multicolumn{1}{c|}{ReLU} &
  \multicolumn{1}{c|}{ReLU} \\ \cline{2-4} 
\multicolumn{1}{r|}{\textbf{random-seed}} &
  \multicolumn{1}{c|}{0} &
  \multicolumn{1}{c|}{0} &
  \multicolumn{1}{c|}{0} \\ \cline{2-4} 
\multicolumn{1}{r|}{\textbf{scheduler-factor}} &
  \multicolumn{1}{c|}{0.95} &
  \multicolumn{1}{c|}{0.95} &
  \multicolumn{1}{c|}{0.2} \\ \cline{2-4} 
\multicolumn{1}{r|}{\textbf{scheduler-min-lr}} &
  \multicolumn{1}{c|}{0.0} &
  \multicolumn{1}{c|}{0.0} &
  \multicolumn{1}{c|}{0.0} \\ \cline{2-4} 
\multicolumn{1}{r|}{\textbf{scheduler-patience}} &
  \multicolumn{1}{c|}{25} &
  \multicolumn{1}{c|}{50} &
  \multicolumn{1}{c|}{40} \\ \cline{2-4} 
\multicolumn{1}{r|}{\textbf{scheduler-threshold}} &
  \multicolumn{1}{c|}{0.1} &
  \multicolumn{1}{c|}{0.1} &
  \multicolumn{1}{c|}{0.1} \\ \cline{2-4} 
\multicolumn{1}{r|}{\textbf{sequencer}} &
  \multicolumn{1}{c|}{LSTM} &
  \multicolumn{1}{c|}{LSTM} &
  \multicolumn{1}{c|}{LSTM} \\ \cline{2-4} 
\multicolumn{1}{r|}{\textbf{stride}} &
  \multicolumn{1}{c|}{14} &
  \multicolumn{1}{c|}{56} &
  \multicolumn{1}{c|}{6} \\ \cline{2-4} 
\multicolumn{1}{r|}{\textbf{seed}} &
  \multicolumn{1}{c|}{0} &
  \multicolumn{1}{c|}{0} &
  \multicolumn{1}{c|}{0} \\ \cline{2-4} 
\multicolumn{1}{r|}{\textbf{validation-stride}} &
  \multicolumn{1}{c|}{7} &
  \multicolumn{1}{c|}{28} &
  \multicolumn{1}{c|}{6} \\ \cline{2-4} 
\multicolumn{1}{r|}{\textbf{watch-axis}} &
  \multicolumn{1}{c|}{2} &
  \multicolumn{1}{c|}{2} &
  \multicolumn{1}{c|}{2} \\ \cline{2-4} 
\multicolumn{1}{r|}{\textbf{window}} &
  \multicolumn{1}{c|}{7} &
  \multicolumn{1}{c|}{84} &
  \multicolumn{1}{c|}{12} \\ \cline{2-4} 
\multicolumn{1}{r|}{\textbf{}} &
  \multicolumn{3}{c|}{\multirow{2}{*}{\textbf{---------------| MLCNN |---------------}}} \\
\multicolumn{1}{l|}{} &
  \multicolumn{3}{c|}{} \\ \cline{2-4} 
\multicolumn{1}{l|}{} &
  \multicolumn{1}{c|}{\textbf{SARS-CoV-2}} &
  \multicolumn{1}{c|}{\textbf{Brazilian Weather}} &
  \multicolumn{1}{c|}{\textbf{PhysioNet}} \\ \cline{2-4} 
\multicolumn{1}{r|}{\textbf{batch-size}} &
  \multicolumn{1}{c|}{32} &
  \multicolumn{1}{c|}{64} &
  \multicolumn{1}{c|}{256} \\ \cline{2-4} 
\multicolumn{1}{r|}{\textbf{clip-norm}} &
  \multicolumn{1}{c|}{10} &
  \multicolumn{1}{c|}{10} &
  \multicolumn{1}{c|}{10} \\ \cline{2-4} 
\multicolumn{1}{r|}{\textbf{collaborate-span}} &
  \multicolumn{1}{c|}{2} &
  \multicolumn{1}{c|}{2} &
  \multicolumn{1}{c|}{2} \\ \cline{2-4} 
\multicolumn{1}{r|}{\textbf{collaborate-stride}} &
  \multicolumn{1}{c|}{1} &
  \multicolumn{1}{c|}{1} &
  \multicolumn{1}{c|}{1} \\ \cline{2-4} 
\multicolumn{1}{r|}{\textbf{criterion}} &
  \multicolumn{1}{c|}{MAE} &
  \multicolumn{1}{c|}{MAE} &
  \multicolumn{1}{c|}{MAE} \\ \cline{2-4} 
\multicolumn{1}{r|}{\textbf{dropout}} &
  \multicolumn{1}{c|}{0.2} &
  \multicolumn{1}{c|}{0.2} &
  \multicolumn{1}{c|}{0.2} \\ \cline{2-4} 
\multicolumn{1}{r|}{\textbf{epochs}} &
  \multicolumn{1}{c|}{2500} &
  \multicolumn{1}{c|}{1000} &
  \multicolumn{1}{c|}{100} \\ \cline{2-4} 
\multicolumn{1}{r|}{\textbf{hidden-CNN}} &
  \multicolumn{1}{c|}{100} &
  \multicolumn{1}{c|}{100} &
  \multicolumn{1}{c|}{100} \\ \cline{2-4} 
\multicolumn{1}{r|}{\textbf{hidden-RNN}} &
  \multicolumn{1}{c|}{100} &
  \multicolumn{1}{c|}{100} &
  \multicolumn{1}{c|}{100} \\ \cline{2-4} 
\multicolumn{1}{r|}{\textbf{highway-window}} &
  \multicolumn{1}{c|}{1} &
  \multicolumn{1}{c|}{1} &
  \multicolumn{1}{c|}{1} \\ \cline{2-4} 
\multicolumn{1}{r|}{\textbf{input-size}} &
  \multicolumn{1}{c|}{7} &
  \multicolumn{1}{c|}{84} &
  \multicolumn{1}{c|}{12} \\ \cline{2-4} 
\multicolumn{1}{r|}{\textbf{kernel-size}} &
  \multicolumn{1}{c|}{5} &
  \multicolumn{1}{c|}{5} &
  \multicolumn{1}{c|}{5} \\ \cline{2-4} 
\multicolumn{1}{r|}{\textbf{learning-rate}} &
  \multicolumn{1}{c|}{0.001} &
  \multicolumn{1}{c|}{0.001} &
  \multicolumn{1}{c|}{0.001} \\ \cline{2-4} 
\multicolumn{1}{r|}{\textbf{mode}} &
  \multicolumn{1}{c|}{Continuous} &
  \multicolumn{1}{c|}{Continuous} &
  \multicolumn{1}{c|}{Continuous} \\ \cline{2-4} 
\multicolumn{1}{r|}{\textbf{num-CNN}} &
  \multicolumn{1}{c|}{10} &
  \multicolumn{1}{c|}{10} &
  \multicolumn{1}{c|}{10} \\ \cline{2-4} 
\multicolumn{1}{r|}{\textbf{normalization}} &
  \multicolumn{1}{c|}{1} &
  \multicolumn{1}{c|}{1} &
  \multicolumn{1}{c|}{1} \\ \cline{2-4} 
\multicolumn{1}{r|}{\textbf{optimizer}} &
  \multicolumn{1}{c|}{Adam} &
  \multicolumn{1}{c|}{Adam} &
  \multicolumn{1}{c|}{Adam} \\ \cline{2-4} 
\multicolumn{1}{r|}{\textbf{output-function}} &
  \multicolumn{1}{c|}{ReLU} &
  \multicolumn{1}{c|}{ReLU} &
  \multicolumn{1}{c|}{ReLU} \\ \cline{2-4} 
\multicolumn{1}{r|}{\textbf{output-size}} &
  \multicolumn{1}{c|}{14} &
  \multicolumn{1}{c|}{56} &
  \multicolumn{1}{c|}{6} \\ \cline{2-4} 
\multicolumn{1}{r|}{\textbf{seed}} &
  \multicolumn{1}{c|}{0} &
  \multicolumn{1}{c|}{0} &
  \multicolumn{1}{c|}{0} \\ \cline{2-4} 
\multicolumn{1}{l}{} &
  \multicolumn{1}{l}{} &
  \multicolumn{1}{l}{} &
  \multicolumn{1}{l}{} \\
\multicolumn{1}{l}{} &
  \multicolumn{1}{l}{} &
  \multicolumn{1}{l}{} &
  \multicolumn{1}{l}{} \\
\multicolumn{1}{l}{} &
  \multicolumn{1}{l}{} &
  \multicolumn{1}{l}{} &
  \multicolumn{1}{l}{} \\
\multicolumn{1}{l}{} &
  \multicolumn{1}{l}{} &
  \multicolumn{1}{l}{} &
  \multicolumn{1}{l}{} \\ \cline{2-4} 
\multicolumn{1}{l|}{} &
  \multicolumn{3}{c|}{\multirow{2}{*}{\textbf{---------------| DSANet |---------------}}} \\
\multicolumn{1}{r|}{\textbf{}} &
  \multicolumn{3}{c|}{} \\ \cline{2-4} 
\multicolumn{1}{l|}{} &
  \multicolumn{1}{c|}{\textbf{SARS-CoV-2}} &
  \multicolumn{1}{c|}{\textbf{Brazilian Weather}} &
  \multicolumn{1}{c|}{\textbf{PhysioNet}} \\ \cline{2-4} 
\multicolumn{1}{r|}{\textbf{batch-size}} &
  \multicolumn{1}{c|}{32} &
  \multicolumn{1}{c|}{64} &
  \multicolumn{1}{c|}{256} \\ \cline{2-4} 
\multicolumn{1}{r|}{\textbf{clip-norm}} &
  \multicolumn{1}{c|}{10} &
  \multicolumn{1}{c|}{10} &
  \multicolumn{1}{c|}{10} \\ \cline{2-4} 
\multicolumn{1}{r|}{\textbf{criterion}} &
  \multicolumn{1}{c|}{MAE} &
  \multicolumn{1}{c|}{MAE} &
  \multicolumn{1}{c|}{MAE} \\ \cline{2-4} 
\multicolumn{1}{r|}{\textbf{dim-inner}} &
  \multicolumn{1}{c|}{2048} &
  \multicolumn{1}{c|}{2048} &
  \multicolumn{1}{c|}{2048} \\ \cline{2-4} 
\multicolumn{1}{r|}{\textbf{dim-k}} &
  \multicolumn{1}{c|}{64} &
  \multicolumn{1}{c|}{64} &
  \multicolumn{1}{c|}{64} \\ \cline{2-4} 
\multicolumn{1}{r|}{\textbf{dim-model}} &
  \multicolumn{1}{c|}{512} &
  \multicolumn{1}{c|}{512} &
  \multicolumn{1}{c|}{512} \\ \cline{2-4} 
\multicolumn{1}{r|}{\textbf{dim-v}} &
  \multicolumn{1}{c|}{64} &
  \multicolumn{1}{c|}{64} &
  \multicolumn{1}{c|}{64} \\ \cline{2-4} 
\multicolumn{1}{r|}{\textbf{dropout}} &
  \multicolumn{1}{c|}{0.1} &
  \multicolumn{1}{c|}{0.1} &
  \multicolumn{1}{c|}{0.1} \\ \cline{2-4} 
\multicolumn{1}{r|}{\textbf{early-stop}} &
  \multicolumn{1}{c|}{250} &
  \multicolumn{1}{c|}{100} &
  \multicolumn{1}{c|}{10} \\ \cline{2-4} 
\multicolumn{1}{r|}{\textbf{epochs}} &
  \multicolumn{1}{c|}{2500} &
  \multicolumn{1}{c|}{1000} &
  \multicolumn{1}{c|}{100} \\ \cline{2-4} 
\multicolumn{1}{r|}{\textbf{horizon}} &
  \multicolumn{1}{c|}{1} &
  \multicolumn{1}{c|}{1} &
  \multicolumn{1}{c|}{1} \\ \cline{2-4} 
\multicolumn{1}{r|}{\textbf{learning-rate}} &
  \multicolumn{1}{c|}{0.001} &
  \multicolumn{1}{c|}{0.001} &
  \multicolumn{1}{c|}{0.001} \\ \cline{2-4} 
\multicolumn{1}{r|}{\textbf{local}} &
  \multicolumn{1}{c|}{3} &
  \multicolumn{1}{c|}{3} &
  \multicolumn{1}{c|}{3} \\ \cline{2-4} 
\multicolumn{1}{r|}{\textbf{num-heads}} &
  \multicolumn{1}{c|}{8} &
  \multicolumn{1}{c|}{8} &
  \multicolumn{1}{c|}{8} \\ \cline{2-4} 
\multicolumn{1}{r|}{\textbf{num-kernels}} &
  \multicolumn{1}{c|}{32} &
  \multicolumn{1}{c|}{32} &
  \multicolumn{1}{c|}{32} \\ \cline{2-4} 
\multicolumn{1}{r|}{\textbf{num-layers}} &
  \multicolumn{1}{c|}{6} &
  \multicolumn{1}{c|}{6} &
  \multicolumn{1}{c|}{6} \\ \cline{2-4} 
\multicolumn{1}{r|}{\textbf{normalization}} &
  \multicolumn{1}{c|}{2} &
  \multicolumn{1}{c|}{2} &
  \multicolumn{1}{c|}{2} \\ \cline{2-4} 
\multicolumn{1}{r|}{\textbf{optimizer}} &
  \multicolumn{1}{c|}{Adam} &
  \multicolumn{1}{c|}{Adam} &
  \multicolumn{1}{c|}{Adam} \\ \cline{2-4} 
\multicolumn{1}{r|}{\textbf{output-function}} &
  \multicolumn{1}{c|}{ReLU} &
  \multicolumn{1}{c|}{ReLU} &
  \multicolumn{1}{c|}{ReLU} \\ \cline{2-4} 
\multicolumn{1}{r|}{\textbf{seed}} &
  \multicolumn{1}{c|}{0} &
  \multicolumn{1}{c|}{0} &
  \multicolumn{1}{c|}{0} \\ \cline{2-4} 
\multicolumn{1}{r|}{\textbf{w-kernel}} &
  \multicolumn{1}{c|}{1} &
  \multicolumn{1}{c|}{1} &
  \multicolumn{1}{c|}{1} \\ \cline{2-4} 
\multicolumn{1}{r|}{\textbf{window}} &
  \multicolumn{1}{c|}{7} &
  \multicolumn{1}{c|}{84} &
  \multicolumn{1}{c|}{12} \\ \cline{2-4} 
\multicolumn{1}{l|}{} &
  \multicolumn{3}{c|}{\multirow{2}{*}{\textbf{---------------| LSTNet |---------------}}} \\
\multicolumn{1}{r|}{\textbf{}} &
  \multicolumn{3}{c|}{} \\ \cline{2-4} 
\multicolumn{1}{l|}{} &
  \multicolumn{1}{c|}{\textbf{SARS-CoV-2}} &
  \multicolumn{1}{c|}{\textbf{Brazilian Weather}} &
  \multicolumn{1}{c|}{\textbf{PhysioNet}} \\ \cline{2-4} 
\multicolumn{1}{r|}{\textbf{batch-size}} &
  \multicolumn{1}{c|}{32} &
  \multicolumn{1}{c|}{64} &
  \multicolumn{1}{c|}{256} \\ \cline{2-4} 
\multicolumn{1}{r|}{\textbf{clip-norm}} &
  \multicolumn{1}{c|}{10} &
  \multicolumn{1}{c|}{10} &
  \multicolumn{1}{c|}{10} \\ \cline{2-4} 
\multicolumn{1}{r|}{\textbf{CNN-kernel}} &
  \multicolumn{1}{c|}{6} &
  \multicolumn{1}{c|}{6} &
  \multicolumn{1}{c|}{6} \\ \cline{2-4} 
\multicolumn{1}{r|}{\textbf{criterion}} &
  \multicolumn{1}{c|}{MAE} &
  \multicolumn{1}{c|}{MAE} &
  \multicolumn{1}{c|}{MAE} \\ \cline{2-4} 
\multicolumn{1}{r|}{\textbf{dropout}} &
  \multicolumn{1}{c|}{0.2} &
  \multicolumn{1}{c|}{0.2} &
  \multicolumn{1}{c|}{0.2} \\ \cline{2-4} 
\multicolumn{1}{r|}{\textbf{early-stop}} &
  \multicolumn{1}{c|}{250} &
  \multicolumn{1}{c|}{100} &
  \multicolumn{1}{c|}{10} \\ \cline{2-4} 
\multicolumn{1}{r|}{\textbf{epochs}} &
  \multicolumn{1}{c|}{2500} &
  \multicolumn{1}{c|}{1000} &
  \multicolumn{1}{c|}{100} \\ \cline{2-4} 
\multicolumn{1}{r|}{\textbf{hidden-CNN}} &
  \multicolumn{1}{c|}{100} &
  \multicolumn{1}{c|}{100} &
  \multicolumn{1}{c|}{100} \\ \cline{2-4} 
\multicolumn{1}{r|}{\textbf{hidden-RNN}} &
  \multicolumn{1}{c|}{100} &
  \multicolumn{1}{c|}{100} &
  \multicolumn{1}{c|}{100} \\ \cline{2-4} 
\multicolumn{1}{r|}{\textbf{hidden-Skip}} &
  \multicolumn{1}{c|}{7} &
  \multicolumn{1}{c|}{84} &
  \multicolumn{1}{c|}{12} \\ \cline{2-4} 
\multicolumn{1}{r|}{\textbf{highway-window}} &
  \multicolumn{1}{c|}{7} &
  \multicolumn{1}{c|}{84} &
  \multicolumn{1}{c|}{12} \\ \cline{2-4} 
\multicolumn{1}{r|}{\textbf{horizon}} &
  \multicolumn{1}{c|}{1} &
  \multicolumn{1}{c|}{1} &
  \multicolumn{1}{c|}{1} \\ \cline{2-4} 
\multicolumn{1}{r|}{\textbf{learning-rate}} &
  \multicolumn{1}{c|}{0.001} &
  \multicolumn{1}{c|}{0.001} &
  \multicolumn{1}{c|}{0.001} \\ \cline{2-4} 
\multicolumn{1}{r|}{\textbf{normalization}} &
  \multicolumn{1}{c|}{2} &
  \multicolumn{1}{c|}{2} &
  \multicolumn{1}{c|}{2} \\ \cline{2-4} 
\multicolumn{1}{r|}{\textbf{optimizer}} &
  \multicolumn{1}{c|}{Adam} &
  \multicolumn{1}{c|}{Adam} &
  \multicolumn{1}{c|}{Adam} \\ \cline{2-4} 
\multicolumn{1}{r|}{\textbf{output-function}} &
  \multicolumn{1}{c|}{ReLU} &
  \multicolumn{1}{c|}{ReLU} &
  \multicolumn{1}{c|}{ReLU} \\ \cline{2-4} 
\multicolumn{1}{r|}{\textbf{seed}} &
  \multicolumn{1}{c|}{0} &
  \multicolumn{1}{c|}{0} &
  \multicolumn{1}{c|}{0} \\ \cline{2-4} 
\multicolumn{1}{r|}{\textbf{skip-steps}} &
  \multicolumn{1}{c|}{2} &
  \multicolumn{1}{c|}{2} &
  \multicolumn{1}{c|}{2} \\ \cline{2-4} 
\multicolumn{1}{r|}{\textbf{window}} &
  \multicolumn{1}{c|}{7} &
  \multicolumn{1}{c|}{84} &
  \multicolumn{1}{c|}{12} \\ \cline{2-4} 
\end{longtable}
\thispagestyle{empty}\clearpage
\loadgeometry{geom}

\clearpage\newpage%
\begin{titlepage}
    \vspace*{\fill}
        \begin{center}
          \Huge{\bf Hyperparameters\\\rule[5px]{2cm}{.1cm}~for the~\rule[5px]{2cm}{.1cm}\\Ablation Tests}
        \end{center}
    \vspace*{\fill}
\end{titlepage}

\thispagestyle{empty}\setcounter{page}{12}
\begin{table}
\centering
\caption
[Hyperparameters for the ablation tests.]%
{List of hyperparameters used during the ablation experiments.}
\label{tab:hyperparameters-2}
\begin{tabular}{r|c|c|c|}
\cline{2-4}
\multicolumn{1}{l|}{}        & \ReGENN & \textbf{PyTorch' Deafult} & \textbf{Literature's Default} \\ \cline{2-4} 
\textbf{clip-norm}           & ---    & 0                & 10                   \\ \cline{2-4} 
\textbf{dropout}             & ---    & 0                & 0.1                  \\ \cline{2-4} 
\textbf{learning-rate}       & ---    & 0.001            & 0.001                \\ \cline{2-4} 
\textbf{scheduler-factor}    & ---    & 0.1              & 0.95                 \\ \cline{2-4} 
\textbf{scheduler-patience}  & ---    & 10               & 25                   \\ \cline{2-4} 
\textbf{scheduler-threshold} & ---    & 0.001            & 0.1                   \\
\cline{2-4} 
\end{tabular}
\end{table}

\paragraph{Note.} The hyperparameters from above are shared across all the datasets; the other ones follow as in Tab.~\ref{tab:hyperparameters-1}.\clearpage

\clearpage\newpage%
\begin{titlepage}
    \vspace*{\fill}
        \begin{center}
          \Huge{\bf Detailed Result\\\rule[5px]{2cm}{.1cm}~on the~\rule[5px]{2cm}{.1cm}\\Main Experiments}
        \end{center}
    \vspace*{\fill}
\end{titlepage}

\savegeometry{geom}
\newgeometry{margin=0pt}
\thispagestyle{empty}\setcounter{page}{14}
\begin{landscape}
\begin{table}
\centering
\caption[Detailed baseline results.]%
{Detailed results for the main experiments.\\%
{\footnotesize\hspace{1.4cm}\textbf{Legend:} Algorithms with best performance are in \textbf{bold}, the ones noted as --- yielded exceptions, and others as *** were suppressed due to poor performance.}}
\label{tab:baselines-1}
\resizebox{1.35\textwidth}{!}{%
\begin{tabular}{r|cccccc|cccccc|cccccc|}
\cline{2-19}
\textbf{} &
  \multicolumn{6}{c|}{\textbf{SARS-CoV-2}} &
  \multicolumn{6}{c|}{\textbf{Brazilian Weather}} &
  \multicolumn{6}{c|}{\textbf{PhysioNet}} \\ \cline{2-19} 
 &
  \textbf{MAE} &
  \textbf{$\pm$ STD} &
  \textbf{RMSE} &
  \textbf{$\pm$ STD} &
  \textbf{MSLE} &
  \textbf{$\pm$ STD} &
  \textbf{MAE} &
  \textbf{$\pm$ STD} &
  \textbf{RMSE} &
  \textbf{$\pm$ STD} &
  \textbf{MSLE} &
  \textbf{$\pm$ STD} &
  \textbf{MAE} &
  \textbf{$\pm$ STD} &
  \textbf{RMSE} &
  \textbf{$\pm$ STD} &
  \textbf{MSLE} &
  \textbf{$\pm$ STD} \\ \cline{2-19} 
\multicolumn{1}{l|}{} &
  \multicolumn{18}{c|}{\textbf{Deep Learning Algorithms}} \\ \cline{2-19} 
\ReGENN &
  \textbf{165.41} &
  \textbf{30.37} &
  \textbf{915.92} &
  \textbf{294.06} &
  \textbf{0.05} &
  \textbf{0.02} &
  \textbf{1.92} &
  \textbf{0.02} &
  \textbf{4.86} &
  \textbf{0.13} &
  \textbf{0.28} &
  \textbf{0.01} &
  \textbf{18.22} &
  \textbf{1.04} &
  \textbf{47.31} &
  \textbf{6.27} &
  \textbf{1.37} &
  \textbf{0.12} \\
\textbf{MLCNN} &
  5298.30 &
  22152.36 &
  5779.21 &
  23677.93 &
  16.27 &
  32.19 &
  3.46 &
  3.32 &
  5.74 &
  4.25 &
  0.70 &
  1.52 &				
  20.81 &
  13.49 &
  43.23 &
  28.42 &
  3.32 &
  3.00 \\
\textbf{DSANet} &
  18428.63 &
  81222.22 &
  20131.28 &
  91468.66 &
  52.90 &
  36.86 &
  18.20 &
  2.42 &
  21.65 &
  2.67 &
  7.73 &
  0.79 &
  50.31 &
  17.32 &
  74.96 &
  28.57 &
  10.82 &
  3.39 \\
\textbf{LSTNet} &
  10749.65 &
  64733.89 &
  13172.62 &
  79729.95 &
  15.55 &
  23.33 &
  4.16 &
  5.48 &
  6.34 &
  6.21 &
  1.17 &
  2.47 &
  28.25 &
  18.15 &
  50.70 &
  31.89 &
  4.35 &
  4.04 \\ \cline{2-19} 
 &
  \multicolumn{18}{c|}{\textbf{Multi-Output and Multi-Task Algorithms}} \\ \cline{2-19} 
\textbf{CCA} &
  5,566.74 &
  618.65 &
  48,941.52 &
  3,308.99 &
  0.50 &
  0.22 &
  3.60 &
  0.86 &
  5.33 &
  1.85 &
  0.49 &
  0.71 &
  21.38 &
  20.70 &
  39.26 &
  34.60 &
  2.00 &
  1.27 \\
\textbf{Decision Tree} &
  1,098.77 &
  163.13 &
  6,055.55 &
  2,122.73 &
  0.27 &
  0.05 &
  3.47 &
  0.98 &
  6.05 &
  2.96 &
  0.47 &
  0.65 &
  25.25 &
  22.65 &
  48.81 &
  41.02 &
  3.19 &
  1.49 \\
\textbf{Extra Tree} &
  851.65 &
  12.55 &
  3,316.89 &
  276.68 &
  0.25 &
  0.05 &
  3.62 &
  0.74 &
  6.30 &
  2.42 &
  0.44 &
  0.58 &
  25.04 &
  22.69 &
  48.29 &
  40.68 &
  3.15 &
  1.54 \\
\textbf{Extra Trees} &
  771.33 &
  120.86 &
  3,823.39 &
  5.44 &
  0.14 &
  0.08 &
  \textbf{2.82} &
  \textbf{1.02} &
  \textbf{4.42} &
  \textbf{2.00} &
  \textbf{0.37} &
  \textbf{0.53} &
  \textbf{20.35} &
  \textbf{18.11} &
  \textbf{36.36} &
  \textbf{29.24} &
  \textbf{2.01} &
  \textbf{1.21} \\
\textbf{Gaussian Process} &
  2,730.59 &
  309.13 &
  14,295.71 &
  3,034.59 &
  0.57 &
  0.47 &
  15.51 &
  8.24 &
  17.51 &
  6.55 &
  4.69 &
  2.77 &
  132.00 &
  122.99 &
  681.47 &
  708.10 &
  5.00 &
  2.92 \\
\textbf{Historical Average} &
  2,204.09 &
  496.63 &
  9,540.29 &
  2,570.76 &
  0.16 &
  0.09 &
  3.20 &
  1.10 &
  4.84 &
  1.86 &
  0.32 &
  0.42 &
  29.58 &
  16.93 &
  42.34 &
  28.20 &
  2.24 &
  1.33 \\
\textbf{Kernel Ridge} &
  797.81 &
  78.97 &
  2,917.93 &
  162.53 &
  0.11 &
  0.04 &
  3.20 &
  1.12 &
  4.91 &
  2.13 &
  0.43 &
  0.61 &
  --- &
  --- &
  --- &
  --- &
  --- &
  --- \\
\textbf{KNeighbors} &
  967.37 &
  36.99 &
  5,414.50 &
  1,014.23 &
  0.13 &
  0.07 &
  3.09 &
  1.02 &
  4.84 &
  2.12 &
  0.46 &
  0.69 &
  21.33 &
  20.01 &
  39.04 &
  32.76 &
  2.19 &
  1.28 \\
\textbf{Lars} &
  *** &
  *** &
  *** &
  *** &
  *** &
  *** &
  *** &
  *** &
  *** &
  *** &
  *** &
  *** &
  20.48 &
  18.94 &
  36.77 &
  29.78 &
  2.03 &
  1.31 \\
\textbf{Lasso-Lars} &
  2,204.09 &
  496.63 &
  9,540.29 &
  2,570.76 &
  0.16 &
  0.09 &
  3.20 &
  1.10 &
  4.84 &
  1.86 &
  0.32 &
  0.42 &
  29.58 &
  16.93 &
  42.34 &
  28.20 &
  2.24 &
  1.33 \\
\textbf{Linear Regression} &
  *** &
  *** &
  *** &
  *** &
  *** &
  *** &
  3.59 &
  1.14 &
  5.42 &
  2.19 &
  0.51 &
  0.74 &
  20.47 &
  18.94 &
  36.77 &
  29.78 &
  2.03 &
  1.31 \\
\textbf{Multi-Task Elastic-Net} &
  2,204.09 &
  496.63 &
  9,540.29 &
  2,570.76 &
  0.16 &
  0.09 &
  3.20 &
  1.10 &
  4.84 &
  1.86 &
  0.32 &
  0.42 &
  29.25 &
  17.04 &
  41.98 &
  28.35 &
  2.24 &
  1.33 \\
\textbf{Multi-Task Lasso} &
  2,204.09 &
  496.63 &
  9,540.29 &
  2,570.76 &
  0.16 &
  0.09 &
  3.20 &
  1.10 &
  4.84 &
  1.86 &
  0.32 &
  0.42 &
  29.58 &
  16.93 &
  42.34 &
  28.20 &
  2.24 &
  1.33 \\
\textbf{Orthogonal Matching Pursuit} &
  2,142.70 &
  897.09 &
  19,881.75 &
  10,964.98 &
  0.17 &
  0.09 &
  5.79 &
  1.99 &
  9.86 &
  5.90 &
  0.83 &
  1.12 &
  20.89 &
  18.75 &
  37.23 &
  29.50 &
  2.03 &
  1.31 \\
\textbf{PLS Canonical} &
  8,743.59 &
  2,964.01 &
  84,481.55 &
  58,065.34 &
  0.39 &
  0.11 &
  6.01 &
  1.07 &
  8.33 &
  1.47 &
  0.76 &
  0.82 &
  26.31 &
  20.30 &
  46.43 &
  34.86 &
  2.41 &
  1.22 \\
\textbf{PLS} &
  1,341.96 &
  155.97 &
  6,461.25 &
  322.42 &
  0.50 &
  0.01 &
  2.89 &
  0.97 &
  4.51 &
  1.88 &
  0.38 &
  0.55 &
  21.04 &
  18.71 &
  36.75 &
  29.63 &
  2.07 &
  1.32 \\
\textbf{Radius Neighbors} &
  --- &
  --- &
  --- &
  --- &
  --- &
  --- &
  --- &
  --- &
  --- &
  --- &
  --- &
  --- &
  --- &
  --- &
  --- &
  --- &
  --- &
  --- \\
\textbf{Random Forest} &
  \textbf{736.11} &
  \textbf{53.37} &
  \textbf{3,623.38} &
  \textbf{690.48} &
  \textbf{0.14} &
  \textbf{0.07} &
  2.89 &
  1.02 &
  4.53 &
  2.00 &
  0.38 &
  0.55 &
  20.39 &
  18.08 &
  36.22 &
  29.51 &
  1.99 &
  1.21 \\
\textbf{RANSAC} &
  --- &
  --- &
  --- &
  --- &
  --- &
  --- &
  --- &
  --- &
  --- &
  --- &
  --- &
  --- &
  --- &
  --- &
  --- &
  --- &
  --- &
  --- \\
\textbf{Ridge} &
  804.49 &
  86.67 &
  2,957.27 &
  191.77 &
  0.14 &
  0.08 &
  3.13 &
  1.12 &
  4.86 &
  2.15 &
  0.45 &
  0.66 &
  20.47 &
  18.94 &
  36.77 &
  29.78 &
  2.03 &
  1.31 \\ \cline{2-19} 
 &
  \multicolumn{18}{c|}{\textbf{Single-Target Algorithms on Chain Ensemble}} \\ \cline{2-19} 
\textbf{AdaBoost} &
  1,099.84 &
  337.60 &
  7,184.23 &
  5,335.13 &
  0.15 &
  0.09 &
  2.82 &
  0.97 &
  4.51 &
  1.96 &
  0.34 &
  0.47 &
  --- &
  --- &
  --- &
  --- &
  --- &
  --- \\
\textbf{ARD} &
  889.47 &
  30.26 &
  3,368.50 &
  207.97 &
  0.16 &
  0.08 &
  4.35 &
  1.42 &
  6.65 &
  2.97 &
  0.63 &
  0.91 &
  --- &
  --- &
  --- &
  --- &
  --- &
  --- \\
\textbf{Bagging} &
  862.94 &
  115.31 &
  4,244.29 &
  116.49 &
  0.17 &
  0.06 &
  3.05 &
  1.12 &
  4.73 &
  2.29 &
  0.44 &
  0.66 &
  21.28 &
  18.96 &
  37.78 &
  31.30 &
  2.02 &
  1.24 \\
\textbf{Bayesian Ridge} &
  828.14 &
  12.06 &
  3,031.29 &
  145.24 &
  0.16 &
  0.06 &
  3.14 &
  1.06 &
  4.88 &
  1.96 &
  0.40 &
  0.57 &
  20.47 &
  18.93 &
  36.74 &
  29.74 &
  2.03 &
  1.31 \\
\textbf{CatBoost} &
  1,115.88 &
  85.29 &
  4,129.19 &
  139.31 &
  0.14 &
  0.08 &
  3.40 &
  0.73 &
  4.97 &
  1.57 &
  0.35 &
  0.50 &
  19.95 &
  18.03 &
  35.94 &
  29.16 &
  1.98 &
  1.19 \\
\textbf{Gradient Boosting} &
  \textbf{800.29} &
  \textbf{1.54} &
  \textbf{4,722.74} &
  \textbf{1,271.82} &
  \textbf{0.16} &
  \textbf{0.06} &
  3.05 &
  1.13 &
  4.84 &
  2.44 &
  0.45 &
  0.66 &
  --- &
  --- &
  --- &
  --- &
  --- &
  --- \\
\textbf{Histogram Grad. Boosting} &
  866.30 &
  38.80 &
  4,612.11 &
  44.68 &
  0.16 &
  0.08 &
  3.09 &
  1.15 &
  4.75 &
  2.28 &
  0.47 &
  0.70 &
  20.90 &
  18.49 &
  37.57 &
  30.70 &
  2.00 &
  1.19 \\
\textbf{Huber} &
  2,032.72 &
  41.78 &
  17,770.19 &
  610.28 &
  0.37 &
  0.08 &
  3.43 &
  1.09 &
  5.21 &
  2.13 &
  0.48 &
  0.70 &
  19.87 &
  18.04 &
  38.10 &
  28.49 &
  2.14 &
  1.19 \\
\textbf{Isotonic} &
  --- &
  --- &
  --- &
  --- &
  --- &
  --- &
  --- &
  --- &
  --- &
  --- &
  --- &
  --- &
  --- &
  --- &
  --- &
  --- &
  --- &
  --- \\
\textbf{Lasso-Lars-IC} &
  810.03 &
  30.97 &
  3,929.81 &
  129.59 &
  0.14 &
  0.08 &
  2.91 &
  1.07 &
  4.59 &
  1.94 &
  0.31 &
  0.42 &
  20.46 &
  18.94 &
  36.72 &
  29.75 &
  2.03 &
  1.32 \\
\textbf{LGBM} &
  1,110.17 &
  198.23 &
  7,650.02 &
  3,822.07 &
  0.28 &
  0.01 &
  --- &
  --- &
  --- &
  --- &
  --- &
  --- &
  20.12 &
  18.14 &
  36.30 &
  29.41 &
  1.98 &
  1.19 \\
\textbf{Linear SVR} &
  875.51 &
  97.79 &
  3,540.90 &
  247.08 &
  0.16 &
  0.01 &
  3.44 &
  1.19 &
  5.21 &
  2.30 &
  0.50 &
  0.73 &
  \textbf{19.72} &
  \textbf{17.71} &
  \textbf{38.65} &
  \textbf{29.06} &
  \textbf{2.18} &
  \textbf{1.08} \\
\textbf{NuSVR} &
  809.97 &
  19.17 &
  2,973.63 &
  21.61 &
  0.12 &
  0.06 &
  \textbf{2.79} &
  \textbf{0.97} &
  \textbf{4.40} &
  \textbf{1.96} &
  \textbf{0.34} &
  \textbf{0.49} &
  --- &
  --- &
  --- &
  --- &
  --- &
  --- \\
\textbf{Passive Aggressive} &
  1,730.12 &
  113.18 &
  8,310.46 &
  432.48 &
  0.18 &
  0.12 &
  3.22 &
  0.89 &
  4.82 &
  1.98 &
  0.46 &
  0.69 &
  31.27 &
  31.35 &
  55.82 &
  58.27 &
  2.76 &
  1.84 \\
\textbf{SGD} &
  815.97 &
  11.37 &
  3,106.47 &
  40.02 &
  0.13 &
  0.06 &
  *** &
  *** &
  *** &
  *** &
  *** &
  *** &
  20.53 &
  19.00 &
  36.78 &
  29.88 &
  2.04 &
  1.32 \\
\textbf{SVR} &
  1,344.80 &
  244.56 &
  5,511.70 &
  1,539.93 &
  0.12 &
  0.06 &
  3.38 &
  1.20 &
  4.63 &
  2.05 &
  0.56 &
  0.86 &
  --- &
  --- &
  --- &
  --- &
  --- &
  --- \\
\textbf{Theil-Sen} &
  91,149.14 &
  59,396.23 &
  1,292,430.04 &
  858,201.91 &
  6.45 &
  2.47 &
  3.65 &
  1.22 &
  5.44 &
  2.21 &
  0.51 &
  0.73 &
  20.61 &
  19.56 &
  37.32 &
  31.23 &
  2.05 &
  1.31 \\
\textbf{Transformed Target} &
  68,530.56 &
  55,751.11 &
  868,586.61 &
  989,030.54 &
  16.39 &
  1.00 &
  3.58 &
  1.13 &
  5.42 &
  2.19 &
  0.51 &
  0.74 &
  20.47 &
  18.94 &
  36.77 &
  29.78 &
  2.03 &
  1.31 \\
\textbf{XGBoost} &
  806.03 &
  37.02 &
  5,033.10 &
  1,599.32 &
  0.16 &
  0.06 &
  3.04 &
  1.12 &
  4.81 &
  2.41 &
  0.44 &
  0.66 &
  19.94 &
  18.05 &
  35.92 &
  29.08 &
  1.98 &
  1.19 \\ \cline{2-19} 
 &
  \multicolumn{18}{c|}{\textbf{Time Series Algorithms}} \\ \cline{2-19} 
\textbf{Autoregressive} &
  11,529.79 &
  6,752.42 &
  76,118.62 &
  48,374.82 &
  9.15 &
  6.14 &
  2.69 &
  1.44 &
  4.68 &
  1.73 &
  0.37 &
  0.54 &
  24.44 &
  9.68 &
  53.92 &
  16.37 &
  2.08 &
  0.95 \\
\textbf{ARIMA} &
  11,578.80 &
  6,798.84 &
  76,044.64 &
  48,478.91 &
  9.22 &
  5.23 &
  2.68 &
  1.49 &
  4.81 &
  1.78 &
  0.34 &
  0.49 &
  *** &
  *** &
  *** &
  *** &
  *** &
  *** \\
\textbf{ARMA} &
  12,578.49 &
  6,091.86 &
  79,263.71 &
  45,555.46 &
  16.71 &
  4.79 &
  3.21 &
  0.82 &
  6.22 &
  1.71 &
  0.58 &
  0.47 &
  182.01 &
  189.23 &
  20,124.41 &
  29,967.38 &
  2.61 &
  0.87 \\
\textbf{Moving Average} &
  7,757.33 &
  2,887.33 &
  73,013.22 &
  42,555.86 &
  2.54 &
  1.17 &
  2.70 &
  1.48 &
  4.49 &
  1.75 &
  0.47 &
  0.73 &
  --- &
  --- &
  --- &
  --- &
  --- &
  --- \\
\textbf{SARIMA} &
  *** &
  *** &
  *** &
  *** &
  *** &
  *** &
  2.68 &
  1.50 &
  4.82 &
  1.78 &
  0.34 &
  0.49 &
  25.05 &
  9.48 &
  55.89 &
  16.04 &
  2.18 &
  0.93 \\
\textbf{Exponential Smoothing} &
  \textbf{667.33} &
  \textbf{940.85} &
  \textbf{2,607.29} &
  \textbf{3,634.47} &
  \textbf{0.27} &
  \textbf{0.17} &
  \textbf{2.66} &
  \textbf{1.47} &
  \textbf{4.78} &
  \textbf{1.78} &
  \textbf{0.34} &
  \textbf{0.48} &
  \textbf{23.77} &
  \textbf{9.48} &
  \textbf{52.89} &
  \textbf{15.52} &
  \textbf{2.09} &
  \textbf{0.92} \\
\textbf{Vector Autoregression} &
  6,395.14 &
  2,397.58 &
  75,292.43 &
  39,167.56 &
  3.21 &
  0.05 &
  2.70 &
  1.51 &
  4.46 &
  1.76 &
  0.49 &
  0.76 &
  *** &
  *** &
  *** &
  *** &
  *** &
  *** \\ \cline{2-19} 
\end{tabular}%
}
\end{table}
\end{landscape}\clearpage
\loadgeometry{geom}

\savegeometry{geom}
\newgeometry{margin=0pt}
\thispagestyle{empty}\setcounter{page}{15}
\begin{landscape}
\begin{table}
\centering
\caption[Ablation with PyTorch's Hyperparameters.]%
{Ablation on the main experiments with PyTorch's hyperparameters.\\%
{\footnotesize\textbf{Legend:} Algorithms with the best performance are in \textbf{bold}; besides, we refer to the Transformer Encoder as $\mathrm{E}$, recurrent unit as \textsc{RU}, Bidirectional as \textsc{B}, Unidirectional as \textsc{U}, and Autoregressive as \textsc{AR}.}}
\label{tab:ablation-1}
\resizebox{1.35\textwidth}{!}{%
\begin{tabular}{rr|cccccc|cccccc|cccccc|}
\cline{3-20}
 &
   &
  \multicolumn{6}{c|}{\textbf{SARS-CoV-2}} &
  \multicolumn{6}{c|}{\textbf{Brazilian Weather}} &
  \multicolumn{6}{c|}{\textbf{PhysioNet}} \\ \cline{3-20} 
 &
   &
  \textbf{MAE} &
  \textbf{$\pm$ STD} &
  \textbf{RMSE} &
  \textbf{$\pm$ STD} &
  \textbf{MSLE} &
  \textbf{$\pm$ STD} &
  \textbf{MAE} &
  \textbf{$\pm$ STD} &
  \textbf{RMSE} &
  \textbf{$\pm$ STD} &
  \textbf{MSLE} &
  \textbf{$\pm$ STD} &
  \textbf{MAE} &
  \textbf{$\pm$ STD} &
  \textbf{RMSE} &
  \textbf{$\pm$ STD} &
  \textbf{MSLE} &
  \textbf{$\pm$ STD} \\ \cline{3-20} 
 & 
  &
  \multicolumn{18}{c|}{\textbf{Recurrent Neural Network (RNN)}} \\ \cline{3-20} 
\multicolumn{2}{r|}{${}_{\mathrm{B}}\text{\textsc{RU}}$} &
  434.18 &
  250.54 &
  2678.49 &
  2046.82 &
  0.09 &
  0.05 &
  2.03 &
  0.03 &
  4.91 &
  0.10 &
  0.28 &
  0.01 &
  19.94 &
  1.18 &
  50.34 &
  7.73 &
  1.39 &
  0.13 \\
\multicolumn{2}{r|}{$\mathrm{E} \rightarrow {}_{\mathrm{B}}\text{\textsc{RU}}$} &
  3680.89 &
  2404.90 &
  25773.58 &
  22768.94 &
  7.47 &
  0.80 &
  2.84 &
  0.09 &
  6.49 &
  0.16 &
  0.55 &
  0.02 &
  19.01 &
  0.97 &
  48.70 &
  6.16 &
  1.42 &
  0.13 \\
\multicolumn{2}{r|}{$\left(\mathrm{E} \rightarrow {}_{\mathrm{B}}\text{\textsc{RU}} + {}_{\mathrm{B}}\text{\textsc{RU}}\right) + \mathrm{AR}$} &
  1533.62 &
  949.53 &
  18013.32 &
  16584.35 &
  3.82 &
  0.16 &
  2.44 &
  0.06 &
  6.19 &
  0.10 &
  0.51 &
  0.02 &
  \textbf{18.68} &
  \textbf{1.06} &
  \textbf{48.02} &
  \textbf{5.70} &
  \textbf{1.38} &
  \textbf{0.13} \\
\multicolumn{2}{r|}{$\left(\mathrm{E} \rightarrow {}_{\mathrm{B}}\text{\textsc{RU}} + {}_{\mathrm{U}}\text{\textsc{RU}}\right) + \mathrm{AR}$} &
  \textbf{245.06} &
  \textbf{65.03} &
  \textbf{1359.34} &
  \textbf{462.61} &
  \textbf{0.09} &
  \textbf{0.05} &
  2.71 &
  0.05 &
  6.42 &
  0.13 &
  0.54 &
  0.02 &
  18.81 &
  1.17 &
  48.92 &
  7.67 &
  1.39 &
  0.12 \\
\multicolumn{2}{r|}{$\left(\mathrm{E} \rightarrow {}_{\mathrm{B}}\text{\textsc{RU}}\right) + \mathrm{AR}$} &
  1386.30 &
  859.67 &
  16848.20 &
  15758.33 &
  3.72 &
  0.45 &
  4.56 &
  0.08 &
  9.81 &
  0.11 &
  1.48 &
  0.03 &
  18.86 &
  1.13 &
  49.54 &
  7.22 &
  1.40 &
  0.13 \\
\multicolumn{2}{r|}{$\mathrm{E} \rightarrow {}_{\mathrm{U}}\text{\textsc{RU}}$} &
  3171.99 &
  1445.00 &
  21117.23 &
  16502.10 &
  3.92 &
  0.10 &
  2.37 &
  0.11 &
  5.35 &
  0.22 &
  0.33 &
  0.02 &
  19.25 &
  0.97 &
  49.18 &
  6.65 &
  1.45 &
  0.12 \\
\multicolumn{2}{r|}{$\left(\mathrm{E} \rightarrow {}_{\mathrm{U}}\text{\textsc{RU}} + {}_{\mathrm{B}}\text{\textsc{RU}}\right) + \mathrm{AR}$} &
  350.75 &
  202.78 &
  1985.39 &
  1228.78 &
  0.11 &
  0.03 &
  2.22 &
  0.13 &
  5.61 &
  0.53 &
  0.41 &
  0.03 &
  18.90 &
  1.26 &
  48.77 &
  7.38 &
  1.41 &
  0.15 \\
\multicolumn{2}{r|}{$\left(\mathrm{E} \rightarrow {}_{\mathrm{U}}\text{\textsc{RU}} + {}_{\mathrm{U}}\text{\textsc{RU}}\right) + \mathrm{AR}$} &
  1386.73 &
  641.20 &
  17459.45 &
  14421.01 &
  3.70 &
  0.30 &
  \textbf{1.97} &
  \textbf{0.10} &
  \textbf{4.84} &
  \textbf{0.38} &
  \textbf{0.27} &
  \textbf{0.02} &
  19.07 &
  1.34 &
  48.89 &
  7.24 &
  1.41 &
  0.14 \\
\multicolumn{2}{r|}{$\left(\mathrm{E} \rightarrow {}_{\mathrm{U}}\text{\textsc{RU}}\right) + \mathrm{AR}$} &
  1375.60 &
  988.31 &
  15987.09 &
  16378.54 &
  3.69 &
  0.34 &
  5.07 &
  0.04 &
  10.51 &
  0.04 &
  1.73 &
  0.01 &
  19.04 &
  1.07 &
  50.05 &
  7.24 &
  1.45 &
  0.13 \\
\multicolumn{2}{r|}{${}_{\mathrm{U}}\text{\textsc{RU}}$} &
  1277.24 &
  1053.33 &
  7948.96 &
  7430.65 &
  2.89 &
  0.96 &
  2.13 &
  0.04 &
  5.13 &
  0.05 &
  0.32 &
  0.01 &
  20.52 &
  1.16 &
  49.72 &
  7.29 &
  1.43 &
  0.12 \\ \cline{3-20} 
 &
   &
  \multicolumn{18}{c|}{ \textbf{Gated Recurrent Unit (GRU)}} \\ \cline{3-20} 
\multicolumn{2}{r|}{${}_{\mathrm{B}}\text{\textsc{RU}}$} &
  3167.33 &
  1458.27 &
  28834.18 &
  23983.55 &
  7.80 &
  0.49 &
  2.00 &
  0.09 &
  4.89 &
  0.34 &
  0.27 &
  0.03 &
  24.88 &
  1.04 &
  56.74 &
  6.09 &
  3.01 &
  0.12 \\
\multicolumn{2}{r|}{$\mathrm{E} \rightarrow {}_{\mathrm{B}}\text{\textsc{RU}}$} &
  950.11 &
  433.35 &
  5133.01 &
  2301.34 &
  0.19 &
  0.14 &
  2.35 &
  0.08 &
  5.34 &
  0.21 &
  0.33 &
  0.02 &
  18.93 &
  1.08 &
  49.30 &
  7.19 &
  1.40 &
  0.12 \\
\multicolumn{2}{r|}{$\left(\mathrm{E} \rightarrow {}_{\mathrm{B}}\text{\textsc{RU}} + {}_{\mathrm{B}}\text{\textsc{RU}}\right) + \mathrm{AR}$} &
  311.09 &
  119.17 &
  1873.65 &
  1082.02 &
  0.10 &
  0.07 &
  2.29 &
  0.05 &
  5.68 &
  0.09 &
  0.41 &
  0.01 &
  18.58 &
  1.09 &
  47.94 &
  6.01 &
  1.38 &
  0.13 \\
\multicolumn{2}{r|}{$\left(\mathrm{E} \rightarrow {}_{\mathrm{B}}\text{\textsc{RU}} + {}_{\mathrm{U}}\text{\textsc{RU}}\right) + \mathrm{AR}$} &
  650.06 &
  185.77 &
  3443.73 &
  1270.04 &
  0.10 &
  0.04 &
  2.23 &
  0.05 &
  5.61 &
  0.11 &
  0.41 &
  0.01 &
  18.60 &
  1.07 &
  48.55 &
  5.86 &
  1.38 &
  0.14 \\
\multicolumn{2}{r|}{$\left(\mathrm{E} \rightarrow {}_{\mathrm{B}}\text{\textsc{RU}}\right) + \mathrm{AR}$} &
  1377.91 &
  517.07 &
  17145.02 &
  14067.11 &
  3.66 &
  0.10 &
  4.85 &
  0.12 &
  10.14 &
  0.25 &
  1.60 &
  0.03 &
  \textbf{18.56} &
  \textbf{1.05} &
  \textbf{48.54} &
  \textbf{6.96} &
  \textbf{1.39} &
  \textbf{0.11} \\
\multicolumn{2}{r|}{$\mathrm{E} \rightarrow {}_{\mathrm{U}}\text{\textsc{RU}}$} &
  1360.35 &
  737.84 &
  7611.46 &
  4036.40 &
  0.27 &
  0.11 &
  2.20 &
  0.06 &
  5.11 &
  0.22 &
  0.28 &
  0.02 &
  19.40 &
  1.05 &
  50.73 &
  8.38 &
  1.43 &
  0.13 \\
\multicolumn{2}{r|}{$\left(\mathrm{E} \rightarrow {}_{\mathrm{U}}\text{\textsc{RU}} + {}_{\mathrm{B}}\text{\textsc{RU}}\right) + \mathrm{AR}$} &
  \textbf{240.35} &
  \textbf{57.07} &
  \textbf{1335.96} &
  \textbf{343.59} &
  \textbf{0.08} &
  \textbf{0.04} &
  \textbf{1.98} &
  \textbf{0.11} &
  \textbf{4.87} &
  \textbf{0.51} &
  \textbf{0.28} &
  \textbf{0.03} &
  18.70 &
  1.06 &
  48.74 &
  6.14 &
  1.39 &
  0.10 \\
\multicolumn{2}{r|}{$\left(\mathrm{E} \rightarrow {}_{\mathrm{U}}\text{\textsc{RU}} + {}_{\mathrm{U}}\text{\textsc{RU}}\right) + \mathrm{AR}$} &
  3377.44 &
  1741.87 &
  25335.28 &
  22336.83 &
  7.44 &
  1.07 &
  2.56 &
  0.15 &
  6.30 &
  0.21 &
  0.54 &
  0.02 &
  18.79 &
  0.97 &
  49.56 &
  8.40 &
  1.37 &
  0.12 \\
\multicolumn{2}{r|}{$\left(\mathrm{E} \rightarrow {}_{\mathrm{U}}\text{\textsc{RU}}\right) + \mathrm{AR}$} &
  1592.62 &
  895.15 &
  18207.40 &
  15612.58 &
  3.77 &
  0.39 &
  4.95 &
  0.07 &
  10.32 &
  0.09 &
  1.62 &
  0.02 &
  18.88 &
  0.98 &
  49.72 &
  8.45 &
  1.40 &
  0.11 \\
\multicolumn{2}{r|}{${}_{\mathrm{U}}\text{\textsc{RU}}$} &
  2319.57 &
  1793.72 &
  21536.86 &
  21418.12 &
  6.26 &
  0.55 &
  3.05 &
  0.10 &
  7.24 &
  0.17 &
  0.73 &
  0.02 &
  30.06 &
  0.89 &
  62.71 &
  5.67 &
  4.54 &
  0.09 \\ \cline{3-20} 
 &
   &
  \multicolumn{18}{c|}{\textbf{Long Short-Term Memory (LSTM)}} \\ \cline{3-20} 
\multicolumn{2}{r|}{${}_{\mathrm{B}}\text{\textsc{RU}}$} &
  3000.14 &
  1954.01 &
  25989.65 &
  24313.81 &
  7.65 &
  0.78 &
  \textbf{2.31} &
  \textbf{0.04} &
  \textbf{5.68} &
  \textbf{0.04} &
  \textbf{0.41} &
  \textbf{0.01} &
  19.57 &
  1.23 &
  50.33 &
  7.28 &
  1.40 &
  0.11 \\
\multicolumn{2}{r|}{$\mathrm{E} \rightarrow {}_{\mathrm{B}}\text{\textsc{RU}}$} &
  619.31 &
  356.13 &
  3094.25 &
  1878.42 &
  0.17 &
  0.11 &
  2.32 &
  0.08 &
  5.20 &
  0.27 &
  0.29 &
  0.03 &
  19.08 &
  1.15 &
  49.08 &
  7.15 &
  1.41 &
  0.14 \\
\multicolumn{2}{r|}{$\left(\mathrm{E} \rightarrow {}_{\mathrm{B}}\text{\textsc{RU}} + {}_{\mathrm{B}}\text{\textsc{RU}}\right) + \mathrm{AR}$} &
  \textbf{216.01} &
  \textbf{89.11} &
  \textbf{1146.94} &
  \textbf{465.90} &
  \textbf{0.10} &
  \textbf{0.07} &
  2.52 &
  0.05 &
  6.28 &
  0.18 &
  0.54 &
  0.01 &
  18.66 &
  0.97 &
  49.47 &
  8.82 &
  1.37 &
  0.12 \\
\multicolumn{2}{r|}{$\left(\mathrm{E} \rightarrow {}_{\mathrm{B}}\text{\textsc{RU}} + {}_{\mathrm{U}}\text{\textsc{RU}}\right) + \mathrm{AR}$} &
  2863.94 &
  1345.36 &
  27905.87 &
  22477.61 &
  7.69 &
  0.62 &
  3.68 &
  0.10 &
  8.52 &
  0.21 &
  1.07 &
  0.02 &
  \textbf{18.55} &
  \textbf{0.99} &
  \textbf{49.12} &
  \textbf{6.05} &
  \textbf{1.37} &
  \textbf{0.14} \\
\multicolumn{2}{r|}{$\left(\mathrm{E} \rightarrow {}_{\mathrm{B}}\text{\textsc{RU}}\right) + \mathrm{AR}$} &
  4060.62 &
  2655.67 &
  31538.83 &
  28960.85 &
  11.35 &
  0.99 &
  3.73 &
  0.07 &
  8.53 &
  0.15 &
  1.08 &
  0.02 &
  18.64 &
  1.09 &
  48.76 &
  5.96 &
  1.39 &
  0.12 \\
\multicolumn{2}{r|}{$\mathrm{E} \rightarrow {}_{\mathrm{U}}\text{\textsc{RU}}$} &
  3199.26 &
  2436.81 &
  12581.64 &
  11769.69 &
  0.26 &
  0.07 &
  2.47 &
  0.06 &
  5.10 &
  0.21 &
  0.28 &
  0.02 &
  19.78 &
  1.15 &
  50.21 &
  7.31 &
  1.42 &
  0.13 \\
\multicolumn{2}{r|}{$\left(\mathrm{E} \rightarrow {}_{\mathrm{U}}\text{\textsc{RU}} + {}_{\mathrm{B}}\text{\textsc{RU}}\right) + \mathrm{AR}$} &
  1351.42 &
  644.34 &
  17065.24 &
  14155.37 &
  3.67 &
  0.30 &
  3.38 &
  0.05 &
  7.89 &
  0.06 &
  0.93 &
  0.03 &
  18.69 &
  1.28 &
  49.27 &
  8.58 &
  1.38 &
  0.12 \\
\multicolumn{2}{r|}{$\left(\mathrm{E} \rightarrow {}_{\mathrm{U}}\text{\textsc{RU}} + {}_{\mathrm{U}}\text{\textsc{RU}}\right) + \mathrm{AR}$} &
  1431.78 &
  946.29 &
  17235.49 &
  16431.19 &
  3.79 &
  0.38 &
  5.16 &
  0.08 &
  10.61 &
  0.08 &
  1.74 &
  0.02 &
  18.70 &
  1.13 &
  49.85 &
  6.90 &
  1.38 &
  0.11 \\
\multicolumn{2}{r|}{$\left(\mathrm{E} \rightarrow {}_{\mathrm{U}}\text{\textsc{RU}}\right) + \mathrm{AR}$} &
  222.06 &
  85.96 &
  1265.74 &
  642.42 &
  0.08 &
  0.03 &
  7.48 &
  0.13 &
  13.34 &
  0.14 &
  2.81 &
  0.04 &
  18.86 &
  1.11 &
  49.88 &
  7.80 &
  1.41 &
  0.12 \\
\multicolumn{2}{r|}{${}_{\mathrm{U}}\text{\textsc{RU}}$} &
  4943.62 &
  2892.09 &
  28696.47 &
  24921.66 &
  7.59 &
  0.30 &
  4.80 &
  0.02 &
  9.74 &
  0.08 &
  1.46 &
  0.02 &
  20.80 &
  0.95 &
  50.13 &
  7.28 &
  1.37 &
  0.12 \\ \cline{3-20} 
\end{tabular}%
}
\end{table}
\end{landscape}\clearpage
\loadgeometry{geom}

\savegeometry{geom}
\newgeometry{margin=0pt}
\thispagestyle{empty}\setcounter{page}{16}
\begin{landscape}
\begin{table}
\centering
\caption[Ablation with Literature's Hyperparameters.]%
{Ablation on the main experiments with hyperparameters commonly used in the related literature.\\%
{\footnotesize\textbf{Legend:} Algorithms with the best performance are in \textbf{bold}; besides, we refer to the Transformer Encoder as $\mathrm{E}$, recurrent unit as \textsc{RU}, Bidirectional as \textsc{B}, Unidirectional as \textsc{U}, and Autoregressive as \textsc{AR}.}}
\label{tab:ablation-2}
\resizebox{1.35\textwidth}{!}{%
\begin{tabular}{rr|cccccc|cccccc|cccccc|}
\cline{3-20}
 &
   &
  \multicolumn{6}{c|}{\textbf{SARS-CoV-2}} &
  \multicolumn{6}{c|}{\textbf{Brazilian Weather}} &
  \multicolumn{6}{c|}{\textbf{PhysioNet}} \\ \cline{3-20} 
 &
   &
  \textbf{MAE} &
  \textbf{$\pm$ STD} &
  \textbf{RMSE} &
  \textbf{$\pm$ STD} &
  \textbf{MSLE} &
  \textbf{$\pm$ STD} &
  \textbf{MAE} &
  \textbf{$\pm$ STD} &
  \textbf{RMSE} &
  \textbf{$\pm$ STD} &
  \textbf{MSLE} &
  \textbf{$\pm$ STD} &
  \textbf{MAE} &
  \textbf{$\pm$ STD} &
  \textbf{RMSE} &
  \textbf{$\pm$ STD} &
  \textbf{MSLE} &
  \textbf{$\pm$ STD} \\ \cline{3-20} 
 &
   &
  \multicolumn{18}{c|}{\textbf{Recurrent Neural Network (RNN)}} \\ \cline{3-20} 
\multicolumn{2}{r|}{${}_{\mathrm{B}}\text{\textsc{RU}}$} &
  1,895.13 &
  995.98 &
  7,184.26 &
  5,885.75 &
  0.08 &
  0.03 &
  3.34 &
  0.11 &
  5.49 &
  0.42 &
  0.29 &
  0.02 &
  20.67 &
  0.93 &
  48.09 &
  6.89 &
  1.37 &
  0.11 \\
\multicolumn{2}{r|}{$\mathrm{E} \rightarrow {}_{\mathrm{B}}\text{\textsc{RU}}$} &
  2,510.11 &
  1,324.54 &
  9,521.55 &
  6,720.47 &
  0.19 &
  0.05 &
  4.12 &
  0.21 &
  6.40 &
  0.29 &
  0.40 &
  0.06 &
  19.83 &
  1.04 &
  48.97 &
  6.18 &
  1.44 &
  0.14 \\
\multicolumn{2}{r|}{$\left(\mathrm{E} \rightarrow {}_{\mathrm{B}}\text{\textsc{RU}} + {}_{\mathrm{B}}\text{\textsc{RU}}\right) + \mathrm{AR}$} &
  317.13 &
  184.28 &
  1,806.87 &
  1,269.50 &
  0.10 &
  0.07 &
  2.16 &
  0.11 &
  5.01 &
  0.20 &
  0.28 &
  0.02 &
  18.84 &
  1.14 &
  49.50 &
  6.27 &
  1.37 &
  0.12 \\
\multicolumn{2}{r|}{$\left(\mathrm{E} \rightarrow {}_{\mathrm{B}}\text{\textsc{RU}} + {}_{\mathrm{U}}\text{\textsc{RU}}\right) + \mathrm{AR}$} &
  279.38 &
  115.14 &
  1,569.99 &
  836.28 &
  0.09 &
  0.07 &
  \textbf{2.07} &
  \textbf{0.07} &
  \textbf{4.95} &
  \textbf{0.29} &
  \textbf{0.28} &
  \textbf{0.02} &
  \textbf{18.77} &
  \textbf{1.07} &
  \textbf{48.89} &
  \textbf{6.17} &
  \textbf{1.38} &
  \textbf{0.14} \\
\multicolumn{2}{r|}{$\left(\mathrm{E} \rightarrow {}_{\mathrm{B}}\text{\textsc{RU}}\right) + \mathrm{AR}$} &
  304.52 &
  141.03 &
  2,029.00 &
  1,595.46 &
  0.10 &
  0.04 &
  3.53 &
  0.12 &
  8.05 &
  0.29 &
  0.94 &
  0.04 &
  18.88 &
  1.04 &
  49.21 &
  6.81 &
  1.40 &
  0.13 \\
\multicolumn{2}{r|}{$\mathrm{E} \rightarrow {}_{\mathrm{U}}\text{\textsc{RU}}$} &
  2,692.19 &
  2,133.65 &
  8,934.01 &
  8,459.59 &
  0.29 &
  0.18 &
  3.94 &
  0.21 &
  6.26 &
  0.45 &
  0.35 &
  0.08 &
  20.04 &
  1.09 &
  47.82 &
  7.02 &
  1.45 &
  0.13 \\
\multicolumn{2}{r|}{$\left(\mathrm{E} \rightarrow {}_{\mathrm{U}}\text{\textsc{RU}} + {}_{\mathrm{B}}\text{\textsc{RU}}\right) + \mathrm{AR}$} &
  280.09 &
  142.50 &
  1,669.51 &
  1,096.35 &
  0.10 &
  0.03 &
  2.79 &
  0.11 &
  6.46 &
  0.05 &
  0.55 &
  0.03 &
  18.96 &
  1.00 &
  48.06 &
  6.35 &
  1.42 &
  0.09 \\
\multicolumn{2}{r|}{$\left(\mathrm{E} \rightarrow {}_{\mathrm{U}}\text{\textsc{RU}} + {}_{\mathrm{U}}\text{\textsc{RU}}\right) + \mathrm{AR}$} &
  1,423.25 &
  738.45 &
  17,459.46 &
  14,825.39 &
  3.72 &
  0.39 &
  2.16 &
  0.15 &
  4.98 &
  0.45 &
  0.28 &
  0.04 &
  18.99 &
  0.94 &
  49.17 &
  6.33 &
  1.42 &
  0.11 \\
\multicolumn{2}{r|}{$\left(\mathrm{E} \rightarrow {}_{\mathrm{U}}\text{\textsc{RU}}\right) + \mathrm{AR}$} &
  \textbf{240.34} &
  \textbf{93.62} &
  \textbf{1,386.81} &
  \textbf{832.15} &
  \textbf{0.10} &
  \textbf{0.06} &
  4.60 &
  0.05 &
  9.78 &
  0.14 &
  1.47 &
  0.02 &
  19.11 &
  1.18 &
  50.14 &
  6.98 &
  1.46 &
  0.11 \\
\multicolumn{2}{r|}{${}_{\mathrm{U}}\text{\textsc{RU}}$} &
  1,954.90 &
  1,154.82 &
  7,310.14 &
  6,297.24 &
  0.21 &
  0.13 &
  3.31 &
  0.08 &
  5.53 &
  0.29 &
  0.29 &
  0.02 &
  20.93 &
  1.15 &
  48.50 &
  6.71 &
  1.42 &
  0.12 \\ \cline{3-20} 
 &
   &
  \multicolumn{18}{c|}{\textbf{Gated Recurrent Unit (GRU)}} \\ \cline{3-20} 
\multicolumn{2}{r|}{${}_{\mathrm{B}}\text{\textsc{RU}}$} &
  3,094.99 &
  2,293.39 &
  18,854.48 &
  19,286.26 &
  3.89 &
  0.31 &
  3.56 &
  0.05 &
  5.93 &
  0.13 &
  0.34 &
  0.01 &
  30.27 &
  0.89 &
  60.74 &
  5.45 &
  4.53 &
  0.12 \\
\multicolumn{2}{r|}{$\mathrm{E} \rightarrow {}_{\mathrm{B}}\text{\textsc{RU}}$} &
  2,702.04 &
  2,483.15 &
  10,715.26 &
  13,157.31 &
  0.18 &
  0.10 &
  3.93 &
  0.10 &
  6.38 &
  0.16 &
  0.40 &
  0.02 &
  19.86 &
  1.11 &
  48.37 &
  7.14 &
  1.41 &
  0.12 \\
\multicolumn{2}{r|}{$\left(\mathrm{E} \rightarrow {}_{\mathrm{B}}\text{\textsc{RU}} + {}_{\mathrm{B}}\text{\textsc{RU}}\right) + \mathrm{AR}$} &
  219.16 &
  106.06 &
  1,276.78 &
  881.66 &
  0.07 &
  0.03 &
  2.32 &
  0.07 &
  5.69 &
  0.22 &
  0.41 &
  0.02 &
  18.72 &
  0.98 &
  48.11 &
  6.47 &
  1.40 &
  0.11 \\
\multicolumn{2}{r|}{$\left(\mathrm{E} \rightarrow {}_{\mathrm{B}}\text{\textsc{RU}} + {}_{\mathrm{U}}\text{\textsc{RU}}\right) + \mathrm{AR}$} &
  259.91 &
  149.24 &
  1,597.42 &
  1,273.26 &
  0.11 &
  0.07 &
  \textbf{2.08} &
  \textbf{0.18} &
  \textbf{4.94} &
  \textbf{0.41} &
  \textbf{0.28} &
  \textbf{0.04} &
  \textbf{18.56} &
  \textbf{0.99} &
  \textbf{48.32} &
  \textbf{6.78} &
  \textbf{1.37} &
  \textbf{0.13} \\
\multicolumn{2}{r|}{$\left(\mathrm{E} \rightarrow {}_{\mathrm{B}}\text{\textsc{RU}}\right) + \mathrm{AR}$} &
  505.81 &
  582.37 &
  3,916.95 &
  5,297.45 &
  0.11 &
  0.08 &
  3.53 &
  0.13 &
  8.06 &
  0.12 &
  0.94 &
  0.03 &
  18.72 &
  0.97 &
  48.54 &
  6.37 &
  1.37 &
  0.10 \\
\multicolumn{2}{r|}{$\mathrm{E} \rightarrow {}_{\mathrm{U}}\text{\textsc{RU}}$} &
  2,380.63 &
  1,475.81 &
  8,555.18 &
  6,771.74 &
  0.23 &
  0.12 &
  2.92 &
  0.06 &
  5.44 &
  0.12 &
  0.36 &
  0.01 &
  20.29 &
  1.09 &
  50.32 &
  7.57 &
  1.49 &
  0.15 \\
\multicolumn{2}{r|}{$\left(\mathrm{E} \rightarrow {}_{\mathrm{U}}\text{\textsc{RU}} + {}_{\mathrm{B}}\text{\textsc{RU}}\right) + \mathrm{AR}$} &
  245.62 &
  159.01 &
  1,367.82 &
  966.48 &
  0.07 &
  0.03 &
  2.93 &
  0.07 &
  6.93 &
  0.17 &
  0.67 &
  0.02 &
  18.68 &
  1.09 &
  48.85 &
  6.29 &
  1.37 &
  0.14 \\
\multicolumn{2}{r|}{$\left(\mathrm{E} \rightarrow {}_{\mathrm{U}}\text{\textsc{RU}} + {}_{\mathrm{U}}\text{\textsc{RU}}\right) + \mathrm{AR}$} &
  \textbf{216.81} &
  \textbf{107.29} &
  \textbf{1,280.58} &
  \textbf{846.01} &
  \textbf{0.11} &
  \textbf{0.04} &
  3.17 &
  0.08 &
  7.46 &
  0.09 &
  0.81 &
  0.02 &
  18.71 &
  1.09 &
  49.42 &
  7.21 &
  1.38 &
  0.11 \\
\multicolumn{2}{r|}{$\left(\mathrm{E} \rightarrow {}_{\mathrm{U}}\text{\textsc{RU}}\right) + \mathrm{AR}$} &
  248.10 &
  166.88 &
  1,508.53 &
  1,184.28 &
  0.09 &
  0.04 &
  4.60 &
  0.08 &
  9.87 &
  0.03 &
  1.49 &
  0.03 &
  19.01 &
  1.15 &
  49.86 &
  7.30 &
  1.46 &
  0.12 \\
\multicolumn{2}{r|}{${}_{\mathrm{U}}\text{\textsc{RU}}$} &
  1,875.92 &
  1,328.11 &
  6,933.49 &
  6,460.49 &
  0.11 &
  0.08 &
  4.00 &
  0.05 &
  6.72 &
  0.20 &
  0.52 &
  0.01 &
  35.91 &
  0.93 &
  66.81 &
  4.86 &
  6.21 &
  0.11 \\ \cline{3-20} 
 &
   &
  \multicolumn{18}{c|}{\textbf{Long Short-Term Memory (LSTM)}} \\ \cline{3-20} 
\multicolumn{2}{r|}{${}_{\mathrm{B}}\text{\textsc{RU}}$} &
  3,042.70 &
  1,987.48 &
  18,539.83 &
  17,591.06 &
  3.80 &
  0.41 &
  3.68 &
  0.02 &
  6.23 &
  0.13 &
  0.42 &
  0.01 &
  20.34 &
  0.87 &
  48.42 &
  6.44 &
  1.37 &
  0.10 \\
\multicolumn{2}{r|}{$\mathrm{E} \rightarrow {}_{\mathrm{B}}\text{\textsc{RU}}$} &
  2,309.35 &
  1,269.98 &
  7,953.65 &
  6,114.91 &
  0.23 &
  0.15 &
  3.55 &
  0.01 &
  5.70 &
  0.14 &
  0.30 &
  0.01 &
  19.85 &
  1.13 &
  48.19 &
  7.10 &
  1.41 &
  0.13 \\
\multicolumn{2}{r|}{$\left(\mathrm{E} \rightarrow {}_{\mathrm{B}}\text{\textsc{RU}} + {}_{\mathrm{B}}\text{\textsc{RU}}\right) + \mathrm{AR}$} &
  256.32 &
  59.69 &
  1,412.55 &
  487.27 &
  0.12 &
  0.07 &
  \textbf{2.31} &
  \textbf{0.15} &
  \textbf{5.61} &
  \textbf{0.46} &
  \textbf{0.41} &
  \textbf{0.03} &
  \textbf{18.57} &
  \textbf{1.39} &
  \textbf{48.78} &
  \textbf{7.67} &
  \textbf{1.37} &
  \textbf{0.14} \\
\multicolumn{2}{r|}{$\left(\mathrm{E} \rightarrow {}_{\mathrm{B}}\text{\textsc{RU}} + {}_{\mathrm{U}}\text{\textsc{RU}}\right) + \mathrm{AR}$} &
  \textbf{244.81} &
  \textbf{128.06} &
  \textbf{1,233.11} &
  \textbf{666.28} &
  \textbf{0.18} &
  \textbf{0.21} &
  3.17 &
  0.05 &
  7.51 &
  0.17 &
  0.81 &
  0.01 &
  18.63 &
  1.30 &
  49.18 &
  7.93 &
  1.37 &
  0.12 \\
\multicolumn{2}{r|}{$\left(\mathrm{E} \rightarrow {}_{\mathrm{B}}\text{\textsc{RU}}\right) + \mathrm{AR}$} &
  1,554.31 &
  1,416.54 &
  17,417.78 &
  18,060.72 &
  3.90 &
  0.51 &
  4.95 &
  0.11 &
  10.27 &
  0.18 &
  1.62 &
  0.03 &
  18.69 &
  1.14 &
  48.87 &
  6.62 &
  1.37 &
  0.12 \\
\multicolumn{2}{r|}{$\mathrm{E} \rightarrow {}_{\mathrm{U}}\text{\textsc{RU}}$} &
  3,169.42 &
  2,063.50 &
  12,461.42 &
  11,674.12 &
  0.17 &
  0.05 &
  3.75 &
  0.16 &
  6.26 &
  0.22 &
  0.44 &
  0.04 &
  20.74 &
  1.02 &
  49.10 &
  7.14 &
  1.46 &
  0.14 \\
\multicolumn{2}{r|}{$\left(\mathrm{E} \rightarrow {}_{\mathrm{U}}\text{\textsc{RU}} + {}_{\mathrm{B}}\text{\textsc{RU}}\right) + \mathrm{AR}$} &
  1,375.17 &
  1,053.14 &
  15,406.60 &
  15,929.81 &
  3.69 &
  0.34 &
  4.36 &
  0.10 &
  9.39 &
  0.19 &
  1.34 &
  0.02 &
  18.69 &
  1.18 &
  49.36 &
  7.07 &
  1.38 &
  0.13 \\
\multicolumn{2}{r|}{$\left(\mathrm{E} \rightarrow {}_{\mathrm{U}}\text{\textsc{RU}} + {}_{\mathrm{U}}\text{\textsc{RU}}\right) + \mathrm{AR}$} &
  285.02 &
  121.63 &
  1,802.48 &
  1,118.25 &
  0.11 &
  0.06 &
  5.27 &
  0.05 &
  10.69 &
  0.02 &
  1.76 &
  0.01 &
  18.78 &
  1.10 &
  49.68 &
  6.64 &
  1.39 &
  0.15 \\
\multicolumn{2}{r|}{$\left(\mathrm{E} \rightarrow {}_{\mathrm{U}}\text{\textsc{RU}}\right) + \mathrm{AR}$} &
  265.80 &
  112.59 &
  1,437.64 &
  686.42 &
  0.14 &
  0.09 &
  7.91 &
  0.14 &
  13.67 &
  0.20 &
  2.95 &
  0.04 &
  18.84 &
  1.17 &
  49.71 &
  7.81 &
  1.40 &
  0.11 \\
\multicolumn{2}{r|}{${}_{\mathrm{U}}\text{\textsc{RU}}$} &
  4,942.10 &
  3,425.34 &
  25,691.11 &
  25,415.68 &
  7.38 &
  0.97 &
  4.36 &
  0.02 &
  7.41 &
  0.06 &
  0.69 &
  0.01 &
  21.27 &
  1.06 &
  48.57 &
  6.40 &
  1.37 &
  0.11 \\ \cline{3-20} 
\end{tabular}%
}
\end{table}
\end{landscape}\clearpage
\loadgeometry{geom}

\savegeometry{geom}
\newgeometry{margin=0pt}
\thispagestyle{empty}\setcounter{page}{17}
\begin{landscape}
\begin{table}
\centering
\caption[Ablation with \ReGENN's Hyperparameters.]%
{Ablation on the main experiments with \ReGENN's hyperparameters.\\%
{\footnotesize\textbf{Legend:} Algorithms with the best performance are in \textbf{bold}; besides, we refer to the Transformer Encoder as $\mathrm{E}$, recurrent unit as \textsc{RU}, Bidirectional as \textsc{B}, Unidirectional as \textsc{U}, and Autoregressive as \textsc{AR}.}}
\label{tab:ablation-3}
\resizebox{1.35\textwidth}{!}{%
\begin{tabular}{rr|cccccc|cccccc|cccccc|}
\cline{3-20}
 &
  \multicolumn{1}{r|}{} &
  \multicolumn{6}{c|}{\textbf{SARS-CoV-2}} &
  \multicolumn{6}{c|}{\textbf{Brazilian Weather}} &
  \multicolumn{6}{c|}{\textbf{PhysioNet}} \\ \cline{3-20} 
 &
   &
  \textbf{MAE} &
  \textbf{$\pm$ STD} &
  \textbf{RMSE} &
  \textbf{$\pm$ STD} &
  \textbf{MSLE} &
  \textbf{$\pm$ STD} &
  \textbf{MAE} &
  \textbf{$\pm$ STD} &
  \textbf{RMSE} &
  \textbf{$\pm$ STD} &
  \textbf{MSLE} &
  \textbf{$\pm$ STD} &
  \textbf{MAE} &
  \textbf{$\pm$ STD} &
  \textbf{RMSE} &
  \textbf{$\pm$ STD} &
  \textbf{MSLE} &
  \textbf{$\pm$ STD} \\ \cline{3-20} 
 &
  \multicolumn{1}{r|}{} &
  \multicolumn{18}{c|}{\textbf{Recurrent Neural Network (RNN)}} \\ \cline{3-20} 
\multicolumn{2}{r|}{${}_{\mathrm{B}}\text{\textsc{RU}}$} &
  424.32 &
  201.00 &
  2819.11 &
  2020.00 &
  0.07 &
  0.04 &
  2.56 &
  0.04 &
  5.90 &
  0.10 &
  0.45 &
  0.02 &
  19.95 &
  1.08 &
  50.47 &
  7.86 &
  1.38 &
  0.11 \\
\multicolumn{2}{r|}{$\mathrm{E} \rightarrow {}_{\mathrm{B}}\text{\textsc{RU}}$} &
  4161.35 &
  3030.00 &
  28755.33 &
  28700.00 &
  11.08 &
  1.31 &
  2.70 &
  0.13 &
  5.66 &
  0.19 &
  0.37 &
  0.02 &
  19.11 &
  1.19 &
  49.16 &
  7.55 &
  1.41 &
  0.12 \\
\multicolumn{2}{r|}{$\left(\mathrm{E} \rightarrow {}_{\mathrm{B}}\text{\textsc{RU}} + {}_{\mathrm{B}}\text{\textsc{RU}}\right) + \mathrm{AR}$} &
  1624.44 &
  1220.00 &
  17245.24 &
  17200.00 &
  3.84 &
  0.49 &
  \textbf{2.18} &
  \textbf{0.06} &
  \textbf{5.52} &
  \textbf{0.29} &
  \textbf{0.37} &
  \textbf{0.02} &
  \textbf{18.72} &
  \textbf{0.89} &
  \textbf{48.60} &
  \textbf{6.57} &
  \textbf{1.37} &
  \textbf{0.13} \\
\multicolumn{2}{r|}{$\left(\mathrm{E} \rightarrow {}_{\mathrm{B}}\text{\textsc{RU}} + {}_{\mathrm{U}}\text{\textsc{RU}}\right) + \mathrm{AR}$} &
  \textbf{257.16} &
  \textbf{149.00} &
  \textbf{1438.72} &
  \textbf{823.00} &
  \textbf{0.09} &
  \textbf{0.06} &
  2.84 &
  0.15 &
  6.47 &
  0.15 &
  0.55 &
  0.01 &
  18.97 &
  1.27 &
  49.59 &
  7.78 &
  1.39 &
  0.11 \\
\multicolumn{2}{r|}{$\left(\mathrm{E} \rightarrow {}_{\mathrm{B}}\text{\textsc{RU}}\right) + \mathrm{AR}$} &
  1471.49 &
  953.00 &
  17050.08 &
  15700.00 &
  3.74 &
  0.34 &
  3.59 &
  0.08 &
  8.13 &
  0.08 &
  0.98 &
  0.02 &
  18.81 &
  1.23 &
  50.20 &
  7.62 &
  1.38 &
  0.12 \\
\multicolumn{2}{r|}{$\mathrm{E} \rightarrow {}_{\mathrm{U}}\text{\textsc{RU}}$} &
  1949.65 &
  1190.00 &
  16925.69 &
  16100.00 &
  3.84 &
  0.41 &
  3.12 &
  0.11 &
  5.86 &
  0.04 &
  0.32 &
  0.01 &
  19.47 &
  1.24 &
  50.24 &
  7.48 &
  1.44 &
  0.15 \\
\multicolumn{2}{r|}{$\left(\mathrm{E} \rightarrow {}_{\mathrm{U}}\text{\textsc{RU}} + {}_{\mathrm{B}}\text{\textsc{RU}}\right) + \mathrm{AR}$} &
  372.81 &
  190.00 &
  2155.37 &
  1320.00 &
  0.11 &
  0.07 &
  2.36 &
  0.12 &
  5.69 &
  0.24 &
  0.41 &
  0.03 &
  18.76 &
  1.05 &
  48.81 &
  6.47 &
  1.39 &
  0.14 \\
\multicolumn{2}{r|}{$\left(\mathrm{E} \rightarrow {}_{\mathrm{U}}\text{\textsc{RU}} + {}_{\mathrm{U}}\text{\textsc{RU}}\right) + \mathrm{AR}$} &
  1472.05 &
  917.00 &
  16491.81 &
  15700.00 &
  3.72 &
  0.26 &
  2.52 &
  0.10 &
  5.83 &
  0.25 &
  0.42 &
  0.02 &
  19.02 &
  1.00 &
  49.75 &
  7.80 &
  1.43 &
  0.13 \\
\multicolumn{2}{r|}{$\left(\mathrm{E} \rightarrow {}_{\mathrm{U}}\text{\textsc{RU}}\right) + \mathrm{AR}$} &
  1350.35 &
  818.00 &
  16673.11 &
  15300.00 &
  3.69 &
  0.33 &
  4.30 &
  0.06 &
  9.38 &
  0.06 &
  1.34 &
  0.03 &
  18.98 &
  1.20 &
  48.67 &
  7.06 &
  1.45 &
  0.13 \\
\multicolumn{2}{r|}{${}_{\mathrm{U}}\text{\textsc{RU}}$} &
  1175.99 &
  769.00 &
  6716.42 &
  5340.00 &
  2.08 &
  0.73 &
  2.40 &
  0.07 &
  5.32 &
  0.19 &
  0.32 &
  0.01 &
  20.58 &
  1.00 &
  50.32 &
  7.02 &
  1.42 &
  0.12 \\ \cline{3-20} 
 &
  \multicolumn{1}{r|}{} &
  \multicolumn{18}{c|}{\textbf{Gated Recurrent Unit (GRU)}} \\ \cline{3-20} 
\multicolumn{2}{r|}{${}_{\mathrm{B}}\text{\textsc{RU}}$} &
  3130.47 &
  2070.00 &
  28063.07 &
  24400.00 &
  7.78 &
  0.75 &
  2.19 &
  0.06 &
  5.00 &
  0.26 &
  0.28 &
  0.02 &
  24.80 &
  1.16 &
  56.43 &
  6.27 &
  3.01 &
  0.11 \\
\multicolumn{2}{r|}{$\mathrm{E} \rightarrow {}_{\mathrm{B}}\text{\textsc{RU}}$} &
  596.77 &
  404.00 &
  3446.67 &
  2750.00 &
  0.17 &
  0.16 &
  2.33 &
  0.13 &
  5.19 &
  0.26 &
  0.28 &
  0.01 &
  24.17 &
  0.97 &
  54.81 &
  5.65 &
  3.00 &
  0.11 \\
\multicolumn{2}{r|}{$\left(\mathrm{E} \rightarrow {}_{\mathrm{B}}\text{\textsc{RU}} + {}_{\mathrm{B}}\text{\textsc{RU}}\right) + \mathrm{AR}$} &
  388.93 &
  335.00 &
  1888.70 &
  1750.00 &
  0.10 &
  0.03 &
  2.38 &
  0.13 &
  5.72 &
  0.27 &
  0.41 &
  0.02 &
  18.55 &
  0.93 &
  47.86 &
  6.80 &
  1.39 &
  0.12 \\
\multicolumn{2}{r|}{$\left(\mathrm{E} \rightarrow {}_{\mathrm{B}}\text{\textsc{RU}} + {}_{\mathrm{U}}\text{\textsc{RU}}\right) + \mathrm{AR}$} &
  366.06 &
  175.00 &
  2158.32 &
  1210.00 &
  0.10 &
  0.04 &
  2.24 &
  0.07 &
  5.05 &
  0.40 &
  0.28 &
  0.02 &
  \textbf{18.54} &
  \textbf{0.94} &
  \textbf{48.90} &
  \textbf{5.63} &
  \textbf{1.37} &
  \textbf{0.13} \\
\multicolumn{2}{r|}{$\left(\mathrm{E} \rightarrow {}_{\mathrm{B}}\text{\textsc{RU}}\right) + \mathrm{AR}$} &
  1502.20 &
  1010.00 &
  16799.97 &
  14800.00 &
  3.67 &
  0.39 &
  3.31 &
  0.06 &
  7.57 &
  0.29 &
  0.82 &
  0.03 &
  18.67 &
  0.91 &
  48.59 &
  6.26 &
  1.39 &
  0.11 \\
\multicolumn{2}{r|}{$\mathrm{E} \rightarrow {}_{\mathrm{U}}\text{\textsc{RU}}$} &
  968.83 &
  674.00 &
  5982.35 &
  5430.00 &
  0.22 &
  0.07 &
  2.81 &
  0.15 &
  5.69 &
  0.11 &
  0.34 &
  0.01 &
  19.42 &
  1.26 &
  51.03 &
  7.90 &
  1.44 &
  0.15 \\
\multicolumn{2}{r|}{$\left(\mathrm{E} \rightarrow {}_{\mathrm{U}}\text{\textsc{RU}} + {}_{\mathrm{B}}\text{\textsc{RU}}\right) + \mathrm{AR}$} &
  \textbf{208.88} &
  \textbf{120.00} &
  \textbf{1141.56} &
  \textbf{576.00} &
  \textbf{0.08} &
  \textbf{0.04} &
  2.24 &
  0.22 &
  5.11 &
  0.34 &
  0.28 &
  0.03 &
  18.65 &
  0.96 &
  48.64 &
  6.99 &
  1.38 &
  0.12 \\
\multicolumn{2}{r|}{$\left(\mathrm{E} \rightarrow {}_{\mathrm{U}}\text{\textsc{RU}} + {}_{\mathrm{U}}\text{\textsc{RU}}\right) + \mathrm{AR}$} &
  4052.71 &
  2630.00 &
  31384.80 &
  28700.00 &
  11.30 &
  1.27 &
  \textbf{2.10} &
  \textbf{0.10} &
  \textbf{4.97} &
  \textbf{0.23} &
  \textbf{0.28} &
  \textbf{0.01} &
  18.64 &
  1.28 &
  48.81 &
  8.07 &
  1.38 &
  0.12 \\
\multicolumn{2}{r|}{$\left(\mathrm{E} \rightarrow {}_{\mathrm{U}}\text{\textsc{RU}}\right) + \mathrm{AR}$} &
  2852.28 &
  1740.00 &
  25869.18 &
  23700.00 &
  7.60 &
  0.59 &
  3.25 &
  0.08 &
  7.55 &
  0.12 &
  0.81 &
  0.02 &
  18.82 &
  0.93 &
  49.98 &
  7.61 &
  1.41 &
  0.13 \\
\multicolumn{2}{r|}{${}_{\mathrm{U}}\text{\textsc{RU}}$} &
  1825.66 &
  651.00 &
  20400.04 &
  15800.00 &
  4.95 &
  0.20 &
  3.26 &
  0.15 &
  7.45 &
  0.35 &
  0.77 &
  0.02 &
  30.86 &
  1.16 &
  63.30 &
  5.42 &
  4.69 &
  0.13 \\ \cline{3-20} 
 &
  \multicolumn{1}{r|}{} &
  \multicolumn{18}{c|}{\textbf{Long Short-Term Memory (LSTM)}} \\ \cline{3-20} 
\multicolumn{2}{r|}{${}_{\mathrm{B}}\text{\textsc{RU}}$} &
  1800.23 &
  929.00 &
  20264.27 &
  17500.00 &
  3.95 &
  0.15 &
  \textbf{2.20} &
  \textbf{0.03} &
  \textbf{5.04} &
  \textbf{0.14} &
  \textbf{0.28} &
  \textbf{0.01} &
  19.49 &
  1.30 &
  50.47 &
  8.71 &
  1.40 &
  0.12 \\
\multicolumn{2}{r|}{$\mathrm{E} \rightarrow {}_{\mathrm{B}}\text{\textsc{RU}}$} &
  704.80 &
  343.00 &
  4088.05 &
  2890.00 &
  0.20 &
  0.10 &
  2.57 &
  0.20 &
  5.42 &
  0.41 &
  0.29 &
  0.02 &
  18.88 &
  1.14 &
  48.55 &
  7.13 &
  1.40 &
  0.14 \\
\multicolumn{2}{r|}{$\left(\mathrm{E} \rightarrow {}_{\mathrm{B}}\text{\textsc{RU}} + {}_{\mathrm{B}}\text{\textsc{RU}}\right) + \mathrm{AR}$} &
  1598.03 &
  884.00 &
  19056.28 &
  16900.00 &
  3.89 &
  0.17 &
  2.44 &
  0.12 &
  5.75 &
  0.25 &
  0.41 &
  0.01 &
  \textbf{18.42} &
  \textbf{1.17} &
  \textbf{47.78} &
  \textbf{6.06} &
  \textbf{1.37} &
  \textbf{0.11} \\
\multicolumn{2}{r|}{$\left(\mathrm{E} \rightarrow {}_{\mathrm{B}}\text{\textsc{RU}} + {}_{\mathrm{U}}\text{\textsc{RU}}\right) + \mathrm{AR}$} &
  4064.52 &
  2700.00 &
  31196.10 &
  28700.00 &
  11.30 &
  1.01 &
  3.17 &
  0.20 &
  7.42 &
  0.26 &
  0.80 &
  0.03 &
  18.46 &
  0.85 &
  48.33 &
  6.48 &
  1.36 &
  0.13 \\
\multicolumn{2}{r|}{$\left(\mathrm{E} \rightarrow {}_{\mathrm{B}}\text{\textsc{RU}}\right) + \mathrm{AR}$} &
  \textbf{211.93} &
  \textbf{83.40} &
  \textbf{1181.31} &
  \textbf{570.00} &
  \textbf{0.08} &
  \textbf{0.06} &
  3.61 &
  0.10 &
  8.19 &
  0.10 &
  0.96 &
  0.01 &
  18.54 &
  1.16 &
  47.35 &
  6.65 &
  1.38 &
  0.11 \\
\multicolumn{2}{r|}{$\mathrm{E} \rightarrow {}_{\mathrm{U}}\text{\textsc{RU}}$} &
  2778.55 &
  1110.00 &
  12830.03 &
  10200.00 &
  0.21 &
  0.06 &
  2.96 &
  0.15 &
  5.54 &
  0.12 &
  0.30 &
  0.01 &
  19.78 &
  1.02 &
  51.29 &
  8.67 &
  1.44 &
  0.14 \\
\multicolumn{2}{r|}{$\left(\mathrm{E} \rightarrow {}_{\mathrm{U}}\text{\textsc{RU}} + {}_{\mathrm{B}}\text{\textsc{RU}}\right) + \mathrm{AR}$} &
  1380.99 &
  864.00 &
  16370.41 &
  15000.00 &
  3.66 &
  0.32 &
  2.47 &
  0.06 &
  5.77 &
  0.14 &
  0.41 &
  0.03 &
  18.55 &
  0.87 &
  48.28 &
  6.58 &
  1.38 &
  0.12 \\
\multicolumn{2}{r|}{$\left(\mathrm{E} \rightarrow {}_{\mathrm{U}}\text{\textsc{RU}} + {}_{\mathrm{U}}\text{\textsc{RU}}\right) + \mathrm{AR}$} &
  1713.72 &
  1030.00 &
  18531.68 &
  15900.00 &
  3.81 &
  0.39 &
  4.88 &
  0.05 &
  10.24 &
  0.07 &
  1.61 &
  0.01 &
  18.63 &
  1.20 &
  48.93 &
  7.08 &
  1.37 &
  0.14 \\
\multicolumn{2}{r|}{$\left(\mathrm{E} \rightarrow {}_{\mathrm{U}}\text{\textsc{RU}}\right) + \mathrm{AR}$} &
  260.66 &
  215.00 &
  1513.92 &
  1430.00 &
  0.10 &
  0.03 &
  5.25 &
  0.08 &
  10.69 &
  0.12 &
  1.75 &
  0.01 &
  18.88 &
  0.99 &
  50.31 &
  7.53 &
  1.39 &
  0.10 \\
\multicolumn{2}{r|}{${}_{\mathrm{U}}\text{\textsc{RU}}$} &
  3723.88 &
  2540.00 &
  21022.60 &
  20000.00 &
  2.83 &
  0.31 &
  2.83 &
  0.02 &
  6.17 &
  0.07 &
  0.50 &
  0.02 &
  24.85 &
  0.95 &
  56.41 &
  6.61 &
  2.92 & 
  0.10 \\ \cline{3-20}
\end{tabular}%
}
\end{table}
\end{landscape}\clearpage
\loadgeometry{geom}


\clearpage\newpage%
\begin{titlepage}
    \vspace*{\fill}
        \begin{center}
          \Huge{\bf Hyperparameters\\\rule[5px]{1.7cm}{.1cm}~used for~\rule[5px]{1.7cm}{.1cm}\\Transfer Learning}
        \end{center}
    \vspace*{\fill}
\end{titlepage}

\savegeometry{geom}
\newgeometry{margin=0pt}
\thispagestyle{empty}\setcounter{page}{19}
\begin{table}
\centering
\caption[Hyperparameters used during Transfer Learning.]%
{List of hyperparameters used during Transfer Learning.}
\label{tab:hyperparameters-3}
\begin{tabular}{r|c|c|c|c|c|c|l}
\cline{2-7}
                             & \multicolumn{6}{c|}{\textbf{Transfer Learning}} &                              \\ \cline{2-7}
 &
  \textbf{45 Days} &
  \textbf{60 Days} &
  \textbf{75 Days} &
  \textbf{90 Days} &
  \textbf{105 Days} &
  \textbf{120 Days} &
   \\ \cline{2-7}
\textbf{autoregression}      & True    & True    & True    & True    & True     & True    \\ \cline{2-7}
\textbf{batch-size}          & 32      & 32      & 32      & 32      & 32       & 32      \\ \cline{2-7}
\textbf{bias}                & True    & True    & True    & True    & True     & True    \\ \cline{2-7}
\textbf{bidirectional-gate} &
  False &
  False &
  False &
  False &
  False &
  False &
  \\ \cline{2-7}
\textbf{bidirectional-sequencer} &
  False &
  False &
  False &
  False &
  False &
  False &
  \\ \cline{2-7}
\textbf{clip-norm}           & 15.0    & 0.0     & 10.0    & 0.0     & 0.0      & 85.0    \\ \cline{2-7}
\textbf{criterion}           & MAE     & MAE     & MAE     & MAE     & MAE      & MAE     \\ \cline{2-7}
\textbf{dropout}             & 0.25    & 0.0     & 0.35    & 0.1     & 0.0      & 0.0     \\ \cline{2-7}
\textbf{early-stop}          & 250     & 250     & 250     & 250     & 250      & 250     \\ \cline{2-7}
\textbf{epochs}              & 2500    & 5000    & 2500    & 2500    & 2500     & 2500    \\ \cline{2-7}
\textbf{evolution-function} &
  Identity &
  Identity &
  Identity &
  Identity &
  Identity &
  Identity &
  \\ \cline{2-7}
\textbf{gate}                & LSTM    & LSTM    & LSTM    & LSTM    & LSTM     & LSTM    \\ \cline{2-7}
\textbf{iterator}            & Time    & Time    & Time    & Time    & Time     & time    \\ \cline{2-7}
\textbf{learning-rate}       & 0.006   & 0.009   & 0.001   & 0.002   & 0.0002   & 0.001   \\ \cline{2-7}
\textbf{load-weights}        & False   & True    & True    & True    & True     & True    \\ \cline{2-7}
\textbf{no-encoder}          & False   & False   & False   & False   & False    & False   \\ \cline{2-7}
\textbf{no-sequencer}        & False   & False   & False   & False   & False    & False   \\ \cline{2-7}
\textbf{normalization-axis}  & 2       & 2       & 2       & 2       & 2        & 2       \\ \cline{2-7}
\textbf{normalization-type} &
  Maximum &
  Maximum &
  Maximum &
  Maximum &
  Maximum &
  Maximum &
  \\ \cline{2-7}
\textbf{optimizer}           & Adam    & Adam    & Adam    & Adam    & Adam     & Adam    \\ \cline{2-7}
\textbf{output-function}     & ReLU    & ReLU    & ReLU    & ReLU    & ReLU     & ReLU    \\ \cline{2-7}
\textbf{random-seed}         & 0       & 0       & 0       & 0       & 0        & 0       \\ \cline{2-7}
\textbf{scheduler-factor}    & 0.95    & 0.95    & 0.95    & 0.65    & 0.05     & 0.95    \\ \cline{2-7}
\textbf{scheduler-min-lr}    & 0.0     & 0.0     & 0.0     & 0.0     & 0.0      & 0.0     \\ \cline{2-7}
\textbf{scheduler-patience}  & 25      & 20      & 25      & 60      & 25       & 25      \\ \cline{2-7}
\textbf{scheduler-threshold} & 0.1     & 0.01    & 0.1     & 0.1     & 0.1      & 0.1     \\ \cline{2-7}
\textbf{sequencer}           & LSTM    & LSTM    & LSTM    & LSTM    & LSTM     & LSTM    \\ \cline{2-7}
\textbf{stride}              & 14      & 14      & 14      & 14      & 14       & 14      \\ \cline{2-7}
\textbf{validation-stride}   & 7       & 7       & 7       & 7       & 7        & 7       \\ \cline{2-7}
\textbf{watch-axis}          & 2       & 2       & 2       & 2       & 2        & 2       \\ \cline{2-7}
\textbf{window}              & 7       & 7       & 7       & 7       & 7        & 7       \\ \cline{2-7}
\end{tabular}
\end{table}\clearpage
\loadgeometry{geom}

\savegeometry{geom}
\newgeometry{margin=0pt}
\thispagestyle{empty}\setcounter{page}{20}
\begin{landscape}
\begin{table}
\centering
\caption[Baselines on the first 45, 60, and 75 days of the SARS-CoV-2 dataset.]%
{Detailed results for the first three slices of the SARS-CoV-2.\\%
{\footnotesize\hspace{1.4cm}\textbf{Legend:} Algorithms with best performance are in \textbf{bold}, the ones noted as --- yielded exceptions, and others as *** were suppressed due to poor performance.}}
\label{tab:baselines-2}
\resizebox{1.35\textwidth}{!}{%
\begin{tabular}{r|cccccc|cccccc|cccccc|}
\cline{2-19}
\multicolumn{1}{l|}{} &
  \multicolumn{6}{c|}{\textbf{45 Days}} &
  \multicolumn{6}{c|}{\textbf{60 Days}} &
  \multicolumn{6}{c|}{\textbf{75 Days}} \\ \cline{2-19} 
 &
  \textbf{MAE} &
  \textbf{$\pm$ STD} &
  \textbf{RMSE} &
  \textbf{$\pm$ STD} &
  \textbf{MSLE} &
  \textbf{$\pm$ STD} &
  \textbf{MAE} &
  \textbf{$\pm$ STD} &
  \textbf{RMSE} &
  \textbf{$\pm$ STD} &
  \textbf{MSLE} &
  \textbf{$\pm$ STD} &
  \textbf{MAE} &
  \textbf{$\pm$ STD} &
  \textbf{RMSE} &
  \textbf{$\pm$ STD} &
  \textbf{MSLE} &
  \textbf{$\pm$ STD} \\ \cline{2-19} 
\multicolumn{1}{l|}{} &
  \multicolumn{18}{c|}{\textbf{Deep Learning Algorithms}} \\ \cline{2-19} 
\ReGENN &
  \textbf{32.13} &
  \textbf{57.55} &
  \textbf{371.29} &
  \textbf{676.19} &
  \textbf{0.31} &
  \textbf{0.13} &
  \textbf{20.61} &
  \textbf{23.25} &
  \textbf{179.22} &
  \textbf{210.21} &
  \textbf{0.79} &
  \textbf{0.16} &
  \textbf{67.85} &
  \textbf{66.11} &
  \textbf{490.51} &
  \textbf{493.93} &
  \textbf{0.31} &
  \textbf{0.05} \\
\textbf{MLCNN} &
  151.22 &
  1816.69 &
  181.94 &
  2081.31 &
  1.43 &
  4.54 &
  353.39 &
  1982.21 &
  482.66 &
  2420.29 &
  6.31 &
  10.00 &
  1574.14 &
  6425.31 &
  2246.30 &
  9462.22 &
  14.44 &
  14.70 \\
\textbf{DSANet} &
  378.54 &
  4759.23 &
  411.23 &
  4990.41 &
  2.86 &
  10.69 &
  764.39 &
  5758.71 &
  917.00 &
  5979.91 &
  10.13 &
  16.67 &
  2746.73 &
  11622.21 &
  3471.10 &
  14806.59 &
  22.99 &
  24.39 \\
\textbf{LSTNet} &
  89.65 &
  852.78 &
  110.26 &
  940.48 &
  1.57 &
  5.72 &
  330.07 &
  1538.96 &
  475.37 &
  2137.49 &
  6.98 &
  11.61 &
  2023.90 &
  9475.81 &
  2648.69 &
  12510.47 &
  13.54 &
  18.76 \\ \cline{2-19} 
 &
  \multicolumn{18}{c|}{\textbf{Multi-Output and Multi-Task Algorithms}} \\ \cline{2-19} 
\textbf{CCA} &
  387.21 &
  127.88 &
  3,812.86 &
  670.27 &
  1.96 &
  0.52 &
  572.70 &
  112.76 &
  5,423.93 &
  423.77 &
  2.25 &
  0.49 &
  1,768.42 &
  194.76 &
  11,192.02 &
  2,977.72 &
  1.98 &
  0.52 \\
\textbf{Decision Tree} &
  81.28 &
  1.64 &
  997.48 &
  131.29 &
  0.54 &
  0.31 &
  197.50 &
  34.61 &
  1,503.36 &
  179.76 &
  0.56 &
  0.03 &
  \textbf{606.66} &
  \textbf{257.03} &
  \textbf{3,639.93} &
  \textbf{1,636.15} &
  \textbf{0.67} &
  \textbf{0.12} \\
\textbf{Extra Tree} &
  87.90 &
  6.31 &
  1,010.82 &
  121.86 &
  0.55 &
  0.31 &
  228.55 &
  80.70 &
  1,617.34 &
  23.65 &
  0.70 &
  0.12 &
  819.14 &
  367.86 &
  5,297.72 &
  2,520.36 &
  0.90 &
  0.12 \\
\textbf{Extra Trees} &
  51.88 &
  10.09 &
  515.05 &
  6.04 &
  0.56 &
  0.33 &
  195.56 &
  87.78 &
  1,432.03 &
  402.45 &
  0.54 &
  0.05 &
  696.51 &
  355.07 &
  3,566.17 &
  1,737.84 &
  0.43 &
  0.17 \\
\textbf{Gaussian Process} &
  209.79 &
  104.24 &
  2,710.77 &
  1,301.22 &
  0.50 &
  0.28 &
  274.72 &
  98.87 &
  1,865.20 &
  69.19 &
  0.68 &
  0.16 &
  1,201.62 &
  689.68 &
  6,393.49 &
  3,403.30 &
  2.43 &
  1.08 \\
\textbf{Historical Average} &
  328.54 &
  102.80 &
  4,204.59 &
  1,175.56 &
  0.46 &
  0.07 &
  542.31 &
  141.89 &
  4,808.31 &
  449.91 &
  0.56 &
  0.06 &
  1,471.47 &
  618.25 &
  7,256.21 &
  1,906.30 &
  0.67 &
  0.03 \\
\textbf{Kernel Ridge} &
  57.95 &
  9.15 &
  609.72 &
  0.58 &
  0.67 &
  0.29 &
  230.48 &
  113.90 &
  1,605.85 &
  565.99 &
  1.26 &
  0.50 &
  856.45 &
  395.81 &
  4,384.56 &
  1,488.20 &
  0.81 &
  0.33 \\
\textbf{KNeighbors} &
  55.86 &
  21.10 &
  566.80 &
  125.88 &
  0.59 &
  0.36 &
  208.37 &
  45.67 &
  1,858.56 &
  19.75 &
  0.66 &
  0.10 &
  766.19 &
  361.97 &
  4,176.97 &
  1,480.61 &
  0.54 &
  0.31 \\
\textbf{Lars} &
  --- &
  --- &
  --- &
  --- &
  --- &
  --- &
  *** &
  *** &
  *** &
  *** &
  *** &
  *** &
  *** &
  *** &
  *** &
  *** &
  *** &
  *** \\
\textbf{Lasso-Lars} &
  328.54 &
  102.80 &
  4,204.59 &
  1,175.56 &
  0.46 &
  0.07 &
  542.31 &
  141.89 &
  4,808.31 &
  449.91 &
  0.56 &
  0.06 &
  1,471.47 &
  618.25 &
  7,256.21 &
  1,906.30 &
  0.67 &
  0.03 \\
\textbf{Linear Regression} &
  282.75 &
  159.14 &
  3,846.25 &
  2,141.83 &
  1.56 &
  1.03 &
  *** &
  *** &
  *** &
  *** &
  *** &
  *** &
  *** &
  *** &
  *** &
  *** &
  *** &
  *** \\
\textbf{Multi-Task Elastic-Net} &
  328.54 &
  102.80 &
  4,204.59 &
  1,175.56 &
  0.46 &
  0.07 &
  542.31 &
  141.89 &
  4,808.31 &
  449.91 &
  0.56 &
  0.06 &
  1,471.47 &
  618.25 &
  7,256.21 &
  1,906.30 &
  0.67 &
  0.03 \\
\textbf{Multi-Task Lasso} &
  328.54 &
  102.80 &
  4,204.59 &
  1,175.56 &
  0.46 &
  0.07 &
  542.31 &
  141.89 &
  4,808.31 &
  449.91 &
  0.56 &
  0.06 &
  1,471.47 &
  618.25 &
  7,256.21 &
  1,906.30 &
  0.67 &
  0.03 \\
\textbf{Orthogonal Matching Pursuit} &
  \textbf{41.91} &
  \textbf{4.57} &
  \textbf{426.72} &
  \textbf{47.62} &
  \textbf{0.56} &
  \textbf{0.34} &
  557.52 &
  82.20 &
  7,826.10 &
  4,217.06 &
  0.63 &
  0.09 &
  3,105.51 &
  2,409.81 &
  34,411.01 &
  38,075.02 &
  1.29 &
  0.07 \\
\textbf{PLS Canonical} &
  1,338.79 &
  508.42 &
  22,679.50 &
  10,092.75 &
  1.30 &
  0.52 &
  2,328.60 &
  2,238.32 &
  43,325.47 &
  50,723.61 &
  1.54 &
  0.02 &
  4,986.37 &
  3,773.55 &
  72,919.82 &
  78,510.01 &
  1.91 &
  0.75 \\
\textbf{PLS} &
  360.54 &
  166.60 &
  5,007.97 &
  2,916.95 &
  0.68 &
  0.35 &
  965.53 &
  363.99 &
  10,693.05 &
  6,439.21 &
  0.74 &
  0.02 &
  5,404.75 &
  5,528.56 &
  60,923.11 &
  75,197.91 &
  1.01 &
  0.35 \\
\textbf{Radius Neighbors} &
  --- &
  --- &
  --- &
  --- &
  --- &
  --- &
  --- &
  --- &
  --- &
  --- &
  --- &
  --- &
  --- &
  --- &
  --- &
  --- &
  --- &
  --- \\
\textbf{Random Forest} &
  63.66 &
  21.68 &
  719.19 &
  192.72 &
  0.54 &
  0.32 &
  \textbf{193.86} &
  \textbf{84.33} &
  \textbf{1,405.11} &
  \textbf{372.31} &
  \textbf{0.51} &
  \textbf{0.04} &
  684.15 &
  312.72 &
  3,469.58 &
  1,305.83 &
  0.43 &
  0.16 \\
\textbf{RANSAC} &
  --- &
  --- &
  --- &
  --- &
  --- &
  --- &
  --- &
  --- &
  --- &
  --- &
  --- &
  --- &
  --- &
  --- &
  --- &
  --- &
  --- &
  --- \\
\textbf{Ridge} &
  57.67 &
  9.48 &
  608.22 &
  3.14 &
  0.52 &
  0.31 &
  230.04 &
  113.79 &
  1,608.98 &
  568.38 &
  0.69 &
  0.18 &
  851.08 &
  396.66 &
  4,375.78 &
  1,486.11 &
  0.45 &
  0.17 \\ \cline{2-19} 
 &
  \multicolumn{18}{c|}{\textbf{Single-Target Algorithms on Chain Ensemble}} \\ \cline{2-19} 
\textbf{AdaBoost} &
  64.66 &
  31.74 &
  914.45 &
  573.35 &
  0.58 &
  0.35 &
  232.07 &
  116.02 &
  1,570.25 &
  490.25 &
  0.66 &
  0.17 &
  913.16 &
  468.54 &
  4,555.41 &
  2,216.05 &
  0.50 &
  0.16 \\
\textbf{ARD} &
  249.15 &
  145.03 &
  3,492.20 &
  2,013.14 &
  1.22 &
  0.80 &
  746.99 &
  328.16 &
  9,975.92 &
  8,745.04 &
  1.14 &
  0.06 &
  2,070.55 &
  1,112.22 &
  24,256.52 &
  24,504.36 &
  0.97 &
  0.10 \\
\textbf{Bagging} &
  74.38 &
  18.62 &
  877.74 &
  175.47 &
  0.53 &
  0.32 &
  214.39 &
  106.38 &
  1,632.65 &
  586.02 &
  0.48 &
  0.02 &
  687.52 &
  321.81 &
  3,503.36 &
  1,481.42 &
  0.45 &
  0.17 \\
\textbf{Bayesian Ridge} &
  168.97 &
  88.19 &
  2,732.62 &
  1,495.76 &
  0.94 &
  0.61 &
  385.57 &
  185.21 &
  3,149.31 &
  927.97 &
  1.03 &
  0.17 &
  896.91 &
  414.00 &
  4,532.26 &
  1,632.95 &
  0.46 &
  0.17 \\
\textbf{CatBoost} &
  90.59 &
  31.85 &
  1,031.37 &
  292.43 &
  0.52 &
  0.31 &
  259.99 &
  104.90 &
  2,375.76 &
  716.46 &
  0.46 &
  0.02 &
  606.33 &
  227.25 &
  3,378.25 &
  998.34 &
  0.42 &
  0.15 \\
\textbf{Gradient Boosting} &
  49.80 &
  0.63 &
  546.89 &
  148.35 &
  0.56 &
  0.33 &
  \textbf{190.73} &
  \textbf{83.11} &
  \textbf{1,243.65} &
  \textbf{210.91} &
  \textbf{0.51} &
  \textbf{0.03} &
  \textbf{563.49} &
  \textbf{253.34} &
  \textbf{3,087.26} &
  \textbf{1,419.53} &
  \textbf{0.47} &
  \textbf{0.19} \\
\textbf{Histogram Grad. Boosting} &
  88.14 &
  67.19 &
  1,045.45 &
  1,058.30 &
  0.64 &
  0.20 &
  317.10 &
  181.62 &
  2,144.85 &
  1,149.52 &
  0.82 &
  0.25 &
  573.71 &
  178.47 &
  3,040.38 &
  507.34 &
  0.88 &
  0.19 \\
\textbf{Huber} &
  \textbf{43.58} &
  \textbf{2.91} &
  \textbf{439.65} &
  \textbf{184.34} &
  \textbf{0.69} &
  \textbf{0.30} &
  272.56 &
  143.23 &
  1,911.75 &
  677.08 &
  1.11 &
  0.40 &
  890.99 &
  450.34 &
  4,991.10 &
  2,033.22 &
  1.06 &
  0.23 \\
\textbf{Isotonic} &
  --- &
  --- &
  --- &
  --- &
  --- &
  --- &
  --- &
  --- &
  --- &
  --- &
  --- &
  --- &
  --- &
  --- &
  --- &
  --- &
  --- &
  --- \\
\textbf{Lasso-Lars-IC} &
  139.88 &
  63.08 &
  2,327.89 &
  1,089.33 &
  0.63 &
  0.35 &
  439.78 &
  218.16 &
  4,455.69 &
  958.65 &
  0.92 &
  0.34 &
  2,108.88 &
  1,105.91 &
  24,748.55 &
  25,219.81 &
  0.89 &
  0.09 \\
\textbf{LGBM} &
  100.85 &
  58.20 &
  1,163.55 &
  974.79 &
  0.64 &
  0.20 &
  283.25 &
  151.31 &
  2,066.16 &
  1,015.90 &
  0.61 &
  0.09 &
  681.31 &
  277.35 &
  3,714.49 &
  995.86 &
  0.48 &
  0.11 \\
\textbf{Linear SVR} &
  64.43 &
  9.42 &
  749.39 &
  15.07 &
  0.66 &
  0.27 &
  208.63 &
  104.34 &
  1,330.17 &
  373.68 &
  1.26 &
  0.51 &
  898.33 &
  488.94 &
  4,434.00 &
  2,103.88 &
  0.84 &
  0.35 \\
\textbf{NuSVR} &
  69.58 &
  7.74 &
  668.88 &
  75.24 &
  0.69 &
  0.31 &
  271.07 &
  131.83 &
  2,146.04 &
  748.33 &
  0.63 &
  0.07 &
  821.21 &
  385.28 &
  4,172.45 &
  1,534.14 &
  0.53 &
  0.14 \\
\textbf{Passive Aggressive} &
  109.67 &
  52.64 &
  1,835.38 &
  908.67 &
  0.34 &
  0.13 &
  298.56 &
  135.27 &
  2,121.33 &
  509.49 &
  0.81 &
  0.38 &
  1,192.12 &
  636.62 &
  6,796.09 &
  3,098.40 &
  1.11 &
  0.15 \\
\textbf{SGD} &
  76.74 &
  34.66 &
  815.18 &
  631.38 &
  0.51 &
  0.26 &
  282.62 &
  147.38 &
  2,071.76 &
  952.88 &
  0.46 &
  0.03 &
  934.62 &
  420.51 &
  4,603.72 &
  1,524.15 &
  0.44 &
  0.13 \\
\textbf{SVR} &
  79.75 &
  13.15 &
  784.25 &
  17.13 &
  0.44 &
  0.23 &
  244.21 &
  101.24 &
  1,973.38 &
  513.88 &
  0.71 &
  0.25 &
  812.32 &
  400.33 &
  4,135.83 &
  1,547.69 &
  0.46 &
  0.19 \\
\textbf{Theil-Sen} &
  91.26 &
  48.34 &
  1,047.89 &
  808.99 &
  0.62 &
  0.27 &
  240.39 &
  125.96 &
  2,033.75 &
  963.63 &
  0.60 &
  0.10 &
  842.23 &
  415.84 &
  5,116.64 &
  2,179.12 &
  0.97 &
  0.16 \\
\textbf{Transformed Target} &
  282.75 &
  159.14 &
  3,846.25 &
  2,141.83 &
  1.56 &
  1.03 &
  *** &
  *** &
  *** &
  *** &
  *** &
  *** &
  *** &
  *** &
  *** &
  *** &
  *** &
  *** \\
\textbf{XGBoost} &
  45.40 &
  8.43 &
  499.40 &
  5.96 &
  0.56 &
  0.34 &
  207.96 &
  89.25 &
  1,372.12 &
  233.16 &
  0.49 &
  0.01 &
  604.45 &
  272.50 &
  3,454.03 &
  1,583.76 &
  0.47 &
  0.20 \\ \cline{2-19} 
 &
  \multicolumn{18}{c|}{\textbf{Time Series Algorithms}} \\ \cline{2-19} 
\textbf{Autoregressive} &
  348.60 &
  141.89 &
  4,511.58 &
  1,800.83 &
  1.95 &
  1.04 &
  565.61 &
  276.92 &
  4,912.10 &
  2,148.18 &
  5.19 &
  2.79 &
  2,269.85 &
  1,375.06 &
  13,416.69 &
  8,254.88 &
  7.92 &
  3.76 \\
\textbf{ARIMA} &
  349.35 &
  141.59 &
  4,516.17 &
  1,794.31 &
  1.94 &
  1.04 &
  419.66 &
  171.52 &
  4,625.91 &
  1,941.85 &
  3.31 &
  1.51 &
  1,826.65 &
  1,039.10 &
  12,448.50 &
  7,422.94 &
  6.22 &
  2.46 \\
\textbf{ARMA} &
  73.97 &
  92.88 &
  1,010.59 &
  1,311.22 &
  1.16 &
  0.06 &
  667.46 &
  186.38 &
  5,897.04 &
  763.92 &
  6.87 &
  3.23 &
  1,860.71 &
  818.25 &
  13,319.47 &
  5,341.02 &
  11.30 &
  4.78 \\
\textbf{Moving Average} &
  353.80 &
  136.70 &
  4,589.26 &
  1,691.54 &
  1.50 &
  0.76 &
  485.71 &
  166.86 &
  4,920.53 &
  1,549.24 &
  2.84 &
  1.25 &
  1,357.32 &
  650.09 &
  11,027.22 &
  6,049.53 &
  3.38 &
  0.61 \\
\textbf{SARIMA} &
  \textbf{54.29} &
  \textbf{67.17} &
  \textbf{716.26} &
  \textbf{892.21} &
  \textbf{0.80} &
  \textbf{0.21} &
  72.83 &
  73.86 &
  689.27 &
  839.55 &
  2.37 &
  0.70 &
  145.76 &
  146.20 &
  789.46 &
  799.24 &
  2.38 &
  0.80 \\
\textbf{Exponential Smoothing} &
  54.79 &
  70.22 &
  737.75 &
  926.05 &
  0.68 &
  0.12 &
  \textbf{71.39} &
  \textbf{84.12} &
  \textbf{709.93} &
  \textbf{937.92} &
  \textbf{2.33} &
  \textbf{0.73} &
  \textbf{127.92} &
  \textbf{169.73} &
  \textbf{711.72} &
  \textbf{906.47} &
  \textbf{2.16} &
  \textbf{0.51} \\
\textbf{Vector Autoregression} &
  364.13 &
  111.03 &
  4,811.29 &
  1,376.14 &
  1.73 &
  0.39 &
  478.40 &
  58.18 &
  5,603.67 &
  557.96 &
  2.61 &
  0.42 &
  1,148.39 &
  362.46 &
  12,038.19 &
  4,524.07 &
  3.18 &
  0.22 \\ \cline{2-19} 
\end{tabular}%
}
\end{table}
\end{landscape}\clearpage
\loadgeometry{geom}

\savegeometry{geom}
\newgeometry{margin=0pt}
\thispagestyle{empty}\setcounter{page}{21}
\begin{landscape}
\begin{table}
\centering
\caption[Baselines on 90 and 105 days, and also on the complete SARS-CoV-2 dataset.]%
{Detailed results for the last three slices of the SARS-CoV-2.\\%
{\footnotesize\hspace{1.0cm}\textbf{Legend:} Algorithms with best performance are in \textbf{bold}, the ones noted as --- yielded exceptions, and others as *** were suppressed due to poor performance.}}
\label{tab:baselines-3}
\resizebox{1.35\textwidth}{!}{%
\begin{tabular}{r|cccccc|cccccc|cccccc|}
\cline{2-19}
\multicolumn{1}{l|}{} &
  \multicolumn{6}{c|}{\textbf{90 Days}} &
  \multicolumn{6}{c|}{\textbf{105 Days}} &
  \multicolumn{6}{c|}{\textbf{120 Days}} \\ \cline{2-19} 
 &
  \textbf{MAE} &
  \textbf{$\pm$ STD} &
  \textbf{RMSE} &
  \textbf{$\pm$ STD} &
  \textbf{MSLE} &
  \textbf{$\pm$ STD} &
  \textbf{MAE} &
  \textbf{$\pm$ STD} &
  \textbf{RMSE} &
  \textbf{$\pm$ STD} &
  \textbf{MSLE} &
  \textbf{$\pm$ STD} &
  \textbf{MAE} &
  \textbf{$\pm$ STD} &
  \textbf{RMSE} &
  \textbf{$\pm$ STD} &
  \textbf{MSLE} &
  \textbf{$\pm$ STD} \\ \cline{2-19} 
\multicolumn{1}{l|}{} &
  \multicolumn{18}{c|}{\textbf{Deep Learning Algorithms}} \\ \cline{2-19} 
\ReGENN &
  \textbf{152.61} &
  \textbf{113.07} &
  \textbf{986.61} &
  \textbf{827.13} &
  \textbf{0.19} &
  \textbf{0.03} &
  \textbf{179.53} &
  \textbf{62.95} &
  \textbf{1,241.91} &
  \textbf{594.23} &
  \textbf{0.09} &
  \textbf{0.03} &
  \textbf{165.41} &
  \textbf{30.37} &
  \textbf{915.92} &
  \textbf{294.06} &
  \textbf{0.05} &
  \textbf{0.02} \\
\textbf{MLCNN} &
  4976.73 &
  26967.49 &
  6301.34 &
  34530.94 &
  17.72 &
  18.67 &
  4525.83 &
  14092.01 &
  5551.06 &
  17189.83 &
  12.78 &
  19.26 &
  10749.65 &
  64733.89 &
  13172.62 &
  79729.95 &
  15.55 &
  23.33 \\
\textbf{DSANet} &
  7762.14 &
  35162.14 &
  8875.20 &
  41308.52 &
  36.96 &
  31.45 &
  12466.20 &
  58693.35 &
  13910.18 &
  67326.81 &
  45.43 &
  34.40 &
  18428.63 &
  81222.22 &
  20131.28 &
  91468.66 &
  52.90 &
  36.86 \\
\textbf{LSTNet} &
  5698.94 &
  32591.40 &
  6658.76 &
  39067.23 &
  18.44 &
  27.38 &
  6080.57 &
  53703.70 &
  7050.81 &
  62425.81 &
  15.62 &
  27.35 &
  5298.30 &
  22152.36 &
  5779.21 &
  23677.93 &
  16.27 &
  32.19 \\ \cline{2-19} 
 &
  \multicolumn{18}{c|}{\textbf{Multi-Output and Multi-Task Algorithms}} \\ \cline{2-19} 
\textbf{CCA} &
  2,341.82 &
  651.84 &
  11,696.07 &
  3,087.75 &
  1.32 &
  0.67 &
  2,309.74 &
  394.06 &
  10,585.08 &
  1,772.80 &
  0.97 &
  0.97 &
  5,566.74 &
  618.65 &
  48,941.52 &
  3,308.99 &
  0.50 &
  0.22 \\
\textbf{Decision Tree} &
  \textbf{1,070.55} &
  \textbf{362.37} &
  \textbf{4,702.47} &
  \textbf{1,440.38} &
  \textbf{0.38} &
  \textbf{0.20} &
  1,618.26 &
  206.36 &
  8,656.78 &
  1,525.96 &
  0.27 &
  0.09 &
  1,098.77 &
  163.13 &
  6,055.55 &
  2,122.73 &
  0.27 &
  0.05 \\
\textbf{Extra Tree} &
  1,298.45 &
  457.81 &
  6,664.91 &
  2,554.38 &
  0.38 &
  0.12 &
  1,403.07 &
  566.50 &
  7,279.55 &
  3,330.30 &
  0.26 &
  0.02 &
  851.65 &
  12.55 &
  3,316.89 &
  276.68 &
  0.25 &
  0.05 \\
\textbf{Extra Trees} &
  1,125.67 &
  487.87 &
  5,238.64 &
  2,386.86 &
  0.22 &
  0.10 &
  1,213.77 &
  281.65 &
  5,371.02 &
  1,217.81 &
  0.15 &
  0.02 &
  771.33 &
  120.86 &
  3,823.39 &
  5.44 &
  0.14 &
  0.08 \\
\textbf{Gaussian Process} &
  1,727.28 &
  781.46 &
  7,647.19 &
  2,479.27 &
  2.19 &
  0.99 &
  2,137.51 &
  411.08 &
  8,600.10 &
  541.87 &
  1.38 &
  0.07 &
  2,730.59 &
  309.13 &
  14,295.71 &
  3,034.59 &
  0.57 &
  0.47 \\
\textbf{Historical Average} &
  2,127.99 &
  606.38 &
  10,609.89 &
  3,223.15 &
  0.34 &
  0.19 &
  2,079.79 &
  413.43 &
  9,428.33 &
  2,131.05 &
  0.22 &
  0.00 &
  2,204.09 &
  496.63 &
  9,540.29 &
  2,570.76 &
  0.16 &
  0.09 \\
\textbf{Kernel Ridge} &
  1,371.02 &
  337.03 &
  7,263.13 &
  2,026.30 &
  0.26 &
  0.18 &
  1,295.63 &
  318.42 &
  5,289.67 &
  1,241.99 &
  0.17 &
  0.03 &
  797.81 &
  78.97 &
  2,917.93 &
  162.53 &
  0.11 &
  0.04 \\
\textbf{KNeighbors} &
  1,342.60 &
  537.35 &
  7,314.54 &
  3,096.09 &
  0.22 &
  0.12 &
  1,197.44 &
  249.97 &
  4,798.34 &
  862.09 &
  0.17 &
  0.02 &
  967.37 &
  36.99 &
  5,414.50 &
  1,014.23 &
  0.13 &
  0.07 \\
\textbf{Lars} &
  *** &
  *** &
  *** &
  *** &
  *** &
  *** &
  --- &
  --- &
  --- &
  --- &
  --- &
  --- &
  *** &
  *** &
  *** &
  *** &
  *** &
  *** \\
\textbf{Lasso-Lars} &
  2,127.99 &
  606.38 &
  10,609.89 &
  3,223.15 &
  0.34 &
  0.19 &
  2,079.79 &
  413.42 &
  9,428.33 &
  2,131.05 &
  0.22 &
  0.00 &
  2,204.09 &
  496.63 &
  9,540.29 &
  2,570.76 &
  0.16 &
  0.09 \\
\textbf{Linear Regression} &
  *** &
  *** &
  *** &
  *** &
  *** &
  *** &
  *** &
  *** &
  *** &
  *** &
  *** &
  *** &
  *** &
  *** &
  *** &
  *** &
  *** &
  *** \\
\textbf{Multi-Task Elastic-Net} &
  2,127.99 &
  606.38 &
  10,609.89 &
  3,223.15 &
  0.34 &
  0.19 &
  2,079.79 &
  413.43 &
  9,428.33 &
  2,131.05 &
  0.22 &
  0.00 &
  2,204.09 &
  496.63 &
  9,540.29 &
  2,570.76 &
  0.16 &
  0.09 \\
\textbf{Multi-Task Lasso} &
  2,127.99 &
  606.38 &
  10,609.89 &
  3,223.15 &
  0.34 &
  0.19 &
  2,079.79 &
  413.43 &
  9,428.33 &
  2,131.05 &
  0.22 &
  0.00 &
  2,204.09 &
  496.63 &
  9,540.29 &
  2,570.76 &
  0.16 &
  0.09 \\
\textbf{Orthogonal Matching Pursuit} &
  5,124.70 &
  3,058.36 &
  62,230.48 &
  56,134.60 &
  0.28 &
  0.18 &
  3,123.49 &
  802.13 &
  29,161.34 &
  5,194.85 &
  0.19 &
  0.00 &
  2,142.70 &
  897.09 &
  19,881.75 &
  10,964.98 &
  0.17 &
  0.09 \\
\textbf{PLS Canonical} &
  7,129.60 &
  3,477.83 &
  106,376.76 &
  78,979.43 &
  0.84 &
  0.82 &
  9,943.51 &
  5,019.18 &
  135,467.09 &
  100,867.91 &
  0.63 &
  0.28 &
  8,743.59 &
  2,964.01 &
  84,481.55 &
  58,065.34 &
  0.39 &
  0.11 \\
\textbf{PLS} &
  3,818.23 &
  2,606.97 &
  39,133.11 &
  37,842.39 &
  0.29 &
  0.23 &
  2,316.76 &
  1,496.43 &
  20,758.84 &
  22,298.29 &
  0.19 &
  0.02 &
  1,341.96 &
  155.97 &
  6,461.25 &
  322.42 &
  0.50 &
  0.01 \\
\textbf{Radius Neighbors} &
  --- &
  --- &
  --- &
  --- &
  --- &
  --- &
  --- &
  --- &
  --- &
  --- &
  --- &
  --- &
  --- &
  --- &
  --- &
  --- &
  --- &
  --- \\
\textbf{Random Forest} &
  1,144.34 &
  416.79 &
  5,292.91 &
  1,902.90 &
  0.21 &
  0.10 &
  \textbf{949.42} &
  \textbf{132.46} &
  \textbf{4,309.11} &
  \textbf{704.34} &
  \textbf{0.15} &
  \textbf{0.00} &
  \textbf{736.11} &
  \textbf{53.37} &
  \textbf{3,623.38} &
  \textbf{690.48} &
  \textbf{0.14} &
  \textbf{0.07} \\
\textbf{RANSAC} &
  --- &
  --- &
  --- &
  --- &
  --- &
  --- &
  --- &
  --- &
  --- &
  --- &
  --- &
  --- &
  --- &
  --- &
  --- &
  --- &
  --- &
  --- \\
\textbf{Ridge} &
  1,373.02 &
  355.28 &
  7,295.49 &
  2,075.05 &
  0.21 &
  0.11 &
  1,307.54 &
  317.33 &
  5,339.62 &
  1,231.87 &
  0.16 &
  0.00 &
  804.49 &
  86.67 &
  2,957.27 &
  191.77 &
  0.14 &
  0.08 \\ \cline{2-19} 
 &
  \multicolumn{18}{c|}{\textbf{Single-Target Algorithms on Chain Ensemble}} \\ \cline{2-19} 
\textbf{AdaBoost} &
  1,335.46 &
  518.11 &
  6,967.35 &
  2,706.46 &
  0.21 &
  0.11 &
  1,129.36 &
  200.58 &
  4,745.23 &
  771.75 &
  0.19 &
  0.02 &
  1,099.84 &
  337.60 &
  7,184.23 &
  5,335.13 &
  0.15 &
  0.09 \\
\textbf{ARD} &
  3,673.37 &
  1,994.35 &
  43,202.60 &
  40,595.97 &
  0.27 &
  0.19 &
  1,765.81 &
  130.95 &
  13,621.08 &
  1,163.70 &
  0.18 &
  0.00 &
  889.47 &
  30.26 &
  3,368.50 &
  207.97 &
  0.16 &
  0.08 \\
\textbf{Bagging} &
  \textbf{1,120.38} &
  \textbf{425.26} &
  \textbf{5,256.17} &
  \textbf{2,099.54} &
  \textbf{0.22} &
  \textbf{0.11} &
  1,024.42 &
  124.09 &
  4,538.16 &
  604.34 &
  0.15 &
  0.02 &
  862.94 &
  115.31 &
  4,244.29 &
  116.49 &
  0.17 &
  0.06 \\
\textbf{Bayesian Ridge} &
  1,354.00 &
  341.24 &
  7,238.96 &
  2,090.06 &
  0.22 &
  0.13 &
  1,288.36 &
  308.33 &
  5,354.24 &
  1,223.32 &
  0.17 &
  0.01 &
  828.14 &
  12.06 &
  3,031.29 &
  145.24 &
  0.16 &
  0.06 \\
\textbf{CatBoost} &
  1,360.99 &
  437.43 &
  6,909.28 &
  1,865.61 &
  0.22 &
  0.12 &
  1,328.43 &
  303.93 &
  5,494.43 &
  940.83 &
  0.16 &
  0.00 &
  1,115.88 &
  85.29 &
  4,129.19 &
  139.31 &
  0.14 &
  0.08 \\
\textbf{Gradient Boosting} &
  1,182.57 &
  409.10 &
  5,680.21 &
  2,013.45 &
  0.23 &
  0.13 &
  1,055.45 &
  99.35 &
  5,995.73 &
  1,150.59 &
  0.16 &
  0.02 &
  \textbf{800.29} &
  \textbf{1.54} &
  \textbf{4,722.74} &
  \textbf{1,271.82} &
  \textbf{0.16} &
  \textbf{0.06} \\
\textbf{Histogram Grad. Boosting} &
  1,284.27 &
  330.98 &
  6,143.09 &
  1,249.58 &
  0.43 &
  0.37 &
  1,341.97 &
  277.69 &
  9,450.66 &
  3,843.12 &
  0.23 &
  0.00 &
  866.30 &
  38.80 &
  4,612.11 &
  44.68 &
  0.16 &
  0.08 \\
\textbf{Huber} &
  1,433.68 &
  438.15 &
  7,571.89 &
  2,061.11 &
  0.93 &
  0.22 &
  1,688.87 &
  117.01 &
  10,555.93 &
  808.14 &
  0.62 &
  0.04 &
  2,032.72 &
  41.78 &
  17,770.19 &
  610.28 &
  0.37 &
  0.08 \\
\textbf{Isotonic} &
  --- &
  --- &
  --- &
  --- &
  --- &
  --- &
  --- &
  --- &
  --- &
  --- &
  --- &
  --- &
  --- &
  --- &
  --- &
  --- &
  --- &
  --- \\
\textbf{Lasso-Lars-IC} &
  3,844.71 &
  2,113.02 &
  46,007.40 &
  43,846.83 &
  0.27 &
  0.18 &
  2,808.14 &
  844.84 &
  30,186.71 &
  9,531.58 &
  0.18 &
  0.01 &
  810.03 &
  30.97 &
  3,929.81 &
  129.59 &
  0.14 &
  0.08 \\
\textbf{LGBM} &
  1,151.12 &
  217.85 &
  5,854.56 &
  1,281.29 &
  0.28 &
  0.21 &
  \textbf{870.05} &
  \textbf{15.88} &
  \textbf{4,611.39} &
  \textbf{516.98} &
  \textbf{0.17} &
  \textbf{0.01} &
  1,110.17 &
  198.23 &
  7,650.02 &
  3,822.07 &
  0.28 &
  0.01 \\
\textbf{Linear SVR} &
  1,349.30 &
  265.00 &
  7,580.89 &
  1,602.24 &
  0.29 &
  0.22 &
  917.52 &
  9.41 &
  3,773.66 &
  127.97 &
  0.19 &
  0.05 &
  875.51 &
  97.79 &
  3,540.90 &
  247.08 &
  0.16 &
  0.01 \\
\textbf{NuSVR} &
  1,505.71 &
  493.60 &
  7,609.47 &
  2,106.44 &
  0.24 &
  0.13 &
  943.65 &
  152.83 &
  3,635.38 &
  312.43 &
  0.16 &
  0.01 &
  809.97 &
  19.17 &
  2,973.63 &
  21.61 &
  0.12 &
  0.06 \\
\textbf{Passive Aggressive} &
  1,513.43 &
  811.55 &
  9,090.06 &
  5,056.45 &
  0.29 &
  0.13 &
  2,131.91 &
  827.72 &
  12,195.75 &
  7,398.69 &
  0.32 &
  0.13 &
  1,730.12 &
  113.18 &
  8,310.46 &
  432.48 &
  0.18 &
  0.12 \\
\textbf{SGD} &
  1,152.69 &
  218.01 &
  6,126.69 &
  1,382.46 &
  0.22 &
  0.14 &
  958.29 &
  93.84 &
  3,742.62 &
  146.27 &
  0.15 &
  0.00 &
  815.97 &
  11.37 &
  3,106.47 &
  40.02 &
  0.13 &
  0.06 \\
\textbf{SVR} &
  1,561.92 &
  562.85 &
  7,624.46 &
  2,272.31 &
  0.23 &
  0.12 &
  1,216.19 &
  298.18 &
  4,831.88 &
  1,087.95 &
  0.16 &
  0.01 &
  1,344.80 &
  244.56 &
  5,511.70 &
  1,539.93 &
  0.12 &
  0.06 \\
\textbf{Theil-Sen} &
  2,330.21 &
  1,126.08 &
  11,210.84 &
  4,841.23 &
  4.04 &
  2.43 &
  4,800.68 &
  900.31 &
  21,180.70 &
  1,794.50 &
  6.07 &
  1.18 &
  91,149.14 &
  59,396.23 &
  1,292,430.04 &
  858,201.91 &
  6.45 &
  2.47 \\
\textbf{Transformed Target} &
  10,464.84 &
  1,595.88 &
  119,669.67 &
  56,077.14 &
  14.32 &
  0.93 &
  36,788.18 &
  37,380.39 &
  721,732.88 &
  945,437.62 &
  16.99 &
  4.01 &
  68,530.56 &
  55,751.11 &
  868,586.61 &
  989,030.54 &
  16.39 &
  1.00 \\
\textbf{XGBoost} &
  1,252.61 &
  426.10 &
  6,089.49 &
  2,145.76 &
  0.30 &
  0.22 &
  1,070.54 &
  142.59 &
  5,820.65 &
  1,367.38 &
  0.16 &
  0.01 &
  806.03 &
  37.02 &
  5,033.10 &
  1,599.32 &
  0.16 &
  0.06 \\ \cline{2-19} 
 &
  \multicolumn{18}{c|}{\textbf{Time Series Algorithms}} \\ \cline{2-19} 
\textbf{Autoregressive} &
  5,621.04 &
  3,216.40 &
  35,169.53 &
  21,541.40 &
  8.58 &
  4.33 &
  8,788.66 &
  5,007.74 &
  57,016.07 &
  35,534.58 &
  8.94 &
  5.44 &
  11,529.79 &
  6,752.42 &
  76,118.62 &
  48,374.82 &
  9.15 &
  6.14 \\
\textbf{ARIMA} &
  3,548.24 &
  1,712.65 &
  32,183.46 &
  19,248.04 &
  5.48 &
  1.49 &
  8,845.58 &
  5,039.42 &
  57,008.21 &
  35,546.27 &
  9.83 &
  4.19 &
  11,578.80 &
  6,798.84 &
  76,044.64 &
  48,478.91 &
  9.22 &
  5.23 \\
\textbf{ARMA} &
  4,913.53 &
  2,525.52 &
  34,314.04 &
  20,370.73 &
  10.11 &
  1.05 &
  8,719.13 &
  3,958.49 &
  58,083.03 &
  32,762.97 &
  14.45 &
  3.20 &
  12,578.49 &
  6,091.86 &
  79,263.71 &
  45,555.46 &
  16.71 &
  4.79 \\
\textbf{Moving Average} &
  3,521.66 &
  1,540.40 &
  31,763.66 &
  18,590.82 &
  3.10 &
  0.27 &
  5,780.07 &
  2,266.98 &
  53,814.10 &
  31,461.03 &
  2.85 &
  0.70 &
  7,757.33 &
  2,887.33 &
  73,013.22 &
  42,555.86 &
  2.54 &
  1.17 \\
\textbf{SARIMA} &
  429.33 &
  489.53 &
  2,047.34 &
  2,073.95 &
  1.18 &
  1.39 &
  11,171.46 &
  6,702.33 &
  279,332.27 &
  192,747.63 &
  1.03 &
  0.40 &
  *** &
  *** &
  *** &
  *** &
  *** &
  *** \\
\textbf{Exponential Smoothing} &
  \textbf{381.55} &
  \textbf{538.84} &
  \textbf{1,692.42} &
  \textbf{2,384.16} &
  \textbf{1.06} &
  \textbf{1.21} &
  \textbf{576.39} &
  \textbf{814.72} &
  \textbf{2,300.48} &
  \textbf{3,242.27} &
  \textbf{0.55} &
  \textbf{0.71} &
  \textbf{667.33} &
  \textbf{940.85} &
  \textbf{2,607.29} &
  \textbf{3,634.47} &
  \textbf{0.27} &
  \textbf{0.17} \\
\textbf{Vector Autoregression} &
  *** &
  *** &
  *** &
  *** &
  *** &
  *** &
  12,985.56 &
  9,735.09 &
  270,265.90 &
  274,777.24 &
  3.23 &
  0.02 &
  6,395.14 &
  2,397.58 &
  75,292.43 &
  39,167.56 &
  3.21 &
  0.05 \\ \cline{2-19} 
\end{tabular}%
}
\end{table}
\end{landscape}\clearpage
\loadgeometry{geom}

\savegeometry{geom}
\newgeometry{margin=0pt}
\thispagestyle{empty}\setcounter{page}{22}
\begin{landscape}
\begin{table}
\centering
\caption[Ablation on the first\ \ 45 days of the SARS-CoV-2 dataset.]%
{Ablation on the first 45 days of the SARS-CoV-2 dataset.\\%
{\footnotesize\textbf{Legend:} Algorithms with the best performance are in \textbf{bold}; besides, we refer to the Transformer Encoder as $\mathrm{E}$, recurrent unit as \textsc{RU}, Bidirectional as \textsc{B}, Unidirectional as \textsc{U}, and Autoregressive as \textsc{AR}.}}
\label{tab:ablation-4}
\resizebox{1.35\textwidth}{!}{%
\begin{tabular}{r|cccccc|cccccc|cccccc|}
\cline{2-19}
\multicolumn{1}{c|}{} &
  \multicolumn{6}{c|}{\textbf{PyTorch's Hyperparameters}} &
  \multicolumn{6}{c|}{\textbf{Literature's Hyperparameters}} &
  \multicolumn{6}{c|}{\ReGENN\textbf{'s Hyperparameters}} \\
\cline{2-19}
\multicolumn{1}{c|}{} &
  \multicolumn{1}{c}{\textbf{MAE}} &
  \multicolumn{1}{c}{\textbf{$\pm$ STD}} &
  \multicolumn{1}{c}{\textbf{MSLE}} &
  \multicolumn{1}{c}{\textbf{$\pm$ STD}} &
  \multicolumn{1}{c}{\textbf{RMSE}} &
  \textbf{$\pm$ STD} &
  \multicolumn{1}{c}{\textbf{MAE}} &
  \multicolumn{1}{c}{\textbf{$\pm$ STD}} &
  \multicolumn{1}{c}{\textbf{MSLE}} &
  \multicolumn{1}{c}{\textbf{$\pm$ STD}} &
  \multicolumn{1}{c}{\textbf{RMSE}} &
  \textbf{$\pm$ STD} &
  \multicolumn{1}{c}{\textbf{MAE}} &
  \multicolumn{1}{c}{\textbf{$\pm$ STD}} &
  \multicolumn{1}{c}{\textbf{MSLE}} &
  \multicolumn{1}{c}{\textbf{$\pm$ STD}} &
  \multicolumn{1}{c}{\textbf{RMSE}} &
  \textbf{$\pm$ STD} \\ \cline{2-19} 
\multicolumn{1}{l|}{} &
  \multicolumn{18}{c|}{\textbf{Recurrent Neural Network (RNN)}} \\ \cline{2-19} 
${}_{\mathrm{B}}\text{\textsc{RU}}$ &
  92.38 &
  168.24 &
  983.14 &
  1,820.04 &
  0.60 &
  0.14 &
  81.60 &
  146.62 &
  759.77 &
  1,410.46 &
  0.44 &
  0.05 &
  112.86 &
  202.94 &
  842.04 &
  1,532.05 &
  0.50 &
  0.19 \\
$\mathrm{E} \rightarrow {}_{\mathrm{B}}\text{\textsc{RU}}$ &
  154.44 &
  294.30 &
  1,312.81 &
  2,517.02 &
  0.87 &
  0.33 &
  134.87 &
  260.04 &
  1,232.46 &
  2,394.62 &
  0.86 &
  0.56 &
  227.58 &
  426.24 &
  1,667.99 &
  3,102.15 &
  1.36 &
  0.84 \\
$\left(\mathrm{E} \rightarrow {}_{\mathrm{B}}\text{\textsc{RU}} + {}_{\mathrm{B}}\text{\textsc{RU}}\right) + \mathrm{AR}$ &
  123.91 &
  236.71 &
  1,081.02 &
  2,072.41 &
  0.78 &
  0.61 &
  \textbf{38.92} &
  \textbf{73.72} &
  \textbf{389.25} &
  \textbf{753.11} &
  \textbf{0.39} &
  \textbf{0.09} &
  43.86 &
  76.12 &
  385.74 &
  689.40 &
  0.39 &
  0.13 \\
$\left(\mathrm{E} \rightarrow {}_{\mathrm{B}}\text{\textsc{RU}} + {}_{\mathrm{U}}\text{\textsc{RU}}\right) + \mathrm{AR}$ &
  130.66 &
  244.62 &
  1,127.02 &
  2,128.93 &
  0.83 &
  0.42 &
  38.93 &
  72.00 &
  401.29 &
  758.97 &
  0.40 &
  0.06 &
  40.10 &
  71.31 &
  369.97 &
  668.56 &
  0.35 &
  0.08 \\
$\left(\mathrm{E} \rightarrow {}_{\mathrm{B}}\text{\textsc{RU}}\right) + \mathrm{AR}$ &
  117.76 &
  224.79 &
  1,096.91 &
  2,125.03 &
  0.78 &
  0.39 &
  86.93 &
  167.43 &
  928.81 &
  1,812.63 &
  0.67 &
  0.31 &
  52.23 &
  94.22 &
  519.21 &
  947.12 &
  0.38 &
  0.11 \\
$\mathrm{E} \rightarrow {}_{\mathrm{U}}\text{\textsc{RU}}$ &
  161.50 &
  301.14 &
  1,393.98 &
  2,596.00 &
  1.03 &
  0.44 &
  114.50 &
  218.48 &
  969.78 &
  1,869.53 &
  0.74 &
  0.44 &
  120.86 &
  213.88 &
  888.87 &
  1,580.65 &
  0.76 &
  0.40 \\
$\left(\mathrm{E} \rightarrow {}_{\mathrm{U}}\text{\textsc{RU}} + {}_{\mathrm{B}}\text{\textsc{RU}}\right) + \mathrm{AR}$ &
  125.39 &
  236.08 &
  968.55 &
  1,836.12 &
  0.73 &
  0.44 &
  60.47 &
  110.75 &
  717.60 &
  1,351.65 &
  0.48 &
  0.08 &
  33.97 &
  60.20 &
  378.05 &
  685.64 &
  0.31 &
  0.08 \\
$\left(\mathrm{E} \rightarrow {}_{\mathrm{U}}\text{\textsc{RU}} + {}_{\mathrm{U}}\text{\textsc{RU}}\right) + \mathrm{AR}$ &
  143.91 &
  252.03 &
  1,365.39 &
  2,416.99 &
  1.11 &
  0.47 &
  42.22 &
  74.12 &
  388.16 &
  689.13 &
  0.40 &
  0.17 &
  \textbf{33.47} &
  \textbf{59.33} &
  \textbf{346.66} &
  \textbf{616.35} &
  \textbf{0.30} &
  \textbf{0.10} \\
$\left(\mathrm{E} \rightarrow {}_{\mathrm{U}}\text{\textsc{RU}}\right) + \mathrm{AR}$ &
  \textbf{73.49} &
  \textbf{133.81} &
  \textbf{895.15} &
  \textbf{1,689.02} &
  \textbf{0.58} &
  \textbf{0.21} &
  41.19 &
  76.65 &
  423.19 &
  810.52 &
  0.36 &
  0.07 &
  33.56 &
  59.79 &
  381.37 &
  697.52 &
  0.30 &
  0.17 \\
${}_{\mathrm{U}}\text{\textsc{RU}}$ &
  152.68 &
  283.80 &
  1,419.21 &
  2,639.93 &
  1.06 &
  0.64 &
  107.50 &
  194.83 &
  961.91 &
  1,773.30 &
  0.63 &
  0.17 &
  132.75 &
  248.92 &
  890.12 &
  1,664.24 &
  0.68 &
  0.36 \\ \cline{2-19} 
\multicolumn{1}{l|}{} &
  \multicolumn{18}{c|}{\textbf{Gated Recurrent Unit (GRU)}} \\ \cline{2-19} 
${}_{\mathrm{B}}\text{\textsc{RU}}$ &
  139.53 &
  251.27 &
  1,351.34 &
  2,425.51 &
  1.06 &
  0.30 &
  110.63 &
  193.52 &
  1,050.93 &
  1,867.69 &
  0.86 &
  0.33 &
  131.26 &
  229.29 &
  1,036.19 &
  1,772.92 &
  0.68 &
  0.10 \\
$\mathrm{E} \rightarrow {}_{\mathrm{B}}\text{\textsc{RU}}$ &
  119.39 &
  234.78 &
  928.39 &
  1,825.68 &
  0.39 &
  0.19 &
  70.22 &
  132.42 &
  612.60 &
  1,172.71 &
  0.48 &
  0.13 &
  71.47 &
  131.55 &
  560.74 &
  1,045.89 &
  0.42 &
  0.14 \\
$\left(\mathrm{E} \rightarrow {}_{\mathrm{B}}\text{\textsc{RU}} + {}_{\mathrm{B}}\text{\textsc{RU}}\right) + \mathrm{AR}$ &
  266.39 &
  493.08 &
  1,837.22 &
  3,355.49 &
  0.92 &
  0.28 &
  \textbf{37.16} &
  \textbf{67.72} &
  \textbf{412.10} &
  \textbf{757.24} &
  \textbf{0.29} &
  \textbf{0.14} &
  42.77 &
  69.71 &
  380.37 &
  639.67 &
  0.38 &
  0.12 \\
$\left(\mathrm{E} \rightarrow {}_{\mathrm{B}}\text{\textsc{RU}} + {}_{\mathrm{U}}\text{\textsc{RU}}\right) + \mathrm{AR}$ &
  252.54 &
  464.95 &
  1,801.24 &
  3,242.68 &
  1.24 &
  0.33 &
  60.86 &
  116.71 &
  705.36 &
  1,373.44 &
  0.50 &
  0.20 &
  \textbf{35.42} &
  \textbf{62.81} &
  \textbf{396.71} &
  \textbf{720.01} &
  \textbf{0.35} &
  \textbf{0.05} \\
$\left(\mathrm{E} \rightarrow {}_{\mathrm{B}}\text{\textsc{RU}}\right) + \mathrm{AR}$ &
  140.96 &
  254.21 &
  1,311.44 &
  2,259.16 &
  0.92 &
  0.29 &
  39.24 &
  73.68 &
  414.09 &
  797.69 &
  0.38 &
  0.06 &
  57.46 &
  106.34 &
  700.14 &
  1,334.55 &
  0.45 &
  0.14 \\
$\mathrm{E} \rightarrow {}_{\mathrm{U}}\text{\textsc{RU}}$ &
  \textbf{110.85} &
  \textbf{202.79} &
  \textbf{1,127.41} &
  \textbf{2,015.28} &
  \textbf{0.92} &
  \textbf{0.39} &
  119.25 &
  217.88 &
  1,044.76 &
  1,833.26 &
  0.70 &
  0.20 &
  137.64 &
  259.08 &
  1,054.46 &
  1,947.95 &
  0.75 &
  0.29 \\
$\left(\mathrm{E} \rightarrow {}_{\mathrm{U}}\text{\textsc{RU}} + {}_{\mathrm{B}}\text{\textsc{RU}}\right) + \mathrm{AR}$ &
  131.94 &
  245.52 &
  1,218.18 &
  2,240.15 &
  1.02 &
  0.43 &
  41.17 &
  73.11 &
  449.23 &
  797.88 &
  0.47 &
  0.13 &
  42.33 &
  73.97 &
  402.23 &
  721.30 &
  0.34 &
  0.10 \\
$\left(\mathrm{E} \rightarrow {}_{\mathrm{U}}\text{\textsc{RU}} + {}_{\mathrm{U}}\text{\textsc{RU}}\right) + \mathrm{AR}$ &
  219.26 &
  408.35 &
  1,698.29 &
  3,173.21 &
  1.66 &
  0.70 &
  62.52 &
  115.16 &
  744.72 &
  1,396.55 &
  0.51 &
  0.15 &
  90.67 &
  163.53 &
  1,007.57 &
  1,829.93 &
  0.75 &
  0.29 \\
$\left(\mathrm{E} \rightarrow {}_{\mathrm{U}}\text{\textsc{RU}}\right) + \mathrm{AR}$ &
  156.25 &
  275.91 &
  1,494.13 &
  2,650.03 &
  1.21 &
  0.49 &
  90.03 &
  163.76 &
  986.54 &
  1,824.13 &
  0.69 &
  0.28 &
  104.06 &
  194.04 &
  1,104.26 &
  2,083.88 &
  0.85 &
  0.34 \\
${}_{\mathrm{U}}\text{\textsc{RU}}$ &
  138.25 &
  252.42 &
  1,321.32 &
  2,399.06 &
  1.05 &
  0.58 &
  180.26 &
  340.49 &
  1,440.16 &
  2,736.77 &
  1.23 &
  0.88 &
  163.53 &
  296.28 &
  1,213.40 &
  2,181.95 &
  0.91 &
  0.42 \\ \cline{2-19} 
\multicolumn{1}{l|}{} &
  \multicolumn{18}{c|}{\textbf{Long Short-Term Memory (LSTM)}} \\ \cline{2-19} 
${}_{\mathrm{B}}\text{\textsc{RU}}$ &
  171.79 &
  328.51 &
  1,457.40 &
  2,772.59 &
  1.24 &
  0.82 &
  152.31 &
  271.20 &
  1,370.13 &
  2,445.64 &
  1.09 &
  0.37 &
  150.75 &
  274.94 &
  1,152.45 &
  2,085.77 &
  0.83 &
  0.17 \\
$\mathrm{E} \rightarrow {}_{\mathrm{B}}\text{\textsc{RU}}$ &
  114.78 &
  206.10 &
  1,068.62 &
  1,939.87 &
  0.79 &
  0.32 &
  70.51 &
  130.64 &
  562.77 &
  1,059.56 &
  0.27 &
  0.05 &
  108.36 &
  193.72 &
  775.35 &
  1,402.93 &
  0.51 &
  0.15 \\
$\left(\mathrm{E} \rightarrow {}_{\mathrm{B}}\text{\textsc{RU}} + {}_{\mathrm{B}}\text{\textsc{RU}}\right) + \mathrm{AR}$ &
  211.74 &
  385.14 &
  1,704.56 &
  3,117.28 &
  1.39 &
  0.88 &
  85.36 &
  162.34 &
  939.39 &
  1,804.93 &
  0.65 &
  0.41 &
  83.51 &
  153.19 &
  952.25 &
  1,770.62 &
  0.62 &
  0.22 \\
$\left(\mathrm{E} \rightarrow {}_{\mathrm{B}}\text{\textsc{RU}} + {}_{\mathrm{U}}\text{\textsc{RU}}\right) + \mathrm{AR}$ &
  \textbf{92.34} &
  \textbf{175.30} &
  \textbf{979.40} &
  \textbf{1,868.96} &
  \textbf{0.58} &
  \textbf{0.29} &
  62.69 &
  115.65 &
  729.74 &
  1,383.14 &
  0.50 &
  0.14 &
  37.11 &
  67.01 &
  410.27 &
  754.33 &
  0.32 &
  0.14 \\
$\left(\mathrm{E} \rightarrow {}_{\mathrm{B}}\text{\textsc{RU}}\right) + \mathrm{AR}$ &
  131.72 &
  247.70 &
  1,277.84 &
  2,410.56 &
  0.89 &
  0.37 &
  81.91 &
  151.00 &
  939.29 &
  1,758.27 &
  0.65 &
  0.28 &
  139.78 &
  258.71 &
  1,325.73 &
  2,444.80 &
  0.98 &
  0.52 \\
$\mathrm{E} \rightarrow {}_{\mathrm{U}}\text{\textsc{RU}}$ &
  200.08 &
  363.84 &
  1,563.10 &
  2,787.79 &
  1.14 &
  0.44 &
  159.26 &
  295.13 &
  1,237.13 &
  2,243.83 &
  0.81 &
  0.32 &
  119.47 &
  206.38 &
  769.78 &
  1,319.61 &
  0.50 &
  0.23 \\
$\left(\mathrm{E} \rightarrow {}_{\mathrm{U}}\text{\textsc{RU}} + {}_{\mathrm{B}}\text{\textsc{RU}}\right) + \mathrm{AR}$ &
  134.88 &
  249.97 &
  1,309.42 &
  2,359.58 &
  1.14 &
  0.42 &
  92.05 &
  167.31 &
  1,054.34 &
  1,881.80 &
  0.87 &
  0.40 &
  39.04 &
  73.06 &
  416.84 &
  798.12 &
  0.33 &
  0.12 \\
$\left(\mathrm{E} \rightarrow {}_{\mathrm{U}}\text{\textsc{RU}} + {}_{\mathrm{U}}\text{\textsc{RU}}\right) + \mathrm{AR}$ &
  113.29 &
  213.19 &
  1,120.50 &
  2,112.08 &
  0.84 &
  0.53 &
  88.22 &
  159.65 &
  1,002.30 &
  1,834.26 &
  0.71 &
  0.19 &
  61.23 &
  113.29 &
  730.30 &
  1,381.64 &
  0.47 &
  0.11 \\
$\left(\mathrm{E} \rightarrow {}_{\mathrm{U}}\text{\textsc{RU}}\right) + \mathrm{AR}$ &
  145.47 &
  262.58 &
  1,432.06 &
  2,606.44 &
  1.13 &
  0.63 &
  \textbf{39.04} &
  \textbf{67.51} &
  \textbf{420.52} &
  \textbf{750.02} &
  \textbf{0.33} &
  \textbf{0.17} &
  \textbf{32.79} &
  \textbf{58.34} &
  \textbf{378.05} &
  \textbf{688.07} &
  \textbf{0.31} &
  \textbf{0.09} \\
${}_{\mathrm{U}}\text{\textsc{RU}}$ &
  155.38 &
  277.69 &
  1,277.61 &
  2,286.99 &
  0.81 &
  0.26 &
  199.23 &
  363.06 &
  1,554.73 &
  2,793.46 &
  1.18 &
  0.37 &
  173.46 &
  321.89 &
  1,247.96 &
  2,344.72 &
  0.75 &
  0.33 \\ \cline{2-19} 
\end{tabular}%
}
\end{table}
\end{landscape}\clearpage
\loadgeometry{geom}

\savegeometry{geom}
\newgeometry{margin=0pt}
\thispagestyle{empty}\setcounter{page}{23}
\begin{landscape}
\begin{table}
\centering
\caption[Ablation on the first\ \ 60 days of the SARS-CoV-2 dataset.]%
{Ablation on the first 60 days of the SARS-CoV-2 dataset.\\%
{\footnotesize\textbf{Legend:} Algorithms with the best performance are in \textbf{bold}; besides, we refer to the Transformer Encoder as $\mathrm{E}$, recurrent unit as \textsc{RU}, Bidirectional as \textsc{B}, Unidirectional as \textsc{U}, and Autoregressive as \textsc{AR}.}}
\label{tab:ablation-5}
\resizebox{1.35\textwidth}{!}{%
\begin{tabular}{r|cccccc|cccccc|cccccc|}
\cline{2-19}
\multicolumn{1}{c|}{} &
  \multicolumn{6}{c|}{\textbf{PyTorch's Hyperparameters}} &
  \multicolumn{6}{c|}{\textbf{Literature's Hyperparameters}} &
  \multicolumn{6}{c|}{\ReGENN\textbf{'s Hyperparameters}} \\
\cline{2-19}
\multicolumn{1}{c|}{} &
  \multicolumn{1}{c}{\textbf{MAE}} &
  \multicolumn{1}{c}{\textbf{$\pm$ STD}} &
  \multicolumn{1}{c}{\textbf{MSLE}} &
  \multicolumn{1}{c}{\textbf{$\pm$ STD}} &
  \multicolumn{1}{c}{\textbf{RMSE}} &
  \textbf{$\pm$ STD} &
  \multicolumn{1}{c}{\textbf{MAE}} &
  \multicolumn{1}{c}{\textbf{$\pm$ STD}} &
  \multicolumn{1}{c}{\textbf{MSLE}} &
  \multicolumn{1}{c}{\textbf{$\pm$ STD}} &
  \multicolumn{1}{c}{\textbf{RMSE}} &
  \textbf{$\pm$ STD} &
  \multicolumn{1}{c}{\textbf{MAE}} &
  \multicolumn{1}{c}{\textbf{$\pm$ STD}} &
  \multicolumn{1}{c}{\textbf{MSLE}} &
  \multicolumn{1}{c}{\textbf{$\pm$ STD}} &
  \multicolumn{1}{c}{\textbf{RMSE}} &
  \textbf{$\pm$ STD} \\ \cline{2-19} 
\multicolumn{1}{l|}{} &
  \multicolumn{18}{c|}{\textbf{Recurrent Neural Network (RNN)}} \\ \cline{2-19} 
${}_{\mathrm{B}}\text{\textsc{RU}}$ &
  \textbf{43.57} &
  \textbf{38.12} &
  \textbf{323.55} &
  \textbf{297.46} &
  \textbf{0.71} &
  \textbf{0.09} &
  144.34 &
  118.88 &
  1,275.31 &
  1,270.70 &
  1.43 &
  0.32 &
  86.60 &
  97.98 &
  446.65 &
  498.60 &
  0.61 &
  0.12 \\
$\mathrm{E} \rightarrow {}_{\mathrm{B}}\text{\textsc{RU}}$ &
  253.53 &
  374.90 &
  1,752.47 &
  2,739.24 &
  2.90 &
  1.18 &
  219.04 &
  304.26 &
  1,575.91 &
  2,457.67 &
  2.09 &
  0.44 &
  241.50 &
  315.27 &
  1,601.64 &
  2,244.13 &
  2.36 &
  0.51 \\
$\left(\mathrm{E} \rightarrow {}_{\mathrm{B}}\text{\textsc{RU}} + {}_{\mathrm{B}}\text{\textsc{RU}}\right) + \mathrm{AR}$ &
  175.65 &
  219.43 &
  1,211.78 &
  1,656.35 &
  1.70 &
  0.24 &
  53.65 &
  59.24 &
  638.43 &
  935.69 &
  0.66 &
  0.07 &
  \textbf{30.77} &
  \textbf{16.65} &
  \textbf{243.40} &
  \textbf{185.19} &
  \textbf{0.92} &
  \textbf{0.20} \\
$\left(\mathrm{E} \rightarrow {}_{\mathrm{B}}\text{\textsc{RU}} + {}_{\mathrm{U}}\text{\textsc{RU}}\right) + \mathrm{AR}$ &
  93.77 &
  121.73 &
  922.43 &
  1,329.31 &
  1.05 &
  0.15 &
  \textbf{39.10} &
  \textbf{35.06} &
  \textbf{346.88} &
  \textbf{331.93} &
  \textbf{0.78} &
  \textbf{0.12} &
  35.03 &
  20.29 &
  239.61 &
  195.37 &
  0.80 &
  0.13 \\
$\left(\mathrm{E} \rightarrow {}_{\mathrm{B}}\text{\textsc{RU}}\right) + \mathrm{AR}$ &
  189.30 &
  229.91 &
  1,693.76 &
  2,266.29 &
  2.38 &
  0.34 &
  133.75 &
  165.63 &
  1,479.53 &
  2,016.71 &
  1.63 &
  0.32 &
  70.69 &
  61.18 &
  961.55 &
  1,201.27 &
  1.30 &
  0.20 \\
$\mathrm{E} \rightarrow {}_{\mathrm{U}}\text{\textsc{RU}}$ &
  83.38 &
  94.75 &
  1,041.24 &
  1,249.44 &
  1.29 &
  0.34 &
  106.04 &
  147.65 &
  495.28 &
  658.96 &
  0.76 &
  0.18 &
  80.69 &
  90.97 &
  430.62 &
  516.05 &
  1.51 &
  0.49 \\
$\left(\mathrm{E} \rightarrow {}_{\mathrm{U}}\text{\textsc{RU}} + {}_{\mathrm{B}}\text{\textsc{RU}}\right) + \mathrm{AR}$ &
  143.76 &
  224.86 &
  1,285.79 &
  2,042.16 &
  1.86 &
  0.46 &
  42.94 &
  21.19 &
  336.45 &
  202.77 &
  0.99 &
  0.10 &
  31.13 &
  22.61 &
  212.93 &
  173.76 &
  0.68 &
  0.22 \\
$\left(\mathrm{E} \rightarrow {}_{\mathrm{U}}\text{\textsc{RU}} + {}_{\mathrm{U}}\text{\textsc{RU}}\right) + \mathrm{AR}$ &
  85.20 &
  117.27 &
  827.96 &
  1,314.00 &
  1.27 &
  0.36 &
  81.56 &
  79.32 &
  924.11 &
  1,224.77 &
  1.37 &
  0.27 &
  31.03 &
  19.91 &
  217.80 &
  169.85 &
  0.71 &
  0.22 \\
$\left(\mathrm{E} \rightarrow {}_{\mathrm{U}}\text{\textsc{RU}}\right) + \mathrm{AR}$ &
  116.63 &
  153.58 &
  1,209.29 &
  1,793.28 &
  1.66 &
  0.34 &
  40.78 &
  35.11 &
  542.23 &
  490.54 &
  0.72 &
  0.15 &
  33.34 &
  19.26 &
  251.21 &
  193.90 &
  0.72 &
  0.09 \\
${}_{\mathrm{U}}\text{\textsc{RU}}$ &
  72.64 &
  84.28 &
  730.01 &
  880.44 &
  1.05 &
  0.19 &
  91.29 &
  68.33 &
  477.49 &
  391.91 &
  0.74 &
  0.10 &
  104.92 &
  81.72 &
  767.81 &
  525.68 &
  1.42 &
  0.50 \\ \cline{2-19} 
\multicolumn{1}{l|}{} &
  \multicolumn{18}{c|}{\textbf{Gated Recurrent Unit (GRU)}} \\ \cline{2-19} 
${}_{\mathrm{B}}\text{\textsc{RU}}$ &
  122.88 &
  106.19 &
  1,335.11 &
  1,227.14 &
  1.87 &
  0.15 &
  196.72 &
  166.02 &
  1,811.57 &
  1,657.98 &
  2.36 &
  0.27 &
  162.40 &
  174.15 &
  1,259.63 &
  1,431.54 &
  1.89 &
  0.13 \\
$\mathrm{E} \rightarrow {}_{\mathrm{B}}\text{\textsc{RU}}$ &
  70.41 &
  83.18 &
  536.49 &
  592.37 &
  0.77 &
  0.11 &
  94.89 &
  74.57 &
  564.44 &
  568.49 &
  0.80 &
  0.19 &
  175.12 &
  229.00 &
  1,371.57 &
  1,934.60 &
  1.16 &
  0.34 \\
$\left(\mathrm{E} \rightarrow {}_{\mathrm{B}}\text{\textsc{RU}} + {}_{\mathrm{B}}\text{\textsc{RU}}\right) + \mathrm{AR}$ &
  132.02 &
  112.80 &
  807.68 &
  900.76 &
  1.08 &
  0.20 &
  36.06 &
  19.09 &
  300.60 &
  198.17 &
  0.82 &
  0.21 &
  29.73 &
  22.89 &
  224.72 &
  220.17 &
  0.90 &
  0.18 \\
$\left(\mathrm{E} \rightarrow {}_{\mathrm{B}}\text{\textsc{RU}} + {}_{\mathrm{U}}\text{\textsc{RU}}\right) + \mathrm{AR}$ &
  154.86 &
  260.17 &
  892.88 &
  1,498.91 &
  1.28 &
  0.14 &
  36.37 &
  23.99 &
  351.25 &
  321.62 &
  0.73 &
  0.09 &
  \textbf{26.37} &
  \textbf{16.57} &
  \textbf{186.60} &
  \textbf{169.99} &
  \textbf{0.76} &
  \textbf{0.15} \\
$\left(\mathrm{E} \rightarrow {}_{\mathrm{B}}\text{\textsc{RU}}\right) + \mathrm{AR}$ &
  101.44 &
  156.22 &
  838.37 &
  1,386.84 &
  1.18 &
  0.29 &
  56.13 &
  67.31 &
  534.99 &
  667.50 &
  0.70 &
  0.07 &
  105.82 &
  146.13 &
  1,109.72 &
  1,768.27 &
  1.65 &
  0.39 \\
$\mathrm{E} \rightarrow {}_{\mathrm{U}}\text{\textsc{RU}}$ &
  153.46 &
  160.30 &
  1,534.67 &
  1,336.70 &
  1.88 &
  0.34 &
  168.26 &
  154.76 &
  1,464.07 &
  1,294.01 &
  1.74 &
  0.31 &
  249.24 &
  213.83 &
  1,989.17 &
  1,935.53 &
  2.41 &
  0.37 \\
$\left(\mathrm{E} \rightarrow {}_{\mathrm{U}}\text{\textsc{RU}} + {}_{\mathrm{B}}\text{\textsc{RU}}\right) + \mathrm{AR}$ &
  \textbf{59.72} &
  \textbf{59.97} &
  \textbf{470.12} &
  \textbf{544.68} &
  \textbf{1.41} &
  \textbf{0.10} &
  \textbf{30.07} &
  \textbf{18.19} &
  \textbf{293.46} &
  \textbf{189.92} &
  \textbf{0.60} &
  \textbf{0.11} &
  102.63 &
  120.04 &
  1,213.07 &
  1,410.27 &
  1.72 &
  0.32 \\
$\left(\mathrm{E} \rightarrow {}_{\mathrm{U}}\text{\textsc{RU}} + {}_{\mathrm{U}}\text{\textsc{RU}}\right) + \mathrm{AR}$ &
  138.05 &
  110.69 &
  1,430.42 &
  1,252.14 &
  2.32 &
  0.35 &
  40.99 &
  45.09 &
  326.81 &
  369.76 &
  0.77 &
  0.21 &
  136.89 &
  117.29 &
  1,702.69 &
  1,589.37 &
  2.20 &
  0.31 \\
$\left(\mathrm{E} \rightarrow {}_{\mathrm{U}}\text{\textsc{RU}}\right) + \mathrm{AR}$ &
  228.16 &
  306.27 &
  1,892.92 &
  2,534.10 &
  3.12 &
  0.82 &
  63.48 &
  91.29 &
  636.59 &
  945.50 &
  0.65 &
  0.08 &
  69.33 &
  89.52 &
  823.69 &
  1,271.88 &
  1.36 &
  0.30 \\
${}_{\mathrm{U}}\text{\textsc{RU}}$ &
  201.01 &
  156.47 &
  1,980.91 &
  1,866.96 &
  2.72 &
  0.83 &
  249.49 &
  377.04 &
  1,712.88 &
  2,439.47 &
  2.88 &
  1.07 &
  203.73 &
  207.45 &
  1,796.38 &
  2,009.88 &
  2.33 &
  0.31 \\ \cline{2-19} 
\multicolumn{1}{l|}{} &
  \multicolumn{18}{c|}{\textbf{Long Short-Term Memory (LSTM)}} \\ \cline{2-19} 
${}_{\mathrm{B}}\text{\textsc{RU}}$ &
  343.04 &
  362.51 &
  2,505.45 &
  2,804.50 &
  4.57 &
  1.22 &
  351.66 &
  352.76 &
  2,597.14 &
  2,669.53 &
  4.41 &
  0.68 &
  160.29 &
  147.31 &
  1,332.31 &
  1,276.44 &
  1.62 &
  0.17 \\
$\mathrm{E} \rightarrow {}_{\mathrm{B}}\text{\textsc{RU}}$ &
  140.21 &
  171.49 &
  1,327.95 &
  1,795.97 &
  1.53 &
  0.34 &
  111.11 &
  127.42 &
  607.34 &
  751.98 &
  0.63 &
  0.27 &
  149.57 &
  204.37 &
  1,034.95 &
  1,524.12 &
  1.24 &
  0.37 \\
$\left(\mathrm{E} \rightarrow {}_{\mathrm{B}}\text{\textsc{RU}} + {}_{\mathrm{B}}\text{\textsc{RU}}\right) + \mathrm{AR}$ &
  137.21 &
  134.69 &
  1,424.55 &
  1,717.85 &
  1.94 &
  0.39 &
  93.71 &
  87.38 &
  1,182.30 &
  1,293.40 &
  1.47 &
  0.18 &
  110.13 &
  140.48 &
  1,269.03 &
  1,768.95 &
  2.12 &
  0.38 \\
$\left(\mathrm{E} \rightarrow {}_{\mathrm{B}}\text{\textsc{RU}} + {}_{\mathrm{U}}\text{\textsc{RU}}\right) + \mathrm{AR}$ &
  137.24 &
  152.13 &
  1,381.15 &
  1,753.82 &
  1.77 &
  0.24 &
  94.28 &
  105.97 &
  1,123.98 &
  1,422.39 &
  1.22 &
  0.17 &
  67.52 &
  54.95 &
  965.37 &
  1,188.71 &
  1.28 &
  0.19 \\
$\left(\mathrm{E} \rightarrow {}_{\mathrm{B}}\text{\textsc{RU}}\right) + \mathrm{AR}$ &
  \textbf{85.07} &
  \textbf{98.50} &
  \textbf{897.68} &
  \textbf{1,271.08} &
  \textbf{1.18} &
  \textbf{0.38} &
  94.10 &
  97.26 &
  1,143.15 &
  1,248.17 &
  1.39 &
  0.17 &
  245.25 &
  323.28 &
  1,898.61 &
  2,562.27 &
  3.47 &
  0.86 \\
$\mathrm{E} \rightarrow {}_{\mathrm{U}}\text{\textsc{RU}}$ &
  211.64 &
  191.72 &
  1,674.10 &
  1,372.85 &
  1.83 &
  0.32 &
  146.44 &
  129.21 &
  819.64 &
  774.25 &
  0.69 &
  0.22 &
  138.36 &
  105.50 &
  776.30 &
  734.64 &
  0.99 &
  0.30 \\
$\left(\mathrm{E} \rightarrow {}_{\mathrm{U}}\text{\textsc{RU}} + {}_{\mathrm{B}}\text{\textsc{RU}}\right) + \mathrm{AR}$ &
  224.49 &
  329.07 &
  1,942.68 &
  2,672.95 &
  2.48 &
  0.77 &
  72.96 &
  82.00 &
  890.21 &
  1,273.83 &
  1.18 &
  0.13 &
  166.00 &
  292.10 &
  1,510.17 &
  2,698.45 &
  1.74 &
  0.84 \\
$\left(\mathrm{E} \rightarrow {}_{\mathrm{U}}\text{\textsc{RU}} + {}_{\mathrm{U}}\text{\textsc{RU}}\right) + \mathrm{AR}$ &
  215.84 &
  307.51 &
  1,943.65 &
  2,726.70 &
  2.47 &
  0.69 &
  135.95 &
  172.25 &
  1,362.01 &
  1,831.27 &
  2.00 &
  0.39 &
  124.01 &
  139.02 &
  1,366.26 &
  1,754.75 &
  2.05 &
  0.42 \\
$\left(\mathrm{E} \rightarrow {}_{\mathrm{U}}\text{\textsc{RU}}\right) + \mathrm{AR}$ &
  177.16 &
  132.98 &
  1,977.65 &
  1,606.18 &
  2.55 &
  0.44 &
  \textbf{47.56} &
  \textbf{25.51} &
  \textbf{529.98} &
  \textbf{269.66} &
  \textbf{0.77} &
  \textbf{0.27} &
  \textbf{26.79} &
  \textbf{11.42} &
  \textbf{212.39} &
  \textbf{150.62} &
  \textbf{0.77} &
  \textbf{0.16} \\
${}_{\mathrm{U}}\text{\textsc{RU}}$ &
  267.83 &
  346.86 &
  1,792.51 &
  2,281.50 &
  2.72 &
  0.57 &
  326.20 &
  433.08 &
  2,150.92 &
  2,754.01 &
  3.38 &
  0.82 &
  150.72 &
  113.44 &
  861.31 &
  708.08 &
  1.03 &
  0.25 \\ \cline{2-19} 
\end{tabular}%
}
\end{table}
\end{landscape}\clearpage
\loadgeometry{geom}

\savegeometry{geom}
\newgeometry{margin=0pt}
\thispagestyle{empty}\setcounter{page}{24}
\begin{landscape}
\begin{table}
\centering
\caption[Ablation on the first\ \ 75 days of the SARS-CoV-2 dataset.]%
{Ablation on the first 75 days of the SARS-CoV-2 dataset.\\%
{\footnotesize\textbf{Legend:} Algorithms with the best performance are in \textbf{bold}; besides, we refer to the Transformer Encoder as $\mathrm{E}$, recurrent unit as \textsc{RU}, Bidirectional as \textsc{B}, Unidirectional as \textsc{U}, and Autoregressive as \textsc{AR}.}}
\label{tab:ablation-6}
\resizebox{1.35\textwidth}{!}{%
\begin{tabular}{r|cccccc|cccccc|cccccc|}
\cline{2-19}
\multicolumn{1}{c|}{} &
  \multicolumn{6}{c|}{\textbf{PyTorch's Hyperparameters}} &
  \multicolumn{6}{c|}{\textbf{Literature's Hyperparameters}} &
  \multicolumn{6}{c|}{\ReGENN\textbf{'s Hyperparameters}} \\
\cline{2-19}
\multicolumn{1}{c|}{} &
  \multicolumn{1}{c}{\textbf{MAE}} &
  \multicolumn{1}{c}{\textbf{$\pm$ STD}} &
  \multicolumn{1}{c}{\textbf{MSLE}} &
  \multicolumn{1}{c}{\textbf{$\pm$ STD}} &
  \multicolumn{1}{c}{\textbf{RMSE}} &
  \textbf{$\pm$ STD} &
  \multicolumn{1}{c}{\textbf{MAE}} &
  \multicolumn{1}{c}{\textbf{$\pm$ STD}} &
  \multicolumn{1}{c}{\textbf{MSLE}} &
  \multicolumn{1}{c}{\textbf{$\pm$ STD}} &
  \multicolumn{1}{c}{\textbf{RMSE}} &
  \textbf{$\pm$ STD} &
  \multicolumn{1}{c}{\textbf{MAE}} &
  \multicolumn{1}{c}{\textbf{$\pm$ STD}} &
  \multicolumn{1}{c}{\textbf{MSLE}} &
  \multicolumn{1}{c}{\textbf{$\pm$ STD}} &
  \multicolumn{1}{c}{\textbf{RMSE}} &
  \textbf{$\pm$ STD} \\ \cline{2-19} 
\multicolumn{1}{l|}{} &
  \multicolumn{18}{c|}{\textbf{Recurrent Neural Network (RNN)}} \\ \cline{2-19} 
${}_{\mathrm{B}}\text{\textsc{RU}}$ &
  161.14 &
  97.42 &
  1,435.75 &
  1,110.46 &
  0.66 &
  0.14 &
  411.17 &
  303.24 &
  2,037.13 &
  1,755.05 &
  0.71 &
  0.07 &
  504.53 &
  388.17 &
  2,589.88 &
  1,795.09 &
  0.78 &
  0.23 \\
$\mathrm{E} \rightarrow {}_{\mathrm{B}}\text{\textsc{RU}}$ &
  718.88 &
  325.71 &
  5,240.10 &
  2,714.46 &
  6.51 &
  0.60 &
  683.60 &
  356.26 &
  4,074.54 &
  2,244.10 &
  4.12 &
  0.37 &
  797.03 &
  273.84 &
  4,669.25 &
  2,198.02 &
  5.29 &
  0.36 \\
$\left(\mathrm{E} \rightarrow {}_{\mathrm{B}}\text{\textsc{RU}} + {}_{\mathrm{B}}\text{\textsc{RU}}\right) + \mathrm{AR}$ &
  318.26 &
  308.94 &
  2,813.68 &
  2,311.01 &
  2.13 &
  0.46 &
  170.41 &
  119.06 &
  1,594.96 &
  1,416.35 &
  1.89 &
  0.35 &
  182.35 &
  143.06 &
  1,692.18 &
  1,759.38 &
  2.15 &
  0.38 \\
$\left(\mathrm{E} \rightarrow {}_{\mathrm{B}}\text{\textsc{RU}} + {}_{\mathrm{U}}\text{\textsc{RU}}\right) + \mathrm{AR}$ &
  \textbf{151.45} &
  \textbf{49.83} &
  \textbf{1,339.26} &
  \textbf{731.89} &
  \textbf{0.71} &
  \textbf{0.11} &
  180.85 &
  183.45 &
  1,609.37 &
  1,852.03 &
  2.01 &
  0.56 &
  183.33 &
  158.36 &
  1,752.28 &
  1,692.14 &
  1.77 &
  0.40 \\
$\left(\mathrm{E} \rightarrow {}_{\mathrm{B}}\text{\textsc{RU}}\right) + \mathrm{AR}$ &
  524.40 &
  236.82 &
  4,032.76 &
  2,150.00 &
  5.09 &
  0.52 &
  289.20 &
  192.77 &
  2,835.31 &
  1,756.82 &
  2.15 &
  0.30 &
  356.14 &
  198.64 &
  3,558.44 &
  1,649.24 &
  2.38 &
  0.50 \\
$\mathrm{E} \rightarrow {}_{\mathrm{U}}\text{\textsc{RU}}$ &
  653.11 &
  115.79 &
  6,928.49 &
  2,979.28 &
  4.48 &
  0.46 &
  448.82 &
  340.43 &
  2,034.42 &
  1,708.50 &
  1.45 &
  0.21 &
  404.58 &
  296.14 &
  1,711.90 &
  1,160.99 &
  1.33 &
  0.38 \\
$\left(\mathrm{E} \rightarrow {}_{\mathrm{U}}\text{\textsc{RU}} + {}_{\mathrm{B}}\text{\textsc{RU}}\right) + \mathrm{AR}$ &
  391.36 &
  282.20 &
  3,203.38 &
  2,431.20 &
  3.31 &
  0.51 &
  \textbf{123.58} &
  \textbf{64.91} &
  \textbf{1,039.56} &
  \textbf{450.26} &
  \textbf{2.10} &
  \textbf{0.47} &
  \textbf{145.41} &
  \textbf{106.91} &
  \textbf{1,100.23} &
  \textbf{810.37} &
  \textbf{1.59} &
  \textbf{0.29} \\
$\left(\mathrm{E} \rightarrow {}_{\mathrm{U}}\text{\textsc{RU}} + {}_{\mathrm{U}}\text{\textsc{RU}}\right) + \mathrm{AR}$ &
  250.54 &
  198.90 &
  2,111.15 &
  1,554.28 &
  1.92 &
  0.14 &
  161.39 &
  100.01 &
  1,668.05 &
  1,432.03 &
  0.81 &
  0.30 &
  157.09 &
  128.19 &
  1,623.64 &
  1,582.04 &
  1.00 &
  0.45 \\
$\left(\mathrm{E} \rightarrow {}_{\mathrm{U}}\text{\textsc{RU}}\right) + \mathrm{AR}$ &
  336.16 &
  169.55 &
  2,928.13 &
  1,643.90 &
  3.43 &
  0.45 &
  134.50 &
  69.89 &
  1,069.85 &
  758.32 &
  0.70 &
  0.15 &
  149.06 &
  72.80 &
  1,392.70 &
  916.71 &
  0.96 &
  0.25 \\
${}_{\mathrm{U}}\text{\textsc{RU}}$ &
  174.78 &
  121.62 &
  1,371.81 &
  905.07 &
  1.65 &
  0.48 &
  518.13 &
  300.14 &
  3,090.42 &
  2,026.79 &
  1.79 &
  0.41 &
  504.14 &
  264.01 &
  3,091.64 &
  1,803.63 &
  1.51 &
  0.14 \\ \cline{2-19} 
\multicolumn{1}{l|}{} &
  \multicolumn{18}{c|}{\textbf{Gated Recurrent Unit (GRU)}} \\ \cline{2-19} 
${}_{\mathrm{B}}\text{\textsc{RU}}$ &
  713.73 &
  305.34 &
  6,470.68 &
  3,376.01 &
  4.39 &
  0.42 &
  1,042.15 &
  894.79 &
  6,395.98 &
  5,717.38 &
  6.21 &
  1.08 &
  818.92 &
  479.63 &
  6,154.47 &
  3,878.30 &
  4.36 &
  0.61 \\
$\mathrm{E} \rightarrow {}_{\mathrm{B}}\text{\textsc{RU}}$ &
  191.61 &
  135.71 &
  1,470.70 &
  1,165.24 &
  0.86 &
  0.18 &
  388.02 &
  258.39 &
  1,624.89 &
  1,018.48 &
  0.99 &
  0.18 &
  459.37 &
  447.34 &
  2,092.11 &
  2,324.37 &
  1.00 &
  0.23 \\
$\left(\mathrm{E} \rightarrow {}_{\mathrm{B}}\text{\textsc{RU}} + {}_{\mathrm{B}}\text{\textsc{RU}}\right) + \mathrm{AR}$ &
  \textbf{163.11} &
  \textbf{136.68} &
  \textbf{1,102.59} &
  \textbf{880.54} &
  \textbf{0.61} &
  \textbf{0.11} &
  \textbf{134.85} &
  \textbf{158.37} &
  \textbf{1,177.67} &
  \textbf{1,542.84} &
  \textbf{0.53} &
  \textbf{0.14} &
  152.27 &
  123.36 &
  1,527.13 &
  1,609.60 &
  0.56 &
  0.16 \\
$\left(\mathrm{E} \rightarrow {}_{\mathrm{B}}\text{\textsc{RU}} + {}_{\mathrm{U}}\text{\textsc{RU}}\right) + \mathrm{AR}$ &
  311.42 &
  161.97 &
  2,924.49 &
  1,658.64 &
  2.11 &
  0.39 &
  149.88 &
  162.32 &
  1,554.70 &
  1,972.45 &
  0.52 &
  0.29 &
  169.54 &
  152.49 &
  1,662.18 &
  1,594.09 &
  1.82 &
  0.42 \\
$\left(\mathrm{E} \rightarrow {}_{\mathrm{B}}\text{\textsc{RU}}\right) + \mathrm{AR}$ &
  227.22 &
  227.73 &
  1,796.87 &
  1,618.45 &
  1.74 &
  0.24 &
  241.01 &
  129.01 &
  2,481.91 &
  1,363.01 &
  1.92 &
  0.33 &
  143.98 &
  102.04 &
  1,397.98 &
  1,238.76 &
  1.24 &
  0.22 \\
$\mathrm{E} \rightarrow {}_{\mathrm{U}}\text{\textsc{RU}}$ &
  195.72 &
  74.66 &
  1,728.65 &
  1,166.72 &
  0.95 &
  0.15 &
  426.13 &
  210.99 &
  2,117.59 &
  1,303.15 &
  0.60 &
  0.16 &
  659.88 &
  438.00 &
  5,309.42 &
  3,948.28 &
  2.37 &
  0.29 \\
$\left(\mathrm{E} \rightarrow {}_{\mathrm{U}}\text{\textsc{RU}} + {}_{\mathrm{B}}\text{\textsc{RU}}\right) + \mathrm{AR}$ &
  206.54 &
  126.84 &
  1,381.13 &
  1,060.64 &
  0.87 &
  0.16 &
  176.50 &
  114.42 &
  1,814.09 &
  1,510.40 &
  1.55 &
  0.33 &
  \textbf{141.97} &
  \textbf{65.56} &
  \textbf{1,386.72} &
  \textbf{807.02} &
  \textbf{1.80} &
  \textbf{0.29} \\
$\left(\mathrm{E} \rightarrow {}_{\mathrm{U}}\text{\textsc{RU}} + {}_{\mathrm{U}}\text{\textsc{RU}}\right) + \mathrm{AR}$ &
  693.87 &
  478.19 &
  5,567.19 &
  4,299.04 &
  4.46 &
  0.83 &
  359.02 &
  295.11 &
  3,667.50 &
  2,315.53 &
  3.23 &
  0.53 &
  309.70 &
  167.80 &
  3,485.05 &
  1,702.31 &
  2.46 &
  0.34 \\
$\left(\mathrm{E} \rightarrow {}_{\mathrm{U}}\text{\textsc{RU}}\right) + \mathrm{AR}$ &
  568.63 &
  279.26 &
  5,961.40 &
  3,709.67 &
  4.14 &
  0.39 &
  393.74 &
  115.74 &
  5,773.68 &
  2,751.32 &
  2.49 &
  0.21 &
  802.04 &
  344.69 &
  7,315.72 &
  4,095.29 &
  6.01 &
  0.97 \\
${}_{\mathrm{U}}\text{\textsc{RU}}$ &
  627.82 &
  328.31 &
  5,932.93 &
  3,530.15 &
  4.56 &
  0.65 &
  359.47 &
  125.72 &
  1,649.94 &
  659.50 &
  0.39 &
  0.10 &
  834.36 &
  512.41 &
  5,994.85 &
  3,940.99 &
  4.75 &
  0.41 \\ \cline{2-19} 
\multicolumn{1}{l|}{} &
  \multicolumn{18}{c|}{\textbf{Long Short-Term Memory (LSTM)}} \\ \cline{2-19} 
${}_{\mathrm{B}}\text{\textsc{RU}}$ &
  138.53 &
  117.29 &
  917.36 &
  840.42 &
  0.47 &
  0.07 &
  649.80 &
  438.23 &
  4,846.82 &
  3,199.61 &
  2.47 &
  0.51 &
  1,050.28 &
  657.98 &
  7,328.83 &
  4,749.73 &
  6.11 &
  1.13 \\
$\mathrm{E} \rightarrow {}_{\mathrm{B}}\text{\textsc{RU}}$ &
  380.34 &
  314.46 &
  2,744.68 &
  2,147.57 &
  3.14 &
  0.52 &
  553.72 &
  549.20 &
  2,518.26 &
  2,639.77 &
  2.19 &
  0.41 &
  516.29 &
  180.14 &
  2,854.97 &
  1,122.73 &
  1.83 &
  0.23 \\
$\left(\mathrm{E} \rightarrow {}_{\mathrm{B}}\text{\textsc{RU}} + {}_{\mathrm{B}}\text{\textsc{RU}}\right) + \mathrm{AR}$ &
  102.62 &
  20.06 &
  836.03 &
  153.62 &
  0.46 &
  0.12 &
  374.85 &
  268.30 &
  3,495.07 &
  2,567.92 &
  3.81 &
  0.45 &
  349.46 &
  202.43 &
  3,512.31 &
  2,416.90 &
  2.52 &
  0.18 \\
$\left(\mathrm{E} \rightarrow {}_{\mathrm{B}}\text{\textsc{RU}} + {}_{\mathrm{U}}\text{\textsc{RU}}\right) + \mathrm{AR}$ &
  296.93 &
  194.02 &
  2,727.79 &
  1,834.36 &
  2.12 &
  0.25 &
  474.71 &
  295.70 &
  4,039.41 &
  2,350.60 &
  4.99 &
  0.75 &
  191.60 &
  154.19 &
  1,769.89 &
  1,568.53 &
  1.83 &
  0.29 \\
$\left(\mathrm{E} \rightarrow {}_{\mathrm{B}}\text{\textsc{RU}}\right) + \mathrm{AR}$ &
  443.01 &
  350.65 &
  3,607.06 &
  3,112.68 &
  3.55 &
  0.45 &
  302.57 &
  61.93 &
  3,981.58 &
  1,071.91 &
  2.27 &
  0.17 &
  307.15 &
  384.35 &
  2,834.08 &
  3,028.40 &
  2.20 &
  0.48 \\
$\mathrm{E} \rightarrow {}_{\mathrm{U}}\text{\textsc{RU}}$ &
  558.91 &
  337.01 &
  2,976.82 &
  1,697.81 &
  1.28 &
  0.18 &
  623.12 &
  396.10 &
  2,943.44 &
  1,825.41 &
  0.49 &
  0.05 &
  550.42 &
  511.73 &
  2,189.06 &
  2,087.18 &
  0.51 &
  0.13 \\
$\left(\mathrm{E} \rightarrow {}_{\mathrm{U}}\text{\textsc{RU}} + {}_{\mathrm{B}}\text{\textsc{RU}}\right) + \mathrm{AR}$ &
  270.76 &
  118.92 &
  2,365.14 &
  1,401.32 &
  2.49 &
  0.26 &
  165.43 &
  141.71 &
  1,456.72 &
  1,484.11 &
  2.31 &
  0.59 &
  157.76 &
  143.09 &
  1,123.17 &
  1,015.65 &
  2.90 &
  0.78 \\
$\left(\mathrm{E} \rightarrow {}_{\mathrm{U}}\text{\textsc{RU}} + {}_{\mathrm{U}}\text{\textsc{RU}}\right) + \mathrm{AR}$ &
  315.88 &
  192.72 &
  2,975.82 &
  1,611.11 &
  2.05 &
  0.09 &
  345.42 &
  219.75 &
  3,470.74 &
  1,963.87 &
  3.36 &
  0.79 &
  194.90 &
  130.19 &
  1,965.46 &
  1,608.19 &
  1.68 &
  0.67 \\
$\left(\mathrm{E} \rightarrow {}_{\mathrm{U}}\text{\textsc{RU}}\right) + \mathrm{AR}$ &
  \textbf{100.52} &
  \textbf{79.84} &
  \textbf{743.39} &
  \textbf{567.97} &
  \textbf{0.46} &
  \textbf{0.08} &
  \textbf{84.27} &
  \textbf{35.68} &
  \textbf{654.90} &
  \textbf{326.59} &
  \textbf{0.44} &
  \textbf{0.06} &
  \textbf{78.74} &
  \textbf{48.82} &
  \textbf{614.42} &
  \textbf{377.49} &
  \textbf{0.39} &
  \textbf{0.12} \\
${}_{\mathrm{U}}\text{\textsc{RU}}$ &
  716.32 &
  508.83 &
  3,639.20 &
  2,751.68 &
  2.78 &
  0.38 &
  821.10 &
  300.26 &
  4,518.70 &
  1,855.99 &
  3.79 &
  0.55 &
  751.09 &
  527.12 &
  3,599.69 &
  2,691.51 &
  2.73 &
  0.58 \\ \cline{2-19} 
\end{tabular}%
}
\end{table}
\end{landscape}\clearpage
\loadgeometry{geom}

\savegeometry{geom}
\newgeometry{margin=0pt}
\thispagestyle{empty}\setcounter{page}{25}
\begin{landscape}
\begin{table}
\centering
\caption[Ablation on the first\ \ 90 days of the SARS-CoV-2 dataset.]%
{Ablation on the first 90 days of the SARS-CoV-2 dataset.\\%
{\footnotesize\textbf{Legend:} Algorithms with the best performance are in \textbf{bold}; besides, we refer to the Transformer Encoder as $\mathrm{E}$, recurrent unit as \textsc{RU}, Bidirectional as \textsc{B}, Unidirectional as \textsc{U}, and Autoregressive as \textsc{AR}.}}
\label{tab:ablation-7}
\resizebox{1.35\textwidth}{!}{%
\begin{tabular}{r|cccccc|cccccc|cccccc|}
\cline{2-19}
\multicolumn{1}{c|}{} &
  \multicolumn{6}{c|}{\textbf{PyTorch's Hyperparameters}} &
  \multicolumn{6}{c|}{\textbf{Literature's Hyperparameters}} &
  \multicolumn{6}{c|}{\ReGENN\textbf{'s Hyperparameters}} \\
\cline{2-19}
\multicolumn{1}{c|}{} &
  \multicolumn{1}{c}{\textbf{MAE}} &
  \multicolumn{1}{c}{\textbf{$\pm$ STD}} &
  \multicolumn{1}{c}{\textbf{MSLE}} &
  \multicolumn{1}{c}{\textbf{$\pm$ STD}} &
  \multicolumn{1}{c}{\textbf{RMSE}} &
  \textbf{$\pm$ STD} &
  \multicolumn{1}{c}{\textbf{MAE}} &
  \multicolumn{1}{c}{\textbf{$\pm$ STD}} &
  \multicolumn{1}{c}{\textbf{MSLE}} &
  \multicolumn{1}{c}{\textbf{$\pm$ STD}} &
  \multicolumn{1}{c}{\textbf{RMSE}} &
  \textbf{$\pm$ STD} &
  \multicolumn{1}{c}{\textbf{MAE}} &
  \multicolumn{1}{c}{\textbf{$\pm$ STD}} &
  \multicolumn{1}{c}{\textbf{MSLE}} &
  \multicolumn{1}{c}{\textbf{$\pm$ STD}} &
  \multicolumn{1}{c}{\textbf{RMSE}} &
  \textbf{$\pm$ STD} \\ \cline{2-19} 
\multicolumn{1}{l|}{} &
  \multicolumn{18}{c|}{\textbf{Recurrent Neural Network (RNN)}} \\ \cline{2-19} 
${}_{\mathrm{B}}\text{\textsc{RU}}$ &
  \textbf{242.94} &
  \textbf{140.17} &
  \textbf{1,541.55} &
  \textbf{1,048.30} &
  \textbf{0.23} &
  \textbf{0.02} &
  1,511.05 &
  738.58 &
  10,258.68 &
  7,629.47 &
  3.21 &
  0.22 &
  4,126.78 &
  1,477.31 &
  19,562.50 &
  13,001.12 &
  3.40 &
  0.15 \\
$\mathrm{E} \rightarrow {}_{\mathrm{B}}\text{\textsc{RU}}$ &
  2,264.19 &
  1,633.01 &
  15,315.27 &
  12,110.71 &
  10.15 &
  1.95 &
  1,886.29 &
  1,371.64 &
  10,610.35 &
  9,311.24 &
  5.73 &
  0.96 &
  3,887.42 &
  2,694.59 &
  15,666.70 &
  14,370.03 &
  3.02 &
  0.45 \\
$\left(\mathrm{E} \rightarrow {}_{\mathrm{B}}\text{\textsc{RU}} + {}_{\mathrm{B}}\text{\textsc{RU}}\right) + \mathrm{AR}$ &
  778.99 &
  310.60 &
  8,999.00 &
  5,753.81 &
  2.76 &
  0.27 &
  869.49 &
  708.33 &
  7,935.13 &
  7,653.96 &
  3.54 &
  0.59 &
  \textbf{193.31} &
  \textbf{176.40} &
  \textbf{1,202.10} &
  \textbf{992.75} &
  \textbf{0.22} &
  \textbf{0.06} \\
$\left(\mathrm{E} \rightarrow {}_{\mathrm{B}}\text{\textsc{RU}} + {}_{\mathrm{U}}\text{\textsc{RU}}\right) + \mathrm{AR}$ &
  318.65 &
  232.28 &
  1,707.38 &
  1,300.59 &
  0.24 &
  0.08 &
  362.09 &
  161.48 &
  3,352.83 &
  1,248.02 &
  1.31 &
  0.26 &
  199.27 &
  165.91 &
  1,348.27 &
  1,154.47 &
  0.21 &
  0.06 \\
$\left(\mathrm{E} \rightarrow {}_{\mathrm{B}}\text{\textsc{RU}}\right) + \mathrm{AR}$ &
  742.97 &
  492.51 &
  7,333.20 &
  6,379.55 &
  2.70 &
  0.45 &
  237.05 &
  118.27 &
  1,876.90 &
  410.60 &
  0.20 &
  0.10 &
  206.64 &
  130.16 &
  1,515.35 &
  1,080.97 &
  0.22 &
  0.11 \\
$\mathrm{E} \rightarrow {}_{\mathrm{U}}\text{\textsc{RU}}$ &
  738.67 &
  474.03 &
  7,120.58 &
  5,497.46 &
  2.84 &
  0.44 &
  1,473.40 &
  1,176.38 &
  5,987.45 &
  5,215.71 &
  0.78 &
  0.32 &
  3,646.65 &
  1,630.25 &
  16,109.81 &
  11,125.85 &
  5.74 &
  1.34 \\
$\left(\mathrm{E} \rightarrow {}_{\mathrm{U}}\text{\textsc{RU}} + {}_{\mathrm{B}}\text{\textsc{RU}}\right) + \mathrm{AR}$ &
  283.06 &
  75.59 &
  2,054.32 &
  903.32 &
  0.28 &
  0.14 &
  279.72 &
  123.54 &
  2,286.41 &
  1,084.17 &
  1.05 &
  0.54 &
  207.86 &
  110.68 &
  1,493.80 &
  917.88 &
  0.19 &
  0.03 \\
$\left(\mathrm{E} \rightarrow {}_{\mathrm{U}}\text{\textsc{RU}} + {}_{\mathrm{U}}\text{\textsc{RU}}\right) + \mathrm{AR}$ &
  673.46 &
  209.93 &
  7,651.99 &
  3,975.96 &
  2.70 &
  0.31 &
  740.28 &
  478.88 &
  7,154.20 &
  5,232.24 &
  3.64 &
  0.92 &
  201.83 &
  177.38 &
  1,256.22 &
  1,254.76 &
  0.23 &
  0.12 \\
$\left(\mathrm{E} \rightarrow {}_{\mathrm{U}}\text{\textsc{RU}}\right) + \mathrm{AR}$ &
  703.48 &
  527.47 &
  7,034.98 &
  5,485.51 &
  2.56 &
  0.21 &
  \textbf{190.42} &
  \textbf{190.32} &
  \textbf{1,144.52} &
  \textbf{1,230.53} &
  \textbf{0.22} &
  \textbf{0.08} &
  197.50 &
  64.81 &
  1,534.59 &
  742.72 &
  0.21 &
  0.07 \\
${}_{\mathrm{U}}\text{\textsc{RU}}$ &
  365.99 &
  326.64 &
  1,992.32 &
  1,814.87 &
  2.11 &
  0.71 &
  906.45 &
  577.09 &
  3,622.20 &
  2,874.36 &
  1.21 &
  0.44 &
  3,926.71 &
  2,129.55 &
  17,143.74 &
  12,706.58 &
  0.72 &
  0.08 \\ \cline{2-19} 
\multicolumn{1}{l|}{} &
  \multicolumn{18}{c|}{\textbf{Gated Recurrent Unit (GRU)}} \\ \cline{2-19} 
${}_{\mathrm{B}}\text{\textsc{RU}}$ &
  890.38 &
  580.01 &
  9,857.40 &
  9,307.51 &
  3.08 &
  0.32 &
  1,442.15 &
  837.67 &
  10,937.08 &
  8,662.15 &
  3.06 &
  0.41 &
  4,308.93 &
  1,709.03 &
  20,693.26 &
  13,930.29 &
  5.33 &
  0.23 \\
$\mathrm{E} \rightarrow {}_{\mathrm{B}}\text{\textsc{RU}}$ &
  426.15 &
  373.83 &
  2,881.20 &
  2,435.46 &
  0.36 &
  0.14 &
  1,096.77 &
  857.15 &
  4,092.96 &
  3,413.09 &
  0.45 &
  0.08 &
  3,427.21 &
  1,147.82 &
  15,347.22 &
  9,743.89 &
  1.26 &
  0.48 \\
$\left(\mathrm{E} \rightarrow {}_{\mathrm{B}}\text{\textsc{RU}} + {}_{\mathrm{B}}\text{\textsc{RU}}\right) + \mathrm{AR}$ &
  249.97 &
  92.04 &
  1,688.83 &
  890.59 &
  0.28 &
  0.15 &
  252.57 &
  136.37 &
  2,052.86 &
  1,448.23 &
  0.24 &
  0.11 &
  190.44 &
  143.42 &
  1,209.26 &
  873.04 &
  0.18 &
  0.07 \\
$\left(\mathrm{E} \rightarrow {}_{\mathrm{B}}\text{\textsc{RU}} + {}_{\mathrm{U}}\text{\textsc{RU}}\right) + \mathrm{AR}$ &
  791.02 &
  517.33 &
  7,613.47 &
  6,661.82 &
  2.72 &
  0.44 &
  461.40 &
  218.69 &
  4,287.87 &
  1,452.17 &
  2.06 &
  0.61 &
  190.61 &
  139.96 &
  1,315.81 &
  981.13 &
  0.21 &
  0.07 \\
$\left(\mathrm{E} \rightarrow {}_{\mathrm{B}}\text{\textsc{RU}}\right) + \mathrm{AR}$ &
  \textbf{228.32} &
  \textbf{117.23} &
  \textbf{1,457.24} &
  \textbf{969.59} &
  \textbf{0.23} &
  \textbf{0.08} &
  \textbf{239.36} &
  \textbf{158.83} &
  \textbf{1,734.29} &
  \textbf{1,297.86} &
  \textbf{0.42} &
  \textbf{0.16} &
  211.31 &
  186.86 &
  1,298.85 &
  1,191.48 &
  0.21 &
  0.09 \\
$\mathrm{E} \rightarrow {}_{\mathrm{U}}\text{\textsc{RU}}$ &
  342.96 &
  259.87 &
  2,110.70 &
  1,719.47 &
  0.40 &
  0.15 &
  2,161.34 &
  1,746.48 &
  11,637.07 &
  11,359.42 &
  1.01 &
  0.37 &
  3,832.96 &
  3,514.14 &
  16,232.39 &
  14,729.52 &
  4.53 &
  0.70 \\
$\left(\mathrm{E} \rightarrow {}_{\mathrm{U}}\text{\textsc{RU}} + {}_{\mathrm{B}}\text{\textsc{RU}}\right) + \mathrm{AR}$ &
  951.85 &
  439.86 &
  11,261.48 &
  8,436.19 &
  3.12 &
  0.26 &
  305.29 &
  309.49 &
  2,145.37 &
  2,199.81 &
  2.11 &
  0.52 &
  848.35 &
  505.80 &
  10,591.76 &
  9,146.85 &
  3.17 &
  0.72 \\
$\left(\mathrm{E} \rightarrow {}_{\mathrm{U}}\text{\textsc{RU}} + {}_{\mathrm{U}}\text{\textsc{RU}}\right) + \mathrm{AR}$ &
  1,839.28 &
  1,245.97 &
  14,492.23 &
  12,161.61 &
  8.00 &
  0.94 &
  723.14 &
  456.53 &
  8,472.29 &
  7,068.90 &
  3.06 &
  0.55 &
  \textbf{179.35} &
  \textbf{102.26} &
  \textbf{1,253.62} &
  \textbf{801.38} &
  \textbf{0.18} &
  \textbf{0.06} \\
$\left(\mathrm{E} \rightarrow {}_{\mathrm{U}}\text{\textsc{RU}}\right) + \mathrm{AR}$ &
  285.47 &
  78.53 &
  2,432.23 &
  871.10 &
  0.24 &
  0.08 &
  869.42 &
  690.38 &
  8,641.86 &
  10,078.24 &
  0.91 &
  0.38 &
  1,238.96 &
  924.89 &
  11,673.27 &
  9,827.31 &
  5.38 &
  0.45 \\
${}_{\mathrm{U}}\text{\textsc{RU}}$ &
  883.60 &
  468.32 &
  10,247.84 &
  7,336.02 &
  3.62 &
  0.46 &
  1,107.06 &
  700.25 &
  6,164.05 &
  4,731.55 &
  1.65 &
  0.18 &
  3,888.44 &
  1,606.01 &
  17,143.93 &
  12,597.02 &
  0.58 &
  0.11 \\ \cline{2-19} 
\multicolumn{1}{l|}{} &
  \multicolumn{18}{c|}{\textbf{Long Short-Term Memory (LSTM)}} \\ \cline{2-19} 
${}_{\mathrm{B}}\text{\textsc{RU}}$ &
  \textbf{257.99} &
  \textbf{143.04} &
  \textbf{1,730.68} &
  \textbf{1,018.27} &
  \textbf{0.22} &
  \textbf{0.02} &
  2,010.31 &
  1,131.95 &
  14,467.37 &
  12,147.42 &
  5.88 &
  0.69 &
  4,069.37 &
  3,919.60 &
  16,311.66 &
  17,267.56 &
  3.37 &
  0.19 \\
$\mathrm{E} \rightarrow {}_{\mathrm{B}}\text{\textsc{RU}}$ &
  365.42 &
  184.23 &
  2,534.62 &
  1,653.02 &
  0.31 &
  0.12 &
  1,367.59 &
  1,110.26 &
  5,612.86 &
  5,986.90 &
  0.62 &
  0.21 &
  3,692.45 &
  2,757.03 &
  14,303.79 &
  13,581.50 &
  1.38 &
  0.49 \\
$\left(\mathrm{E} \rightarrow {}_{\mathrm{B}}\text{\textsc{RU}} + {}_{\mathrm{B}}\text{\textsc{RU}}\right) + \mathrm{AR}$ &
  1,657.74 &
  818.46 &
  13,233.88 &
  9,759.93 &
  7.61 &
  0.85 &
  738.94 &
  399.30 &
  8,983.37 &
  7,134.58 &
  3.28 &
  0.49 &
  194.23 &
  100.19 &
  1,342.74 &
  762.09 &
  0.18 &
  0.06 \\
$\left(\mathrm{E} \rightarrow {}_{\mathrm{B}}\text{\textsc{RU}} + {}_{\mathrm{U}}\text{\textsc{RU}}\right) + \mathrm{AR}$ &
  1,207.08 &
  834.76 &
  10,220.21 &
  8,595.59 &
  5.08 &
  0.56 &
  825.39 &
  382.22 &
  8,795.38 &
  5,960.91 &
  4.06 &
  1.62 &
  184.65 &
  152.93 &
  1,204.69 &
  940.83 &
  0.19 &
  0.09 \\
$\left(\mathrm{E} \rightarrow {}_{\mathrm{B}}\text{\textsc{RU}}\right) + \mathrm{AR}$ &
  882.08 &
  801.18 &
  8,251.39 &
  8,533.21 &
  2.90 &
  0.34 &
  1,274.05 &
  1,083.48 &
  11,276.86 &
  10,900.65 &
  5.61 &
  0.77 &
  183.95 &
  162.25 &
  1,125.99 &
  1,060.32 &
  0.20 &
  0.07 \\
$\mathrm{E} \rightarrow {}_{\mathrm{U}}\text{\textsc{RU}}$ &
  1,848.50 &
  854.54 &
  11,982.81 &
  8,193.23 &
  3.11 &
  0.34 &
  1,988.35 &
  1,111.96 &
  8,604.09 &
  6,648.45 &
  0.63 &
  0.26 &
  4,234.82 &
  2,935.90 &
  19,072.87 &
  16,114.08 &
  6.08 &
  1.08 \\
$\left(\mathrm{E} \rightarrow {}_{\mathrm{U}}\text{\textsc{RU}} + {}_{\mathrm{B}}\text{\textsc{RU}}\right) + \mathrm{AR}$ &
  649.64 &
  489.43 &
  6,142.64 &
  5,304.13 &
  2.57 &
  0.30 &
  191.33 &
  118.30 &
  1,614.64 &
  1,416.72 &
  0.47 &
  0.12 &
  191.16 &
  171.66 &
  1,191.72 &
  1,058.59 &
  0.20 &
  0.08 \\
$\left(\mathrm{E} \rightarrow {}_{\mathrm{U}}\text{\textsc{RU}} + {}_{\mathrm{U}}\text{\textsc{RU}}\right) + \mathrm{AR}$ &
  669.24 &
  423.71 &
  7,508.73 &
  6,691.47 &
  2.72 &
  0.38 &
  856.88 &
  512.69 &
  8,460.02 &
  6,132.46 &
  4.24 &
  1.12 &
  \textbf{183.63} &
  \textbf{103.46} &
  \textbf{1,384.82} &
  \textbf{802.17} &
  \textbf{0.20} &
  \textbf{0.13} \\
$\left(\mathrm{E} \rightarrow {}_{\mathrm{U}}\text{\textsc{RU}}\right) + \mathrm{AR}$ &
  277.93 &
  227.02 &
  1,991.10 &
  1,646.43 &
  0.23 &
  0.07 &
  \textbf{162.88} &
  \textbf{76.44} &
  \textbf{1,160.18} &
  \textbf{752.93} &
  \textbf{0.21} &
  \textbf{0.11} &
  187.40 &
  122.37 &
  1,283.08 &
  921.00 &
  0.21 &
  0.08 \\
${}_{\mathrm{U}}\text{\textsc{RU}}$ &
  2,850.77 &
  2,301.93 &
  15,645.06 &
  15,051.94 &
  9.65 &
  0.93 &
  2,044.83 &
  1,351.42 &
  10,417.92 &
  8,664.39 &
  4.92 &
  0.57 &
  3,769.08 &
  2,845.18 &
  15,525.91 &
  14,128.42 &
  0.55 &
  0.08 \\ \cline{2-19} 
\end{tabular}%
}
\end{table}
\end{landscape}\clearpage
\loadgeometry{geom}

\savegeometry{geom}
\newgeometry{margin=0pt}
\thispagestyle{empty}\setcounter{page}{26}
\begin{landscape}
\begin{table}
\centering
\caption[Ablation on the first 105 days of the SARS-CoV-2 dataset.]%
{Ablation on the first 105 days of the SARS-CoV-2 dataset.\\%
{\footnotesize\textbf{Legend:} Algorithms with the best performance are in \textbf{bold}; besides, we refer to the Transformer Encoder as $\mathrm{E}$, recurrent unit as \textsc{RU}, Bidirectional as \textsc{B}, Unidirectional as \textsc{U}, and Autoregressive as \textsc{AR}.}}
\label{tab:ablation-8}
\resizebox{1.35\textwidth}{!}{%
\begin{tabular}{r|cccccc|cccccc|cccccc|}
\cline{2-19}
\multicolumn{1}{c|}{} &
  \multicolumn{6}{c|}{\textbf{PyTorch's Hyperparameters}} &
  \multicolumn{6}{c|}{\textbf{Literature's Hyperparameters}} &
  \multicolumn{6}{c|}{\ReGENN\textbf{'s Hyperparameters}} \\
\cline{2-19}
\multicolumn{1}{c|}{} &
  \multicolumn{1}{c}{\textbf{MAE}} &
  \multicolumn{1}{c}{\textbf{$\pm$ STD}} &
  \multicolumn{1}{c}{\textbf{MSLE}} &
  \multicolumn{1}{c}{\textbf{$\pm$ STD}} &
  \multicolumn{1}{c}{\textbf{RMSE}} &
  \textbf{$\pm$ STD} &
  \multicolumn{1}{c}{\textbf{MAE}} &
  \multicolumn{1}{c}{\textbf{$\pm$ STD}} &
  \multicolumn{1}{c}{\textbf{MSLE}} &
  \multicolumn{1}{c}{\textbf{$\pm$ STD}} &
  \multicolumn{1}{c}{\textbf{RMSE}} &
  \textbf{$\pm$ STD} &
  \multicolumn{1}{c}{\textbf{MAE}} &
  \multicolumn{1}{c}{\textbf{$\pm$ STD}} &
  \multicolumn{1}{c}{\textbf{MSLE}} &
  \multicolumn{1}{c}{\textbf{$\pm$ STD}} &
  \multicolumn{1}{c}{\textbf{RMSE}} &
  \textbf{$\pm$ STD} \\ \cline{2-19} 
\multicolumn{1}{l|}{} &
  \multicolumn{18}{c|}{\textbf{Recurrent Neural Network (RNN)}} \\ \cline{2-19} 
${}_{\mathrm{B}}\text{\textsc{RU}}$ &
  341.96 &
  160.69 &
  2,004.99 &
  1,108.56 &
  0.12 &
  0.02 &
  1,261.25 &
  1,129.84 &
  4,716.83 &
  4,379.85 &
  0.13 &
  0.05 &
  1,323.09 &
  883.58 &
  5,040.30 &
  4,548.85 &
  0.14 &
  0.05 \\
$\mathrm{E} \rightarrow {}_{\mathrm{B}}\text{\textsc{RU}}$ &
  4,341.20 &
  2,825.80 &
  25,086.49 &
  23,858.93 &
  12.92 &
  0.84 &
  3,126.49 &
  2,291.22 &
  17,483.74 &
  16,935.89 &
  6.56 &
  0.77 &
  3,404.41 &
  1,317.70 &
  20,373.84 &
  14,638.73 &
  6.55 &
  0.55 \\
$\left(\mathrm{E} \rightarrow {}_{\mathrm{B}}\text{\textsc{RU}} + {}_{\mathrm{B}}\text{\textsc{RU}}\right) + \mathrm{AR}$ &
  1,120.61 &
  877.22 &
  12,934.67 &
  12,210.30 &
  3.35 &
  0.36 &
  1,166.00 &
  601.41 &
  13,690.65 &
  11,482.17 &
  3.33 &
  0.17 &
  238.42 &
  90.06 &
  1,582.80 &
  531.20 &
  0.11 &
  0.04 \\
$\left(\mathrm{E} \rightarrow {}_{\mathrm{B}}\text{\textsc{RU}} + {}_{\mathrm{U}}\text{\textsc{RU}}\right) + \mathrm{AR}$ &
  \textbf{246.59} &
  \textbf{148.21} &
  \textbf{1,474.80} &
  \textbf{1,032.02} &
  \textbf{0.11} &
  \textbf{0.04} &
  320.71 &
  171.59 &
  2,169.15 &
  1,011.90 &
  0.10 &
  0.02 &
  227.72 &
  111.45 &
  1,333.91 &
  713.54 &
  0.12 &
  0.03 \\
$\left(\mathrm{E} \rightarrow {}_{\mathrm{B}}\text{\textsc{RU}}\right) + \mathrm{AR}$ &
  2,190.95 &
  1,807.99 &
  17,646.09 &
  16,804.20 &
  6.54 &
  0.69 &
  \textbf{239.77} &
  \textbf{141.32} &
  \textbf{1,556.19} &
  \textbf{959.49} &
  \textbf{0.09} &
  \textbf{0.01} &
  \textbf{226.43} &
  \textbf{90.79} &
  \textbf{1,455.35} &
  \textbf{585.70} &
  \textbf{0.11} &
  \textbf{0.05} \\
$\mathrm{E} \rightarrow {}_{\mathrm{U}}\text{\textsc{RU}}$ &
  2,947.62 &
  2,982.07 &
  20,010.83 &
  24,909.44 &
  3.52 &
  0.43 &
  2,391.16 &
  1,265.33 &
  9,282.64 &
  5,811.69 &
  0.44 &
  0.17 &
  1,993.13 &
  1,126.18 &
  7,824.99 &
  5,893.62 &
  0.64 &
  0.22 \\
$\left(\mathrm{E} \rightarrow {}_{\mathrm{U}}\text{\textsc{RU}} + {}_{\mathrm{B}}\text{\textsc{RU}}\right) + \mathrm{AR}$ &
  278.32 &
  169.27 &
  1,477.52 &
  1,030.39 &
  0.17 &
  0.04 &
  280.10 &
  135.78 &
  2,033.65 &
  1,390.79 &
  0.10 &
  0.02 &
  261.75 &
  120.82 &
  1,512.66 &
  719.22 &
  0.11 &
  0.03 \\
$\left(\mathrm{E} \rightarrow {}_{\mathrm{U}}\text{\textsc{RU}} + {}_{\mathrm{U}}\text{\textsc{RU}}\right) + \mathrm{AR}$ &
  986.76 &
  718.41 &
  11,505.97 &
  10,866.77 &
  3.21 &
  0.40 &
  262.21 &
  116.82 &
  1,723.81 &
  781.72 &
  0.11 &
  0.03 &
  249.63 &
  124.28 &
  1,494.06 &
  836.21 &
  0.11 &
  0.05 \\
$\left(\mathrm{E} \rightarrow {}_{\mathrm{U}}\text{\textsc{RU}}\right) + \mathrm{AR}$ &
  981.33 &
  653.20 &
  11,512.57 &
  10,876.46 &
  3.20 &
  0.23 &
  253.41 &
  132.72 &
  1,621.04 &
  846.46 &
  0.13 &
  0.07 &
  269.68 &
  129.55 &
  1,620.35 &
  848.37 &
  0.12 &
  0.04 \\
${}_{\mathrm{U}}\text{\textsc{RU}}$ &
  997.57 &
  753.72 &
  6,170.68 &
  5,317.17 &
  2.86 &
  0.72 &
  1,374.42 &
  851.34 &
  5,262.59 &
  4,588.21 &
  0.23 &
  0.19 &
  1,344.69 &
  571.12 &
  5,549.24 &
  4,109.40 &
  0.18 &
  0.05 \\ \cline{2-19} 
\multicolumn{1}{l|}{} &
  \multicolumn{18}{c|}{\textbf{Gated Recurrent Unit (GRU)}} \\ \cline{2-19} 
${}_{\mathrm{B}}\text{\textsc{RU}}$ &
  1,197.76 &
  614.55 &
  14,024.42 &
  12,312.72 &
  3.48 &
  0.42 &
  3,138.88 &
  2,547.76 &
  20,082.44 &
  20,202.11 &
  6.86 &
  0.79 &
  2,248.76 &
  1,391.69 &
  15,338.81 &
  15,264.58 &
  3.57 &
  0.42 \\
$\mathrm{E} \rightarrow {}_{\mathrm{B}}\text{\textsc{RU}}$ &
  941.01 &
  859.05 &
  5,899.12 &
  6,879.50 &
  0.28 &
  0.10 &
  3,282.37 &
  2,882.56 &
  17,989.68 &
  22,475.37 &
  0.43 &
  0.22 &
  1,693.16 &
  1,403.68 &
  6,452.83 &
  6,431.29 &
  0.24 &
  0.08 \\
$\left(\mathrm{E} \rightarrow {}_{\mathrm{B}}\text{\textsc{RU}} + {}_{\mathrm{B}}\text{\textsc{RU}}\right) + \mathrm{AR}$ &
  260.26 &
  143.14 &
  1,580.15 &
  877.78 &
  0.12 &
  0.04 &
  362.61 &
  165.44 &
  2,560.90 &
  1,559.77 &
  0.55 &
  0.38 &
  245.13 &
  99.07 &
  1,526.87 &
  747.33 &
  0.12 &
  0.04 \\
$\left(\mathrm{E} \rightarrow {}_{\mathrm{B}}\text{\textsc{RU}} + {}_{\mathrm{U}}\text{\textsc{RU}}\right) + \mathrm{AR}$ &
  242.25 &
  118.67 &
  1,471.29 &
  633.96 &
  0.12 &
  0.04 &
  301.90 &
  108.46 &
  1,978.05 &
  808.96 &
  0.18 &
  0.05 &
  272.90 &
  164.68 &
  1,587.65 &
  817.19 &
  0.12 &
  0.04 \\
$\left(\mathrm{E} \rightarrow {}_{\mathrm{B}}\text{\textsc{RU}}\right) + \mathrm{AR}$ &
  1,064.23 &
  593.75 &
  11,838.02 &
  9,917.30 &
  3.16 &
  0.21 &
  267.71 &
  81.15 &
  2,014.32 &
  663.13 &
  0.24 &
  0.20 &
  251.47 &
  208.25 &
  1,321.24 &
  1,038.05 &
  0.11 &
  0.04 \\
$\mathrm{E} \rightarrow {}_{\mathrm{U}}\text{\textsc{RU}}$ &
  1,141.02 &
  1,270.93 &
  9,951.23 &
  11,855.78 &
  0.39 &
  0.08 &
  2,414.07 &
  1,048.40 &
  9,049.15 &
  4,455.50 &
  0.48 &
  0.10 &
  1,904.25 &
  1,246.60 &
  7,347.94 &
  5,395.31 &
  0.23 &
  0.05 \\
$\left(\mathrm{E} \rightarrow {}_{\mathrm{U}}\text{\textsc{RU}} + {}_{\mathrm{B}}\text{\textsc{RU}}\right) + \mathrm{AR}$ &
  \textbf{223.08} &
  \textbf{130.92} &
  \textbf{1,340.96} &
  \textbf{793.38} &
  \textbf{0.13} &
  \textbf{0.07} &
  247.52 &
  170.20 &
  1,364.31 &
  964.02 &
  0.32 &
  0.07 &
  \textbf{234.74} &
  \textbf{138.74} &
  \textbf{1,405.17} &
  \textbf{857.20} &
  \textbf{0.11} &
  \textbf{0.04} \\
$\left(\mathrm{E} \rightarrow {}_{\mathrm{U}}\text{\textsc{RU}} + {}_{\mathrm{U}}\text{\textsc{RU}}\right) + \mathrm{AR}$ &
  2,950.38 &
  1,927.05 &
  23,175.39 &
  21,293.64 &
  9.89 &
  0.71 &
  \textbf{244.72} &
  \textbf{114.03} &
  \textbf{1,519.01} &
  \textbf{749.52} &
  \textbf{0.15} &
  \textbf{0.06} &
  269.38 &
  100.36 &
  1,605.27 &
  672.70 &
  0.12 &
  0.06 \\
$\left(\mathrm{E} \rightarrow {}_{\mathrm{U}}\text{\textsc{RU}}\right) + \mathrm{AR}$ &
  2,162.26 &
  1,541.75 &
  18,386.77 &
  18,374.57 &
  6.64 &
  0.44 &
  1,155.48 &
  1,005.94 &
  12,959.38 &
  14,123.98 &
  3.50 &
  0.56 &
  248.97 &
  184.04 &
  1,327.23 &
  1,018.99 &
  0.11 &
  0.04 \\
${}_{\mathrm{U}}\text{\textsc{RU}}$ &
  1,780.31 &
  1,059.76 &
  17,555.20 &
  14,679.39 &
  6.35 &
  0.73 &
  1,327.66 &
  808.13 &
  5,282.80 &
  4,617.99 &
  0.10 &
  0.03 &
  1,840.21 &
  1,247.34 &
  11,737.09 &
  10,978.18 &
  2.33 &
  0.18 \\ \cline{2-19} 
\multicolumn{1}{l|}{} &
  \multicolumn{18}{c|}{\textbf{Long Short-Term Memory (LSTM)}} \\ \cline{2-19} 
${}_{\mathrm{B}}\text{\textsc{RU}}$ &
  1,124.29 &
  522.77 &
  13,528.89 &
  10,943.31 &
  3.30 &
  0.21 &
  2,267.79 &
  1,935.99 &
  14,917.69 &
  15,686.46 &
  3.56 &
  0.45 &
  1,328.14 &
  647.15 &
  5,485.95 &
  4,212.59 &
  0.12 &
  0.04 \\
$\mathrm{E} \rightarrow {}_{\mathrm{B}}\text{\textsc{RU}}$ &
  1,339.27 &
  904.16 &
  7,096.00 &
  4,733.68 &
  0.30 &
  0.15 &
  3,029.54 &
  3,088.49 &
  17,588.19 &
  23,093.53 &
  0.71 &
  0.24 &
  2,013.99 &
  1,002.01 &
  8,203.49 &
  6,971.06 &
  0.27 &
  0.03 \\
$\left(\mathrm{E} \rightarrow {}_{\mathrm{B}}\text{\textsc{RU}} + {}_{\mathrm{B}}\text{\textsc{RU}}\right) + \mathrm{AR}$ &
  1,463.09 &
  1,012.26 &
  14,503.84 &
  13,505.02 &
  3.45 &
  0.21 &
  \textbf{229.85} &
  \textbf{135.67} &
  \textbf{1,304.71} &
  \textbf{746.95} &
  \textbf{0.26} &
  \textbf{0.15} &
  \textbf{221.61} &
  \textbf{110.78} &
  \textbf{1,317.11} &
  \textbf{712.53} &
  \textbf{0.12} &
  \textbf{0.03} \\
$\left(\mathrm{E} \rightarrow {}_{\mathrm{B}}\text{\textsc{RU}} + {}_{\mathrm{U}}\text{\textsc{RU}}\right) + \mathrm{AR}$ &
  1,114.13 &
  875.69 &
  12,659.32 &
  12,044.33 &
  3.32 &
  0.30 &
  253.15 &
  154.09 &
  1,506.05 &
  1,005.71 &
  0.25 &
  0.10 &
  250.02 &
  115.48 &
  1,536.47 &
  733.91 &
  0.12 &
  0.03 \\
$\left(\mathrm{E} \rightarrow {}_{\mathrm{B}}\text{\textsc{RU}}\right) + \mathrm{AR}$ &
  \textbf{230.72} &
  \textbf{66.84} &
  \textbf{1,430.87} &
  \textbf{375.47} &
  \textbf{0.11} &
  \textbf{0.04} &
  1,163.75 &
  620.16 &
  14,356.48 &
  11,907.68 &
  3.68 &
  0.35 &
  257.26 &
  225.31 &
  1,263.43 &
  1,122.21 &
  0.11 &
  0.04 \\
$\mathrm{E} \rightarrow {}_{\mathrm{U}}\text{\textsc{RU}}$ &
  2,901.71 &
  1,821.21 &
  16,836.92 &
  13,897.86 &
  3.52 &
  0.45 &
  3,828.85 &
  3,702.78 &
  16,994.16 &
  20,532.17 &
  0.43 &
  0.11 &
  3,420.16 &
  2,384.23 &
  14,959.58 &
  16,020.56 &
  0.35 &
  0.08 \\
$\left(\mathrm{E} \rightarrow {}_{\mathrm{U}}\text{\textsc{RU}} + {}_{\mathrm{B}}\text{\textsc{RU}}\right) + \mathrm{AR}$ &
  1,030.73 &
  841.67 &
  10,553.23 &
  11,022.68 &
  3.16 &
  0.22 &
  251.33 &
  93.95 &
  1,592.97 &
  672.39 &
  0.19 &
  0.09 &
  1,016.56 &
  532.07 &
  12,012.26 &
  9,602.11 &
  3.16 &
  0.26 \\
$\left(\mathrm{E} \rightarrow {}_{\mathrm{U}}\text{\textsc{RU}} + {}_{\mathrm{U}}\text{\textsc{RU}}\right) + \mathrm{AR}$ &
  1,143.99 &
  922.78 &
  12,695.98 &
  12,119.30 &
  3.33 &
  0.31 &
  1,083.09 &
  848.17 &
  11,669.35 &
  12,669.18 &
  3.42 &
  0.26 &
  247.24 &
  76.05 &
  1,545.03 &
  440.14 &
  0.11 &
  0.06 \\
$\left(\mathrm{E} \rightarrow {}_{\mathrm{U}}\text{\textsc{RU}}\right) + \mathrm{AR}$ &
  305.72 &
  199.80 &
  2,147.26 &
  1,840.07 &
  0.12 &
  0.07 &
  243.91 &
  138.62 &
  1,573.45 &
  874.97 &
  0.33 &
  0.30 &
  242.51 &
  106.62 &
  1,505.87 &
  750.02 &
  0.11 &
  0.06 \\
${}_{\mathrm{U}}\text{\textsc{RU}}$ &
  4,313.01 &
  2,677.42 &
  24,016.09 &
  22,526.52 &
  10.75 &
  1.40 &
  3,370.24 &
  2,602.69 &
  16,814.62 &
  16,794.95 &
  6.01 &
  0.87 &
  3,074.70 &
  2,227.76 &
  16,504.10 &
  15,659.24 &
  4.31 &
  0.44 \\ \cline{2-19} 
\end{tabular}%
}
\end{table}
\end{landscape}\clearpage
\loadgeometry{geom}

\savegeometry{geom}
\newgeometry{margin=0pt}
\thispagestyle{empty}\setcounter{page}{27}
\begin{landscape}
\begin{table}
\centering
\caption[Ablation on the complete SARS-CoV-2 dataset.]%
{Ablation on the complete SARS-CoV-2 dataset.\\%
{\footnotesize\textbf{Legend:} Algorithms with the best performance are in \textbf{bold}; besides, we refer to the Transformer Encoder as $\mathrm{E}$, recurrent unit as \textsc{RU}, Bidirectional as \textsc{B}, Unidirectional as \textsc{U}, and Autoregressive as \textsc{AR}.}}
\label{tab:ablation-9}
\resizebox{1.35\textwidth}{!}{%
\begin{tabular}{r|cccccc|cccccc|cccccc|}
\cline{2-19}
\multicolumn{1}{c|}{} &
  \multicolumn{6}{c|}{\textbf{PyTorch's Hyperparameters}} &
  \multicolumn{6}{c|}{\textbf{Literature's Hyperparameters}} &
  \multicolumn{6}{c|}{\ReGENN\textbf{'s Hyperparameters}} \\ \cline{2-19} 
\multicolumn{1}{c|}{} &
  \multicolumn{1}{c}{\textbf{MAE}} &
  \multicolumn{1}{c}{\textbf{$\pm$ STD}} &
  \multicolumn{1}{c}{\textbf{MSLE}} &
  \multicolumn{1}{c}{\textbf{$\pm$ STD}} &
  \multicolumn{1}{c}{\textbf{RMSE}} &
  \textbf{$\pm$ STD} &
  \multicolumn{1}{c}{\textbf{MAE}} &
  \multicolumn{1}{c}{\textbf{$\pm$ STD}} &
  \multicolumn{1}{c}{\textbf{MSLE}} &
  \multicolumn{1}{c}{\textbf{$\pm$ STD}} &
  \multicolumn{1}{c}{\textbf{RMSE}} &
  \textbf{$\pm$ STD} &
  \multicolumn{1}{c}{\textbf{MAE}} &
  \multicolumn{1}{c}{\textbf{$\pm$ STD}} &
  \multicolumn{1}{c}{\textbf{MSLE}} &
  \multicolumn{1}{c}{\textbf{$\pm$ STD}} &
  \multicolumn{1}{c}{\textbf{RMSE}} &
  \textbf{$\pm$ STD} \\ \cline{2-19} 
\multicolumn{1}{l|}{} &
  \multicolumn{18}{c|}{\textbf{Recurrent Neural Network (RNN)}} \\ \cline{2-19} 
${}_{\mathrm{B}}\text{\textsc{RU}}$ &
  434.18 &
  250.54 &
  2,678.49 &
  2,046.82 &
  0.09 &
  0.05 &
  1,895.13 &
  995.98 &
  7,184.26 &
  5,885.75 &
  0.08 &
  0.03 &
  424.32 &
  200.68 &
  2,819.11 &
  2,021.66 &
  0.07 &
  0.04 \\
$\mathrm{E} \rightarrow {}_{\mathrm{B}}\text{\textsc{RU}}$ &
  3,680.89 &
  2,404.90 &
  25,773.58 &
  22,768.94 &
  7.47 &
  0.80 &
  2,510.11 &
  1,324.54 &
  9,521.55 &
  6,720.47 &
  0.19 &
  0.05 &
  4,161.35 &
  3,028.85 &
  28,755.33 &
  28,712.61 &
  11.08 &
  1.31 \\
$\left(\mathrm{E} \rightarrow {}_{\mathrm{B}}\text{\textsc{RU}} + {}_{\mathrm{B}}\text{\textsc{RU}}\right) + \mathrm{AR}$ &
  1,533.62 &
  949.53 &
  18,013.32 &
  16,584.35 &
  3.82 &
  0.16 &
  317.13 &
  184.28 &
  1,806.87 &
  1,269.50 &
  0.10 &
  0.07 &
  1,624.44 &
  1,221.57 &
  17,245.24 &
  17,218.81 &
  3.84 &
  0.49 \\
$\left(\mathrm{E} \rightarrow {}_{\mathrm{B}}\text{\textsc{RU}} + {}_{\mathrm{U}}\text{\textsc{RU}}\right) + \mathrm{AR}$ &
  \textbf{245.06} &
  \textbf{65.03} &
  \textbf{1,359.34} &
  \textbf{462.61} &
  \textbf{0.09} &
  \textbf{0.05} &
  279.38 &
  115.14 &
  1,569.99 &
  836.28 &
  0.09 &
  0.07 &
  \textbf{257.15} &
  \textbf{149.23} &
  \textbf{1,438.72} &
  \textbf{823.36} &
  \textbf{0.09} &
  \textbf{0.06} \\
$\left(\mathrm{E} \rightarrow {}_{\mathrm{B}}\text{\textsc{RU}}\right) + \mathrm{AR}$ &
  1,386.30 &
  859.67 &
  16,848.20 &
  15,758.33 &
  3.72 &
  0.45 &
  304.52 &
  141.03 &
  2,029.00 &
  1,595.46 &
  0.10 &
  0.04 &
  1,471.49 &
  952.84 &
  17,050.08 &
  15,672.77 &
  3.74 &
  0.34 \\
$\mathrm{E} \rightarrow {}_{\mathrm{U}}\text{\textsc{RU}}$ &
  3,171.99 &
  1,445.00 &
  21,117.23 &
  16,502.10 &
  3.92 &
  0.10 &
  2,692.19 &
  2,133.65 &
  8,934.01 &
  8,459.59 &
  0.29 &
  0.18 &
  1,949.65 &
  1,188.93 &
  16,925.69 &
  16,149.04 &
  3.84 &
  0.41 \\
$\left(\mathrm{E} \rightarrow {}_{\mathrm{U}}\text{\textsc{RU}} + {}_{\mathrm{B}}\text{\textsc{RU}}\right) + \mathrm{AR}$ &
  350.75 &
  202.78 &
  1,985.39 &
  1,228.78 &
  0.11 &
  0.03 &
  280.09 &
  142.50 &
  1,669.51 &
  1,096.35 &
  0.10 &
  0.03 &
  372.80 &
  190.41 &
  2,155.36 &
  1,322.94 &
  0.11 &
  0.07 \\
$\left(\mathrm{E} \rightarrow {}_{\mathrm{U}}\text{\textsc{RU}} + {}_{\mathrm{U}}\text{\textsc{RU}}\right) + \mathrm{AR}$ &
  1,386.73 &
  641.20 &
  17,459.45 &
  14,421.01 &
  3.70 &
  0.30 &
  1,423.25 &
  738.45 &
  17,459.46 &
  14,825.39 &
  3.72 &
  0.39 &
  1,472.05 &
  916.89 &
  16,491.81 &
  15,674.45 &
  3.72 &
  0.26 \\
$\left(\mathrm{E} \rightarrow {}_{\mathrm{U}}\text{\textsc{RU}}\right) + \mathrm{AR}$ &
  1,375.60 &
  988.31 &
  15,987.09 &
  16,378.54 &
  3.69 &
  0.34 &
  \textbf{240.34} &
  \textbf{93.62} &
  \textbf{1,386.81} &
  \textbf{832.15} &
  \textbf{0.10} &
  \textbf{0.06} &
  1,350.35 &
  817.74 &
  16,673.11 &
  15,280.76 &
  3.69 &
  0.33 \\
${}_{\mathrm{U}}\text{\textsc{RU}}$ &
  1,277.24 &
  1,053.33 &
  7,948.96 &
  7,430.65 &
  2.89 &
  0.96 &
  1,954.90 &
  1,154.82 &
  7,310.14 &
  6,297.24 &
  0.21 &
  0.13 &
  1,175.99 &
  768.83 &
  6,716.42 &
  5,337.00 &
  2.08 &
  0.73 \\ \cline{2-19} 
\multicolumn{1}{l|}{} &
  \multicolumn{18}{c|}{\textbf{Gated Recurrent Unit (GRU)}} \\ \cline{2-19} 
${}_{\mathrm{B}}\text{\textsc{RU}}$ &
  3,167.33 &
  1,458.27 &
  28,834.18 &
  23,983.55 &
  7.80 &
  0.49 &
  3,094.99 &
  2,293.39 &
  18,854.48 &
  19,286.26 &
  3.89 &
  0.31 &
  3,130.47 &
  2,065.13 &
  28,063.07 &
  24,375.98 &
  7.78 &
  0.75 \\
$\mathrm{E} \rightarrow {}_{\mathrm{B}}\text{\textsc{RU}}$ &
  950.11 &
  433.35 &
  5,133.01 &
  2,301.34 &
  0.19 &
  0.14 &
  2,702.04 &
  2,483.15 &
  10,715.26 &
  13,157.31 &
  0.18 &
  0.10 &
  596.77 &
  403.89 &
  3,446.67 &
  2,746.08 &
  0.17 &
  0.16 \\
$\left(\mathrm{E} \rightarrow {}_{\mathrm{B}}\text{\textsc{RU}} + {}_{\mathrm{B}}\text{\textsc{RU}}\right) + \mathrm{AR}$ &
  311.09 &
  119.17 &
  1,873.65 &
  1,082.02 &
  0.10 &
  0.07 &
  219.16 &
  106.06 &
  1,276.78 &
  881.66 &
  0.07 &
  0.03 &
  388.93 &
  335.26 &
  1,888.70 &
  1,754.53 &
  0.09 &
  0.03 \\
$\left(\mathrm{E} \rightarrow {}_{\mathrm{B}}\text{\textsc{RU}} + {}_{\mathrm{U}}\text{\textsc{RU}}\right) + \mathrm{AR}$ &
  650.06 &
  185.77 &
  3,443.73 &
  1,270.04 &
  0.10 &
  0.04 &
  259.91 &
  149.24 &
  1,597.42 &
  1,273.26 &
  0.11 &
  0.07 &
  366.06 &
  175.38 &
  2,158.32 &
  1,206.14 &
  0.10 &
  0.04 \\
$\left(\mathrm{E} \rightarrow {}_{\mathrm{B}}\text{\textsc{RU}}\right) + \mathrm{AR}$ &
  1,377.91 &
  517.07 &
  17,145.02 &
  14,067.11 &
  3.66 &
  0.10 &
  505.81 &
  582.37 &
  3,916.95 &
  5,297.45 &
  0.11 &
  0.08 &
  1,502.20 &
  1,005.35 &
  16,799.96 &
  14,836.87 &
  3.67 &
  0.39 \\
$\mathrm{E} \rightarrow {}_{\mathrm{U}}\text{\textsc{RU}}$ &
  1,360.35 &
  737.84 &
  7,611.46 &
  4,036.40 &
  0.27 &
  0.11 &
  2,380.63 &
  1,475.81 &
  8,555.18 &
  6,771.74 &
  0.23 &
  0.12 &
  968.83 &
  673.75 &
  5,982.35 &
  5,430.49 &
  0.22 &
  0.07 \\
$\left(\mathrm{E} \rightarrow {}_{\mathrm{U}}\text{\textsc{RU}} + {}_{\mathrm{B}}\text{\textsc{RU}}\right) + \mathrm{AR}$ &
  \textbf{240.35} &
  \textbf{57.07} &
  \textbf{1,335.96} &
  \textbf{343.59} &
  \textbf{0.08} &
  \textbf{0.04} &
  245.62 &
  159.01 &
  1,367.82 &
  966.48 &
  0.07 &
  0.03 &
  \textbf{208.88} &
  \textbf{119.65} &
  \textbf{1,141.56} &
  \textbf{575.75} &
  \textbf{0.08} &
  \textbf{0.04} \\
$\left(\mathrm{E} \rightarrow {}_{\mathrm{U}}\text{\textsc{RU}} + {}_{\mathrm{U}}\text{\textsc{RU}}\right) + \mathrm{AR}$ &
  3,377.44 &
  1,741.87 &
  25,335.28 &
  22,336.83 &
  7.44 &
  1.07 &
  \textbf{216.81} &
  \textbf{107.29} &
  \textbf{1,280.58} &
  \textbf{846.01} &
  \textbf{0.11} &
  \textbf{0.04} &
  4,052.71 &
  2,625.95 &
  31,384.80 &
  28,694.03 &
  11.30 &
  1.27 \\
$\left(\mathrm{E} \rightarrow {}_{\mathrm{U}}\text{\textsc{RU}}\right) + \mathrm{AR}$ &
  1,592.62 &
  895.15 &
  18,207.40 &
  15,612.58 &
  3.77 &
  0.39 &
  248.10 &
  166.88 &
  1,508.53 &
  1,184.28 &
  0.09 &
  0.04 &
  2,852.28 &
  1,735.94 &
  25,869.18 &
  23,701.61 &
  7.60 &
  0.59 \\
${}_{\mathrm{U}}\text{\textsc{RU}}$ &
  2,319.57 &
  1,793.72 &
  21,536.86 &
  21,418.12 &
  6.26 &
  0.55 &
  1,875.92 &
  1,328.11 &
  6,933.49 &
  6,460.49 &
  0.11 &
  0.08 &
  1,825.66 &
  650.78 &
  20,400.04 &
  15,797.17 &
  4.95 &
  0.20 \\ \cline{2-19} 
\multicolumn{1}{l|}{} &
  \multicolumn{18}{c|}{\textbf{Long Short-Term Memory (LSTM)}} \\ \cline{2-19} 
${}_{\mathrm{B}}\text{\textsc{RU}}$ &
  3,000.14 &
  1,954.01 &
  25,989.65 &
  24,313.81 &
  7.65 &
  0.78 &
  3,042.70 &
  1,987.48 &
  18,539.83 &
  17,591.06 &
  3.80 &
  0.41 &
  1,800.23 &
  928.80 &
  20,264.27 &
  17,544.01 &
  3.95 &
  0.15 \\
$\mathrm{E} \rightarrow {}_{\mathrm{B}}\text{\textsc{RU}}$ &
  619.31 &
  356.13 &
  3,094.25 &
  1,878.42 &
  0.17 &
  0.11 &
  2,309.35 &
  1,269.98 &
  7,953.65 &
  6,114.91 &
  0.23 &
  0.15 &
  704.80 &
  343.17 &
  4,088.05 &
  2,893.91 &
  0.20 &
  0.10 \\
$\left(\mathrm{E} \rightarrow {}_{\mathrm{B}}\text{\textsc{RU}} + {}_{\mathrm{B}}\text{\textsc{RU}}\right) + \mathrm{AR}$ &
  \textbf{216.01} &
  \textbf{89.11} &
  \textbf{1,146.94} &
  \textbf{465.90} &
  \textbf{0.10} &
  \textbf{0.07} &
  256.32 &
  59.69 &
  1,412.55 &
  487.27 &
  0.12 &
  0.07 &
  1,598.03 &
  884.02 &
  19,056.28 &
  16,854.23 &
  3.89 &
  0.17 \\
$\left(\mathrm{E} \rightarrow {}_{\mathrm{B}}\text{\textsc{RU}} + {}_{\mathrm{U}}\text{\textsc{RU}}\right) + \mathrm{AR}$ &
  2,863.94 &
  1,345.36 &
  27,905.87 &
  22,477.61 &
  7.69 &
  0.62 &
  \textbf{244.81} &
  \textbf{128.06} &
  \textbf{1,233.11} &
  \textbf{666.28} &
  \textbf{0.18} &
  \textbf{0.21} &
  4,064.52 &
  2,702.07 &
  31,196.10 &
  28,670.34 &
  11.30 &
  1.01 \\
$\left(\mathrm{E} \rightarrow {}_{\mathrm{B}}\text{\textsc{RU}}\right) + \mathrm{AR}$ &
  4,060.62 &
  2,655.67 &
  31,538.83 &
  28,960.85 &
  11.35 &
  0.99 &
  1,554.31 &
  1,416.54 &
  17,417.78 &
  18,060.72 &
  3.90 &
  0.51 &
  \textbf{211.93} &
  \textbf{83.35} &
  \textbf{1,181.31} &
  \textbf{570.10} &
  \textbf{0.08} &
  \textbf{0.06} \\
$\mathrm{E} \rightarrow {}_{\mathrm{U}}\text{\textsc{RU}}$ &
  3,199.26 &
  2,436.81 &
  12,581.64 &
  11,769.69 &
  0.26 &
  0.07 &
  3,169.42 &
  2,063.50 &
  12,461.42 &
  11,674.12 &
  0.17 &
  0.05 &
  2,778.55 &
  1,112.16 &
  12,830.03 &
  10,175.60 &
  0.21 &
  0.06 \\
$\left(\mathrm{E} \rightarrow {}_{\mathrm{U}}\text{\textsc{RU}} + {}_{\mathrm{B}}\text{\textsc{RU}}\right) + \mathrm{AR}$ &
  1,351.42 &
  644.34 &
  17,065.24 &
  14,155.37 &
  3.67 &
  0.30 &
  1,375.17 &
  1,053.14 &
  15,406.60 &
  15,929.81 &
  3.69 &
  0.34 &
  1,380.99 &
  864.20 &
  16,370.41 &
  14,977.34 &
  3.66 &
  0.32 \\
$\left(\mathrm{E} \rightarrow {}_{\mathrm{U}}\text{\textsc{RU}} + {}_{\mathrm{U}}\text{\textsc{RU}}\right) + \mathrm{AR}$ &
  1,431.78 &
  946.29 &
  17,235.49 &
  16,431.19 &
  3.79 &
  0.38 &
  285.02 &
  121.63 &
  1,802.48 &
  1,118.25 &
  0.11 &
  0.06 &
  1,713.72 &
  1,025.91 &
  18,531.68 &
  15,888.72 &
  3.81 &
  0.39 \\
$\left(\mathrm{E} \rightarrow {}_{\mathrm{U}}\text{\textsc{RU}}\right) + \mathrm{AR}$ &
  222.06 &
  85.96 &
  1,265.74 &
  642.42 &
  0.08 &
  0.03 &
  265.80 &
  112.59 &
  1,437.64 &
  686.42 &
  0.14 &
  0.09 &
  260.66 &
  215.01 &
  1,513.92 &
  1,430.67 &
  0.10 &
  0.03 \\
${}_{\mathrm{U}}\text{\textsc{RU}}$ &
  4,943.62 &
  2,892.09 &
  28,696.47 &
  24,921.66 &
  7.59 &
  0.30 &
  4,942.10 &
  3,425.34 &
  25,691.11 &
  25,415.68 &
  7.38 &
  0.97 &
  3,723.88 &
  2,536.94 &
  21,022.60 &
  20,012.26 &
  2.83 &
  0.31 \\ \cline{2-19} 
\end{tabular}%
}
\end{table}
\end{landscape}\clearpage
\loadgeometry{geom}